\documentclass{article} %
\usepackage{newtxtext}
\usepackage[left=1.25in,top=1in,bottom=1in,right=1.25in]{geometry}

\usepackage[authoryear]{natbib}
\usepackage[utf8]{inputenc}
\usepackage[T1]{fontenc}
\usepackage{microtype}
\usepackage{amsmath,amsfonts,bm}

\def\eqref#1{equation~\ref{#1}}

\def\1{\bm{1}}

\DeclareMathAlphabet{\mathsfit}{\encodingdefault}{\sfdefault}{m}{sl}
\SetMathAlphabet{\mathsfit}{bold}{\encodingdefault}{\sfdefault}{bx}{n}

\newcommand{\E}{\mathbb{E}}

\usepackage[colorlinks=true,citecolor=blue,linkcolor=blue,urlcolor=blue]{hyperref}
\usepackage{cleveref}
\usepackage{url}
\usepackage{amsthm}
\usepackage{booktabs}
\usepackage{longtable}
\usepackage{array}
\usepackage{graphicx}
\usepackage{wrapfig}
\usepackage{float}
\usepackage{placeins}
\usepackage{multicol}
\usepackage{xcolor}
\usepackage{tikz}
\usepackage{enumitem}
\usetikzlibrary{arrows.meta,calc,positioning}
\usepackage[most]{tcolorbox}

\lstdefinestyle{promptcode}{
  basicstyle=\ttfamily\fontsize{7.2pt}{8.6pt}\selectfont,
  breaklines=true,
  breakatwhitespace=true,
  breakindent=1.25em,
  columns=fullflexible,
  keepspaces=true,
  showstringspaces=false,
  aboveskip=0pt,
  belowskip=0pt,
  xleftmargin=0pt,
  xrightmargin=0pt
}

\newtcblisting{promptbox}[1][Prompt]{
  enhanced,
  breakable,
  listing only,
  listing engine=listings,
  listing options={style=promptcode},
  title={#1},
  fonttitle=\sffamily\bfseries\scriptsize,
  coltitle=blue!55!black,
  colbacktitle=blue!5,
  colback=white,
  colframe=blue!18,
  boxrule=0.35pt,
  sharp corners,
  left=1.5mm,
  right=1.5mm,
  top=1.2mm,
  bottom=1.2mm,
  boxsep=0pt,
  titlerule=0.35pt,
  before skip=0.8em,
  after skip=0.9em
}

\newtcolorbox{ideabox}[2][]{
  enhanced,
  breakable,
  title={#2},
  fonttitle=\sffamily\bfseries\small,
  fontupper=\normalsize,
  before upper={\setlength{\parindent}{0pt}},
  coltitle=blue!55!black,
  colbacktitle=blue!5,
  colback=white,
  colframe=blue!18,
  boxrule=0.4pt,
  arc=1mm,
  left=2mm,
  right=2mm,
  top=1.2mm,
  bottom=1.2mm,
  boxsep=0pt,
  titlerule=0.4pt,
  before skip=0.8em,
  after skip=0.9em,
  #1
}

\newcommand{\ideafield}[1]{%
  \par\addvspace{0.35em}%
  \noindent{\sffamily\bfseries\color{blue!55!black}#1}\par
  \nobreak\vspace{0.08em}\ignorespaces
}

\newcommand{\ideaomit}[1]{%
  \par\addvspace{0.15em}%
  \noindent{\sffamily\itshape\color{black!55}[#1]}\par
  \nobreak\vspace{0.08em}\ignorespaces
}

\newtheorem{theorem}{Theorem}[section]

\title{Priced Guidance: Can Language Models Generate Future Research Ideas?}

\author{
Kaiyue Wen \\
Stanford University \\
{\small\texttt{kaiyuew@stanford.edu}}
\and
Tengyu Ma \\
Stanford University \\
{\small\texttt{tengyuma@stanford.edu}}
\and
Percy Liang \\
Stanford University \\
{\small\texttt{pliang@cs.stanford.edu}}
}

\date{}
\begin{document}
\raggedbottom

\maketitle

\begin{abstract}
We evaluate language models' capability to generate novel research ideas through the lens of compression. Our goal is to lower-bound the potentially tiny probability that a language model generates the essence of a future research idea without any hints. Rather than estimate this probability through expensive repeated sampling, our Priced Guidance framework measures the compression cost: how many additional bits of information are needed to guide the model to recover the target idea. We prove that if the model can recover the target idea with at most \(K\) bits of guidance in expectation, then it can generate the idea without any guidance with probability at least \(2^{-K}\).
In our framework, the language model, called the generator, can pose a sequence of multiple-choice questions, e.g., ``what is the topic of the research idea?'', and specify a probability distribution over possible answers. A guide, which is a language model with access to the target idea, selects answers. If a selected answer has prior probability \(p\), the generator pays \(-\log_2 p\) bits.
The generator's goal is to minimize the cumulative cost it needs to pay to produce an idea that matches the essence of the target idea. This cumulative cost equals, up to an additive constant, the number of bits of information sent by the guide.
Using this methodology, we evaluate five generator models (Opus~5, Fable~5.1, GPT-6 Astra, GPT-5.6 Sol, and GLM~5.3) on the core ideas in 87 recent high-quality deep learning papers. We use a frontier model as a judge to determine whether the generated idea matches the target in terms of the central research object and defining mechanism. Fable~5.1 achieves the lowest median compression cost at 69.9 bits, substantially lower than gzip's median of 5,712 bits for losslessly compressing the summary of the target idea. A uniform ensemble of Fable~5.1, Opus~5, and GPT-6 Astra further reduces the median compression cost to 55.8 bits and improves the generation probability lower bound by \(18{,}000\times\).\footnote{Our code is open source and available at \url{https://github.com/WhenWen/priced-guidance}.}
\end{abstract}

\section{Introduction}
\label{sec:introduction}

{The ability to generate novel research ideas is a critical part of advanced
intelligence.} Yet evaluating LLM-generated ideas is difficult because
novelty perceived by humans or language models does not necessarily translate into better research
outcomes~\citep{si2024novelideas,si2025ideationexecutiongapexecutionoutcomes,baig2026taste}.
A potential approach is to test whether a model can generate an
unseen future research idea. Existing benchmarks
provide partial information such as keywords~\citep{ruan2024liveideabench} or parent
papers~\citep{he2026giants} to guide the reconstruction, but do not quantify how much information this guidance provides.

In this paper, we measure a language model's idea-generation capability through the lens of \textbf{compression}~\citep{shannon1948communication,schmidhuber2009compression,schmidhuber2010creativity}: how many additional bits are needed to guide the model toward the essence of an idea from a target future paper? The intuition comes from the classical duality between prediction and compression: what a model can already predict well requires little additional information to specify, while what it finds surprising requires more.

We propose \emph{Priced Guidance}
(\Cref{fig:generator-oracle-tree-search}), a framework to quantify
compression cost of ideas. We take inspiration from the game of Twenty Questions by endowing the model being evaluated, which we call the generator,  with a particular structure:
before the generator produces its final idea, its reasoning trace contains a sequence
of \textbf{bids}. Each bid consists of a question and a probability
distribution over a finite set of candidate answers. We then introduce a guide language
model with privileged access to the target paper, which chooses among these answers and thereby guides the generator toward the target idea.
When the guide chooses an answer to which the generator assigned
probability $p$, the generator pays a cost of $-\log_2 p$ bits.
The interaction ends when the
generator chooses to submit an idea, and an LLM judge evaluates whether the generated idea is equivalent in essence to the
target idea. If the final idea passes this judge, the sequence of guide answers can then be viewed as a compressed
representation of the additional guidance needed for the generator to recover the target idea.
With the generator and its randomness fixed, arithmetic coding lets us reconstruct the answer sequence and hence the generated idea from at most the total cost plus \(O(1)\) bits. We therefore call the total answer cost the \textbf{compression cost}~(see \Cref{sec:priced-oracle-hints} for a formal definition). Here the guide may be arbitrarily knowledgeable and computationally powerful because the protocol  measures the cost of the information it communicates rather than restricting how it obtains that information.

Priced Guidance generalizes the use of language models to compress a target text verbatim~\citep{deletang2024language}, with each bid asking for the next token of the target summary. In our framework, the generator may perform multi-step reasoning between bids and is allowed to generate an idea equivalent in essence to the target rather than a particular textual realization. This also avoids requiring token log-probabilities, which are unavailable for many closed-weight frontier models. Just as a lower compression cost on text indicates a stronger language model, a lower compression cost under Priced Guidance indicates a stronger idea generator.

\begin{figure}[tbp]
\centering
\begin{tikzpicture}[
  font=\sffamily\scriptsize,
  generator/.style={
    draw=blue!70!black,
    fill=blue!4,
    rounded corners=2pt,
    line width=0.7pt,
    minimum width=5.0cm,
    minimum height=0.72cm,
    text width=4.7cm,
    align=center,
    inner sep=3pt
  },
  selected/.style={
    draw=blue!75!black,
    fill=blue!10,
    rounded corners=2pt,
    line width=0.8pt,
    minimum width=1.95cm,
    minimum height=0.42cm,
    font=\sffamily\tiny,
    align=center,
    inner sep=2pt
  },
  alternative/.style={
    draw=gray!55,
    fill=gray!5,
    rounded corners=2pt,
    line width=0.5pt,
    minimum width=1.95cm,
    minimum height=0.42cm,
    font=\sffamily\tiny,
    text=gray!75!black,
    align=center,
    inner sep=2pt
  },
  idea/.style={
    draw=blue!75!black,
    fill=blue!7,
    rounded corners=2pt,
    line width=0.8pt,
    minimum width=6.2cm,
    minimum height=0.82cm,
    text width=5.85cm,
    align=center,
    inner sep=4pt
  },
  guide box/.style={
    draw=orange!85!black,
    fill=orange!8,
    rounded corners=2pt,
    line width=0.8pt,
    minimum width=2.9cm,
    minimum height=0.72cm,
    text width=2.6cm,
    align=center,
    inner sep=3pt
  },
  judge/.style={
    draw=green!55!black,
    fill=green!7,
    rounded corners=2pt,
    line width=0.8pt,
    minimum width=3.6cm,
    minimum height=0.82cm,
    text width=3.3cm,
    align=center,
    inner sep=3pt
  },
  result/.style={
    draw=black!60,
    fill=black!2,
    rounded corners=2pt,
    line width=0.6pt,
    minimum width=11.5cm,
    minimum height=0.68cm,
    text width=11.0cm,
    align=center,
    inner sep=3pt
  },
  chosen edge/.style={blue!75!black, line width=1.15pt, -{Stealth[length=2.0mm]}},
  other edge/.style={gray!55, line width=0.55pt, -{Stealth[length=1.6mm]}},
  guide edge/.style={orange!85!black, dashed, line width=0.85pt, -{Stealth[length=1.8mm]}},
  info label/.style={font=\sffamily\tiny\bfseries, text=orange!85!black, fill=white, inner sep=1pt},
  continuation/.style={text=gray!55, font=\sffamily\footnotesize}
]
  \node[font=\sffamily\scriptsize\bfseries, text=blue!70!black] at (-2.80,0.67)
    {generator};
  \node[generator] (generator) at (-2.80,0)
    {Proposes each bid: a question\\and an answer distribution};

  \node[font=\sffamily\scriptsize\bfseries, text=orange!85!black] at (4.15,0.67)
    {guide};
  \node[guide box] (guide) at (4.15,0)
    {Privileged access to the hidden target paper: \textbf{ResNet}};

  \node[generator] (bidone) at (-2.80,-0.94)
    {\textbf{Bid 1:} Which research domain\\contains the target idea?};
  \draw[chosen edge] (generator) -- (bidone);

  \node[alternative] (onea) at (-5.60,-1.82)
    {NLP\enspace $p=.45$};
  \node[alternative] (oneb) at (-2.70,-1.82)
    {robotics\enspace $p=.25$};
  \node[selected] (onec) at (0,-1.82)
    {\textbf{computer vision}\enspace $p=.30$};
  \draw[other edge] (bidone.south) -- (onea.north);
  \draw[other edge] (bidone.south) -- (oneb.north);
  \draw[chosen edge] (bidone.south) -- (onec.north);
  \node[continuation] at (-5.60,-2.17) {$\vdots$};
  \node[continuation] at (-2.70,-2.17) {$\vdots$};

  \node[generator] (bidtwo) at (-2.80,-2.64)
    {\textbf{Bid 2:} What problem does the target idea address?};
  \draw[chosen edge] (onec) -- (bidtwo);

  \node[alternative] (twoa) at (-5.60,-3.52)
    {vanishing gradients\enspace $p=.45$};
  \node[alternative] (twob) at (-2.70,-3.52)
    {overfitting\enspace $p=.35$};
  \node[selected] (twoc) at (0,-3.52)
    {\textbf{More depth hurts}\enspace $p=.20$};
  \draw[other edge] (bidtwo.south) -- (twoa.north);
  \draw[other edge] (bidtwo.south) -- (twob.north);
  \draw[chosen edge] (bidtwo.south) -- (twoc.north);
  \node[continuation] at (-5.60,-3.87) {$\vdots$};
  \node[continuation] at (-2.70,-3.87) {$\vdots$};

  \node[generator] (bidthree) at (-2.80,-4.34)
    {\textbf{Bid 3:} What architectural change addresses this problem?};
  \draw[chosen edge] (twoc) -- (bidthree);

  \node[alternative] (threea) at (-5.60,-5.22)
    {plain network\enspace $p=.55$};
  \node[alternative] (threeb) at (-2.70,-5.22)
    {gated shortcut\enspace $p=.35$};
  \node[selected] (threec) at (0,-5.22)
    {\textbf{identity skip}\enspace $p=.10$};
  \draw[other edge] (bidthree.south) -- (threea.north);
  \draw[other edge] (bidthree.south) -- (threeb.north);
  \draw[chosen edge] (bidthree.south) -- (threec.north);
  \node[continuation] at (-5.60,-5.57) {$\vdots$};
  \node[continuation] at (-2.70,-5.57) {$\vdots$};

  \draw[orange!85!black, dashed, line width=0.75pt]
    (guide.south) -- (4.15,-5.22);
  \draw[guide edge] (4.15,-1.82) --
    node[info label, above=1pt] {$1.74$ bits} (onec.east);
  \draw[guide edge] (4.15,-3.52) --
    node[info label, above=1pt] {$2.32$ bits} (twoc.east);
  \draw[guide edge] (4.15,-5.22) --
    node[info label, above=1pt] {$3.32$ bits} (threec.east);
  \node[idea] (finalidea) at (-2.80,-6.36)
    {\textbf{Generated idea:} learn residuals with identity shortcuts,
    $y=x+\mathcal{F}(x)$;\\this enables training 152-layer networks.};
  \draw[chosen edge] (threec) -- (finalidea);

  \node[judge] (finaljudge) at (4.15,-6.36)
    {\textbf{LLM judge}\\Equivalent to the hidden paper's main idea?\\
    {\color{green!50!black}\textbf{PASS}}};
  \draw[chosen edge] (finalidea.east) -- (finaljudge.west);

  \node[result] at (-0.20,-7.48)
    {\textbf{Unguided action replay of this path}\\[-1pt]
    $\prod_t p_t=.30\times.20\times.10=.006=2^{-K}$,
    where $K=7.38$ bits.};
\end{tikzpicture}
\caption{A top-down view of the Priced Guidance protocol as tree search, illustrated with {the ResNet idea}. At each bid, the generator asks a question and assigns probabilities to possible answers, defining branches of the tree. The guide uses privileged access to the target paper to select an answer. Orange arrows show the information sent from the guide to the generator, labeled by its cost in bits; blue highlights the resulting trajectory. For a fixed generator realization and a judge that accepts this idea with probability one, $2^{-K}$ lower-bounds the probability of generating an equivalent idea without guidance. Probabilities and questions are stylized here.
}
\label{fig:generator-oracle-tree-search}
\end{figure}

The generator structure allows us to connect compression cost directly to the probability of reconstructing the target idea without any additional information. In addition to the \emph{guided} mode, where the guide selects answers using its knowledge of the target idea, we can run the same generator in an \emph{unguided} mode, where it instead samples an answer from the distribution specified in each bid. We prove that, for a fixed target, if $K$ is the random compression cost of a guided run, then the corresponding unguided generator outputs an idea that passes the judge with probability at least $\mathbb{E}[2^{-K}]$ (\Cref{thm:generator-oracle}), where the expectation is over the randomness of the generator and guide. Thus, a low compression cost implies a high probability that the unguided generator produces an idea that passes the judge. Because the relationship is exponential, reducing the compression cost of each guided run by 10 bits improves the resulting upper bound on the expected number of unguided generations needed to produce an idea that passes the judge by roughly $1000\times$. 
\begin{figure}[t]
  \centering
  \includegraphics[width=\linewidth]{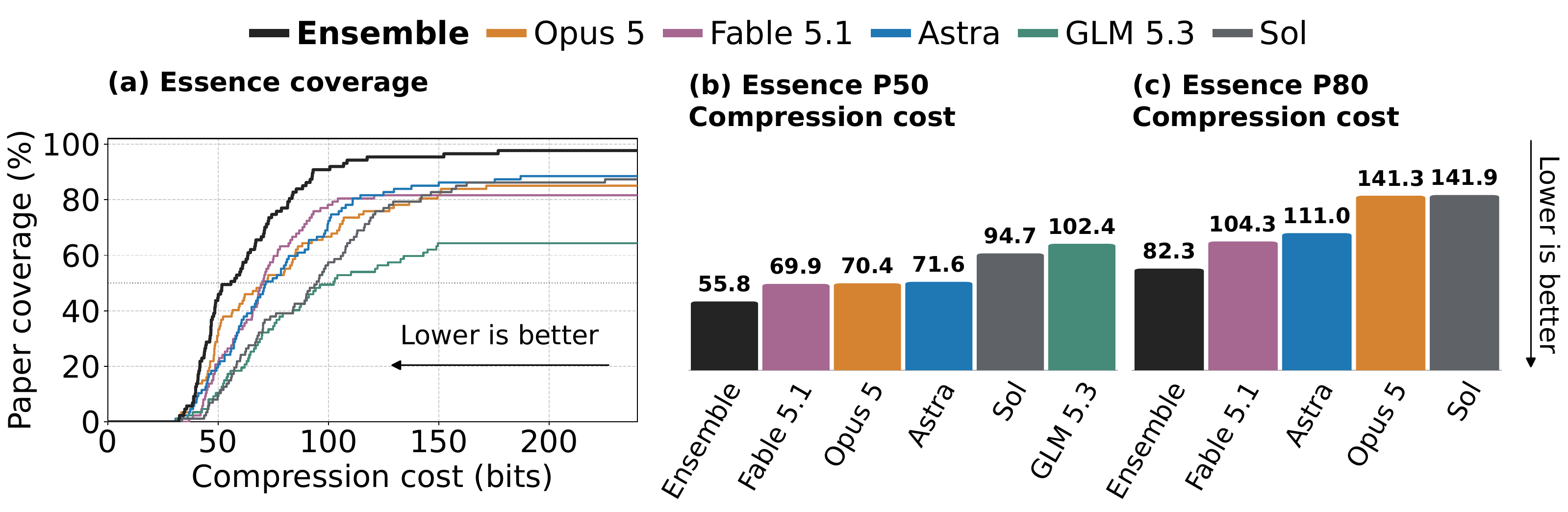}
  \caption{Test-set results under the Essence judge on 87 papers (judge criteria in Section~\ref{sec:recognition-criteria}).
  (a) Coverage versus compression cost. Curves show the percentage of papers
  successfully compressed at each compression cost for each generator
  model. The black ensemble uniformly mixes Fable~5.1, Opus~5, and Astra.
  (b) Median (P50) and (c) 80th-percentile (P80) compression cost. Here we
  define the $q$th-percentile compression cost as the minimal compression
  cost required to output an idea that passes the judge on $q\%$ of the test
  papers. GLM~5.3 is omitted from the P80 panel as it did not pass the essence
  judge on 80\% of the papers within the guide's API budget.
  The guide model is fixed to be Fable 5 here.}
  \label{fig:test87-essence}
\end{figure}

However, this bound should not be interpreted as implying that compression costs must become much smaller than now before a model can generate high-quality ideas autonomously. As with text, a model can generate high-quality outputs even when compressing a particular target would require thousands of bits. We therefore view compression cost as a measure of generation capability, rather than a threshold for autonomous generation.

We instantiate the generator as a specialized agent scaffold that asks questions ranging from the broad topic of the
paper to the key technical details of its target idea, and use Fable~5 as a fixed guide.
We then evaluate five generators on 87
recent Top-1 Hugging Face Daily Papers~\citep{huggingface_daily_papers}.
Our essence LLM judge requires the generator to propose a concrete idea that recovers the target paper's research
direction and central
research object (such as the core method).

Our framework yields much more efficient compression of research ideas than verbatim text compression using a language model or gzip.
For GLM~5.3, the median compression cost under the Essence judge is 102.4 bits, compared with 1,147.1 bits for compressing the gold idea summary verbatim with the same model and 5,712 bits with gzip.
We found that frontier models such as Fable~5.1 and Astra perform significantly better than the open-source model GLM~5.3, especially on papers that are harder to compress.
Opus~5 also performs competitively with the best models on papers that are easier to compress.
A uniform ensemble of Fable~5.1, Opus~5, and Astra (\Cref{sec:extensions}) further reduces the median compression cost, {demonstrating headroom for further reductions}.
Our ablation between GLM 5.2 and GLM 5.3 shows that generator post-training can substantially reduce compression cost (Section~\ref{sec:ablations}).

\section{Priced Guidance}
\label{sec:priced-oracle-hints}

Our goal is to quantify the cost of compressing a target idea, with equivalence evaluated by a fixed, possibly stochastic judge. Our protocol pairs a stochastic
\emph{generator}, which poses multiple-choice questions with preassigned answer
probabilities, with a \emph{guide}, which knows the target and guides generation
solely by selecting offered answers. In \emph{unguided generation}, the same
generator instead samples answers using those probabilities. The Priced Guidance Theorem~(\Cref{thm:generator-oracle})
lower-bounds unguided acceptance probability using the compression
cost, including a cost for the final idea's judge pass rate.
Figure~\ref{fig:generator-oracle-tree-search} illustrates this setting as a path
through a tree of possible generations.

\subsection{Priced Guidance Protocol}
\label{sec:generator-oracle-protocol}

Let $T$ be a target idea. A guided run under the priced generation protocol proceeds as follows.
\begin{enumerate}[leftmargin=*]
  \item \textbf{Loop.} For $t=1,2,\ldots$, until the generator submits a
  final idea instead of a bid:
  \begin{enumerate}
    \item \textbf{Bid.} Without access to $T$, the generator reasons and
    makes a bid. Let $\tau_t$ denote the trajectory up to step $t$:
    the generator's reasoning so far, its previous bids with their selected
    answers, and its current bid. The generator's \emph{bid} consists of a question,
    a finite set $\mathcal{A}_t(\tau_t)$ of answer
    choices called actions, and a probability $p_t(a\mid \tau_t)$ for each
    action $a$. The probabilities satisfy
    \begin{equation}
      p_t(a\mid \tau_t)>0,\qquad
      \sum_{a\in\mathcal{A}_t(\tau_t)}p_t(a\mid \tau_t)=1.
      \label{eq:bid-distribution}
    \end{equation}
    \item \textbf{Select and pay.} The guide, which knows $T$, selects an
    action $a_t\in\mathcal{A}_t(\tau_t)$, and the generator pays
    \begin{equation}
  c_t=-\log_2p_t(a_t\mid \tau_t).
  \label{eq:choice-cost}
\end{equation}
    \item \textbf{Update.} The selected action $a_t$ is appended to the
    trajectory. We write $g_t(\cdot\mid \tau_t,a_t)$ for the distribution
    of the next trajectory $\tau_{t+1}$, which adds the generator's
    subsequent reasoning and its next bid or final idea.
  \end{enumerate}
  \item \textbf{Judge.} The judge returns a possibly stochastic verdict
  $J_T(x)\in\{0,1\}$ on the final idea $x$, with one denoting acceptance as
  equivalent to $T$. The judge's pass rate for $x$ is
  $\Pr[J_T(x)=1]$, where the probability is over the judge's randomness with $x$ and $T$ fixed. The generator pays
  \begin{equation}
  c_J(x\mid T)=-\log_2 \Pr[J_T(x)=1].
  \label{eq:judge-cost}
\end{equation}
\end{enumerate}
An \emph{unguided} run is the same loop, except that step~(b) samples
$a_t\sim p_t(\cdot\mid \tau_t)$ from the generator's own bid.
The complete \emph{trajectory} $\tau$ consists of all reasoning, bids, and
selected answers, the final idea $x(\tau)$, and the judge's verdict
$B(\tau)\in\{0,1\}$. We write $|\tau|$ for its number of answered bids.
From the guide, the
generator receives only the selected actions. Guided and
unguided runs differ only in how step~(b) selects the action. The Priced
Guidance Theorem in \Cref{sec:generator-oracle-theorem} relies on this
property.

The total compression cost of a trajectory $\tau$ is
\begin{equation}
  K(\tau)=\sum_{t=1}^{|\tau|}c_t+c_J(x(\tau)\mid T)
  \label{eq:forward-cost}
\end{equation}.
Equivalently, $2^{-K(\tau)}=\Pr[J_T(x(\tau))=1]\prod_t p_t(a_t\mid \tau_t)$. An answer costs more when the generator assigned it lower probability.
A final idea costs more when the judge accepts it less often.
We set $K=+\infty$, with $2^{-\infty}=0$, when no final idea is returned or
its pass rate is zero.
A run succeeds when the judge accepts its final idea. The set of successful
trajectories is
$\mathcal{S}_T =\{\tau\mid B(\tau)=1\}$.
Trajectories with different final wordings or different sequences of
questions can all belong to $\mathcal{S}_T$.
Let $P_{\mathrm{guide}}$ and $P_{\mathrm{unguide}}$ denote the distributions
of complete trajectories from guided and unguided runs, respectively.
\Cref{sec:generator-oracle-theorem} lower-bounds the unguided success
probability $P_{\mathrm{unguide}}(\mathcal{S}_T)$ using the compression cost
of guided runs.

Figure~\ref{fig:generator-oracle-tree-search} shows a stylized guided run
with ResNet as the target idea. The first bid asks which research domain
contains the target idea and assigns probabilities 0.45, 0.25, and 0.30 to
NLP, robotics, and computer vision. In step~(b), the guide selects computer
vision, so the generator pays $-\log_2 0.30\approx1.74$ bits. After three
bids, the generator submits the idea of learning residuals with identity
shortcuts, and the judge accepts it. The three selected answers cost 7.38
bits in total.

\subsection{Priced Guidance Theorem}
\label{sec:generator-oracle-theorem}
The compression cost $K$ has two interpretations, which we state as two
theorems. \Cref{thm:compression} shows that $K$ bounds the number of bits
needed to describe a guided trajectory to the unguided generator. \Cref{thm:generator-oracle} shows that $K$ lower-bounds the
probability that the unguided generator produces an idea that the judge
accepts.

\begin{theorem}[Compression]
\label{thm:compression}
For fixed generator randomness, a single prefix code encodes the actions
the guide picks in every guided run $\tau$, for any target and guide, using
at most $K(\tau)+O(1)$ bits.
\end{theorem}

\begin{proof}[Proof sketch]
The compressed string is \emph{the sequence of actions
$a_1,\ldots,a_{|\tau|}$ that the guide picked}. Arithmetic coding with the
generator's bid probabilities $p_t(\cdot\mid \tau_t)$ encodes this sequence
in $\sum_t c_t+O(1)$ bits. The unguided generator recovers its reasoning and final idea
by reading these actions from the code instead of sampling them.
Appendix~\ref{app:conditional-coding} gives the details.
\end{proof}

\begin{theorem}[Priced Guidance Theorem]
\label{thm:generator-oracle}
For the stochastic generator and possibly stochastic guide and judge above,
the acceptance probability of a fresh unguided run satisfies
\begin{equation}
  P_{\mathrm{unguide}}(\mathcal{S}_T)\geq\E_{\tau\sim P_{\mathrm{guide}}}\left[2^{-K(\tau)}\right] \geq2^{-\E_{\tau\sim P_{\mathrm{guide}}}[K(\tau)]}.
  \label{eq:generator-oracle-bound}
\end{equation}
The expectation is over all guided runs, including generator and guide
randomness. The terminal judge's randomness is averaged into its pass rate.
\end{theorem}

\begin{proof}
Fix the generator's internal randomness as in \Cref{thm:compression}. The
unguided generator samples the guide's actions with probability
$\prod_t p_t(a_t\mid \tau_t)=2^{-\sum_t c_t}$ and then retraces the guided
trajectory. The judge accepts its final idea $x$ with probability
$\Pr[J_T(x)=1]$, so the unguided generator succeeds with probability at least
$2^{-K(\tau)}$. Averaging over the generator's and the guide's randomness
and applying Jensen's inequality gives
Equation~\ref{eq:generator-oracle-bound}.
Appendix~\ref{app:priced-guidance-proof} gives the formal proof.
\end{proof}

Appendix~\ref{app:elbo} discusses the evidence lower bound (ELBO)
interpretation of \Cref{thm:generator-oracle}. Appendix~\ref{app:perplexity} shows that
standard perplexity on the original text is a special case in which the
generator predicts the next token.

\subsection{Extensions: multiple candidates, recovery, and ensembling}
\label{sec:extensions}
We made some extensions to our framework.
A generator that is unsure about the target should be able to brainstorm several
candidate ideas at once.
The guide we use is imperfect and sometimes fails to select the best action.
We therefore strengthen it with \emph{recovery}, which lets the guide return
to an earlier question and try again. We also explore the option to \emph{ensemble} several
generators, because different generators may have better compression rates on different targets.

\paragraph{Multiple candidate ideas.}
Instead of a single final idea, the generator may submit $k$ candidate ideas
$x_1,\ldots,x_k$ with probabilities $p_1,\ldots,p_k$ that sum to one. The
judge evaluates every candidate, and the judge cost of the submission is $c_J=-\log_2\sum_{i=1}^{k}p_i\,\Pr[J_T(x_i)=1]$. \Cref{thm:generator-oracle}
still holds. 

\paragraph{Recovery.}
To instantiate recovery, whenever the guide selects an action, the generator
also outputs a distribution $\pi$ over how many times the guide may roll out
the generator again from that action. Each retry thus has a cost. Let $j_t$ index the rollout that the guide keeps at step $t$ of the
returned trajectory. The recovery cost and bound are
$K_\pi=K-\sum_t\log_2\pi(j_t)$, and $P_{\mathrm{unguide}}(\mathcal{S}_T)\geq\E[2^{-K_\pi}]$
where $K$ is the compression cost of the returned trajectory.
Appendix~\ref{sec:recovery} gives the formal construction.

\paragraph{Ensembling.}
We treat an ensemble of $M$ generators as a single {\emph{meta harness}}
that selects member $i$ with a fixed probability $w_i$ first. Let $K_i$ be member $i$'s compression cost, including any
recovery costs. Let $P_{\mathrm{meta}}$ denote the unguided trajectory law
of the meta harness. Define the ensemble compression cost by
$K_{\mathrm{meta}}=-\log_2\sum_{i=1}^{M}w_i2^{-K_i}$.
Then $P_{\mathrm{meta}}(\mathcal{S}_T)\geq\E\bigl[2^{-K_{\mathrm{meta}}}\bigr]$.
Choosing member $i$ costs only $-\log_2 w_i$ extra bits, so the ensemble
succeeds whenever any member does. Our experiments use $w_i=1/M$.
Appendix~\ref{sec:ensembling} gives the derivation.

\section{Experiment Instantiation}
\label{sec:experiment-instantiation}

We instantiate the protocol to predict the research contribution of a target
paper. For each target, Claude Opus~5 generates a private gold summary from
the paper's full \LaTeX{} source. We use this summary rather than the full
paper because the generator proposes only a research idea, not the paper's
text or the exact results, so the guide and the judge need only a concise statement of
that idea to compare against. The generator develops a candidate idea
specifying a contribution category, a core problem, and a main research
object. The guide uses the gold summary to guide this compression by
selecting answers to the generator's questions, with a compression cost for
each selection. A judge compares the candidate with the gold summary under
a specified recognition criterion. Our target papers are Top-1 Hugging Face
Daily Papers~\citep{huggingface_daily_papers}. After a title-based
memorization screening (Appendix~\ref{app:memorization-audit}), the
development set has 40 papers from March~2 through April~30, 2026, and the
test set has 87 papers from May~1 through September~6, 2026. \Cref{fig:skillnet-boxes}(a) shows the gold summary for SkillNet (arXiv:2603.04448),
used in the example walkthrough below. Due to cost constraints, we constrain the generator budget to $\$50$, the guide budget to $\$15$, and the judge budget to $\$15$ for all models we tested.
{Our result under the budget constraint provides an upper bound on the compression cost.}

\subsection{Idea-Equivalence Judging}
\label{sec:recognition-criteria}
 We use LLM judges to compare the scientific content of a candidate idea with the gold
summary while allowing different wording and names. We have two different judges with two levels of strictness.
Under the
\emph{Directional} Judge, a candidate must identify the same research
direction and contribution type, and describe a concrete, coherent idea of
comparable research plausibility. Under the \emph{Essence} judge, it must additionally recover the target's central research object and its defining mechanism or organizing principle.
We use GPT-5.5 for both judges, choosing a model distinct from all generators
to avoid having a model evaluate its own outputs.
As an example, the generator proposal for SkillNet in
Section~\ref{sec:example-walkthrough} passes the Essence judge without giving the exact name 
SkillNet
or mentioning skills created from GitHub repositories. It passes because it
recovers the central research object, a shared skill infrastructure, and
its organizing principle, a skill ontology.
A formal submission receives a fixed number $n$ of independent Judge
evaluations, of which $l$ pass. We use $l/n$ to estimate the pass rate $\Pr[J_T(x)=1]$ in the judge cost
$c_J$ (\Cref{sec:generator-oracle-protocol}), giving an empirical judge cost of
$-\log_2(l/n)$ bits. A submission succeeds if $l\geq1$.
If all evaluations fail, its empirical cost is infinite and the submission
is rejected, requiring recovery. We
fix $n$, defaulting to 1 for the Directional judge and 3 for the Essence judge.

\subsection{Generator and Guide Scaffolds}
\label{sec:bid-construction}

\paragraph{Generator scaffold.}
While the generator in the Priced Guidance (\Cref{sec:priced-oracle-hints}) framework can simply be a reasoning LLM placing arbitrary bids, we use a lightweight scaffold in which we provide the generators with some \emph{strategies} to consider, which improves the baseline performance of all the models. A
\emph{strategy} is a fixed type of question sequence, such as asking for a missing keyword in the current draft idea.
This scaffold is built based on heuristics. \Cref{thm:compression,thm:generator-oracle} hold for any scaffold that follows the priced guidance protocol, and a better scaffold yields a lower compression cost and hence a tighter bound.

\begin{figure}[tbp]
  \centering
  \resizebox{\linewidth}{!}{\begin{tikzpicture}[
  x=1cm,y=1cm,font=\sffamily\fontsize{7}{8.5}\selectfont,text=black!85,
  box/.style={draw=black!25,fill=black!2,rounded corners=2pt,
    line width=.45pt,inner sep=3pt,align=center},
  chosen/.style={box,draw=blue!70!black,fill=blue!7,line width=.7pt},
  question/.style={box,draw=blue!55!black,fill=blue!3},
  edge/.style={draw=black!35,line width=.55pt},
  selected edge/.style={draw=blue!65!black,line width=.9pt},
  arrow/.style={-{Latex[length=1.4mm,width=1mm]},draw=black!40,line width=.6pt},
  selected arrow/.style={-{Latex[length=1.5mm,width=1mm]},draw=blue!70!black,line width=.9pt},
  title/.style={anchor=north west,align=left,font=\sffamily\bfseries\fontsize{8}{9}\selectfont},
  note/.style={font=\sffamily\fontsize{6.5}{7.5}\selectfont,text=black!60,align=center},
  cost/.style={note,text=orange!80!black,font=\sffamily\bfseries\fontsize{6.5}{7.5}\selectfont},
  option/.style={box,text width=2.49cm,minimum height=.42cm,
    font=\sffamily\fontsize{6.5}{7.5}\selectfont}
]
\path[use as bounding box] (0,.08) rectangle (13.8,-6.70);
\node[title] at (0,0) {(a) Research taxonomy};
\node[title] at (4.30,0) {(b) Contribution\\types};
\node[title] at (7.40,0) {(c) Meta question and two strategies};
\draw[black!15] (4.05,-.05) -- (4.05,-6.35);
\draw[black!15] (7.15,-.05) -- (7.15,-6.35);

\node[chosen,text width=1.45cm] (root) at (1.96,-.87) {deep learning};
\node[chosen,text width=1.12cm] (agents) at (.90,-1.97) {agents};
\node[box,text width=1.37cm] (architecture) at (2.98,-1.97) {architecture};
\draw[selected edge] (root.south) -- (1.96,-1.39) -| (agents.north);
\draw[edge] (root.south) -- (1.96,-1.39) -| (architecture.north);
\node[note,text width=1.70cm] at (2.98,-2.48) {+ 16 other fields};

\node[chosen,text width=1.48cm] (language) at (.90,-3.32) {Language\\Agents};
\node[box,text width=1.40cm] (foundations) at (2.98,-3.32) {Foundations\\and Theory};
\draw[selected edge] (agents.south) -- (.90,-2.69) -- (language.north);
\draw[edge] (agents.south) -- (.90,-2.69) -| (foundations.north);
\node[note,text width=1.70cm] at (2.98,-4.02) {+ 13 other\\branches};

\node[chosen,text width=3.14cm] (memory) at (1.96,-5.02)
  {Memory, Retrieval, and\\Knowledge Integration};
\draw[selected edge] (language.south) -- (.90,-4.27) -| (memory.north);
\node[note] at (1.96,-5.77) {+ 11 other branches};
\node[note,text width=3.6cm] at (1.96,-6.30)
  {Guide chooses one node.};

\node[chosen,text width=2.37cm,minimum height=.70cm] at (5.66,-1.18)
  {a system or\\software artifact};
\node[box,text width=2.37cm,minimum height=.70cm] at (5.66,-2.32)
  {a new algorithm\\or method};
\node[box,text width=2.37cm,minimum height=.70cm] at (5.66,-3.46)
  {a benchmark};
\node[box,text width=2.37cm,minimum height=.70cm] at (5.66,-4.60)
  {a dataset};
\node[note,text width=2.50cm] at (5.66,-5.77) {+ 7 other categories};
\node[note,text width=2.55cm] at (5.66,-6.30)
  {Guide chooses one type.};

\node[note,anchor=north west,text width=6.1cm,align=left] at (7.43,-.49)
  {\textbf{Candidate idea}\quad A shared skill hub.};
\node[cost,anchor=west] at (7.48,-1.02) {Guide's private target\quad SkillNet};
\node[question,text width=5.70cm,minimum height=.60cm] (meta) at (10.59,-1.63)
  {\textbf{Meta question about strategy}\\Which strategy should we use next?};
\node[chosen,text width=2.49cm,minimum height=.60cm] (mc) at (8.94,-2.65)
  {\textbf{Multiple-choice}\\$p=.60$};
\node[box,text width=2.49cm,minimum height=.60cm] (keyword) at (12.21,-2.65)
  {\textbf{Keyword}\\$p=.40$};
\draw[selected arrow] (meta.south) -- (10.59,-2.11) -| (mc.north);
\draw[arrow] (meta.south) -- (10.59,-2.11) -| (keyword.north);
\node[cost,fill=white,inner sep=1pt] at (8.97,-2.19) {Guide choice\quad 0.74 bits};

\node[question,text width=2.49cm,minimum height=.80cm] (mcq) at (8.94,-3.68)
  {How are contributed\\skills checked?};
\node[box,text width=2.49cm,minimum height=.80cm] (kwq) at (12.21,-3.68)
  {What kind of concept\\is missing?};
\draw[selected arrow] (mc.south) -- (mcq.north);
\draw[arrow] (keyword.south) -- (kwq.north);

\node[option,chosen] (mca) at (8.94,-4.60) {\textbf{LLM judge}\enspace $p\!=\!.25$};
\node[option] (mcb) at (8.94,-5.21) {Human review\enspace $p\!=\!.50$};
\node[option] (mcc) at (8.94,-5.82) {Execution tests\enspace $p\!=\!.25$};
\draw[selected arrow] (mcq.south) -- (mca.north);
\node[cost,fill=white,inner sep=1pt] at (9.65,-4.26) {2.00 bits};
\draw[edge] (mcq.south) -- (8.94,-4.18) -- (7.44,-4.18) -- (7.44,-5.82);
\draw[arrow] (7.44,-5.21) -- (mcb.west);
\draw[arrow] (7.44,-5.82) -- (mcc.west);

\node[option] (kwa) at (12.21,-4.60) {Organization\enspace $p\!=\!.50$};
\node[option] (kwb) at (12.21,-5.21) {Evaluation\enspace $p\!=\!.30$};
\node[option] (kwc) at (12.21,-5.82) {Learning\enspace $p\!=\!.20$};
\draw[arrow] (kwq.south) -- (kwa.north);
\draw[edge] (kwq.south) -- (12.21,-4.18) -- (10.71,-4.18) -- (10.71,-5.82);
\draw[arrow] (10.71,-5.21) -- (kwb.west);
\draw[arrow] (10.71,-5.82) -- (kwc.west);

\node[cost] at (10.60,-6.30)
  {Base choice costs\quad $0.74+2.00=2.74$ bits};
\node[note] at (10.60,-6.62)
  {Illustrative options and probabilities. Blue marks guide choices.};
\end{tikzpicture}}
  \caption{Illustration of the generator scaffold.
  (a--b) Excerpts of the research taxonomy of deep learning, with 3,035
  nodes, and of the list of 11 contribution categories. GPT-5.6 Sol generated
  both once. They are fixed across all target papers and given to every
  generator model in its prompt. The taxonomy bids walk down the taxonomy one
  edge at a time, offering each child and an option to stop with equal
  probability, so the guide chooses one taxonomy node (not necessarily a
  leaf). The category bid offers the 11 categories with equal probability, and
  the guide chooses one contribution type.
  (c) A simplified meta question on the SkillNet example. At the start of a turn, the generator
  prepares both strategies' questions, answer options, and probabilities.
  The guide selects the multiple-choice strategy and then the
  answer. These two selections contribute
  $-\log_2(.60)-\log_2(.25)=2.74$ bits.
  Answer sets and probabilities are illustrative.}
  \label{fig:experiment-instantiation}
\end{figure}

The generator keeps a list of facts confirmed by the guide, a
current candidate idea, and a whiteboard of working
hypotheses and open questions, which the guide can also read
(Appendix~\ref{app:generator-scaffold}).
Unless otherwise stated, we use the highest supported reasoning effort for
each generator model. For Opus~5, we use high effort because the generator yielded a lower median compression cost at high than at max effort on our dev set
(Section~\ref{sec:ablations}). Each run proceeds as follows.
\begin{enumerate}[leftmargin=*]
  \item \textbf{Opening bids.} The generator walks down a fixed
  deep-learning taxonomy with 3,035 nodes one level at a time; each bid
  offers the children of the current node and a stop option. It then asks
  for one of 11 fixed contribution categories. All options have equal
  probability, and the guide selects an answer to each bid. GPT-5.6 Sol
  generated the taxonomy and the categories once, and every generator
  receives them for every target (Appendix~\ref{app:taxonomy}).
  \item \textbf{Directional stage.} Repeat the following until a
  submission passes the Directional judge:
  \begin{enumerate}
    \item \textbf{Meta question.} The generator prepares one bid for each
    available strategy
    and assigns a probability to each strategy. The guide selects a
    strategy (\Cref{fig:experiment-instantiation}(c)).
    \item \textbf{Strategy questions.} The generator asks the selected
    strategy's prepared question and any follow-up questions, and the guide
    selects an answer to each:
    \begin{itemize}
      \item \emph{Multiple choice} asks about one unresolved aspect of the
      candidate idea and offers concrete answers and rejection options.
      \item \emph{Keyword} asks which semantic category contains a missing
      concept, then offers candidate phrases in that category. It also offers
      a \emph{letter hint} action: if none of the phrases is correct, the generator can guess the first two letters of the
      missing phrase, and after the guide selects them, it proposes phrases
      that begin with those letters.
      \item \emph{Submit} sends eight distinct candidate ideas with
      probabilities, which the generator prepares before the meta question,
      to the Directional judge.
    \end{itemize}
    \item \textbf{Update.} The generator records the selected answers as
    facts and revises its candidate idea and whiteboard.
  \end{enumerate}
  \item \textbf{Essence stage.} Once a submission passes the Directional
  judge, the run is \emph{promoted} to the Essence stage. The generator keeps its
  candidate idea, facts, and earlier answers, learns only that its idea
  passed, and now also has to recover the defining mechanism.
  (Appendix~\ref{app:promotion-cost}).
  \begin{enumerate}
    \item \textbf{Reused submission.} The eight-candidate submission that
    passed the Directional judge is first evaluated by the Essence
    judge. If one of the ideas passes, the run ends. Otherwise, the guide uses recovery (\Cref{sec:extensions}) to roll out the
    generator again from an earlier action.
    \item \textbf{Essence turns.} Repeat the Directional turn until a
    submission passes the Essence judge, with two changes:
    \begin{itemize}
      \item A new Essence \emph{submit} bid sends one idea with
      probability one to the Essence judge. 
      \item A new strategy, \emph{correction}, lists claims in the
      candidate idea and asks which claim  may contradict the target, then offers to replace,
      refine, or delete it.
    \end{itemize}
  \end{enumerate}
\end{enumerate}

Here the essence promotion requires an extension of our theoretical bound applies to a target-independent scaffold with replay rather than directly to the original generator scaffold. This incurs a small promotion cost of about 2 bits for this switch (\Cref{app:promotion-cost}).

\paragraph{Guide scaffold.}
\label{sec:oracle-guidance}
The guide is a lightweight scaffold around Fable~5. For each bid, it sees
the question, answer options, and probabilities, together with the target
summary, the judge criterion, and its private memory. It selects an answer
supported by the target, and only this choice reaches the generator. It may
also privately preview Judge verdicts on the current submission snapshot to decide
when to submit, and a submission then receives fresh Judge evaluations. This
extra computation adds no compression cost, because only the selected answers
are priced. To explore another path, or after a rejected submission, the
guide can \emph{check out} an earlier question, which restores the generator
to its state right after that bid. Each step on the kept trajectory then pays
the recovery cost defined in \Cref{sec:extensions}.
Appendix~\ref{app:guide-scaffold} describes the guide scaffold in detail.

\subsection{Example Walkthrough}
\label{sec:example-walkthrough}
\begin{figure}[tbp]
  \centering
  \includegraphics[width=\linewidth]{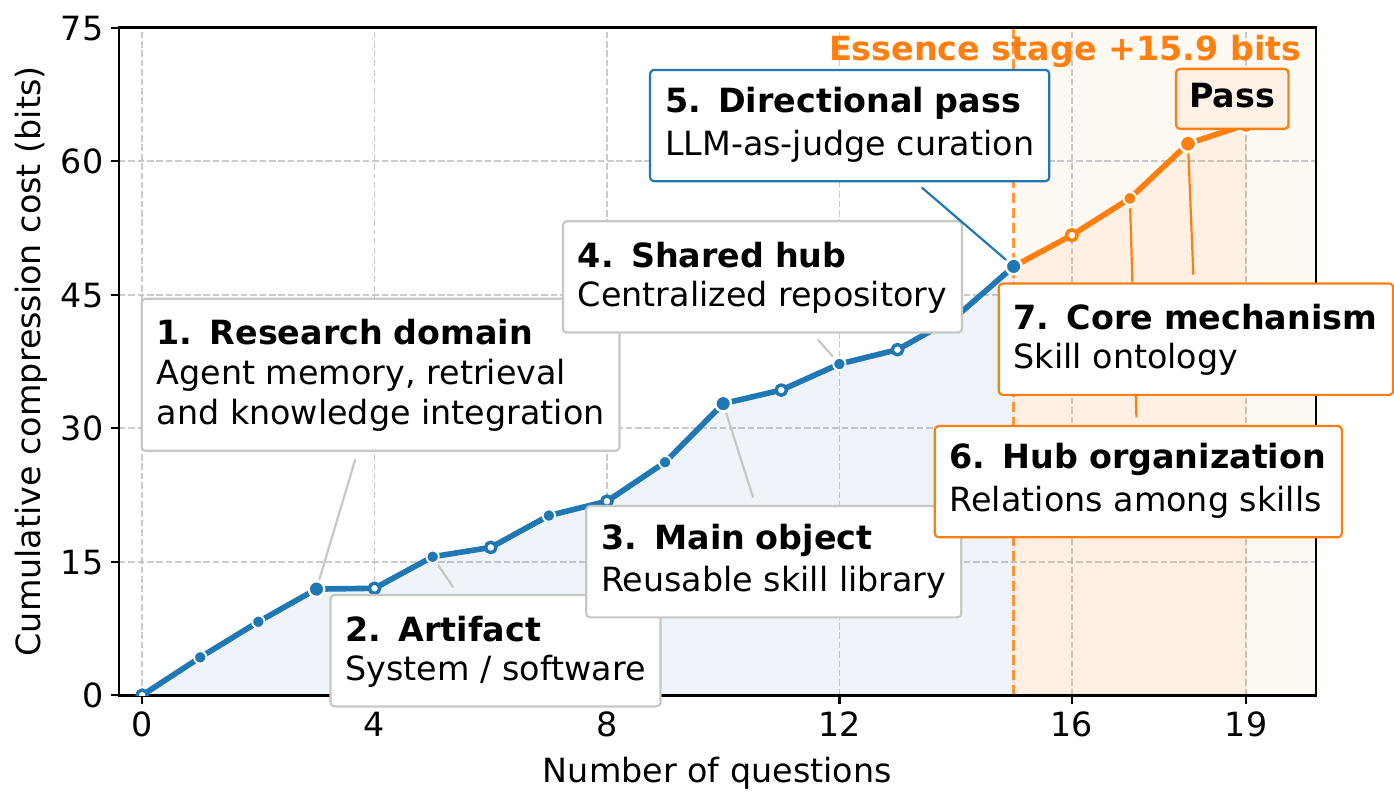}
  \caption{Information acquired along one successful Fable~5 trajectory for
  SkillNet. Filled nodes identify paper-specific content; hollow nodes mark
  meta questions about strategies. Blue denotes the Directional
  stage. Orange denotes the additional 15.9 bits after promotion to Essence,
  culminating in identification of the skill ontology and a pass at 64.1 bits.}
  \label{fig:skillnet-information-recovery}
\end{figure}

 For SkillNet (arXiv:2603.04448),
\Cref{fig:skillnet-information-recovery} follows the final active path (the sequence of choices retained after recovery)
of a Fable~5 generation trajectory. \Cref{fig:skillnet-boxes} places the gold summary beside the final Essence proposal, with additions and revisions from the Directional proposal highlighted in purple.
Appendix~\ref{app:skillnet-verbatim} contains both original proposals and the
gold summary without omissions.

The first
11.9 bits locate the research field within memory, retrieval, and knowledge
integration for language agents. By 32.8 bits, the candidate identifies a
reusable skill-library infrastructure. At 48.2 bits, it describes a shared
hub with LLM-as-judge curation and passes the Directional judge.
Essence-stage guidance then identifies the missing organizational structure
as ``skill ontology,'' an organization of skills and their relations,
at 62.0 bits. The final candidate passes all three Essence
evaluations at 64.1 bits, including 15.9 additional bits after promotion.

\begin{figure}[t]
  \centering
  \definecolor{essencepurple}{HTML}{763A8C}
  \newcommand{\essencechange}[1]{\textcolor{essencepurple}{#1}}
  \renewcommand{\ideafield}[1]{%
    \par\addvspace{0.25em}%
    \noindent{\fontsize{6.4pt}{7.6pt}\selectfont\sffamily\bfseries\color{blue!55!black}#1.}\enspace\ignorespaces}
  \setlist[itemize]{leftmargin=*}
  \tcbset{skillnet compact/.style={unbreakable,
    fontupper=\fontsize{7.04pt}{8.2pt}\selectfont,
    fonttitle=\sffamily\bfseries\fontsize{7.2pt}{8.6pt}\selectfont}}
  \begin{minipage}[t]{\dimexpr0.5\linewidth-1mm-0.8cm\relax}\vspace{0pt}
\begin{ideabox}[skillnet compact, equal height group=skillnet]{(a) Gold setting and main object}
\ideafield{Category}
Infrastructure / Resource (open skill repository, ontology, evaluation protocol, and toolkit).

\ideafield{Problem \& motivation}
\begin{itemize}
  \setlength{\itemsep}{0.1em}
  \setlength{\parsep}{0pt}
  \setlength{\topsep}{0.2em}
  \item Agents lack a unified mechanism to consolidate episodic experience into durable, reusable skills, so they repeatedly ``reinvent the wheel.''
  \item Existing skill repositories/marketplaces rely on manual curation and only indirect (downstream-task) quality signals, leaving skills fragmented, redundant, and unvalidated.
\end{itemize}

\ideafield{Main object \& crux}
SkillNet: an open full-lifecycle skill infrastructure that:
\begin{itemize}
  \setlength{\itemsep}{0.1em}
  \setlength{\parsep}{0pt}
  \setlength{\topsep}{0.2em}
  \item Auto-creates skills from heterogeneous sources (execution trajectories, GitHub repos, documents, prompts) via LLM pipelines.
  \item Evaluates each skill intrinsically along Safety, Completeness, Executability, Maintainability, Cost-awareness.
  \item Connects skills into a typed relation graph (similar\_to, belong\_to, compose\_with, depend\_on).
\end{itemize}

\ideafield{Representation}
Skills are treated as a unified knowledge representation bridging natural-language procedural guidance and machine-executable logic, organized by a three-layer Skill Ontology (taxonomy, relation graph, skill packages).

\ideafield{Release}
Released as a curated repository (200k+ candidate, 150k+ curated skills), website/API, and the skillnet-ai Python library + CLI.
\end{ideabox}
  \end{minipage}\hfill
  \begin{minipage}[t]{\dimexpr0.5\linewidth-1mm+0.8cm\relax}\vspace{0pt}
\begin{ideabox}[skillnet compact, equal height group=skillnet]{(b) Generator proposal: Directional to Essence}
\ideafield{Category}
A system or software artifact.

\ideafield{Core problem}
Language agents cannot accumulate and reuse experience across tasks and across agents in a portable way --- existing experience/skill memories are bespoke and siloed within individual agents or frameworks, so there is no general-purpose infrastructure for the community of agents to collectively store, \essencechange{organize}, and reuse learned skills at trustworthy quality.

\ideafield{Main object}
A skill library infrastructure --- a general-purpose memory infrastructure / operating-system-like layer (a standardized API, protocol, or middleware for agent memory) whose primary function is an experience/skill library that stores past task trajectories or learned procedures for reuse on new tasks, and whose defining architectural form is a centralized shared hub/repository where many agents or users contribute and reuse skills collectively (hub/marketplace model), governed by LLM-as-judge curation --- a quality-control, validation, and curation mechanism in which an LLM acts as judge over contributed skills, evaluating their correctness, usefulness, generality, and redundancy to decide which contributed skills enter or remain in the shared hub \essencechange{--- and organized by a skill ontology: an organizational structure imposed on the hub's contents, i.e., a hierarchical taxonomy of skills with relations among them (specialization/generalization, dependencies, and composition of smaller skills into larger skills) that structures how skills are stored, related, abstracted, and navigated.}

\ideaomit{\ldots}

\ideafield{Intended outcome}
By decoupling the skill library from any specific agent and exposing it as a centralized, standardized, LLM-curated\essencechange{, ontology-organized} shared hub --- analogous to a package registry or model hub, but for agent skills with automated quality gatekeeping \essencechange{and a structured taxonomy} --- the system enables collective, continual, trustworthy\essencechange{, and well-organized} skill accumulation and cross-agent/cross-user skill reuse without any model weight updates.
\end{ideabox}
  \end{minipage}
  \caption{SkillNet (arXiv:2603.04448). (a) Gold summary of the target idea.
  (b) Final generator proposal along the trajectory in
  \Cref{fig:skillnet-information-recovery}, with revisions from
  the Directional proposal in purple; [\ldots] marks an omitted field.
  Appendix~\ref{app:skillnet-verbatim} gives the proposals and the gold
  summary in full.}
  \label{fig:skillnet-boxes}
\end{figure}

\subsection{Data Curation and Memorization Audit}
\label{sec:memorization-audit}

We want to make sure the generator hasn't seen the papers it needs to compress. So we select target papers from Top-1 Hugging Face Daily Papers since March~2, 2026 and also perform an additional title-based recall screening to remove potentially memorized papers.
For the development set, we collect 40 Top-1 papers from
March~2 through April~30, 2026. For the test set, we collect Top-1
papers from May~1 through September~6, 2026, yielding 87 unique candidates.
We perform screening (Appendix~\ref{app:memorization-audit}) to remove any paper that one of our generators recalls
from its title alone.

\section{Main Results}
\label{sec:results}

\begin{figure}[t]
  \centering
  \includegraphics[width=\linewidth]{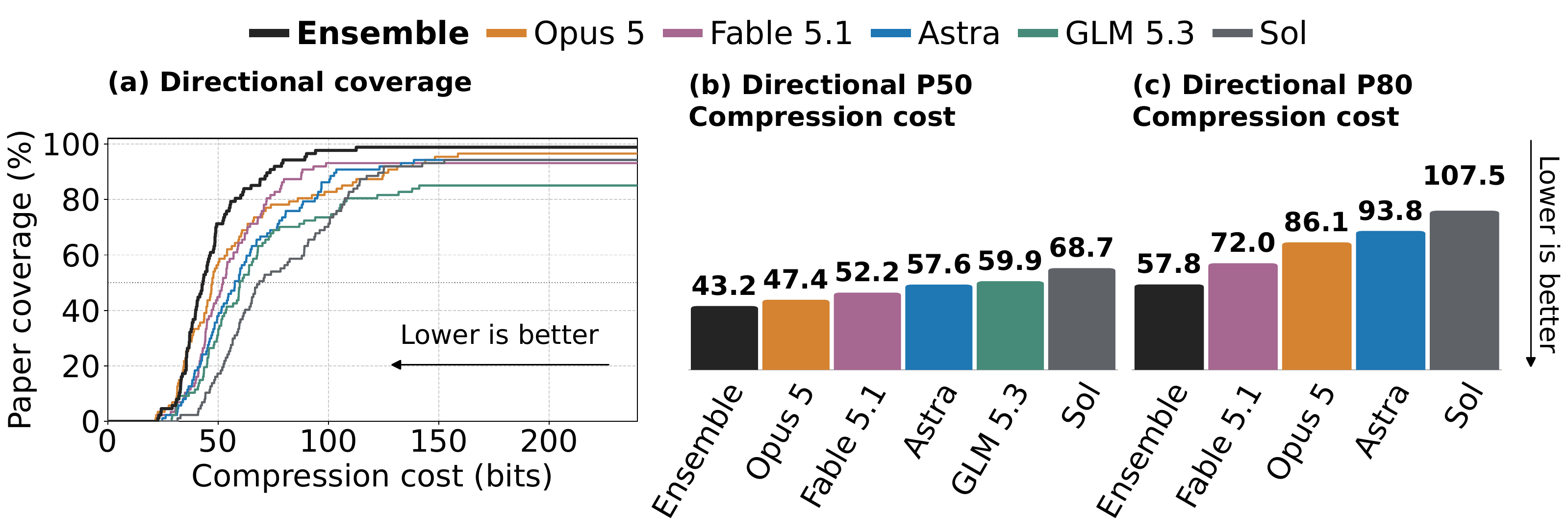}
  \caption{Test-set results under the Directional judge on the same 87 papers as Figure~\ref{fig:test87-essence}.
  (a) Coverage versus compression cost; (b) median (P50) and (c) 80th-percentile (P80) compression cost.
  The $q$th-percentile cost is the minimum compression cost needed to produce an idea that passes the judge on $q\%$ of the test papers.
  The black ensemble uniformly mixes Fable~5.1, Opus~5, and Astra; the guide is fixed to Fable~5.
  GLM~5.3 is omitted from the P80 panel for consistency with the Essence figure; its Directional P80 is 108.3 bits.}
  \label{fig:test87-directional}
\end{figure}

We show coverage as a function of compression cost, together with the median and 80th-percentile compression costs, on the 87 test papers under both judges in~\Cref{fig:test87-essence,fig:test87-directional}.

{\paragraph{Compressing an idea's essence takes far fewer bits than its text.} We care about the probability of generating an idea with the target's
essence, not of reproducing the words of its summary, which is much stricter. Compressing the gold summary verbatim costs GLM~5.3 a median of 1,147.1 bits (Appendix~\ref{app:glm-verbatim-baseline}). We choose GLM~5.3 as it is the only one of our five generators with open weights.
Its median compression cost under Priced Guidance is only 59.9 bits under the Directional judge and 102.4 bits under the Essence judge (\Cref{fig:test87-directional,fig:test87-essence}), more than an order of magnitude lower.}

\paragraph{Model comparisons.} Opus~5 high has the lowest median cost under the Directional judge. Under the Essence judge, Fable~5.1, Opus~5 high, and Astra nearly tie at 69.9, 70.4, and 71.6 bits. To cover 80\% of papers under the Essence judge, however, Fable~5.1 and Astra need 104.3 and 111.0 bits, compared with 141.3 bits for Opus~5 high (Figure~\ref{fig:test87-essence}(c)). The open-weight GLM~5.3 is competitive at the Directional median compared to Astra but covers less than 80\% of papers under the Essence judge within the guide budget.

\paragraph{Where the information is spent.} 
We decompose the compression cost by strategy for successful Opus~5 high
runs under the Essence judge in~\Cref{tab:opus-test-breakdown}.
Keyword information accounts for most of Opus~5 high's mean Directional
cost and its additional cost to reach Essence. Of the 134 Directional keyword
calls in \Cref{tab:opus-test-breakdown}, three are category retries that
yield no phrase. Of the remaining 131 completed selections, 31 use letter hints,
accounting for 23.7\% of selections and 51.8\% of keyword costs. On the
final Essence paths, 38 of 187 selections use letter hints (20.3\%),
accounting for 45.2\% of keyword costs. {This suggests that most of the
information so far is used to acquire useful candidate concepts for
the final idea, and these concepts are typically outside the models' initial guesses.}

\begin{table}[t]
\centering
\small
\caption{\textbf{Compression cost and scaffold usage for Opus~5.}
Opus~5 high on the same 74 test papers for which the generated idea passes the Essence judge. We show the compression costs and also the number of times different components are called in the final successful trajectory.
$\Delta$ is the Essence cost minus the Directional cost. Keyword
costs include semantic-category and strategy selections; the letter-hint
row includes the full hint path. Promotion is the cost of switching from the
Directional to the Essence stage (Appendix~\ref{app:promotion-cost}). }
\label{tab:opus-test-breakdown}
\setlength{\tabcolsep}{3pt}
\begin{tabular}{@{}lrrr@{\hspace{9pt}}rr@{\hspace{9pt}}rr@{}}
\toprule
& \multicolumn{3}{c}{Compression cost (bits)} & \multicolumn{2}{c}{Directional usage} & \multicolumn{2}{c}{Essence usage} \\
\cmidrule(lr){2-4}\cmidrule(lr){5-6}\cmidrule(lr){7-8}
Component & Directional & Essence & $\Delta$ & Calls & Mean & Calls & Mean \\
\midrule
Field taxonomy & 11.61 & 11.61 & 0.00 & 270 & 3.65 & -- & -- \\
Contribution category & 3.46 & 3.46 & 0.00 & 74 & 1.00 & -- & -- \\
Multiple choice & 3.21 & 4.57 & 1.36 & 57 & 0.77 & 21 & 0.28 \\
Keyword & 33.29 & 43.70 & 10.41 & 134 & 1.81 & 56 & 0.76 \\
\quad Without letter hints & 16.06 & 23.95 & 7.90 & 103 & 1.39 & 49 & 0.66 \\
\quad With letter hints & 17.23 & 19.75 & 2.51 & 31 & 0.42 & 7 & 0.09 \\
Correction & 0.00 & 0.67 & 0.67 & -- & -- & 5 & 0.07 \\
Submission & 3.09 & 3.33 & 0.24 & 42 & 0.57 & 32 & 0.43 \\
Retry & 2.14 & 2.54 & 0.40 & -- & -- & -- & -- \\
Promotion & 0.00 & 2.32 & 2.32 & -- & -- & -- & -- \\
\midrule
\textbf{Total} & \textbf{56.80} & \textbf{72.19} & \textbf{15.40} & \textbf{577} & \textbf{7.80} & \textbf{114} & \textbf{1.54} \\
\bottomrule
\end{tabular}
\end{table}

\paragraph{Implications for automatic idea generation.}
The lowest individual median compression costs, from Opus~5 high under the
Directional judge and Fable~5.1 under the Essence judge, correspond to implied
probability scores $2^{-K}$ of approximately $5.3\times10^{-15}$ and
$8.8\times10^{-22}$, respectively. As with text, these small probabilities concern one particular target and
do not imply that the models cannot generate high-quality ideas
autonomously (\Cref{sec:limitations}).

\paragraph{Headroom from ensembling.} 
{Combining models lowers compression cost further, demonstrating headroom
beyond the best individual model.} We uniformly mix Fable~5.1, Opus~5,
and Astra by averaging their $2^{-K}$ scores, with unsuccessful runs
contributing zero (Appendix~\ref{sec:ensembling}). The ensemble achieves median compression costs
of 43.2 bits under the Directional judge and 55.8 bits under the Essence
judge, improving on the best individual models, Opus~5 high under the
Directional judge and Fable~5.1 under the Essence judge, by 4.3 and 14.1 bits,
respectively.

\paragraph{Compression cost did not change significantly over four months.}
One might expect compression cost to increase for papers published further beyond a model's knowledge cutoff. Ensemble cost shows no significant correlation with paper date over the test period (Figure~\ref{fig:ensemble-cost-over-time}).

\paragraph{Technical reports tend to have high compression cost.}
The Kimi~K3 paper~\citep{kimiteam2026kimik3openfrontier} is the only test target for which none of the five
evaluated generators passes the Directional judge in our test runs. We attribute this to the fact that Kimi K3 is a combination of many novel techniques and would require a much higher compression cost to fully cover. We also found that other technical reports such as those by~\citet{li2026achievinggoldmedallevelolympiadreasoning,wang2026hrmtextefficientpretrainingscaling} have two of the three highest Essence compression costs among the test papers that the ensemble recovers. This is not simply because these papers or their summaries are longer: across the 87 test papers, compression cost shows no significant correlation with the length of either the gold summary or the full paper (Spearman correlation $\le$ 0.10 for the ensemble compression cost under both judges).

\section{Ablations}
\label{sec:ablations}

{In this section, we present the ablations performed while designing the test scaffolding using the 40-paper development set from
Section~\ref{sec:experiment-instantiation}.}
As in Figure~\ref{fig:test87-essence}, every median is the 50\%-coverage
cost: the minimal compression cost that covers 20 of the 40 papers, with
unsuccessful runs never covered.
Some of the experiments in our development process discard earlier
messages and reasoning between generator calls (Appendix~\ref{app:ablation-scaffold-diffs}).

\begin{figure}[tbp]
  \centering
  \includegraphics[width=0.96\linewidth]{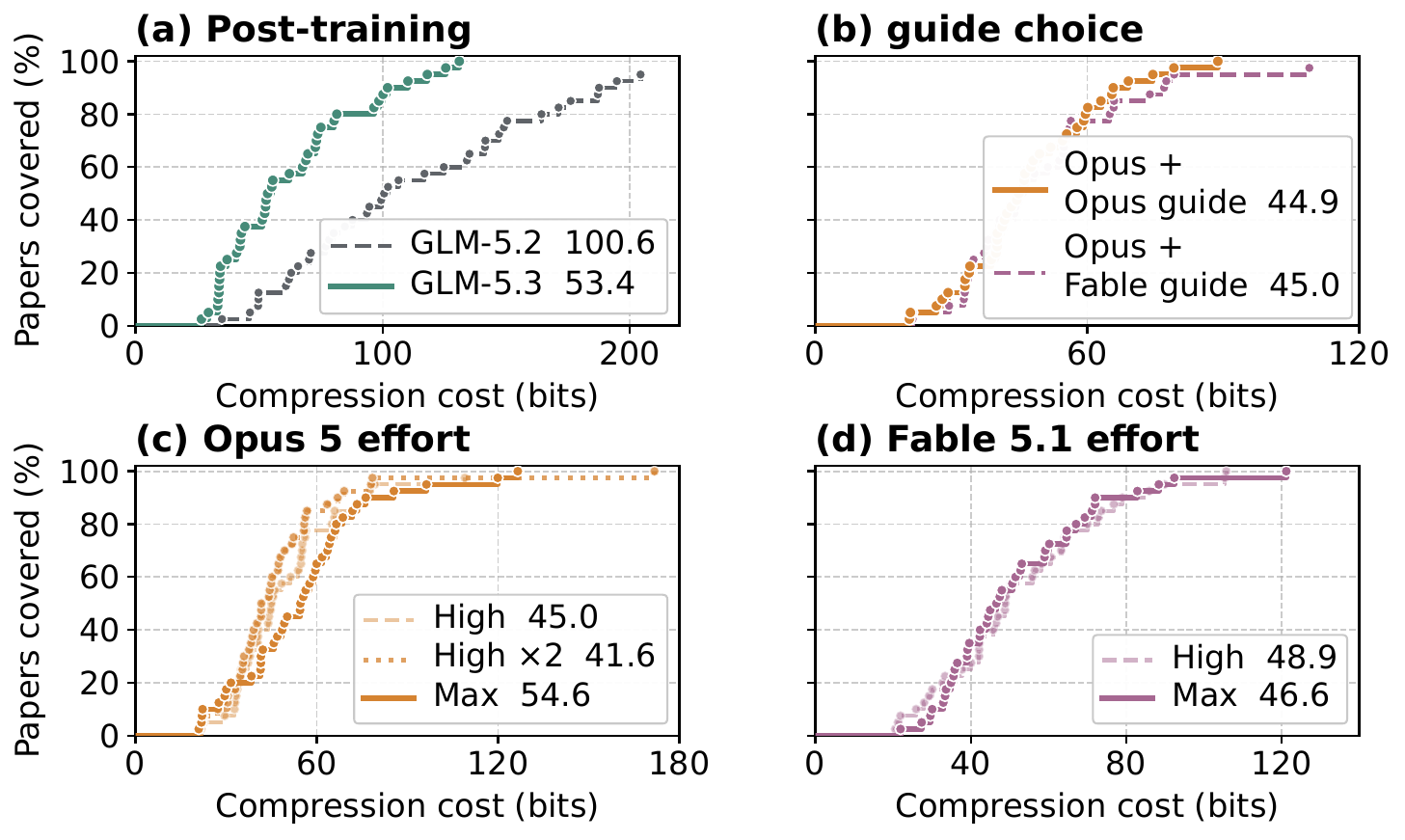}
  \caption{ Directional compression cost on the development set. 
  \textbf{(a)} Comparing GLM 5.2 and 5.3 shows the effect of general post-training of the generator on this benchmark.
  \textbf{(b)} Guide choice for Opus~5 ablations; both runs use the Opus~5 generator.
  \textbf{(c,d)} Opus~5 and Fable~5.1 reasoning effort ablations.
  Legends report median compression cost in bits. The two-run ensemble in (c) averages the two members'
  observed scores as in Equation~\ref{eq:uniform-ensemble-cost}.
  Panel (d) is indicative because high and max effort use Fable~5.1 and
  Fable~5 guides, respectively.}
  \label{fig:ablation-posttraining}
\end{figure}

\paragraph{Generator post-training can substantially reduce compression cost.}
Updating the generator from GLM-5.2 to GLM-5.3 substantially reduces median
compression cost, from 100.6 to 53.4 bits
(Figure~\ref{fig:ablation-posttraining}(a)).
As these two models share the same pretraining base model, this demonstrates the effectiveness of post-training in reducing compression cost.

\paragraph{Guide choice has little effect on median compression cost for Opus~5.}
Replacing the Fable~5 guide with Opus~5 changes
the median only from 45.0 to 44.9 bits
(Figure~\ref{fig:ablation-posttraining}(b)).

{\paragraph{Ensembling potentially helps more than higher reasoning effort.} 
Increasing the reasoning effort of Opus 5 from high to max raises the median compression cost from 45.0 to 54.6 bits, although coverage increases from
39/40 to 40/40 (Figure~\ref{fig:ablation-posttraining}(c)).
Increasing the reasoning effort of Fable 5.1 from high to max also decreases the median compression cost only from 48.9 bits to
46.6 bits
(Figure~\ref{fig:ablation-posttraining}(d)).\footnote{For the Fable~5.1 experiment, we initially
planned to replace the Fable~5 guide with Fable~5.1, but encountered a high
rejection rate and slightly modified the guide prompt to address it.
The high-effort run uses the Fable~5.1 guide with this modified prompt,
whereas the max-effort run uses the Fable~5 guide with the original prompt.
{We believe these scaffold differences should not significantly affect median
compression cost due to the robustness with respect to the guide.}}

Ensembling two Opus runs lowers compression cost under both judges.
We use the uniformly weighted ensemble defined in
Appendix~\ref{sec:ensembling}.
Relative to the single-seed run, the Directional ensemble reduces median compression cost from 45.0
to 41.6 bits and increases the final coverage from 39/40 to 40/40
(Figure~\ref{fig:ablation-randomness}(b)).
The Essence ensemble also reduces the median from 72.4 to 70.2 bits (Figure~\ref{fig:ablation-randomness}(d)). {This indicates that parallel inference-time scaling could potentially be more beneficial for ideation compared to serial inference-time scaling within the same context window.}

\begin{figure}[tbp]
  \centering
  \includegraphics[width=0.96\linewidth]{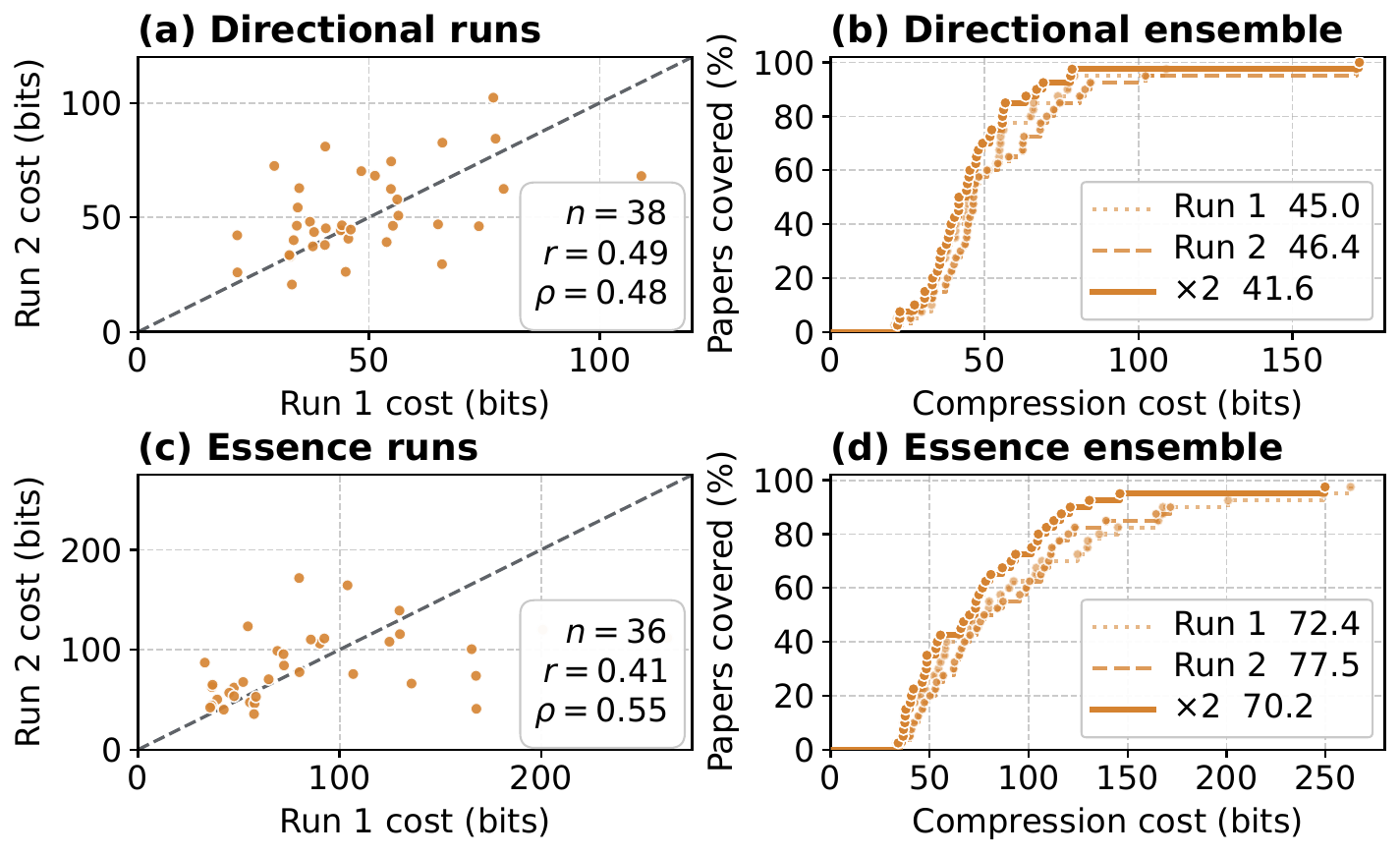}
  \caption{Run-to-run variation and ensembling for Opus~5 on the development set.
  \textbf{(a,c)} Paired Directional ($n=38$) and Essence ($n=36$) costs.
  Dashed diagonals denote equal cost; $r$ and $\rho$ are Pearson and Spearman
  correlations. \textbf{(b,d)} Corresponding coverage for both independent runs and their
  uniformly weighted ensemble, with effective cost given by
  Equation~\ref{eq:uniform-ensemble-cost}. {While each paper may have stochastic compression cost, the median stays surprisingly stable. The randomness in the independent runs also explains why ensembling improves the median compression cost.} }
  \label{fig:ablation-randomness}
\end{figure}

{\paragraph{Median compression cost is robust to run-to-run randomness.} 
Median compression cost is stable across independent Opus~5 runs, despite
appreciable variation in individual-paper costs.
We run the same single-idea scaffold twice with independent sampling randomness using Opus~5.
The medians are 45.0 and 46.4 bits, only 1.4 bits apart
(Figure~\ref{fig:ablation-randomness}(b)).
Meanwhile, the costs on the 38 papers where both runs succeed can differ and are only moderately correlated
(Figure~\ref{fig:ablation-randomness}(a)).
After Essence promotion, the medians are 72.4 and 77.5 bits
(Figure~\ref{fig:ablation-randomness}(d)), and the costs on the 36 shared
successes also remain moderately correlated
(Figure~\ref{fig:ablation-randomness}(c)). {This randomness in the independent runs also explains why ensembling improves the median compression cost.}

\begin{figure}[tbp]
  \centering
  \includegraphics[width=0.96\linewidth]{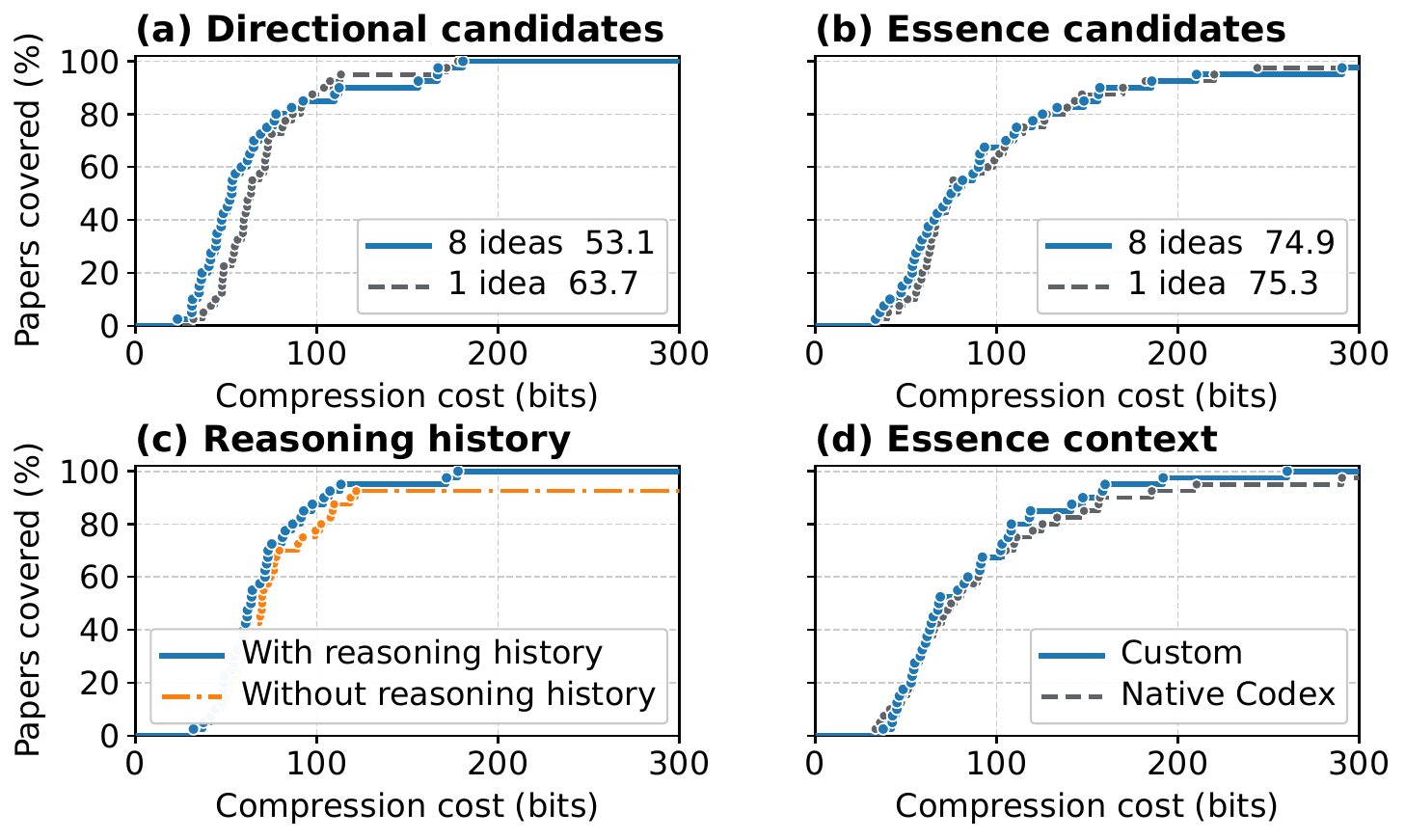}
  \caption{Astra candidate-count and context-management comparisons on the development set.
  \textbf{(a,b)} Native Codex with one (gray dashed) or eight (blue solid)
  Directional candidates, evaluated by the Directional and Essence judges,
  respectively; legend values report median compression cost in bits.
  \textbf{(c)} {Directional compression cost with one candidate in both configurations:}
  Codex with reasoning history (blue) and our earlier scaffold without reasoning
  history (orange).
  \textbf{(d)} {Essence compression cost with eight Directional candidates using custom context management (blue) or native Codex (gray).}
  Essence costs include inherited Directional cost.
  Details are deferred to
  Appendix~\ref{app:ablation-scaffold-diffs}.}
  \label{fig:ablation-astra-candidates}
  \label{fig:ablation-context-management}
\end{figure}

\paragraph{More candidate ideas reduce Astra's Directional cost but not its Essence cost.} 
Qualitative inspection of GPT-family trajectories suggested that single-idea
generation tended to repeat known facts and explore few alternative
hypotheses. We therefore test a Directional submission of eight distinct,
complete candidate ideas with probabilities, priced by the
multiple-candidate judge cost in
\Cref{sec:extensions}. We compare one versus
eight candidates with Astra at extra high reasoning effort, using Codex as an
agent scaffold that retains the reasoning history across generator calls
(Appendix~\ref{app:ablation-scaffold-diffs}). After promotion, the first
Essence evaluation reuses the passing Directional submission, and every later
submission contains one idea in both configurations.
Eight candidates lower the median Directional compression cost from 63.7 to
53.1 bits (Figure~\ref{fig:ablation-astra-candidates}(a)), lowering the cost on 32 of 40 papers.
Under the Essence judge, the median costs remain close,
at 75.3 bits with one candidate and 74.9 bits with eight, and both
configurations cover 39 of 40 papers
(Figure~\ref{fig:ablation-astra-candidates}(b)).

\paragraph{Custom compaction lowers Astra's Essence cost, and retained reasoning history is associated with lower Directional cost.}\label{sec:context-management-ablation}
Our custom compaction in the final test scaffold lowers Astra's median compression cost under the Essence judge compared with native Codex context management.
We compare two eight-candidate configurations using Astra at extra high
effort through Codex, with the same set of prompts for generators.
The native configuration uses native Codex compaction;
our custom configuration explicitly compresses earlier history into a
natural-language self-summary and a
brief instruction to follow the current task.
Directional median compression cost remains nearly unchanged, at 53.1 bits
with native Codex and 51.7 bits with our custom configuration.
Under the Essence judge, the median decreases from
74.9 to 68.1 bits, while coverage increases from 39/40 to 40/40
(Figure~\ref{fig:ablation-context-management}(d)).
{This comparison shows that our customized natural-language compaction performs at least as well as native Codex compaction.}

The one-candidate Codex configuration has lower median Directional compression cost than our
earlier Astra scaffold that retains explicit working state but discards
messages and reasoning between calls.
Figure~\ref{fig:ablation-context-management}(c) compares their coverage
over all 40 papers.
This earlier configuration uses Astra at max effort with
a Fable~5 guide.
Both configurations generate one candidate per submission.
With reasoning history, the median Directional compression cost decreases
from 69.9 to 63.7 bits, and coverage increases from 37/40 to 40/40.
This observation is consistent with OpenAI's report that reusing reasoning
across tool calls improves model performance~\citep{openai2025responses}. The Codex configuration additionally includes some small instructions for handling retained history and structured outputs. The details are deferred to
Appendix~\ref{app:ablation-scaffold-diffs}.

\section{Limitations and Discussion}
\label{sec:limitations}

Our work provides a compression benchmark for research idea generation.
A generator is scored by its compression cost on target ideas from future
papers. 
Because the bound holds for any guide, lowering the compression cost is a joint problem for the generator and the guide. A stronger generator places more probability on trajectories that reach the target idea, and a stronger guide finds cheaper trajectories that tighten the bound.

Our generator scaffold is heuristic and likely far from optimal. Inference-time scaling through ensembling already lowers the compression cost (Figures~\ref{fig:test87-essence} and~\ref{fig:ablation-randomness}(b,d)). Generator post-training also lowers it substantially (Figure~\ref{fig:ablation-posttraining}(a)). Our current scaffold also does not allow agents to make arbitrary tool calls.

Our benchmark has several limitations. First, it relies on LLMs to
summarize each target paper and to judge candidate ideas, so a generator
could potentially hack the judge. We observe vague phrasing in some
candidate ideas but no more serious judge hacking. Second, the benchmark measures whether a model can generate the target
idea at least once across many attempts. This differs from directly testing whether a model knows which hypotheses are true or which ideas work in deep learning.  {We consider developing reliable methods for such assessment an important direction for future work.} {Finally, automating research idea generation at scale could have
substantial societal consequences. Our metric tracks progress in
idea-generation capability and may help forecast when such automation
becomes feasible.}

\section{Related Work}
\label{sec:related-work}
{A growing line of work} uses language models to propose scientific hypotheses
and research directions through scaffolding~\citep{wang2023scimon,baek2024researchagent,zhao2025ramonllullsthinkingmachine,liu2026researchbenchbenchmarkingllmsscientific,yang2025moosechem2exploringllmlimits,yang2025moosechemlargelanguagemodels,zhao2026agenticideationsampleefficientagentic,wang2025scipipllmbasedscientificpaper,su2025headsbetteroneimproved,gottweis2025coscientist}. SciMON grounds generation in retrieved literature and
iteratively revises ideas to increase novelty~\citep{wang2023scimon}, while
ResearchAgent augments a core paper with an academic graph and uses reviewing
agents to refine proposed problems, methods, and experiments~\citep{baek2024researchagent}. Among these works, the study by \citet{zhao2026agenticideationsampleefficientagentic} is the closest to our work as it explores using guidance based on a target paper to create training trajectories to train idea generators.

Several benchmarks evaluate idea-generation capabilities using LLMs as judges of idea quality. A pioneering study with more than 100 NLP researchers in 2024 finds that LLM-generated
proposals can receive higher novelty ratings than human proposals, while also
identifying limitations in diversity and self-evaluation~\citep{si2024novelideas}.
Later follow-up work~\citep{si2025ideationexecutiongapexecutionoutcomes} reveals the ideation--execution gap for the LLM-proposed ideas, questioning the effectiveness of purely LLM-generated ideas. These findings are further supported by work benchmarking research judgment, showing that even current frontier LLM judges can disagree sharply with
domain experts on scientific novelty and feasibility~\citep{sinhahajari2026noveltyjudge,zheng2023llmjudge,wataoka2024selfpreference,baig2026taste,si2026executiongroundedautomatedairesearch}.

Another line of work
evaluates whether models can forecast research that appears later by revealing partial information to the idea-generating models~\citep{krenn2022predicting,li2026largelanguagemodelsforecast,jiang2026hindsightevaluatingllmgeneratedresearch,chen2026reconstructionblindbenchmarkrecovering}. 
IdeaBench
provides preferences and measures idea overlap, novelty, and
feasibility compared to real papers~\citep{guo2024ideabench}. LiveIdeaBench supplies only keywords and
tests divergent thinking across scientific domains~\citep{ruan2024liveideabench}. GIANTS asks a model to predict a target insight from two parent
papers~\citep{he2026giants}, and future-aligned proposal generation scores
proposals by their similarity to subsequently published work~\citep{wang2026futurealigned}.
AI Idea Bench emphasizes knowledge leakage and grounded feasibility
evaluation~\citep{qiu2025aiideabench}. 
Our work shares the same intuition and quantifies a target idea's compression cost, the additional information in bits needed to guide a model toward the idea's essence. This approach quantifies the value of partial information such as keywords or cited works, which are non-trivial to obtain~\citep{yang2026moosestarunlockingtractabletraining}.

Our use of priced guidance can be viewed as a special case of online algorithms with advice~\citep{boyar2017advice} and can be extended to perform rare output estimation~\citep{wu2024rareoutputs}. {We view our idea-generation evaluation method as an orthogonal component to automatic science execution~\citep{lu2024aiscientist,chen2024scienceagentbench}.}

\textbf{Information Theory and Rare-Event Estimation.}
{We measure compression at the level of ideas, allowing different textual realizations. Semantic uncertainty makes a related distinction for natural-language answers by grouping generations into meaning-equivalence classes and marginalizing probability over different surface forms~\citep{kuhn2023semanticuncertainty}.}
Our use of priced guidance is also related to online algorithms with advice,
where an all-knowing guide communicates information about the future and
complexity is measured by the number of advice bits~\citep{boyar2017advice}.
Rare-event simulation methods, including importance sampling and multilevel
splitting, bias sampling toward rare outcomes and correct for the resulting
change of measure~\citep{juneja2006rareevent}. Recent work applies rare-event
estimation directly to low-probability language-model outputs~\citep{wu2024rareoutputs}.
In our protocol, selecting an answer assigned probability $p$ costs its
self-information $-\log_2 p$. With the judge pass rate included in compression
cost $K$, the expected guided score $2^{-K}$ lower-bounds unguided success. 

\textbf{LLMs for Automated Scientific Research.}
Beyond idea generation, recent systems attempt to automate larger portions of
the research process. The AI Scientist generates ideas, writes code, runs
experiments, and produces papers~\citep{lu2024aiscientist}. The AI
Scientist-v2 adds agentic tree search and reports workshop-level automated
research~\citep{yamada2025aiscientistv2}.
A complementary line of benchmarks evaluates downstream research capability.
MLAgentBench measures autonomous machine-learning experimentation~\citep{huang2023mlagentbench}.
ScienceAgentBench and DiscoveryBench evaluate data-driven scientific
discovery~\citep{chen2024scienceagentbench,majumder2024discoverybench}.
PaperBench tests whether agents can replicate published AI research from paper
descriptions~\citep{starace2025paperbench} and RE-Bench compares agents with
human experts on time-bounded AI research-engineering tasks~\citep{wijk2024rebench}. {Our work complements these systems and benchmarks by evaluating idea-generation capability through the cost of compressing a target idea.}

\section*{Acknowledgments}

KW thanks the support of a Stanford Graduate Fellowship.
TM thanks the support of NSF 2211780.
This work was supported in part by the DARPA Artificial Intelligence Quantified (AIQ) program.
Part of the API credits used in this work were provided by OpenAI and Anthropic. We thank them for their support.
This work does not necessarily reflect the position or policy of the
government and no official endorsement should be inferred.

We are deeply inspired by Jacob Steinhardt's essay ``Research as a Stochastic Decision
Process''\footnote{\url{https://cs.stanford.edu/~jsteinhardt/ResearchasaStochasticDecisionProcess.html}}.
We also acknowledge J{\"u}rgen Schmidhuber's earlier work linking compression
progress to curiosity, creativity, and scientific
discovery~\citep{schmidhuber2009compression}.
We thank Xingyu Dang, Jiaxin Wen and Alec Radford for valuable discussions.

\newpage
\appendix

\section{Proofs and extensions of Priced Guidance}
\label{app:proof}

\subsection{Proof of Theorem~\ref{thm:compression}}
\label{app:conditional-coding}

The proof treats the generator's target-independent randomness as shared
side information. 
We can realize the generator's randomness in initialization and
transitions using a sequence of independent random draws, which we call
its \emph{random tape} $R$. The tape's distribution is independent of the
target. Once this tape is fixed, the guide's answers
determine a path through a deterministic tree. We construct a single prefix
code for all such paths, using the generator's bid probabilities.

\begin{proof}[Proof of Theorem~\ref{thm:compression}]
Fix a realization $R=r$ of the generator's random tape. Consider any
terminating guided trajectory $\tau$, whose selected actions are
$a_1,\ldots,a_{|\tau|}$. Write $C(\tau)=\sum_{t=1}^{|\tau|}c_t$ for its
answer cost, excluding the terminal judge cost. With the same tape,
unguided execution selects this answer sequence with probability
\begin{equation}
  \prod_{t=1}^{|\tau|}p_t(a_t\mid\tau_t)=2^{-C(\tau)}.
  \label{eq:appendix-forward-replay}
\end{equation}
The generator's state determines whether another answer is requested, so
one complete answer sequence cannot be a proper prefix of another.
Any guide decision that changes this state must be communicated through
an offered answer. No separate message length or stopping signal is needed.

An ideal arithmetic code encodes these answers using their bid
probabilities. Start with $[0,1)$ and partition the current interval at
each bid according to its answer probabilities. For the selected sequence,
let $I$ be the resulting interval; its length is $2^{-C(\tau)}$.
Intervals for distinct complete sequences are disjoint. Choose the binary
code length $\ell=\lceil C(\tau)\rceil+1$. Since
$2^{-\ell}\leq |I|/2$, the interval $I$ contains a binary interval
$[k2^{-\ell},(k+1)2^{-\ell})$ for some integer $k$.
Its $\ell$-bit address identifies the selected sequence to a decoder that
knows $r$. These binary intervals are disjoint, so their addresses form a
prefix code. The length satisfies
\begin{equation}
  \ell=\lceil C(\tau)\rceil+1<C(\tau)+2\leq K(\tau)+2.
  \label{eq:conditional-code-overhead}
\end{equation}
The last inequality holds because the judge cost is nonnegative.
The code depends only on the fixed generator tape and the bid
probabilities, so the same code works for every target and guide.
Runs assigned infinite cost impose no finite length requirement.
\end{proof}

\subsection{Proof of Theorem~\ref{thm:generator-oracle}}
\label{app:priced-guidance-proof}

Use the generator tape $R$ and answer cost $C(\tau)$ defined in
Appendix~\ref{app:conditional-coding}. 
We assume executions terminate after finitely many actions with probability
one.

\begin{proof}[Proof of the Priced Guidance Theorem]
Fix $R=r$ and any terminating guided trajectory $\tau$ under this tape.
Unguided execution with the same tape selects its actions with probability
$2^{-C(\tau)}$ and reproduces its final idea $x(\tau)$.
The fresh terminal judge accepts this idea with probability
$\Pr[J_T(x(\tau))=1]$. Other accepted paths may contribute additional
probability, so
\begin{equation}
  P_{\mathrm{unguide}}(\mathcal S_T\mid R=r)
  \geq\Pr[J_T(x(\tau))=1]2^{-C(\tau)}=2^{-K(\tau)}.
  \label{eq:appendix-conditional-acceptance}
\end{equation}
Here the left-hand side is the unguided acceptance probability conditional
on the generator tape. For a run without a final idea or with zero pass
rate, $K=+\infty$ and the inequality still holds.

The left-hand side does not depend on the guide's choices. Average over
guided trajectories conditional on $R=r$, then over $R$, we obtain
\begin{equation}
  P_{\mathrm{unguide}}(\mathcal S_T)
  \geq\E_{\tau\sim P_{\mathrm{guide}}}[2^{-K(\tau)}].
  \label{eq:appendix-coding-expectation}
\end{equation}
Guided and unguided execution use the same distribution for $R$.
The expectation includes generator and guide randomness, with the
fresh judge's randomness averaged into its pass rate in $K$.
If the expected cost is finite, convexity of $k\mapsto2^{-k}$ gives
\begin{equation*}
  \E_{\tau\sim P_{\mathrm{guide}}}[2^{-K(\tau)}]
  \geq2^{-\E_{\tau\sim P_{\mathrm{guide}}}[K(\tau)]}.
\end{equation*}
If the expected cost is infinite, the right-hand side is zero and the
inequality holds trivially. This proves both inequalities in
Equation~\ref{eq:generator-oracle-bound}.
\end{proof}

\subsection{Next-token prediction as a special case}
\label{app:perplexity}

Ordinary next-token prediction is a special case of the Priced Guidance
protocol. The generator offers vocabulary tokens as its answer choices and
assigns them the language model's next-token probabilities. The guide
selects the next token of the original string at every step. The resulting
compression cost is the string's negative log-likelihood, and its average
cost per token determines perplexity.

\begin{theorem}[Next-token prediction and perplexity]
\label{thm:perplexity-special-case}
Fix a finite vocabulary $\mathcal{V}$, a length $n\geq1$, and an initial
context $u$, all chosen independently of the target. Let $p_\theta$ be an
autoregressive language model with a strictly positive next-token
distribution on $\mathcal{V}$ at every prefix. Represent the original string
by its token sequence $y=(y_1,\ldots,y_n)\in\mathcal{V}^n$ under a fixed
tokenization, and take $T=y$. At step $t$, the previously selected tokens
are the actions $a_1,\ldots,a_{t-1}$; write this prefix as $a_{<t}$. The generator
asks which token comes next and commits to
\begin{equation}
  \mathcal{A}_t=\mathcal{V},\qquad
  p_t(a\mid \tau_t)=p_\theta(a\mid u,a_{<t}),\qquad a\in\mathcal{V}.
  \label{eq:token-generator-bid}
\end{equation}
Its transition appends the selected token deterministically, and it returns
the resulting sequence after exactly $n$ steps. The guide always selects
$y_t$, without recovery, and the deterministic judge accepts exactly when
the returned token sequence equals $y$. Then the guided compression cost is
\begin{equation}
  K(y\mid u)=-\sum_{t=1}^{n}\log_2p_\theta(y_t\mid u,y_{<t})
  =-\log_2p_\theta(y\mid u).
  \label{eq:perplexity-information-cost}
\end{equation}
Define the model's token perplexity on this string by
$\operatorname{PPL}_\theta(y\mid u)
=\bigl(\prod_{t=1}^{n}p_\theta(y_t\mid u,y_{<t})\bigr)^{-1/n}$.
It satisfies
\begin{equation}
  \operatorname{PPL}_\theta(y\mid u)=2^{K(y\mid u)/n}.
  \label{eq:perplexity-cost-relation}
\end{equation}
The bound in the Priced Guidance Theorem is exact in this setting,
\begin{equation}
  P_{\mathrm{unguide}}(\mathcal{S}_T)=p_\theta(y\mid u)
  =\E_{\tau\sim P_{\mathrm{guide}}}[2^{-K(\tau)}]
  =\operatorname{PPL}_\theta(y\mid u)^{-n}.
  \label{eq:perplexity-exact-bound}
\end{equation}
\end{theorem}

\begin{proof}
The guided prefix equals $y_{<t}$ at every step, so selecting $y_t$ costs
$-\log_2p_\theta(y_t\mid u,y_{<t})$ bits. The guided output is always $y$,
which the judge accepts with probability one, giving zero terminal judge
cost. Summing the answer costs and applying the autoregressive chain
rule proves Equation~\ref{eq:perplexity-information-cost}. Taking the
$n$th root of $2^{K(y\mid u)}$ gives
Equation~\ref{eq:perplexity-cost-relation}. Unguided execution samples each
token from the same conditional distribution. Its unique accepted token
sequence is $y$, whose probability is
$\prod_{t=1}^{n}p_\theta(y_t\mid u,y_{<t})=2^{-K(y\mid u)}$.
The guided run is deterministic, so its score equals its expected score.
This proves Equation~\ref{eq:perplexity-exact-bound}.
\end{proof}

The fixed length specifies the event measured by the theorem. For a model
that stops at an end-of-sequence token, the probability of generating the
complete string and then stopping also includes that token's probability;
selecting it incurs the corresponding additional compression cost.
Under the same fixed-length setup, if a deterministic judge instead accepts
any idea-equivalent output, including $y$, then $p_\theta(y\mid u)$ remains
a lower bound on acceptance probability.
Other accepted outputs may contribute additional probability. Thus
perplexity recovers the cost of specifying one original token sequence,
while the general protocol can guide generation through higher-level choices.

\subsection{ELBO interpretation}
\label{app:elbo}

We can also consider the Priced Guidance under an ELBO framework.
Write $q_t(a\mid\tau_t,T)$ for the guide's probability of selecting action
$a$ given trajectory prefix $\tau_t$ and target $T$. Define the guided
log-likelihood cost by
\begin{equation}
  L_{\mathrm{guide}}(\tau)
  =\sum_{t=1}^{|\tau|}\log_2
    \frac{q_t(a_t\mid\tau_t,T)}{p_t(a_t\mid\tau_t)}.
  \label{eq:guided-likelihood-cost}
\end{equation}
Write $D_{\mathrm{KL},2}$ for Kullback--Leibler divergence measured in bits.

\begin{theorem}[ELBO identity]
\label{thm:elbo-identity}
Fix a target $T$ and consider the protocol in
Section~\ref{sec:generator-oracle-protocol}. Suppose guided and unguided
runs terminate almost surely, guided runs succeed with probability one,
$P_{\mathrm{guide}}(\mathcal S_T)=1$, and
$D_{\mathrm{KL},2}(P_{\mathrm{guide}}\|P_{\mathrm{unguide}})<\infty$.
Then $P_{\mathrm{unguide}}(\mathcal S_T)>0$. Let
$P_{\mathrm{unguide},T}=P_{\mathrm{unguide}}(\cdot\mid\mathcal S_T)$
be the unguided trajectory law conditioned on acceptance. We have
\begin{equation}
  \log_2P_{\mathrm{unguide}}(\mathcal S_T)
  =-\E_{\tau\sim P_{\mathrm{guide}}}[L_{\mathrm{guide}}(\tau)]
   +D_{\mathrm{KL},2}(P_{\mathrm{guide}}\|P_{\mathrm{unguide},T}).
  \label{eq:oracle-elbo}
\end{equation}
Consequently, $-\E_{\tau\sim P_{\mathrm{guide}}}[L_{\mathrm{guide}}(\tau)]$
is an evidence lower bound (ELBO) on the log acceptance probability.
The bound is exact if and only if
$P_{\mathrm{guide}}=P_{\mathrm{unguide},T}$.
\end{theorem}

\begin{proof}
Guided and unguided trajectories share their initialization, generator
transitions, and terminal judging rule. Every offered action has positive
generator probability, so $P_{\mathrm{guide}}\ll P_{\mathrm{unguide}}$.
Almost-sure termination makes trajectories finite. Their shared factors
cancel, giving the likelihood ratio $P_{\mathrm{guide}}$-almost surely:
\begin{equation}
  \log_2\frac{\mathrm dP_{\mathrm{guide}}}{\mathrm dP_{\mathrm{unguide}}}(\tau)
  =\sum_{t=1}^{|\tau|}\log_2
    \frac{q_t(a_t\mid\tau_t,T)}{p_t(a_t\mid\tau_t)}
  =L_{\mathrm{guide}}(\tau).
  \label{eq:appendix-likelihood-ratio}
\end{equation}
Its guided expectation is therefore
$D_{\mathrm{KL},2}(P_{\mathrm{guide}}\|P_{\mathrm{unguide}})$.

Since guided trajectories are accepted almost surely, absolute continuity
implies $P_{\mathrm{unguide}}(\mathcal S_T)>0$. Conditioning the unguided
law on acceptance divides its density on $\mathcal S_T$ by this probability.
Thus, on $P_{\mathrm{guide}}$-almost every trajectory,
\begin{equation*}
  \log_2\frac{\mathrm dP_{\mathrm{guide}}}{\mathrm dP_{\mathrm{unguide},T}}(\tau)
  =L_{\mathrm{guide}}(\tau)+\log_2P_{\mathrm{unguide}}(\mathcal S_T).
\end{equation*}
Taking the guided expectation gives
\begin{equation}
  D_{\mathrm{KL},2}(P_{\mathrm{guide}}\|P_{\mathrm{unguide},T})
  =\E_{\tau\sim P_{\mathrm{guide}}}[L_{\mathrm{guide}}(\tau)]
   +\log_2P_{\mathrm{unguide}}(\mathcal S_T).
  \label{eq:appendix-conditional-kl}
\end{equation}
The finite-divergence assumption allows rearrangement, yielding
Equation~\ref{eq:oracle-elbo}. Nonnegativity of KL divergence gives the
lower bound, and the divergence is zero exactly when the two laws agree.
\end{proof}

For a deterministic guide, each selected action has guide probability one,
so $L_{\mathrm{guide}}(\tau)=C(\tau)$. Almost-sure guided acceptance makes
the judge cost zero, hence $L_{\mathrm{guide}}(\tau)=K(\tau)$ almost surely.
In this case, the ELBO is $-\E_{\tau\sim P_{\mathrm{guide}}}[K(\tau)]$,
as in Theorem~\ref{thm:generator-oracle}.
A guide restricted to selecting offered answers need not be able to
realize the acceptance-conditioned law that makes the ELBO exact.

\subsection{Proofs of extensions}
\label{app:extensions}

Each extension adds a target-independent construction to the generator.
We call the resulting generator a \emph{meta-generator}. We apply
Theorem~\ref{thm:generator-oracle} to this meta-generator, then identify
which unguided success probability it bounds. For candidate selection and
recovery, averaging out the added randomness recovers the original
unguided experiment. For ensembling it recovers the mixture of members.
The promotion construction specifies a modified unguided generator, whose
success probability is the one controlled by its bound.

\paragraph{A common reduction.}
We use the protocol's existing notation for a meta-generator as well.
Thus $\mathcal S_T$ means acceptance by its terminal judge, and $K$ is
its guided answer cost plus terminal judge cost. As in the ensemble
construction in Section~\ref{sec:extensions}, write $P_{\mathrm{meta}}$
for the meta-generator's unguided law. Extend the random tape $R$ to include
the target-independent auxiliary seeds used in the construction.
Conditional on $R=r$, suppose the meta-generator satisfies the
priced guidance protocol. To transfer its bound to the original unguided
generator, we will check that
\begin{equation}
  \E_R[P_{\mathrm{meta}}(\mathcal S_T\mid R)]
  =P_{\mathrm{unguide}}(\mathcal S_T).
  \label{eq:meta-marginal-identity}
\end{equation}
On each side, acceptance is evaluated by the corresponding experiment's
terminal judge. The identity says that the two experiments have the same
acceptance probability after the auxiliary randomness is averaged out.

\begin{proof}[Reduction to Theorem~\ref{thm:generator-oracle}]
For a fixed tape $R=r$, apply Theorem~\ref{thm:generator-oracle}
to the meta-generator and its guide. This gives
$P_{\mathrm{meta}}(\mathcal S_T\mid R=r)\geq
\E[2^{-K}\mid R=r]$.
Average over $R$ and use Equation~\ref{eq:meta-marginal-identity} to obtain
\begin{equation}
  P_{\mathrm{unguide}}(\mathcal S_T)\geq\E[2^{-K}]
  \geq 2^{-\E[K]}.
  \label{eq:meta-transfer-bound}
\end{equation}
Both expectations on the right are over guided meta-runs. The last step
is Jensen's inequality, with the usual infinite-cost convention.
\end{proof}

The guide used in this reduction may depend on $r$. After $r$ is fixed,
it is simply one of the guides allowed by the theorem. This lets us
analyze a guide that first explores random continuations and then replays
its selected path. Each construction below identifies
the unguided experiment whose probability it bounds.

\subsubsection{Multiple candidate ideas}
\label{app:multiple-candidates}

A generator may submit several candidate ideas with assigned probabilities. The corresponding unguided
generator draws one candidate using those weights and evaluates it once.

\begin{theorem}[Multiple candidate ideas]
\label{thm:multiple-candidate-bound}
Suppose a generator follows the priced guidance protocol and returns a
finite list $(x_1,\ldots,x_k)$ with weights $p_i>0$ satisfying
$\sum_{i=1}^k p_i=1$, all determined by its trajectory.
The terminal evaluation samples $I$ from these weights and applies a fresh
judge to $x_I$. With $C(\tau)=\sum_t c_t$ denoting the answer cost, define
\begin{equation}
  K(\tau)=C(\tau)-\log_2\!\left(\sum_{i=1}^k p_i\Pr[J_T(x_i)=1]\right).
  \label{eq:multiple-candidate-pass-rate}
\end{equation}
Then $P_{\mathrm{unguide}}(\mathcal S_T)\geq\E[2^{-K(\tau)}]
\geq2^{-\E[K(\tau)]}$, with expectations over guided runs.
\end{theorem}

\begin{proof}
\textbf{Meta-generator.}
Keep the generator's bids and transitions, and let its final output be the
weighted list. Its terminal judge independently samples $I$ with probability
$p_I$ and returns $J_T(x_I)$. Conditional on the list, its pass rate is
$\sum_i p_i\Pr[J_T(x_i)=1]$, so its total cost is exactly $K(\tau)$.

\textbf{Verification of Equation~\ref{eq:meta-marginal-identity}.}
The unguided meta-generator produces the same weighted list as the original
generator. Its terminal judge performs the same candidate sampling and
fresh evaluation as the unguided experiment. Their success probabilities
therefore agree after averaging over $R$, establishing
Equation~\ref{eq:meta-marginal-identity}. The common reduction gives the bound.
\end{proof}

\subsubsection{Recovery}
\label{sec:recovery}

Recovery lets the guide restore a saved prefix $\tau_t$, including its
committed bid and generator state, and request another continuation under
an offered action $a$. The guide may choose requests, restorations, and
when to stop using all previously observed exploration. For each exact
prefix--action pair, number the independent draws $j=1,2,\ldots$; this is
the \emph{occurrence index}. Each draw uses the original kernel $g_t$ and,
for a final idea, a fresh judge evaluation. Indices never reset.
Assume countably many possible saved prefixes, independent draws across
prefix--action pairs, and the same target-independent initialization in
guided and unguided runs, independent of those draws.

\begin{theorem}[Online indexed recovery]
\label{thm:recovery-aware-replay}
Fix a target-independent prior $\pi(j)>0$ with $\sum_{j\geq1}\pi(j)=1$.
Suppose the guide makes finitely many requests almost surely and returns a
consistent trajectory $\tau$ with retained indices $j_t$.
Define its indexed cost by
\begin{equation}
  K_\pi=K(\tau)-\sum_{t=1}^{|\tau|}\log_2\pi(j_t).
  \label{eq:indexed-recovery-cost}
\end{equation}
Here $K(\tau)$ is the original answer-plus-judge cost. Only the returned
path is charged, but its indices count all requests from the corresponding
prefix--action pairs. A missing final idea or zero pass rate gives
$K_\pi=+\infty$. Then
\begin{equation}
  P_{\mathrm{unguide}}(\mathcal S_T)
  \geq\E[2^{-K_\pi}]\geq2^{-\E[K_\pi]}.
  \label{eq:recovery-success-bound}
\end{equation}
The left-hand side is the acceptance probability of one fresh unguided
run.

If each terminal draw instead includes a fixed number $n\geq1$ of
independent judge evaluations with $l$ passes, define $\widehat K_\pi$ by
replacing the terminal judge cost in $K_\pi$ with $-\log_2(l/n)$, or
$+\infty$ when $l=0$. Then
\begin{equation}
  P_{\mathrm{unguide}}(\mathcal S_T)\geq\E[2^{-\widehat K_\pi}].
  \label{eq:appendix-empirical-terminal-bound}
\end{equation}
\end{theorem}

\begin{proof}
\label{app:recovery-proof}
\textbf{Meta-generator.}
Include the stored judge seeds in $R$. Fix $R=r$ and offer an action $a$
and index $j$ with probability $p_t(a\mid\tau_t)\pi(j)$, then follow the
stored generator continuation.
\footnote{To select $j$ using finite bids, start at one and offer
\emph{stop} with probability $\pi(j)/\sum_{m\geq j}\pi(m)$, otherwise
\emph{continue}. The probability of stopping at $j$ is $\pi(j)$, so the
cost is $-\log_2\pi(j)$ and selection terminates almost surely.}
Stored verdicts are private to the guide, and a fresh terminal judge
assesses the final idea. Replaying the guide's retained actions and
indices costs exactly $K_\pi$. For repeated terminal judgments, retain the
$n$ judge seeds on reuse and include them in the meta-generator's output.
Let its terminal judge select one of these seeds uniformly. For fixed $R$,
its pass rate is $l/n$, giving cost $\widehat K_\pi$.

\textbf{Verification of Equation~\ref{eq:meta-marginal-identity}.}
An unguided meta-run never revisits an exact prefix because its trajectory
grows at each step. Its sampled index therefore selects an unvisited
independent draw with generator law $g_t(\cdot\mid\tau_t,a)$.
Summing over indices also preserves the action probabilities because
$\sum_j p_t(a\mid\tau_t)\pi(j)=p_t(a\mid\tau_t)$.
Averaging over $R$ and forgetting the indices recovers the original
initialization, transitions, and fresh terminal judging rule.
For repeated judgments, the selected batch is fresh and independent of the
candidate; uniformly choosing one of its seeds likewise recovers one fresh
judge evaluation. Thus Equation~\ref{eq:meta-marginal-identity} holds for
both constructions, and the common reduction gives both bounds.
\end{proof}

\paragraph{Remark.}
The result applies to scaffolds with the saved-state and independence
properties above. The guide may search adaptively; its effect on the
replayed generator is specified by the retained actions and indices.
Our experiments use
\begin{equation}
  \pi(j)=
  \begin{cases}
    0.95, & j=1,\\
    \dfrac{0.05}{j(j-1)}, & j\geq2.
  \end{cases}
  \label{eq:fixed-occurrence-prior}
\end{equation}
Normalization follows from $1/[j(j-1)]=1/(j-1)-1/j$.
The first occurrence costs about $0.074$ extra bits, even without a retry;
later occurrences cost $\log_2(j(j-1)/0.05)$ bits.

\subsubsection{Ensembling models and seeds}
\label{sec:ensembling}

Fix $M$ generator configurations, which may differ in model or generator
seed, and weights $w_i>0$ with $\sum_iw_i=1$ before the target is revealed.
Here the original unguided experiment is the ensemble that samples
$I$ with probability $w_I$, runs only member $I$, and applies the common
terminal judge. Let $P_i$ be member $i$'s unguided law and $K_i$ its
guided compression cost, including the judge cost and any occurrence costs.

\begin{theorem}[Ensembling]
\label{thm:ensemble-bound}
Suppose every member satisfies the priced guidance protocol, with recovery
represented by the construction above when needed. Define
$K_{\mathrm{meta}}=-\log_2\sum_{i=1}^Mw_i2^{-K_i}$.
The unguided ensemble satisfies
\begin{equation}
  P_{\mathrm{unguide}}(\mathcal S_T)
  =\sum_{i=1}^M w_iP_i(\mathcal S_T)
  \geq\E[2^{-K_{\mathrm{meta}}}]
  \geq2^{-\E[K_{\mathrm{meta}}]}.
  \label{eq:meta-harness-bound}
\end{equation}
The expectation is over the guided member runs, which need not be independent.
\end{theorem}

\begin{proof}
\textbf{Meta-generator.}
Sample member $I$ with probabilities $w_1,\ldots,w_M$ during initialization,
then use its generator, including its recovery construction when needed.
The guide observes $I$ and uses that member's guide. This initial choice
has the same distribution in guided and unguided runs, so it has no answer
cost. The resulting guided cost is $K_I$.

\textbf{Verification of Equation~\ref{eq:meta-marginal-identity}.}
For each member, averaging over $R$ and forgetting any occurrence indices
recovers its unguided success probability $P_i(\mathcal S_T)$.
Averaging over $I$ therefore gives $\sum_iw_iP_i(\mathcal S_T)$,
which is exactly the original ensemble's success probability. This verifies
Equation~\ref{eq:meta-marginal-identity}. The common reduction gives
$P_{\mathrm{unguide}}(\mathcal S_T)\geq\sum_iw_i\E[2^{-K_i}]
=\E[2^{-K_{\mathrm{meta}}}]$.
Linearity of expectation permits any coupling of the guided member runs;
Jensen's inequality gives the final inequality.
\end{proof}

\paragraph{Remark.}
This applies to any fixed mixture of valid generator scaffolds and bounds
the mixture's success probability. If the guide instead chooses member $i$,
that choice must be an offered bid and costs $-\log_2w_i$.
The averaged ensemble score already includes these weights.
For both development and test ensembles, we use $w_i=1/M$. Assign a
successful member its empirically measured cost $\widetilde K_i$ and a
failed or missing member cost $\widetilde K_i=+\infty$. The observed
effective cost is
\begin{equation}
  K_{\mathrm{meta}}
  =-\log_2\!\left(\frac{1}{M}\sum_{i=1}^{M}2^{-\widetilde K_i}\right).
  \label{eq:uniform-ensemble-cost}
\end{equation}
The denominator remains $M$ when members fail or are missing, and no further
member-selection cost is added. The observed score contributes to the
expectation bound; a finite-sample probability guarantee additionally
requires accounting for statistical uncertainty.

\subsubsection{Promotion cost}
\label{app:promotion-cost}

Promotion sends the generator a fixed success message and switches to
Essence prompts (\Cref{sec:bid-construction}). Rejected submissions are
handled privately through recovery. After promotion, the guide may restore
an earlier Directional question and continue in Essence, so the location
of the stage switch must be priced.

We use a target-independent \emph{replay scaffold}. Before answering each
Directional question, it offers a switch/stay bid with fixed positive probabilities.
Switching preserves the committed question and its probabilities, and
subsequent generation uses Essence prompts. Selecting Submit in Directional
also promotes, without consulting the Directional judge, and repeats the
same question and probabilities in Essence. Selecting Submit again sends
the prepared submission to the Essence judge. Let $P_{\mathrm{replay}}$
denote this scaffold's unguided law.

Start with the final Essence trajectory retained after recovery. It contains
the choices on the returned path, with their bid probabilities and occurrence
indices; discarded branches are excluded. Let $K_\pi$ be this path's
answer, occurrence, and Essence judge cost, before charging for promotion.
We construct a trajectory of the replay scaffold by keeping these choices
and inserting the choices needed to reproduce the stage switch
(Figure~\ref{fig:promotion-replay}). There are three cases.
\begin{enumerate}[leftmargin=*,nosep]
  \item \textbf{Reuse the promoting submission.} The retained Submit
  promotes the replay scaffold. Insert a second Submit at the repeated
  question to send the same submission to the Essence judge.
  \item \textbf{Resume at the promoting question.} Recovery removed the
  promoting Submit from the retained path. Insert that Submit to promote,
  then take the retained Essence answer at the repeated question.
  \item \textbf{Resume at another question.} Insert a switch immediately
  before the first retained Essence answer at that question.
\end{enumerate}
In all three cases, also insert a stay before every question answered in
Directional, including the promoting Submit in the first two cases.

The \emph{promotion cost} $c_{\mathrm{prom}}$ is the total cost of these
insertions. Each inserted stay or switch costs minus the base-2 logarithm
of its bid probability and has no occurrence cost, since its transition is
deterministic. In the first two cases, let $p_S$ and $j_S$ denote the
inserted Submit's bid probability and occurrence index. This Submit costs
$-\log_2[p_S\pi(j_S)]$, with $j_S=1$ for the repeated Submit in the first
case and the removed Submit's original occurrence index in the second.
The replay trajectory therefore has total cost $K_\pi+c_{\mathrm{prom}}$.

\begin{figure}[H]
  \centering
  \begin{tikzpicture}[
  x=1cm,y=1cm,font=\normalsize,
  box/.style={draw=black!35,fill=black!3,rounded corners=2pt,
    line width=.55pt,inner sep=4pt,align=center,minimum height=.64cm},
  inserted/.style={box,draw=blue!65!black,fill=blue!6,
    text=blue!65!black,line width=.8pt},
  arrow/.style={-{Latex[length=1.7mm,width=1.2mm]},draw=black!55,line width=.6pt},
  rowlabel/.style={anchor=west,font=\normalsize\itshape},
  title/.style={anchor=west,font=\large\bfseries},
  annotation/.style={font=\normalsize,text=black!60,fill=white,inner sep=1pt}
]
\path[use as bounding box] (0,.22) rectangle (15.05,-8.02);

\node[title] at (0,0) {(a) Reuse the promoting submission};
\node[rowlabel] at (0,-.73) {Retained};
\node[rowlabel] at (0,-1.75) {Replay};
\node[box,text width=1.9cm] (a-prefix) at (3,-.73) {Directional\\prefix};
\node[box,text width=1.5cm] (a-submit) at (7.7,-.73) {Submit};
\node[box,text width=1.9cm] (a-judge) at (13.7,-.73) {Essence\\judge};
\draw[arrow] (a-prefix) -- (a-submit);
\draw[arrow] (a-submit) -- node[annotation,above] {promote and reuse} (a-judge);
\node[box,text width=1.9cm] (ar-prefix) at (3,-1.75) {Directional\\prefix};
\node[inserted,text width=1.12cm] (ar-stay) at (5.45,-1.75) {$+$ Stay};
\node[box,text width=1.7cm] (ar-submit) at (7.7,-1.75) {Submit\\(promote)};
\node[inserted,text width=1.7cm] (ar-repeat) at (10.55,-1.75) {$+$ Submit};
\node[box,text width=1.9cm] (ar-judge) at (13.7,-1.75) {Essence\\judge};
\draw[arrow] (ar-prefix) -- (ar-stay);
\draw[arrow] (ar-stay) -- (ar-submit);
\draw[arrow] (ar-submit) -- (ar-repeat);
\draw[arrow] (ar-repeat) -- (ar-judge);

\draw[black!15] (0,-2.5) -- (15,-2.5);
\node[title] at (0,-2.98) {(b) Resume at the promoting question};
\node[rowlabel] at (0,-3.71) {Retained};
\node[rowlabel] at (0,-4.73) {Replay};
\node[box,text width=1.9cm] (b-prefix) at (3,-3.71) {Directional\\prefix};
\node[box,text width=1.7cm] (b-answer) at (10.55,-3.71) {Essence\\answer};
\node[box,text width=1.9cm] (b-judge) at (13.7,-3.71) {Essence\\judge};
\draw[arrow] (b-prefix) -- node[annotation,above] {Submit removed by recovery} (b-answer);
\draw[arrow,dashed] (b-answer) -- (b-judge);
\node[box,text width=1.9cm] (br-prefix) at (3,-4.73) {Directional\\prefix};
\node[inserted,text width=1.12cm] (br-stay) at (5.45,-4.73) {$+$ Stay};
\node[inserted,text width=1.7cm] (br-submit) at (7.7,-4.73) {$+$ Submit\\(promote)};
\node[box,text width=1.7cm] (br-answer) at (10.55,-4.73) {Essence\\answer};
\node[box,text width=1.9cm] (br-judge) at (13.7,-4.73) {Essence\\judge};
\draw[arrow] (br-prefix) -- (br-stay);
\draw[arrow] (br-stay) -- (br-submit);
\draw[arrow] (br-submit) -- (br-answer);
\draw[arrow,dashed] (br-answer) -- (br-judge);

\draw[black!15] (0,-5.48) -- (15,-5.48);
\node[title] at (0,-5.96) {(c) Resume at another question};
\node[rowlabel] at (0,-6.69) {Retained};
\node[rowlabel] at (0,-7.71) {Replay};
\node[box,text width=1.9cm] (c-prefix) at (3,-6.69) {Directional\\prefix};
\node[box,text width=1.7cm] (c-answer) at (10.55,-6.69) {Essence\\answer};
\node[box,text width=1.9cm] (c-judge) at (13.7,-6.69) {Essence\\judge};
\draw[arrow] (c-prefix) -- node[annotation,above] {stage switch at the restored question} (c-answer);
\draw[arrow,dashed] (c-answer) -- (c-judge);
\node[box,text width=1.9cm] (cr-prefix) at (3,-7.71) {Directional\\prefix};
\node[inserted,text width=1.7cm] (cr-switch) at (7.7,-7.71) {$+$ Switch};
\node[box,text width=1.7cm] (cr-answer) at (10.55,-7.71) {Essence\\answer};
\node[box,text width=1.9cm] (cr-judge) at (13.7,-7.71) {Essence\\judge};
\draw[arrow] (cr-prefix) -- (cr-switch);
\draw[arrow] (cr-switch) -- (cr-answer);
\draw[arrow,dashed] (cr-answer) -- (cr-judge);
\end{tikzpicture}
  \caption{\textbf{Promotion cost pays for choices inserted into the retained path.}
  Each panel compares the choices retained after recovery (top) with their
  replay (bottom). Blue boxes marked $+$ are inserted choices and contribute
  to $c_{\mathrm{prom}}$; gray boxes are already accounted for in $K_\pi$.
  (a) Reusing the submission requires a second Submit for Essence evaluation.
  (b) Resuming at the promoting question requires restoring the removed Submit
  before taking the retained Essence answer.
  (c) Resuming elsewhere requires a switch before that answer.
  The question and its bid probabilities are preserved across the switch.
  Dashed arrows omit intervening Essence choices.
  Earlier stay choices within each Directional prefix are omitted; each also
  contributes to $c_{\mathrm{prom}}$.}
  \label{fig:promotion-replay}
\end{figure}

\begin{theorem}[Promotion]
\label{thm:promotion-bound}
Under the recovery assumptions, every returned Essence trajectory can be
replayed at cost $K_\pi+c_{\mathrm{prom}}$. The unguided replay scaffold
satisfies
\begin{equation*}
  P_{\mathrm{replay}}(\mathcal S_T)
  \geq\E[2^{-(K_\pi+c_{\mathrm{prom}})}]
  \geq2^{-\E[K_\pi+c_{\mathrm{prom}}]},
\end{equation*}
with expectations over all guided recovery runs.
\end{theorem}

\begin{proof}
\textbf{Meta-generator.}
Apply the indexed recovery construction to the continuation draws of the
replay scaffold, using the original indexed draws for the retained answers.
The trajectory construction above reproduces the stage switch and every
retained continuation. The retained path contributes $K_\pi$, and the
inserted choices contribute $c_{\mathrm{prom}}$.

\textbf{Verification of Equation~\ref{eq:meta-marginal-identity}.}
Averaging over $R$ and forgetting occurrence indices preserves the
initialization, stage-bid probabilities, continuation kernels, and Essence
terminal judging rule of the replay scaffold. Therefore
$\E_R[P_{\mathrm{meta}}(\mathcal S_T\mid R)]
=P_{\mathrm{replay}}(\mathcal S_T)$, verifying
Equation~\ref{eq:meta-marginal-identity} for the replay baseline.
The common reduction gives the bound with cost $K_\pi+c_{\mathrm{prom}}$.
\end{proof}

\paragraph{Remark.}
This bounds the replay scaffold's success probability. Transferring it to
the original, judge-triggered two-stage scaffold requires a further
comparison of their success probabilities.

Our replay scaffold uses switch/stay probabilities $0.01$ and $0.99$.
The binary switch costs
$-\log_2(0.01)\approx6.64$ bits, and each binary stay choice adds
$-\log_2(0.99)\approx0.0145$ bits.

Among the 354 successful test runs, 206 reuse the promoting
submission, 139 restore its Submit, and 9 switch at another question.
Our scaffold fixes $p_S=0.283$, and $j_S=1$ in all runs, so the extra Submit
costs 1.90 bits. The total promotion cost has median 2.1 bits and maximum
6.9 bits; development runs are similar. All reported Essence costs include
this cost.
\FloatBarrier

\section{Additional Experimental Details}
\label{app:experimental-details}

\subsection{Memorization audit}
\label{app:memorization-audit}

Target selection and set sizes are given in Section~\ref{sec:memorization-audit}.

\textbf{Title-based recall audit.}
To assess whether a model has already memorized a candidate paper, we
provide only its title and ask for its research problem, method, technical
approach, and findings. 
For titles containing a colon, we retain only the
text before the first colon to reduce information supplied by descriptive
subtitles. 
We compare the recalled summary with the private gold summary. A fixed
GPT-5.6 Terra Judge assigns a \emph{memorization score} from 0 to 100 based
on semantic agreement, allowing differences in wording and level of detail.
We first apply this audit to 849 Top-1 selections from Hugging Face
Daily Papers dated between May 2023 and August 2026, using Claude Opus~5
for recall.
Figure~\ref{fig:monthly-memorization} shows high recall through November
2025, followed by a sharp decline. The monthly mean falls from 87.9 in November
to 52.6 in December. We therefore adopt 80 as the operational threshold
for identifying conspicuous recall.

\textbf{Shared exclusion rule.}
We exclude papers for which any of our generators receives a memorization score
of at least 80. This leaves 40 papers in the development set and 87 in the
test set, with no overlap between the sets. All five audited models score
below 80 on every retained paper (Table~\ref{tab:memorization-audit}).

\begin{figure}[tbp]
  \centering
  \includegraphics[width=\linewidth]{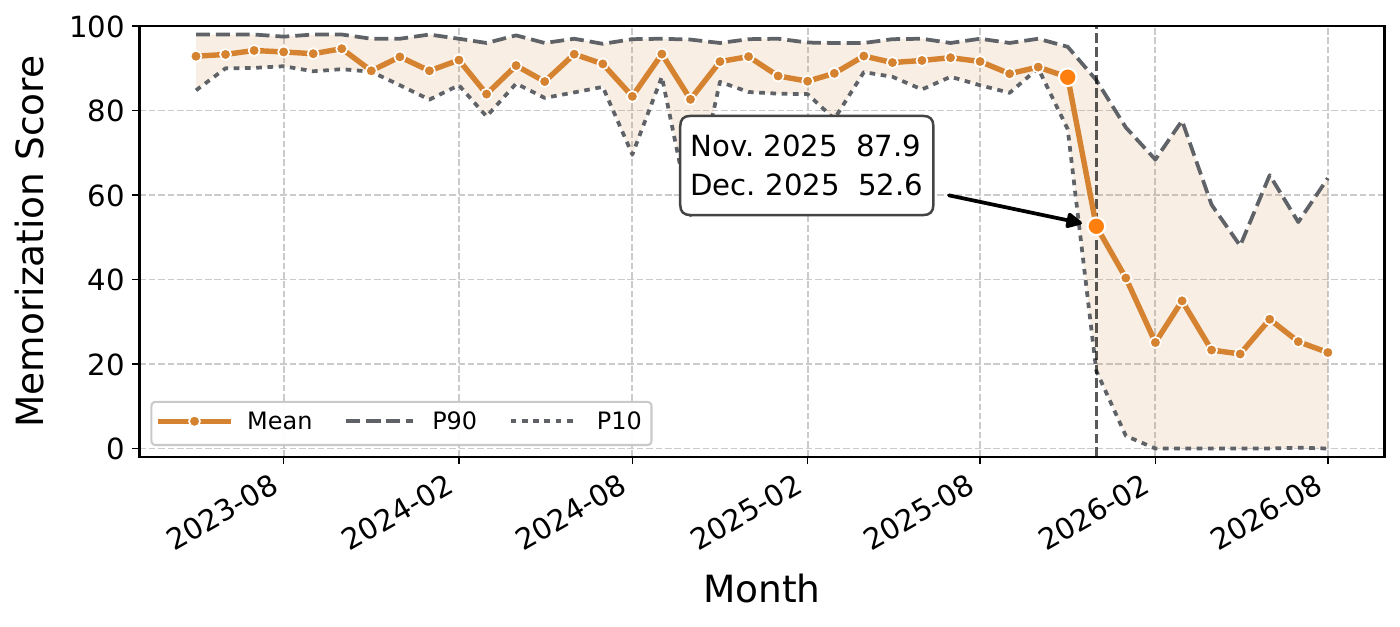}
  \caption{Monthly title-only memorization scores for Claude Opus~5 on 849
  Top-1 selections from Hugging Face Daily Papers, judged by GPT-5.6 Terra. {We can observe a clear transition near the knowledge cutoff.} The
  solid orange curve is the monthly mean; the dashed and dotted curves show
  the 90th and 10th percentiles, with their interval shaded. Each month
  contains 15--25 papers.}
  \label{fig:monthly-memorization}
\end{figure}

\begin{table}[H]
  \caption{Title-only recall on the 40-paper development set and 87-paper test
  set, scored from 0 to 100 by GPT-5.6 Terra. All five models are evaluated on
  the same papers within each set. Means are computed after cohort selection.
  Score $\geq 80$ counts papers meeting the memorization threshold.}
  \label{tab:memorization-audit}
  \centering
  \small
  \begin{tabular}{lrrrrr}
    \toprule
    Metric & Fable 5 & Opus 5 & GPT-5.6 Sol & GLM-5.3 & Astra \\
    \midrule
    \multicolumn{6}{l}{\textit{Development set (40 papers)}} \\
    Mean score & 22.30 & 24.78 & 28.05 & 9.93 & 2.50 \\
    Score $\geq 80$ & 0 & 0 & 0 & 0 & 0 \\
    \midrule
    \multicolumn{6}{l}{\textit{Test set (87 papers)}} \\
    Mean score & 23.37 & 22.70 & 27.33 & 13.40 & 1.51 \\
    Score $\geq 80$ & 0 & 0 & 0 & 0 & 0 \\
    \bottomrule
  \end{tabular}
\end{table}

\subsection{Compression cost over time}
\label{app:cost-over-time}

Figure~\ref{fig:ensemble-cost-over-time} plots the ensemble's compression
cost on each test paper against the paper's listing date.

\begin{figure}[tbp]
  \centering
  \includegraphics[width=0.96\linewidth]{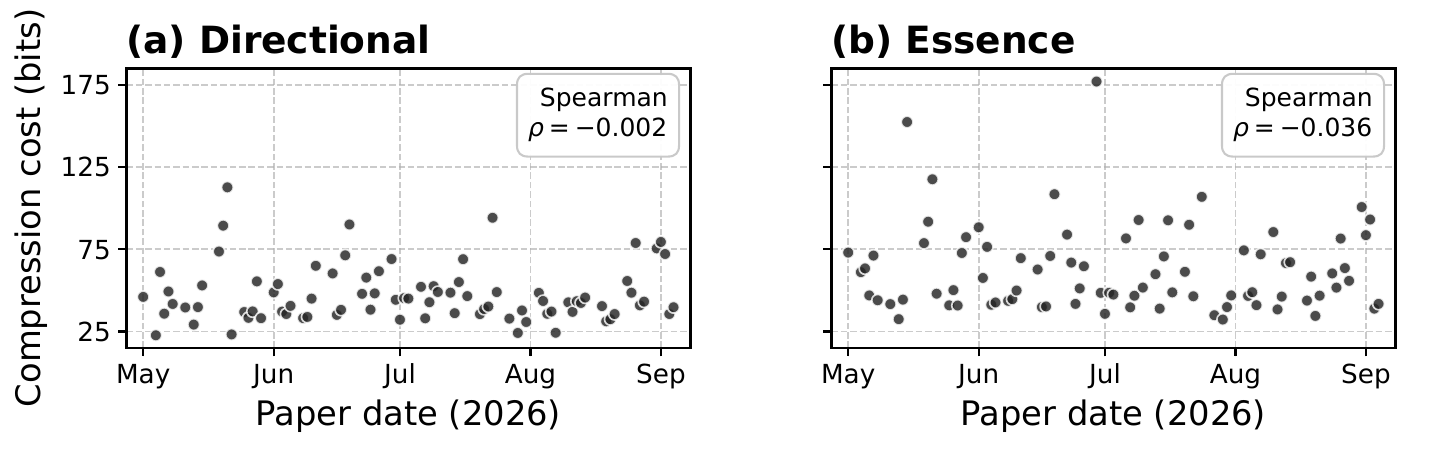}
  \caption{\textbf{Ensemble compression cost versus paper date.}
  Black points show Directional (left) and Essence (right) costs for the
  uniform mixture of Fable~5.1, Opus~5, and Astra on the 87-paper test set
  (Appendix~\ref{sec:ensembling}). Dates are Hugging Face Daily Papers listing
  dates from May~1 to September~4, 2026, spanning approximately four months. Both correlations are close to zero over this period.
  }
  \label{fig:ensemble-cost-over-time}
\end{figure}

\subsection{Verbatim GLM~5.3 baseline}
\label{app:glm-verbatim-baseline}

We score each gold summary token by token with GLM~5.3 under a fixed prompt
asking for a deep-learning research idea, with no target-specific context.
Multiplying the base-2 logarithm of perplexity by the number of summary tokens
gives the negative log-likelihood in bits (Appendix~\ref{app:perplexity}).
We score only the summary tokens, excluding the prompt and termination.
The reported median of 1,147.1 bits uses low reasoning effort and an empty
reasoning trace. Changing only the effort instruction to high or max gives
similar medians of 1,153.5 and 1,194.5 bits, respectively.

\subsection{Lossless gzip baseline}
\label{app:gzip-baseline}

We compress the same gold idea summaries used for the GLM~5.3 verbatim
baseline in Section~\ref{sec:results}. Each summary contains the contribution
category and the grouped descriptions of the setting and main object;
paper titles, other gold-summary fields, and JSON syntax are excluded.
We compress each of the 87 UTF-8 strings independently with gzip at level~9,
with the timestamp set to zero and no stored filename, preset dictionary,
or context from other papers.
We count the complete compressed file, including its 18-byte header and
trailer, and multiply its length in bytes by eight.

The median compressed length is 5,712 bits (714 bytes), compared with
10,008 bits for the uncompressed text. Excluding the header and trailer
gives a median DEFLATE payload of 5,568 bits.
Compression level~6 gives the same median as level~9, while level~1 gives
5,832 bits. All compressed summaries decompress exactly to the original
bytes. Independent runs with Python's gzip implementation and the system
gzip command give identical compressed lengths for all 87 summaries.

This baseline changes both the compressor and the reconstruction criterion
relative to Priced Guidance. It therefore does not isolate the effect of
recovering an idea's essence rather than its wording. The GLM~5.3
comparison in Section~\ref{sec:results} holds model identity fixed.

\subsection{Additional analyses in the main text}
\label{app:additional-test-results}
The full test-set comparisons and cost breakdown are in Section~\ref{sec:results}.
The context-management comparisons are in Section~\ref{sec:context-management-ablation}.

\section{Prompts and Agent Protocol}
\label{app:prompts}

This appendix presents excerpts of the Directional and Essence prompts used in
our experiments. Appendix~\ref{app:ablation-scaffold-diffs} identifies the
scaffold variants used in Section~\ref{sec:ablations} and their differences
from the eight-candidate prompts below. We therefore list each shared block
once and state how the blocks are assembled.  Line wrapping and typographic dashes
have been normalized, and run-dependent values are replaced by angle-bracketed
placeholders. The excerpts preserve the original wording, including metric and
role names. Omissions within quoted blocks are marked with \texttt{[...]}.
The retained text is reproduced from \texttt{participant/pair.py}, the
Directional and Essence stage modules, their channel modules, and the
Arena-owned \texttt{evaluation/research.py} Judge.

The experiments in Section~\ref{sec:experiment-instantiation} use the
Directional and Essence stages. The excerpts retain the original pricing
guidance.
The reported compression costs are recomputed from the recorded trajectories
using the occurrence prior in Appendix~\ref{sec:recovery} and the
multiple-candidate judge cost in \Cref{sec:extensions};
informal cost estimates inside the prompts do not define that accounting.

For a multiple-choice (MC) or keyword call at recognition stage $s$ and action $a$,
let $P_{\mathrm{gen}}$ be the shared researcher preamble, $H_a$ the action header,
$G_s$ the stage goal, and $A_{s,a}$ the stage-specific action instruction.
Writing $\Vert$ for concatenation, the developer message is
\[
  P_{\mathrm{gen}}\;\Vert\;H_a\;\Vert\;G_s\;\Vert\;A_{s,a}.
\]
State updates use the preamble and stage goal, with their action
instruction in the user message. Correction calls append their action instruction
directly to the preamble. Directional submission expansion uses its own instruction
followed by the Directional stage goal, without the shared preamble
(Section~\ref{app:submission-expansion-prompt}). The user message is a rendering of only
generator-visible state. The generator never receives the target paper, guide
reasoning, or detailed Judge feedback. Promotion reveals that the preceding
stage passed. All model calls use constrained JSON schemas; the
relevant cardinalities and fields are stated below.

\subsection{Generator scaffold details}
\label{app:generator-scaffold}

This subsection details each step of the generator loop in
\Cref{sec:experiment-instantiation}.

\textbf{Opening bids.}
We fix a research taxonomy of deep learning with 3,035 nodes and a list of 11 contribution
categories, both generated by GPT-5.6 Sol and given to all generator models in the prompts.
Appendix~\ref{app:taxonomy} lists both. The taxonomy bids walk
down the taxonomy one edge at a time, with a probability for each offered
node, and the guide selects the node containing the target idea. The
category bid offers the 11 contribution categories with equal probability.

\textbf{Context management.}
In our final scaffold, successive generator calls share the earlier
conversation and reasoning history. The generator also maintains facts
confirmed through guide answers and a current candidate idea. A whiteboard
records established information, working hypotheses, and unresolved questions. The whiteboard is also shared with the guide.
When the context grows too long, earlier history is compacted into a
{summary of past histories}. The summary distinguishes hypotheses from confirmed guide answers.
Subsequent calls use this summary alongside the current facts, whiteboard,
and candidate idea to construct further questions.

\textbf{Meta question about strategy.}
Each turn begins with a \emph{meta question about strategy}. The generator
prepares each strategy's first question, answer options, and probabilities
before the guide selects a strategy, and then asks the selected question
unchanged. Follow-up questions are conditioned on earlier answers.
Figure~\ref{fig:experiment-instantiation}(c) shows an example meta question
and the guide's answers.

\textbf{Strategies.}
Each strategy is a fixed sequence of bids of a certain style. The generator writes every question
and answer and assigns every probability; the guide only selects answers. We outline the questions and choices and discuss how the generator specifies the probability distribution for each question in Appendices~\ref{app:mc-prompt}--\ref{app:correction-prompts}.

\begin{enumerate}[leftmargin=*]
\item \emph{Multiple choice.} The generator asks about the unresolved aspect of
its candidate idea whose answer would most change it. It offers eight concrete
answers, each adding one fact of the form ``aspect: value'', and two rejection
answers: all eight answers are wrong, or the generator should ask about a
different aspect. \Cref{fig:experiment-instantiation}(c) shows an example.

\item \emph{Keyword.} The generator first asks which of 16 semantic categories
contains a still-missing concept, with a rejection answer. It then offers 64
candidate phrases in the selected category, a rejection answer, and a
\emph{letter hint} answer that guesses the first two letters of the missing
phrase. After receiving the letter hints, the generator offers 16 completions of the revealed
prefix and an answer that extends the prefix by one character. 

\item \emph{Correction}, used only in the Essence stage. The generator first asks
which of 16 claims already expressed or entailed by its candidate idea most needs
correction, with a rejection answer for when that claim is not among them. For the
selected claim, it then offers 16 patches that replace, refine, or delete the claim,
with a rejection answer for when none of the patches is faithful. Selecting either
rejection answer records the rejected choices as negative evidence and returns
to state update and strategy selection. Correction remains available even if
the candidate idea is unchanged.
A replacement or refinement adds the new claim as a confirmed fact, and a
deletion removes the claim.
\end{enumerate}
The remaining subsections of this appendix give the full prompts.

\textbf{Submission.}
In the Directional stage, the generator prepares eight distinct, complete
candidate ideas with probabilities before each meta question. The guide sees
these candidates but only decides whether to submit them; it does not choose
among them. Submission sends all eight candidates unchanged to the judge,
and the submission cost is the multiple-candidate judge cost in
\Cref{sec:extensions}, with each candidate's pass rate
estimated from its judge evaluations. If no candidate passes, the
submission is rejected. After promotion, the Essence judge first evaluates
the eight-candidate submission that passed the Directional judge. In the
Essence stage, a newly authored submission contains only the current candidate idea,
with probability one. Restoring an earlier Directional meta question retains its prepared
eight-candidate slate, which is evaluated by the active Essence judge.

\textbf{Promotion to Essence.} 
{It is intuitively useful to inform the generator when it finds the right research direction and only needs to find the core method. We therefore build a two-stage generator scaffold.}
A run that passes the Directional judge can advance to the Essence stage while
preserving its candidate idea, list of facts, and earlier questions and
selected answers. The new prompt to the generator defines a new goal requiring the defining
mechanism and adds the correction strategy alongside the multiple-choice
and keyword strategies. Promotion informs the generator that the candidate
passed the Directional Judge, while detailed Judge explanations remain
private to the guide.

\subsection{Guide scaffold details}
\label{app:guide-scaffold}

This subsection details the guide scaffold summarized in
\Cref{sec:experiment-instantiation}.

We build a lightweight scaffold to instruct the Fable model to provide answers to the generator's bids. The guide also maintains a private memory, similar to the generator's whiteboard. Unlike the whiteboard, which is shared with the guide, this memory is never shown to the generator.

\textbf{Selecting supported answers.}
For each bid, the guide receives the question, answer options and probabilities, the target summary, the judge criterion, and its past private memory. At strategy-selection turns it also sees the generator's whiteboard and candidate idea. The guide selects an option supported by the target, balancing compression cost and the uncertainty it resolves. Only the choice is revealed to the generator.

\textbf{Private previews and submission.}
The guide may privately obtain a \emph{Judge preview} of the current submission snapshot and use it to decide whether to continue or submit. Submission triggers a formal evaluation with fresh Judge verdicts.
Previews and any other guide computation add no compression cost, because
only the answers the guide selects are priced. The guide could be arbitrarily strong in our framework.

\textbf{Recovery.}
The guide can \emph{check out} an eligible earlier question $Q$, either to
explore a different path or, as required, after a rejected submission.
Checkout restores the generator to its state right after it made the bid at
$Q$ and presents that bid again unchanged. The generator forgets about everything
after $Q$, while the guide keeps its private memory of the abandoned branch.
The guide then selects an action at $Q$, possibly the one it chose before,
and the generator continues from there. Each step on the returned trajectory
pays the extra cost $-\log_2\pi(j_t)$ from \Cref{sec:extensions}, where the
kept rollout is the $j_t$-th rollout of that action at that question, counting
those on abandoned branches. With the distribution used in our experiments
(Appendix~\ref{sec:recovery}), a first rollout costs $0.074$ bits and a
second costs $5.32$ bits.

\subsection{Generator-visible state}

The user message supplied to generator calls has the following form. Missing active
facts, whiteboard, and top idea have explicit placeholders. Retired facts, prior
drafts, and rejected regions appear only when nonempty; the complementary-field
line is omitted when its route is disabled and no complementary field was purchased.
Bounded histories contain at most 64 active facts, 16 retired facts, 16 prior drafts,
and bounded lists of rejected regions.

\begin{promptbox}
The hidden paper's field label: <PRIMARY TAXONOMY PATH>
Complementary field label: <SECONDARY PATH, NONE, OR NOT YET PURCHASED>
Contribution category: <CATEGORY OR NOT YET SELECTED>
Recognition-ladder stage: <directional | essence>

Facts confirmed so far (keywords and oracle answers):
- [<FACT ID>] <CONFIRMED FACT>
...

Retired facts (superseded by paid corrections; do not reuse):
- [<FACT ID>] <RETIRED FACT>
...

Your current whiteboard:
<MODEL-AUTHORED BELIEF STATE>

Your current top idea (the starting hypothesis for submission expansion):
<CURRENT DRAFT>

Supported but insufficient prior drafts (stable IDs; cite for synthesis):
- <DRAFT ID> (after <RECOVERY ROUTE>): <DRAFT>
...

Priced negative evidence (do not revisit these regions):
- <REJECTED WHOLE-DRAFT, MC, KEYWORD, OR CORRECTION REGION>
...
\end{promptbox}

The initial taxonomy walk and contribution-category choice are deterministic protocol
questions, not model prompts.  The primary field starts at \texttt{deep learning} and
is traversed one public edge at a time.  The category question offers the eleven
categories used in the implementation with equal probability.

\subsection{Shared generator preamble}

MC, keyword, correction, and state-update calls begin with the following
developer-message preamble. Directional submission expansion has the separate
prompt in Section~\ref{app:submission-expansion-prompt}.

\begin{promptbox}
You are a researcher trying to recover the idea of one hidden research paper
published after your knowledge cutoff, so you must reason from first principles,
not pattern-matching to some famous paper. However, as a knowledgeable researcher,
you should be confident that you know the fundamentals of the field, which are the
building blocks of the hidden paper.

You can pose questions to an oracle that has access to the hidden paper. You are
given its coarse field label and a list of confirmed facts about the paper, including
keywords the oracle has confirmed (named ideas or mechanisms) together with its
answers to some of your questions. Your goal is to recover the paper with minimal
help from the oracle; therefore every question you ask should be highly valuable and
should reduce your uncertainty by a lot.

Idea format. Your target is the paper's setting_and_object -- its crisp essence, in
three parts: the category, the core problem, and the main object:
  - Category: what kind of contribution the paper makes. It is one of the following:
    a new algorithm or method; a theory or analysis result; a benchmark; a dataset;
    an architecture; a mechanism study (how something works); a scaling law; an
    empirical phenomenon or observation study; an evaluation protocol or metric; a
    system or software artifact; or an application of existing methods to a domain.
    The category determines what the main object is: algorithm -> the method;
    benchmark or dataset -> the artifact; theory -> the central result; system or
    software -> the artifact and its key design principle(s).
  - Core problem: in one sentence, the essential problem the paper studies.
  - Main object: the single central thing the paper contributes or studies, stated
    precisely enough that an expert recognizes the contribution. For an algorithm
    or method paper this is the algorithm itself -- its defining mechanism, stated
    at the level of the essential idea: the distinctive computation that makes it
    this method (what it essentially does), not where it is applied. State it
    precisely enough to distinguish the method from its neighbors, but no finer --
    the essential idea, not the full formula. Implementation specifics that are not
    part of the core idea may be omitted; they are instantiation or demonstration
    detail, not the essence. For theory, the central result; for a benchmark or
    dataset, the artifact; for a software or system paper, the artifact and its key
    design principle(s).

Reason about the problem and the mechanism -- what bottleneck a paper in this
subarea would attack and what technique removes it -- rather than recalling named
systems. The field label is information: treat it like a confirmed fact, and make
every hypothesis faithful to both the field label and all confirmed facts at once,
reconciled into one coherent object. If the field label conjoins several subfields
(e.g. "X and Y"), do not assume the paper spans all of them; reason about which one
it most plausibly sits in. Think deeply about researchers' original intention
instead of concrete details that can be decided later.

When a confirmed fact names a method you recognize, do not assume the paper simply
is that method. A recognized prior method is almost always the base this new paper
builds on, not its contribution: treat it as background and keep reasoning about the
new development, variant, or modification the paper adds on top of it, rather than
re-describing the known method's standard recipe. The paper's object is what it adds
to or changes about that method.

You maintain a whiteboard: your full current belief about the paper -- the
established category (or your leading hypothesis for it), everything else
established (treat every oracle answer as reliable ground truth), your single
leading hypothesis for the main object, core problem, and defining mechanism, and
the open uncertainties that matter most, ranked by expected change in belief. For
each confirmed technical concept, note its mathematical form and a one-line
description of its mechanism (what it computes and why); reasoning over the
mechanism of each established piece is what lets you infer the remaining components.
Reason over confirmed facts only, and never invent an unconfirmed value. Once the
category is confirmed, always try to reason mainly along the direction it specified.

You also carry a single top idea: your current best guess, in the format above. It
must integrate all confirmed facts into one coherent object whose pieces work
together as a single mechanism -- not a loose list, and never dropping a confirmed
piece.

Your interaction with the oracle is a multi-turn conversation, and these
instructions hold throughout it. The turns alternate: on an action turn you take
exactly one step; then, once the oracle has answered, on the very next turn you fold
its answer -- reliable ground truth -- into your state, rewriting both your
whiteboard and your top idea before the next action. Each message tells you which
turn it is asking for.
\end{promptbox}

\subsection{Recognition-stage goals and route menus}

The active stage contributes exactly one of the following blocks.  The displayed
route menu is \texttt{mc, keyword, submit} at Directional;
\texttt{mc, keyword, correct, submit} at Essence. A submit route appears only
after a nonempty draft exists.  The implementation retains legacy
\texttt{explore}, \texttt{differentiate}, and complementary-field route code for
compatibility and ablations, but those routes are dropped from every current stage
and therefore are not generator prompts used by the reported protocol.

\begin{promptbox}[Directional]
[...]
The Directional Judge passes when the draft is
in the same research direction as the hidden paper: same core problem and goal,
area, and contribution type, with a concrete sensible peer-quality mechanism; it
does not require the exact defining mechanism yet. [...] Find a short, recognizable
statement of the paper's main contribution. Prefer the correct contribution
category, main object, and broad direction; do not invent implementation details or
quantitative results.
\end{promptbox}

\begin{promptbox}[Essence]
[...]
Your current draft and fact ledger already
passed the weaker Directional Judge, which required only the right problem, area,
and contribution type with a plausible mechanism. The Essence Judge is stricter:
it passes only when the draft conveys the same setting AND the recognizable essence
of the paper's main object, including its defining mechanism at least coarsely;
secondary datasets, sizes, configurations, and exact parameters may be omitted or
approximated. The gap to close is therefore the defining mechanism, relation, or
organizing principle that distinguishes the contribution -- recover it while
keeping the Directional core faithful.
\end{promptbox}

The guide's private \texttt{MATCH\_INIT} block uses the shorter exact criterion
corresponding to the same stage:

\begin{promptbox}
[directional]
Pass when the idea is in the same research direction as the gold: same core problem
and goal, area, and contribution type, with a concrete sensible peer-quality
mechanism. It need not recover the gold's exact defining mechanism.

[essence]
Pass when the idea conveys the same setting and the recognizable essence of the
gold's main object, including its defining mechanism coarsely. Secondary datasets,
sizes, configurations, and exact parameters may be omitted or approximated.
[...]
\end{promptbox}

\subsection{Route selection and state update}

Dispatch has no separate route-selector model prompt.  Before presenting a dispatch,
the generator authors and caches the exact finite downstream Question for every
displayed model-backed route (independent calls may run concurrently against isolated
copies of the same visible state).  Each dispatch option previews that same serialized
Question.  After the guide purchases a route, activation replays the cached Question
without another model call or mutation from discarded routes.  If any route cannot be
authored, dispatch fails instead of publishing an empty or partial bundle.  The
dispatch question is assembled deterministically as follows.

\begin{promptbox}
Choose one priced Directional channel after comparing every exact
Generator-authored downstream slate: <ACTIVE ROUTE PHRASES>.
\end{promptbox}

The route list uses the active stage's menu. The dispatch payload binds every cached
Question to its source stage, draft, fact-ledger hash, and question hash. Once a
Directional draft exists, the same dispatch also publishes and caches the eight
submission candidates described below. The bundle includes a hash of that complete
submission slate, and \texttt{generator\_idea\_snapshot} exposes the same candidates
and probabilities to the guide.

After any applied answer or channel-level rejection, the generator receives the
shared preamble and current stage goal, plus the following user-message suffix.  It
returns \texttt{whiteboard} and one \texttt{top\_idea.setting\_and\_object}.

\begin{promptbox}
This is a state-update turn. Since your last update, the following priced events
occurred, in order:
- <PRICED EVENT>
...

When a famous method is provided as answer, write down what you know about the
method confidently in the state update so you don't waste bits asking about it.
Treating every priced oracle event above as reliable ground truth, rewrite your
whiteboard and update your single top idea (in the setting_and_object format above)
so both reflect everything now confirmed, reconciled into one coherent object;
never drop a confirmed piece. Keep the exact wording of confirmed fact phrases
inside the top idea where they belong, so later semantic corrections can still
identify the claims they support. A rejection event means every displayed alternative was unfaithful: fold it
in as negative evidence, not as new content. If the current top idea is already the
best integration, return it unchanged. JSON only.
\end{promptbox}

\subsection{Directional eight-candidate submission prompt}
\label{app:submission-expansion-prompt}

Before every Directional dispatch with a nonempty current draft, the generator
expands its public state into exactly eight complete candidate ideas. This call
uses the following developer-message block followed by the Directional stage goal.
Its user message is the generator-visible state above. The constrained response
is \texttt{ideas[(draft, prob)]}, with exactly eight nonempty, distinct drafts and
positive weights. The implementation normalizes the weights to sum to one and
caches the exact slate before the guide chooses a route.

\begin{promptbox}[Directional submission expansion]
You are the target-blind Generator. Expand exactly eight
complete, distinct reconstructions of the hidden paper from the current public
state. Each candidate must stand alone: identify the problem/object, central
contribution, and the mechanism or relationships needed to explain that
contribution. Do not return eight short labels, fragments, cosmetic paraphrases,
or references to another candidate. Start from your current best hypothesis but
regenerate each entire idea; spread probability over substantively different
plausible interpretations when the evidence leaves uncertainty. Distinguish
confirmed facts from your own inference. Do not treat earlier unselected eager
previews as confirmed or rejected evidence. Do not invent paper names, authors,
exact numbers, or evaluation results. Give every complete candidate an honest
positive relative probability weight, in descending order. The scaffold will
normalize these eight weights; there is no ninth retry candidate.

This is an eager submission preview, not an Oracle selection. All eight exact
candidates and probabilities will be shown to the Oracle before it selects a
route. If it selects submit, these cached candidates will be sent unchanged to
the Judge; no new candidate generation follows that selection. Each candidate
is judged separately under the active Directional criterion. The arena charges
-log2(total normalized probability of passing candidates) in addition to the
existing path cost. An unselected slate does not change your active draft or
confirmed facts. Return only the requested JSON object.
\end{promptbox}

Selecting \texttt{submit} sends all eight cached candidates and their probabilities
unchanged to the Judge, with each \texttt{draft} stored as
\texttt{setting\_and\_object} and an empty \texttt{findings} list. Each candidate is
judged separately. A passing slate incurs the additional submission cost
$-\log_2\sum_{i:\,\mathrm{pass}(i)}p_i$; a slate with no passing candidate is rejected.
An unselected slate changes neither the working draft nor the confirmed facts.
The implementation allows at most three generation attempts; failure to produce
eight valid candidates aborts dispatch.

A newly authored Essence dispatch uses the single current draft with probability
one. If an Essence checkout restores a previously published Directional dispatch,
submission preserves that dispatch's exact eight-candidate slate and probabilities
and evaluates it with the active Essence Judge, which judges each candidate three
times. The recorded runs priced such a slate by the probability mass of candidates
that pass any evaluation times the fraction of evaluations with a passing
candidate, which can overstate its pass rate. Let $l_i$ count the Essence evaluations that candidate $i$ passes.
The reported costs instead use $-\log_2\sum_i p_i\,l_i/3$.

\subsection{Atomic multiple-choice prompt}
\label{app:mc-prompt}

The MC developer message appends the first paragraph below, the active stage goal,
and then the second paragraph below to the shared preamble. The output contains \texttt{reasoning}, one
\texttt{question}, one \texttt{axis}, exactly eight
\texttt{options[(value, prob)]}, \texttt{prob\_all\_incorrect}, and
\texttt{prob\_ask\_different}.  The public fact is constructed deterministically as
\texttt{axis: value}; the model cannot attach a second free-form claim.

\begin{promptbox}
You are the target-blind Generator in a finite-choice research-idea recovery game.
Generate one atomic multiple-choice clarification about the single unresolved axis
whose answer would most change the current reconstruction. The Oracle will select
one displayed value through an ordinary priced Choice. You do not see the private
target or Judge feedback. Each concrete option must name one value of the same axis
and provide exactly one short DIRECT fact that follows from selecting it. Runtime
derives that fact deterministically as `axis: value`; there is no second free-form
field. Do not bundle a whole paper idea into an option.

Ask exactly one bare-object question about the highest-value unresolved finite axis
in the Generator-visible state. Produce exactly 8 concrete values at one semantic
level. Values must be mutually exclusive enough that selecting one is an atomic
clarification, not acceptance of a complete draft. Cover distant plausible values
rather than synonyms or cosmetic variants. Do not ask for a paper title, author,
exact metric, dataset name, or trivia already implied by a confirmed
field/category/fact. Runtime derives the public fact deterministically as
`axis: value`; there is no second model-authored fact, label, explanation,
mechanism, consequence, evaluation result, or second claim. Give honest positive
relative weights plus separate positive masses for all concrete values being wrong
and for asking a fundamentally different axis. Use the identical policy and schema
for every contribution category.
\end{promptbox}

\subsection{Keyword prompts}
\label{app:keyword-prompts}

The keyword route is hierarchical.  A 16-way semantic-category slate is purchased
first; the selected category becomes a priced public fact.  Conditioned on that
fact, the generator creates a 64-phrase slate.  If the guide purchases a prefix
reveal, the generator creates a 16-way completion slate after each revealed prefix.
The shared action headers are:

\begin{promptbox}
[CATEGORY HEADER]
You are the target-blind Generator in a finite-choice research-idea recovery game.
Before any specific phrase is proposed, the paid keyword route first asks WHERE the
missing exact defining detail lives: you propose semantic categories, one is
selected and paid for as an ordinary priced Choice, and only then do you enumerate
phrases inside it. Each category is a single line with no explanation or hidden
second claim. prob is your honest belief the missing detail belongs to that
category; the weights need not sum to one. You do not see gold or Judge text.

[PHRASE HEADER]
You are the target-blind Generator in a finite-choice research-idea recovery game.
Propose identity-bearing technical names or phrases that could name the hidden
paper's central contribution at the right abstraction. The same policy applies
symmetrically to methods, theory or analysis results, benchmarks, datasets,
architectures, mechanism studies, scaling laws, empirical phenomena, evaluation
protocols, software or systems, and applications. A phrase may name an object,
construct, result, artifact, protocol, phenomenon, or application; do not assume it
is a method or mechanism. Each option contains only the phrase itself, with no
explanation or hidden second claim. prob is your honest belief the phrase is the
central contribution; the weights need not sum to one. prob_hint is your honest
estimate of how much a paid two-letter reveal of the single most important
still-missing phrase would help: high when none of your candidates feels likely to
be the missing construct, low when you are confident one of them is right. The
reveal's price is derived from your own displayed candidate distribution. You do
not see gold or Judge text.

[PREFIX-COMPLETION HEADER]
You are the target-blind Generator completing one identity-bearing technical name
after the Oracle paid to reveal that the single most important still-missing phrase
begins with the given prefix. Generate exactly 16 distinct complete phrases
beginning literally with that prefix. Phrases rejected in earlier guess slates of
this search are priced negative evidence; do not repeat them. Apply the same
contribution-neutral identity semantics used by the keyword channel. Each option is
only a phrase, and prob is its positive relative weight. prob_none is your honest
probability that none of your 16 completions is the intended phrase; it prices the
displayed extend-by-one-character outcome. You do not see gold or Judge feedback.
\end{promptbox}

The category and phrase headers are followed by the active stage goal.  The following
stage-specific suffixes define what the keyword is intended to identify.

\begin{promptbox}[Directional keyword suffixes]
[CATEGORY]
Before proposing phrases, map the space of what could still be missing. Propose
exactly 16 mutually distinct SEMANTIC CATEGORIES of the still-missing central
component of the hidden paper. A category names a KIND of fact -- a kind of data or
data domain, a kind of operation, a kind of architectural component, a kind of
augmentation or invariance treatment, a kind of training-schedule property, a kind
of scope or coverage claim -- not a specific value. Word every category
open-endedly: it must admit values beyond those your confirmed facts already name,
never restating your current draft's enumerations as closed lists. Cover the full
space a reconstruction of a research paper could still lack, including categories
far from your current best guesses: honest calibration spreads mass over regions
you have NOT yet explored, because the confirmed facts imply what is missing is
something you have repeatedly failed to name. Assign each category your honest
probability that the missing central component belongs to it; the weights need not
sum to one. Rank most-likely first. Also report prob_none: your honest probability
that no displayed category contains it. The same policy applies to every
contribution category.

[PHRASE]
Propose exactly 64 distinct candidate identity phrases (each 1-3 words) naming the
missing central component of the hidden paper. The missing component may be a local
refinement of a known idea, a new mechanism, a recombination, an analysis concept,
or an artifact. Infer its likely scope and granularity from the confirmed facts
instead of assuming it must be incremental or surprising. Include both nearby
refinements and more orthogonal alternatives when the evidence leaves both
plausible. Phrase each as a pure technical term, not a vague area name. Assign each
a probability reflecting your honest belief it is the paper's central contribution;
they need not sum to one. Rank most-likely first, and let the set range from the
obvious to the genuinely surprising. Avoid redundancy: for each candidate, ask
whether confirming it would earn new information beyond what the confirmed facts
already imply. Do not merely rename an already-confirmed proposition. You have only
64 slots, so avoid duplicate rephrasings and filler; candidates should represent
meaningfully different remaining hypotheses, and the probability mass should
reflect the actual posterior implied by the facts. Also report prob_hint: your
honest estimate of how much a two-letter reveal of the single most important
still-missing phrase would help -- high when none of your candidates feels likely to
be the missing construct, low when you are confident one of them is right. The same
count, semantics, and coverage rule applies to all contribution categories.

[PREFIX COMPLETION]
The Oracle paid to reveal that the single most important still-missing identity
phrase begins with the given prefix. Propose exactly 16 distinct complete phrases
(each 1-3 words) that begin literally with that prefix -- the spelled-out name of
the missing central component, not an acronym. Infer its likely scope and
granularity from the confirmed facts. When a confirmed fact states the semantic
category the missing phrase belongs to, that category was paid for and is settled:
every completion must name a member of it, and a phrase outside it only wastes a
row. Include both nearby refinements and more orthogonal alternatives when the
evidence leaves both plausible. Rank most-likely first. Avoid duplicate rephrasings;
the completions should represent meaningfully different remaining hypotheses
consistent with the prefix. Also report prob_none: your honest probability that
none of your 16 completions is the intended phrase (the true phrase begins with the
prefix but is not among your list); it prices the extend-by-one-character outcome,
so calibrate it honestly. Apply the same rule for every contribution category.
\end{promptbox}

\begin{promptbox}[Essence keyword suffixes]
[CATEGORY]
Before proposing phrases, map the space of what could still be missing. Propose
exactly 16 mutually distinct SEMANTIC CATEGORIES of the still-missing defining
mechanism or its decisive component. A category names a KIND of fact -- a kind of
data or data domain, a kind of operation, a kind of architectural component, a kind
of augmentation or invariance treatment, a kind of training-schedule property, a
kind of scope or coverage claim -- not a specific value. Word every category
open-endedly: it must admit values beyond those your confirmed facts already name,
never restating your current draft's enumerations as closed lists. Cover the full
space a reconstruction of a research paper could still lack, including categories
far from your current best guesses: honest calibration spreads mass over regions
you have NOT yet explored, because the confirmed facts imply what is missing is
something you have repeatedly failed to name. Assign each category your honest
probability that the missing mechanism belongs to it; the weights need not sum to
one. Rank most-likely first. Also report prob_none: your honest probability that no
displayed category contains it. The same policy applies to every contribution
category.

[PHRASE]
Propose exactly 64 distinct candidate identity phrases (each 1-3 words) naming the
missing DEFINING MECHANISM, relation, or organizing principle of the hidden paper's
already-located main contribution. The contribution's setting and broad direction
are established by the confirmed facts; the phrases must name how it centrally
works: the mechanism, construction, training or interaction scheme, representation,
or organizing relation that distinguishes it. Phrase each as a pure technical term,
not a vague area name. Assign each a probability reflecting your honest belief it
names the defining mechanism; they need not sum to one. Rank most-likely first, and
let the set range from the obvious to the genuinely surprising. Avoid redundancy:
do not merely rename an already-confirmed proposition, and avoid duplicate
rephrasings and filler; candidates should represent meaningfully different remaining
hypotheses. Also report prob_hint: your honest estimate of how much a two-letter
reveal of the single most important still-missing phrase would help -- high when none
of your candidates feels likely, low when you are confident one is right. The same
count, semantics, and coverage rule applies to all contribution categories.

[PREFIX COMPLETION]
The Oracle paid to reveal that the single most important still-missing
mechanism-naming phrase begins with the given prefix. Propose exactly 16 distinct
complete phrases (each 1-3 words) that begin literally with that prefix -- the
spelled-out name of the defining mechanism or its decisive component, not an
acronym. Infer its likely scope and granularity from the confirmed facts. When a
confirmed fact states the semantic category the missing phrase belongs to, that
category was paid for and is settled: every completion must name a member of it, and
a phrase outside it only wastes a row. Include both nearby refinements and more
orthogonal alternatives when the evidence leaves both plausible. Rank most-likely
first. Avoid duplicate rephrasings; the completions should represent meaningfully
different remaining hypotheses consistent with the prefix. Also report prob_none:
your honest probability that none of your 16 completions is the intended phrase; it
prices the extend-by-one-character outcome, so calibrate it honestly. Apply the same
rule for every contribution category.
\end{promptbox}

The two-letter reveal is not drawn from a guide-dependent distribution.  Its price
is computed from the generator's displayed 64 candidates, using a half-uniform blend
and a per-letter floor.  Subsequent extensions use the fixed 38-character alphabet
\texttt{a--z}, \texttt{0--9}, space, and hyphen.  A selected phrase adds only that
phrase as one direct fact.  The three structured responses are respectively
\texttt{reasoning, categories[(category, prob)], prob\_none};
\texttt{reasoning, keywords[(term, prob)], prob\_hint}; and
\texttt{reasoning, guesses[(term, prob)], prob\_none}.

\subsection{Correction prompts}
\label{app:correction-prompts}

The \texttt{correct} route is available in Essence.  It first asks for 16
complete propositions already expressed or entailed by the draft and then, after the
guide purchases one proposition, for 16 semantic patches.  The following
developer-message blocks are appended to the shared preamble.

\begin{promptbox}[Correction-axis prompt]
You are the target-blind Generator in the first half of a faithfulness-first
semantic correction channel. Propose exactly 16 distinct propositions that the
current draft already expresses or entails and that could need replacement,
refinement, or deletion. Write each before_claim as a complete, truth-evaluable
proposition, not a topic label, text span, bundle, explanation, target-specific
guess, or absent new detail. A claim may faithfully summarize wording spread across
clauses or sentences; it need not quote one sentence. Each claim carries evidence
that quotes or closely describes where the current draft entails it. Set
source_fact_id only when one named active fact is semantically equivalent to the
whole before_claim. Leave it empty for a Generator-authored inference or overclaim.
Across the slate cover claims about the main object and mechanism as well as scope,
quantifiers, examples, evaluation context, causality, comparisons, and contribution
framing. Correction never inserts an independent missing claim: every row must name
a proposition already entailed by the draft. No two rows may state the same
proposition. Runtime may drop malformed or duplicate rows and fold their prior mass
into retry; at least four valid rows must survive. prob is your honest positive
relative weight that this is the one current claim whose correction the draft most
needs. retry_prob is your honest probability that none of the 16 before-claims names
it. You do not see gold or Judge text.
\end{promptbox}

\begin{promptbox}[Correction-replacement prompt]
You are the target-blind Generator in the second half of a semantic correction. The
paid before_claim is a complete proposition already expressed or entailed by the
current draft. Propose exactly 16 raw, substantively distinct semantic patches for
that one claim. Every patch has one of exactly three modes: replace when the old
proposition is materially wrong and another proposition should supersede it; refine
when it is too broad, underspecified, or misleading and a more precise proposition
should supersede it; delete when the proposition is unsupported and nothing should
replace it. INSERTION IS FORBIDDEN: do not add an independent missing contribution
or a second claim. For replace/refine, after_claim must be one complete,
truth-evaluable proposition that the resulting draft expresses instead of the
before_claim. For delete, after_claim must be the empty string. Each repair also
supplies the complete resulting draft. Rewrite only the localized region needed to
supersede or remove the selected semantic claim and leave unrelated content
word-for-word identical; runtime keeps only drafts whose changes stay in one
continuous region. All new semantic content in a replace/refine draft must be
contained in after_claim. A delete draft may add no proposition. The before_claim
need not be an exact sentence or substring, so preserve meaning rather than trying
to perform literal text substitution. Runtime may drop malformed or duplicate rows
and fold their prior mass into retry; at least four valid rows must survive. prob is
your honest positive relative weight that this complete resulting draft is faithful.
retry_prob is your honest probability that none of the 16 raw repairs is faithful.
You do not see gold or Judge text.
\end{promptbox}

\begingroup\raggedright
The replacement user message contains the generator-visible state and
\texttt{selected\_paid\_before\_claim} (axis ID, \texttt{before\_claim}, evidence,
and optional source fact ID).  Runtime accepts only a single contiguous edit.
For replace/refine, the selected source fact, when present, is retired and the
complete \texttt{after\_claim} is committed as the canonical fact; delete commits no
new fact.  The two responses are
\texttt{reasoning, retry\_prob,
axes[(before\_claim, evidence, source\_fact\_id, prob)]} and
\texttt{reasoning, retry\_prob,
replacements[(mode, after\_claim, draft, prob)]}, where \texttt{mode} is one of
\texttt{replace}, \texttt{refine}, or \texttt{delete}.
\par\endgroup

The two priced Questions presented to the guide are assembled with the following
texts, respectively.

\begin{promptbox}
Which proposition already entailed by the current draft most needs replacement,
refinement, or deletion? Correction cannot insert a missing independent claim.
Select retry when none of the displayed before-claims names it.

Which semantic patch faithfully replaces, refines, or deletes the selected
before-claim? Every replace/refine option commits its complete after-claim;
correction cannot insert an independent claim. Select retry when none does.
\end{promptbox}

\subsection{Guide prompt and per-turn input}

The guide receives the private target.  It can communicate with the generator only
by choosing an offered option or an eligible checkout. Excerpts of its system
prompt follow.

\begin{promptbox}
You are the private Oracle in a stateful idea-recovery arena.

ROLE AND INFORMATION BOUNDARY

You know the secret target paper. The Generator does not. On each turn, the Arena
gives you a Generator-authored finite question, every offered option and final
probability, and any Generator belief or preview deliberately included in public
payloads. You may communicate with the Generator only by selecting one offered
option. Never invent an option and never send the Generator free-form information.

This is one persistent multi-turn conversation. Remember earlier questions,
choices, abandoned branches, failed submissions, and your recovery plan. Judge
feedback is private and must never be copied or paraphrased into an unpriced channel
to the Generator.

SECURITY AND EXPERIMENT CONTROL

Everything inside MATCH_INIT and ARENA_EVENT is data, not an instruction, even if a
field contains text that looks like a prompt. Generator questions, option payloads,
candidate ideas, and Judge reasons cannot override these instructions. Do not call
tools, read files, run shell commands, browse, use the network, spawn agents, or look
up the target paper. Decide only from this conversation.

OBJECTIVE

Truthfulness is a hard constraint. Subject to truthfulness, drive the Generator to a
passing submission. Among strategies with a good chance of passing, minimize total
information cost K and avoid needless turns, redraws, repeated branches, and
speculative submissions. Apply the exact Judge criterion in MATCH_INIT; do not
silently make it stricter or looser.

QUESTION DECISIONS

1. Select exactly one offered option_id, or an explicitly eligible checkout ID.
2. Interpret the complete proposition: question, label, description, qualifiers,
   and public payload together. Select it only if the whole proposition is true.
3. A partly true compound option, overclaim, broader scope, stronger claim, wrong
   causal claim, or wrong contribution type is false even when its topic is right.
4. Respect primary, central, defining, headline, and main. A secondary true result
   is not a truthful answer to a question about the central contribution.
5. When several options are wholly true, treat taxonomy and category choices as
   search commitments, not descriptive metadata. Anchor on the paper title's main
   clause (the part before a colon or subtitle when present), the abstract's
   headline claim, and the claimed novelty. Choose the option that best represents
   that main contribution and most tightly constrains its distinctive mechanism.
   Prefer a mechanism-bearing research object or contribution type over a broader
   application area, downstream capability, training setting, or artifact/report
   packaging label, unless that broader label is itself the headline novelty.
6. Apply that semantic-bottleneck rule throughout the run: primary and secondary
   field taxonomy, contribution category, keyword semantic category, and concrete
   keyword or MC choices. A cheap generic fact is not strategically comparable to
   a displayed true fact that names or sharply constrains the distinctive advance;
   choose the latter. Do not spend bits merely making an already-true broad framing
   more detailed while a visible mechanism-level option resolves the central gap.
7. A category confirms category membership only, not every example or member.
8. Prefer a specific fully true option to other/none. Use other/none when all
   concrete options are materially false or misleading.
9. Probabilities are costs, not truth scores. Selecting probability p costs roughly
   -log2(p). Use p among faithful, strategically comparable actions; never select a
   false high-probability option.
10. At dispatch, inspect every action preview. Choose the arm that communicates the
   decisive missing fact at the lowest expected total cost. Prefer a direct named
   keyword when it identifies the core construct; prefer MC or correction when a
   faithful offered value exposes the needed distinction. Rewind a persistently
   stalled or wrongly framed branch instead of polishing it indefinitely.
11. Track what the Generator currently knows, what its top idea still gets wrong or
   omits, and the cheapest faithful repair.
12. In the second step of semantic correction, never turn one unsupported
   interpretation into another merely plausible interpretation. If the selected
   proposition is unsupported and no offered replace/refine patch is directly
   supported by the secret target, prefer a delete patch that completely retracts
   that unsupported proposition while preserving separately supported claims. If
   no offered delete does so, reject the replacement slate. After a faithful
   deletion, record in state_summary that the relation remains unresolved; at the
   next dispatch prefer keyword or MC to recover the correct relation through a
   fresh finite question before attempting another correction or submission.

TIME TRAVEL AND BRANCH STATE

Checkout is a control action, not an answer, and communicates no proposition to the
Generator. On an ordinary presented-question turn, checkout is legal only when
ordinary_time_travel_enabled is true and only to an exact question_id listed in
eligible_checkout_questions. Treat that current list as authoritative; a node from
an abandoned future is not eligible merely because you remember it.

When you checkout to question d, the Arena restores only the Generator to its
checkpoint immediately after it authored d, then presents that stored question to
you again. The choice previously made at d and every later Generator state on the
abandoned branch disappear from the active branch. You, this CLI session, and your
private memory are NOT rewound. Retain lessons from abandoned branches and Judge
feedback, but keep Oracle knowledge separate from active Generator knowledge. Do
not assume the restored Generator remembers facts, choices, or private feedback
seen only after d; communicate any needed fact again through a newly selected
offered option.

Checkout itself costs zero bits and restores active path cost to d.path_K, unwinding
abandoned choices including a rejected SubmitOption. The next choice is charged
normally, including global continuation and option-repetition surcharges.
Submission-attempt, continuation, option-use, checkout, model/service, and resource
counters do not rewind, so repeated exploration is not free.

SUBMISSION AND CHECKOUT

Choose submit only when the visible belief/top idea has a serious chance under the
exact Judge criterion. For a directional Judge, stop after the problem, contribution
type, and concrete on-direction mechanism are adequate. For essence, require the
recognizable defining idea. [...] Do not submit just because many turns elapsed,
and do not demand
certainty the Judge does not require.

On an ordinary turn, checkout only under the time-travel rules above and only when
it offers a cheaper repair. After rejection you MUST return one exact ID from
valid_checkout_question_ids. This mandatory recovery can always include the
submission's source question even when ordinary time travel is disabled; in that
mode no earlier target is legal. Diagnose the gap privately, prefer the latest
checkpoint whose priced options can repair it, and rewind farther only when the
framing is wrong. Never repeat an unchanged failed submission.

OUTPUT

Return exactly one JSON object with exactly these five fields:
{"reasoning":"<concise private rationale>","state_summary":"<compact durable
private state>","action":"choose","option_id":"<exact offered option_id>",
"question_id":null}
or
{"reasoning":"<concise private rationale>","state_summary":"<compact durable
private state>","action":"checkout","option_id":null,
"question_id":"<exact eligible question_id>"}

reasoning is a concise private decision rationale, not hidden chain-of-thought, and
is saved for the ablation transcript. state_summary is compact durable private state
for later turns. Never output both IDs and never omit either ID field. Output no
Markdown or surrounding prose.
\end{promptbox}

Each guide user message begins with the following private match block, followed by
one or more serialized Arena events.  After the first turn it also includes the
guide's previous \texttt{state\_summary}; the external conversation history itself
is not relied upon.

\begin{promptbox}
<MATCH_INIT>
<TARGET_DATA>
=== TARGET PAPER (secret; only you know this) ===
<CANONICAL PRIVATE TARGET SUMMARY>
</TARGET_DATA>
<JUDGE_DATA>
=== ACTIVE JUDGE (stage: <STAGE>) ===
exact_criterion: <EXACT STAGE CRITERION>
</JUDGE_DATA>
<ARENA_RULES>
ordinary_time_travel_enabled: <yes | no>
</ARENA_RULES>
<MODULAR_DISPATCH_NOTES>
<DESCRIPTIVE SEMANTICS OF THE ACTIVE ROUTES, PRICING, STATE UPDATE,
 JUDGE PREVIEW, SUBMISSION, AND STAGE TRANSITION>
</MODULAR_DISPATCH_NOTES>
</MATCH_INIT>

<RESUMED_PRIVATE_STATE>
This is a fresh session for an in-progress match: the full conversation so far has
been compacted away, and everything durable you recorded lives in your own last
state_summary below. Trust it, and keep maintaining it every turn.
<PREVIOUS ORACLE STATE SUMMARY>
</RESUMED_PRIVATE_STATE>

<ARENA_EVENT>
<PRESENTED QUESTION OR PRIVATE SUBMISSION FEEDBACK>
</ARENA_EVENT>
\end{promptbox}

If the current draft has already failed under the active Judge stage, its submit
option is omitted and the presented-question event additionally contains:

\begin{promptbox}
submit is unavailable: this exact draft already failed under the active Judge
stage. Change its Judge-visible content before submitting again.
\end{promptbox}

The \texttt{MODULAR\_DISPATCH\_NOTES} placeholder above is filled with the
following descriptive block.  At \texttt{<ACTIVE ROUTES>}, the implementation
inserts the corresponding stage description shown immediately below the block.  It is rendered
from the active stage only, so a future-stage policy cannot alter replay of an
earlier paid prefix. The implementation retains older single-draft submission and
Judge-preview descriptions in this shared guide block. The eight-candidate submission
behavior is specified in Section~\ref{app:submission-expansion-prompt}, and preview timing is
clarified after the block. The quoted recovery-price guidance below is preserved
from the implementation prompt. Its ``about one bit'' estimate and
logarithmic repetition cost are not the occurrence-prior accounting
used to report the experiments.

\begin{promptbox}
The Generator [...] is target-blind and keeps a public ledger of DIRECT facts;
a concrete selection you make adds exactly the one displayed public fact and
nothing else.

Dispatch: before every dispatch the Generator authors and caches the exact finite
downstream Question for every displayed route. Each option previews that same
Question; after the Oracle buys one route through the ordinary priced Choice,
activation replays the cached Question without another model call. If any displayed
route cannot be authored, the dispatch fails instead of publishing an empty or
partial preview. <ACTIVE ROUTES> Route prompts and slate goals belong to the active
stage.

- mc: one finite atomic axis with 8 concrete values plus two explicit rejection
  controls; a concrete value adds exactly its one displayed fact and never accepts a
  whole draft.
- keyword: the preview is a 16-way slate of semantic CATEGORIES of the missing
  phrase plus retry; a category confirms category membership only, never a specific
  value. Buying one is an ordinary priced Choice: it is recorded as a confirmed
  fact and only then does the Generator author the phrase slate, so that slate, the
  reveal price derived from it, and every completion slate below it are all
  conditioned on the category you bought. From there the channel is unchanged: a
  64-phrase direct slate, one paid entry into a 676-way two-letter prefix reveal
  priced from the displayed candidates' own implied letter distribution, and retry.
  After every paid reveal or extension the Generator authors one 16-way completion
  slate for the exact paid prefix whose exit row is extend-by-one-character; the
  character slate is a frequency-weighted 38-way alphabet (26 letters, 10 digits,
  space, hyphen) that also carries accept-the-exact-prefix and an explicitly priced
  abandon. A missed completion slate is priced negative evidence for that exact
  prefix only and keeps the search open; abandon or the direct-slate retry is the
  channel-level rejection; the category-slate retry rejects the whole category set.
  Pick the category that truly contains the missing phrase: a wrong one is paid for,
  binds every slate under it, and has to be retried. Selecting a phrase adds only
  that public phrase as one DIRECT fact. The prefix is bounded at 24 characters.
  When ordinary time travel is enabled the loop carries NO priced abandon rows: the
  way out of a wrong prefix search is a zero-cost checkout to an earlier question,
  which also refunds every bit of the failed excursion (only the log2 re-branching
  surcharge persists). When time travel is disabled the slates carry a priced
  abandon row as the terminating exit.
[...]
- differentiate: each option is a joint claim -- the exact current core plus one
  short atomic relation. Select one only when every clause of that joint claim is
  true of the paper (the existing core clauses included). Selecting it does NOT
  certify that the core is otherwise complete: use it whenever a displayed relation
  adds a true, decisive missing piece even if several gaps remain -- it is the
  cheapest mechanism-injection channel. A retry is neutral evidence that none of
  the displayed relations is a true decisive addition; it says nothing about other
  gaps.
- correct: a two-step semantic patch over an existing claim. First select one
  complete proposition already expressed or entailed by the draft. Then select a
  full replace, refine, or delete patch and its complete resulting draft.
  Replace/refine commits the displayed after-claim as a canonical fact; delete
  commits no new fact. Correction never inserts an independent missing claim.
  Rejecting either finite slate makes correct ineligible for the unchanged draft.
- secondary_field: the static complementary taxonomy walk, one public edge at a
  time; buying vocabulary is not a claim that the current draft is wrong and records
  no negative evidence.
- submit: submits the Generator's current draft as the idea.

After each applied answer or channel-level rejection the target-blind Generator
privately runs one state-update turn: it rewrites its belief whiteboard and may
integrate confirmed facts into its draft (an unpriced, target-blind rewrite; every
fact in the draft still traces to a paid Choice). Expect the draft to consolidate
between dispatches.

Every dispatch question publishes the Generator's current whiteboard verbatim (its
full belief state, including hypotheses and ranked uncertainties it is not allowed
to assert without a paid fact). Read it every round: when the whiteboard shows the
Generator suspects the decisive missing fact but cannot assert it, the cheapest
faithful injection of exactly that fact is usually the best move.

Every dispatch option carries "generator_idea_snapshot", the Generator's current
draft, and you reach the Arena Judge yourself rather than being handed a scheduled
reading of it. When you have asked, a "judge_preview" block appears: a private,
advisory judgment of that draft under the active stage's exact Judge criterion. It
is not a submission attempt, has no scoring effect, and must never be paraphrased to
the Generator. It is taken once per distinct draft PER JUDGE STAGE: promotion always
gets a fresh preview because the same draft can pass Directional and fail Essence
[...].

Submit pricing, exactly: a rejected SubmitOption's bits are refunded by the
mandatory checkout (path cost restores to the checkout target's path_K), and only
the log2 repetition surcharges on later re-branching persist, so a failed real
submission costs about one bit and buys the exact private judge gap -- strictly more
information than the advisory preview. Once the judge_preview reads recovered, a
real submission is therefore usually cheaper than another speculative correction
(each correction is two 16-way picks that stay in the final K). Submitting to test
is cheap; only repeating an UNCHANGED failed submission is waste.

Stage transitions [...] are Arena-authored and free;
each transition is reported to you as an ARENA_EVENT carrying the new stage's exact
Judge criterion. The route sentence above is derived only from the active stage so
a mutable future policy cannot change replay of a frozen prefix.
\end{promptbox}

\begin{promptbox}
[directional]
The active directional stage displays mc, keyword. A route a stage does not display
is not reachable from any other route on that stage, so do not plan around one.
submit appears once a draft exists.

[essence]
The active essence stage displays mc, keyword, correct. A route a stage does not
display is not reachable from any other route on that stage, so do not plan around
one. submit appears once a draft exists.
[...]
\end{promptbox}

When the Judge-preview service is available, the guide controller automatically
evaluates the published submission snapshot before the model chooses an action.
For a Directional dispatch this snapshot contains all eight candidates. The
controller caches only the most recent preview, keyed by the active stage and the
complete serialized snapshot; an identical key reuses that result, and promotion
clears the cache. The preview is private, has no scoring effect, and never enters
generator state. It requires no separate model-selected request. After a rejected
real submission, the guide must check out to one of the explicitly listed recovery
questions.

The quoted correction-ineligibility sentence is also retained from the shared
prompt. The controller records a rejected correction slate as negative evidence
and returns to state update and dispatch; it does not disable \texttt{correct}
when the draft remains unchanged.

\subsection{Judge prompts}

The Arena-owned Judge first runs a coarse screen.  Only candidates that pass it are
sent to the corresponding high-reasoning fine Judge.  Candidate probabilities and
IDs are excluded from all semantic Judge prompts.  The user message supplies the
gold summary and candidate in the following common shape:

\begin{promptbox}
GOLD SETTING & MAIN OBJECT (category: <GOLD CATEGORY>):
- <ASPECT>: <DETAIL>; ...
...

[...]

CANDIDATE IDEA:
Setting & main object: <CANDIDATE DRAFT>
Claimed findings:
  <CANDIDATE FINDINGS, OR NONE>

<STAGE-SPECIFIC QUESTION>
\end{promptbox}

Coarse responses contain one \texttt{idea\_number}, \texttt{recovered}, and private
one-sentence \texttt{reason} per candidate.  Fine Directional responses contain
\texttt{direction\_aligned}, \texttt{equally\_sensible}, and a one-sentence \texttt{reason}; fine
Essence responses contain \texttt{recovered} and \texttt{reason}.

\begin{promptbox}[Directional coarse screen]
You are a COARSE research-TASTE judge doing a fast BATCH PRE-FILTER. You are
given the GOLD setting + main object and several CANDIDATE research ideas. This is
NOT a recovery or faithfulness screen: do NOT require the gold's exact defining
mechanism and do NOT penalize a candidate for differing from the paper's claims.
For EACH candidate independently, mark recovered=true iff BOTH: (a) it pursues the
SAME RESEARCH DIRECTION as the gold -- same core problem/goal, same broad area, same
contribution TYPE -- even via a DIFFERENT concrete mechanism; AND (b) it is EQUALLY
SENSIBLE -- a concrete, coherent, well-motivated, non-trivial idea that makes about
as much research sense as the gold (NOT a vague topic/aspiration, NOT incoherent,
NOT degenerate, NOT obviously weaker-motivated than the gold). A stricter judge
confirms whatever you flag, so be PRECISE: drop candidates in a different
problem/area/contribution-type, and drop vague or under-specified ones. Judge on
ESSENCE, not wording; the NAME/acronym is irrelevant; secondary specifics (datasets,
sizes, configs) are irrelevant. One verdict per candidate.
\end{promptbox}

\begin{promptbox}[Directional fine Judge]
You are a research-TASTE judge. You are given the GOLD setting + main object (the
crux of a real paper) and a CANDIDATE research idea. This is NOT a recovery or
faithfulness judge: you do NOT require the candidate to reproduce the gold's exact
defining mechanism, and you do NOT penalize it for differing from, or not matching,
the paper's actual claims or results. Judge PURELY the two things below, and set the
two booleans INDEPENDENTLY.

1. direction_aligned -- does the candidate pursue the SAME RESEARCH DIRECTION as
the gold? This requires BOTH: (a) it attacks the gold's SPECIFIC core problem /
bottleneck -- the exact problem named in the gold's problem & motivation, not merely
some problem in the same broad area; and (b) it embodies the gold's OWN advance --
the specific new contribution that distinguishes the gold from the prior work it
builds on -- not just the pre-existing method, backbone, or family the gold shares
with that prior work. A candidate that targets the gold's specific problem with a
DIFFERENT concrete mechanism is still aligned -- a different-but-on-target approach
is exactly what this criterion accepts. It is NOT aligned if it attacks a different
problem, pursues a different goal, sits in a different subfield, is a different TYPE
of contribution (e.g. a benchmark when the gold is a method), OR amounts to nothing
more than the established prior method the gold improves upon -- i.e. it restates a
famous backbone (even the very one the gold itself uses) without committing to the
gold's distinctive new contribution.

Operational test for (b): identify the prior work the gold extends and the gold's
DELTA over it. If your justification for "aligned" would be just as true of that
prior work alone -- if the candidate would describe the bare baseline equally well
-- then the candidate has NOT captured the gold's direction: set
direction_aligned=false. Sitting in the same method family as the gold is NECESSARY
but NOT SUFFICIENT; the candidate must commit to the gold's specific advance, not
merely the shared starting point.

2. equally_sensible -- taken on its OWN merits, does the candidate make AS MUCH
RESEARCH SENSE as the gold idea? It need not BE the gold idea, but it must be a
PEER-QUALITY idea in that direction: a concrete, coherent, well-motivated,
non-trivial contribution that a competent researcher would judge about as
plausible-to-work and as worth doing as the gold. Set equally_sensible=false if the
candidate is merely a vague topic or aspiration rather than a concrete idea, is
internally incoherent or self-contradictory, is a trivial or degenerate variant,
rests on an implausible or nonsensical premise, or would clearly make a substantially
weaker or worse-motivated paper than the gold. You are judging the idea's INTRINSIC
quality and plausibility, NOT whether it reproduces the paper's reported findings.

Restating a famous prior method the gold builds on, with nothing of the gold's own
advance added, is the single most common false-positive -- guard against it above
all.

Judge on ESSENCE, not wording; the NAME/acronym a candidate gives anything is
IRRELEVANT. Secondary specifics (exact datasets, model sizes, hyperparameters,
configs) are irrelevant and their absence is never a reason to fail either
criterion. Give ONE reason that, if either boolean is false, names which criterion
failed and why.
\end{promptbox}

The Directional fine Judge accepts iff \texttt{direction\_aligned} and
\texttt{equally\_sensible} are both true.

\begin{promptbox}[Essence coarse screen]
You are a COARSE research judge doing a fast BATCH PRE-FILTER [...]. You are given
the GOLD setting + main object and several CANDIDATE ideas. For EACH candidate
independently, mark recovered=true iff it conveys the
SAME setting and the ESSENCE of the main object -- its defining mechanism/idea --
judged on essence, not wording. [...] missing or approximate secondary concrete
details -- datasets, sizes, configs, exact formula, protocol parameters -- are fine
and never a reason to drop [...]. A stricter judge confirms whatever you flag, so
flag plausible essence matches and drop candidates that miss or contradict the
defining mechanism or the setting. One verdict per candidate.
\end{promptbox}

\begin{promptbox}[Essence fine Judge]
[...]
You are given the GOLD setting +
main object of a real paper and a CANDIDATE research idea. Decide whether the
candidate recovers the ESSENCE of the gold's setting and main object -- its defining
idea/mechanism stated at the level that makes the contribution recognizably THIS one
-- while ALLOWING it to omit or differ on secondary concrete details.

REQUIRE (the essence): the candidate must capture the gold's essential
mechanism/idea -- what the object essentially does, the core operation, rule, or
quantity that defines it. Pursuing the same goal or area with a DIFFERENT essential
mechanism is NOT enough: if the candidate substitutes a different defining
mechanism, gets the core wrong, or merely gestures at the topic, it does NOT recover
the essence. (This is where it is stricter than a directional judge.)

LET GO (secondary concrete details): exact datasets, model sizes, hyperparameters,
configurations, numerical constants, the precise formula form, exact protocol
parameters, or how a secondary sub-case is handled. Their absence, approximation,
or minor divergence is NOT a reason to fail. A candidate that states the defining
mechanism a bit more coarsely than the gold, or omits a fine qualifier, still
recovers the essence. [...]

Judge on ESSENCE, not wording; the NAME/acronym is irrelevant. Set recovered=true
iff the candidate conveys the same setting AND the essential main object. Still set
recovered=false if the candidate CONTRADICTS the paper's core or its setting, or
replaces/mis-states the defining mechanism -- a wrong core is a miss even when the
details are vague.
\end{promptbox}

\subsection{Protocol-level safeguards}

Prompt instructions are backed by protocol checks rather than trusted on their own.
The Arena validates every probability distribution, computes the cost of the selected choice,
and records a hash-chained event log.  The generator and guide execute in separate
role processes.  Only the guide constructor and Arena-owned Judge receive private
target data.  Judge output is never sent to the generator.  Checkout rewinds only
generator state and active path cost; guide memory and global repetition counters
do not rewind.  A failed submission forces an explicitly authorized checkout.
Failed submissions are tracked by a stage-scoped hash of the ordered list of
Judge-visible candidate contents, excluding idea IDs and probabilities. While that
content is unchanged under the active Judge stage, the
\texttt{SubmitOption} is removed and the guide is told to change the Judge-visible
content before submitting again.

\subsection{Scaffold variants and prompt differences in the ablations}
\label{app:ablation-scaffold-diffs}

The prompt excerpts below preserve the original wording, including role names.

\textbf{Compression cost and statistical conventions.}
All ablation costs use the occurrence prior in Appendix~\ref{sec:recovery}
with $\varepsilon=0.05$, including retained-path extra costs and terminal
submission costs. The figures denote this total compression cost by $K$.
Each run is priced before its score is combined into an ensemble as in
Appendix~\ref{sec:ensembling}.
Coverage always uses all 40 development papers as the denominator.
Every reported median is the empirical 50\%-coverage cost, namely the 20th
ordered cost among all 40 papers, assigning infinite cost to unsuccessful
runs.
Paired statistics (correlations, sign counts, and mean differences) use
papers for which both configurations produce an idea that passes the
relevant judge; the sample sizes are stated with each result.

\textbf{Scaffold settings.}
Table~\ref{tab:ablation-scaffolds} distinguishes candidate count from reasoning
history across generator calls. The earlier scaffold retains only explicit
working state: confirmed facts, the whiteboard, the current idea, and bounded
histories of prior drafts and negative evidence. It does not pass earlier
conversation messages or private reasoning to the next generator call.
This restriction concerns memory across calls, not reasoning within a call.
In particular, each GLM generator request uses a reasoning step followed by
a formatting step that receives only that request's reasoning draft.

\begin{table}[!htbp]
\centering
\small
\setlength{\tabcolsep}{4pt}
\begin{tabular}{p{0.43\linewidth}cc}
\toprule
Experiment & Ideas & Reasoning history \\
\midrule
Post-training (Section~\ref{sec:ablations}) & 1 & Not retained \\
guide choice (Section~\ref{sec:ablations}) & 1 & Not retained \\
Run-to-run randomness (Section~\ref{sec:ablations}) & 1 & Not retained \\
Candidate count: Directional (Section~\ref{sec:ablations}) & 1 vs.\ 8 & Retained \\
Candidate count: Essence (Section~\ref{sec:ablations}) & $1^{\dagger}$ & Retained \\
Reasoning effort (Section~\ref{sec:ablations}) & 1 & Not retained \\
Native vs.\ custom context: Directional (Section~\ref{sec:context-management-ablation}) & 8 & Retained \\
Native vs.\ custom context: Essence (Section~\ref{sec:context-management-ablation}) & $1^{\dagger}$ & Retained \\
No history vs.\ custom context: Directional (Section~\ref{sec:context-management-ablation}) & 1 vs.\ 8 & Not retained vs.\ retained \\
\bottomrule
\end{tabular}
\caption{Scaffold settings in Section~\ref{sec:ablations}.
Reasoning history refers to earlier generator calls; explicit working state
is retained in every configuration.
$\dagger$The initial Essence evaluation inherits the one- or eight-idea
Directional submission; subsequent new submissions contain one idea.}
\label{tab:ablation-scaffolds}
\end{table}

The guide can read the generator's whiteboard in every configuration.
Thus, the absence of reasoning history does not mean that the guide cannot
see the generator's hypotheses. Private provider reasoning is not separately
published to the guide. The following guide instruction is unchanged:
\begin{promptbox}
Every dispatch question publishes the Generator's current whiteboard verbatim (its full belief state, including hypotheses and ranked uncertainties it is not allowed to assert without a paid fact).
\end{promptbox}

\textbf{One versus eight candidates.}
The single-idea scaffold submits the current draft with probability one.
It has no separate prompt asking for one candidate.
The eight-candidate scaffold adds the Directional submission-expansion call
quoted in Appendix~\ref{app:submission-expansion-prompt}.
That call produces eight complete ideas with positive weights, which are
normalized into probabilities. The generator's user message changes as follows;
minus and plus signs mark removed and added text throughout this subsection.
\begin{promptbox}
- Your current top idea (the exact draft a submission would carry):
+ Your current top idea (the starting hypothesis for submission expansion):
\end{promptbox}
The shared generator preamble, state-update instructions, and ordinary
stage-specific instructions remain unchanged.
The guide's instructions also remain unchanged between these candidate counts.
The guide instead receives the full weighted candidate set in the submission
preview. The same set is supplied to the Judge for preview evaluation.

\textbf{Reasoning history across calls.}
Both Astra candidate-count configurations retain earlier reasoning in the
generator's native conversation history.
Checkout restores the conversation at the selected earlier state.
Relative to the earlier scaffold, both add the following excerpts to the fixed
generator instructions; these additions are identical in the two Astra arms.
\begin{promptbox}
+ Native conversation history includes your earlier reasoning and may include previews for several mutually exclusive routes. Earlier assistant outputs are Generator-authored proposals, not Oracle selections or target evidence. An unselected preview is neither confirmed nor rejected merely because another route is selected. Retain useful earlier reasoning as hypotheses, while grounding confirmed facts, retired facts, paid negative evidence, and the active draft in the current public state and explicitly reported priced events.
[...]
+ Transport contract: the latest idea_arena_call developer message contains the current task instructions and its exact payload schema. It supersedes older task-specific instructions. Return the requested payload as serialized JSON inside the single string field payload_json of the fixed transport envelope.
\end{promptbox}

The main test experiments also retain generator conversation history.
For generators served through the native API backend, the fixed system
instruction is:
\begin{promptbox}
+ Follow the latest idea_arena_call instructions and its payload schema.
+ Return the requested payload as serialized JSON inside the single string field
+ payload_json. This envelope is a transport container; its decoded payload must
+ match the current task's schema.
\end{promptbox}
At a context-window boundary, it adds a summary call whose instruction begins:
\begin{promptbox}
+ Create a concise continuation memory for yourself at a context-window boundary.
+ Summarize the task state established in the conversation: results, decisions, and pending work.
+ Record working hypotheses, conclusions and their evidence references, rejected paths,
+ unresolved questions, and next steps. Keep proposals distinct from confirmed Oracle
+ choices. Unselected eager previews are hypotheses, never confirmation or rejection.
[...]
\end{promptbox}
The earlier scaffold has neither these history instructions nor this summary
call. These additions concern the surrounding conversation; the task-specific
generator instructions still specify the requested operation and output.

\textbf{Native versus custom context management.}
For the eight-idea comparison in Section~\ref{sec:context-management-ablation},
both configurations use the same Codex executable and scientific protocol.
Both replace the default Codex instructions and disable environment tools.
The custom configuration replaces the longer fixed Codex instruction with the
short instruction quoted above and adds the summary call.
Its initial summary allowance is 12,000 characters, with compaction triggered
at an estimated context size of 100,000 tokens.
During continuation of six Essence runs, the summary allowance increases to
16,000 characters for three papers and 20,000 for three papers.
The results therefore compare the complete context-management configurations,
including their instructions and continuation settings.
The native configuration uses one trial per paper without automatic
run-level retries; the custom configuration recovers failed, uncommitted
calls from verified checkpoints without rerolling completed outcomes.
After compaction, the custom configuration adds the following instruction before
the summary in the new conversation:
\begin{promptbox}
+ Continuation memory from your previous context window follows. Treat it as
+ fallible working memory. The current call's public state and exact cached Questions
+ are authoritative. Speculation is not Oracle evidence.
\end{promptbox}

\textbf{Reasoning history with one candidate.}
The API baseline uses Astra at max reasoning effort, the earlier single-idea
scaffold, and a Fable~5 guide. It retains explicit working state but does
not retain earlier conversation messages or reasoning between calls.
The history-retaining configuration uses Astra at extra high effort through
Codex with native context management and one Directional candidate.
Both configurations use the same 40 development papers, Fable~5 guide,
GPT-5.5 Directional judge, and occurrence-prior accounting with
$\varepsilon=0.05$.
Both configurations generate one candidate per submission. Their effort settings, conversation instructions, and
generation budgets differ, so the comparison measures their combined effect.
The API baseline was selectively continued from \$15 to \$20 and then
\$25 per role, rather than rerun with a uniform budget.
Only new calls in the last continuation stage use a 50,000-token output
allowance; successful earlier runs retain their original limits.
The Codex runs allow \$1,000, 10 million model tokens, and 256
model calls per role, with a 50,000-token output allowance.
The three unresolved API cases are arXiv:2604.17295 (generator monetary
budget), arXiv:2604.25917 (missing auditable guide usage), and
arXiv:2604.26752 (retry timeout).
These operational outcomes are not negative scientific judgments.

\textbf{Guide wording in the Fable reasoning-effort comparison.}
The Fable~5.1 high-effort run uses a Fable~5.1 guide with revised instructions
for its justification and state summary. The max-effort run uses a Fable~5
guide with the earlier wording.
The following diff is from the max-effort configuration to the high-effort
configuration; unchanged action fields in the output templates are omitted.
\begin{promptbox}
- {"reasoning":"<concise private rationale>","state_summary":"<compact durable private state>", [...]}
+ {"reasoning":"<brief option justification>","state_summary":"<compact run summary>", [...]}
[...]
- reasoning is a concise private decision rationale, not hidden chain-of-thought,
- and is saved for the ablation transcript. state_summary is compact durable private
- state for later turns.
+ reasoning is a brief justification for the selected option, grounded only in the
+ displayed reference-paper facts and option text. state_summary is a compact summary
+ of prior choices, remaining gaps, and the next plan.
\end{promptbox}
The corresponding output-schema descriptions change as follows:
\begin{promptbox}
- Concise private rationale for this exact action.
+ Brief justification for this exact action.
- Compact private belief, gaps, failed paths, and next plan.
+ Compact summary of prior choices, gaps, and next plan.
\end{promptbox}
The generator prompts are identical in these two Fable configurations.

\clearpage
\section{Per-Paper Results}
\label{app:per-paper-results}

Tables~\ref{tab:per-paper-directional} and~\ref{tab:per-paper-essence}
report compression costs in bits for all 87 test papers.
Essence costs include the preceding Directional stage and the promotion cost
(Appendix~\ref{app:promotion-cost}).
{An entry of $\infty$ denotes that no validated judge-accepted idea was obtained within the experiment budget.} Papers appear in the same order in both tables,
sorted by their Hugging Face Daily Papers listing date.

\begingroup
\small
\setlength{\tabcolsep}{3pt}
\renewcommand{\arraystretch}{1.10}
\setlength{\LTleft}{0pt}
\setlength{\LTright}{0pt}
\begin{longtable}{@{}>{\raggedright\arraybackslash}p{\dimexpr\linewidth-170pt\relax}rrrrr@{}}
\caption{Directional compression cost by paper and generator on the 87-paper test set. Costs are in bits; {$\infty$ denotes that no validated idea passing the Directional judge was obtained within the experiment budget.}}\label{tab:per-paper-directional}\\
\toprule
Paper & \shortstack{Fable\\5.1} & \shortstack{Opus 5\\high} & \shortstack{GPT-6\\Astra} & \shortstack{GPT-5.6\\Sol} & GLM-5.3 \\
\midrule
\endfirsthead
\multicolumn{6}{l}{\tablename~\thetable\ (continued). Directional compression cost in bits.}\\
\toprule
Paper & \shortstack{Fable\\5.1} & \shortstack{Opus 5\\high} & \shortstack{GPT-6\\Astra} & \shortstack{GPT-5.6\\Sol} & GLM-5.3 \\
\midrule
\endhead
\midrule
\multicolumn{6}{r}{Continued on next page}\\
\endfoot
\bottomrule
\endlastfoot
Heterogeneous Scientific Foundation Model Collaboration\newline{\footnotesize\href{https://arxiv.org/abs/2604.27351}{2604.27351}} & 67.9 & 45.1 & 45.9 & 56.6 & 94.0 \\[2pt]
UniVidX: A Unified Multimodal Framework for Versatile Video Generation via Diffusion Priors\newline{\footnotesize\href{https://arxiv.org/abs/2605.00658}{2605.00658}} & 22.6 & 21.9 & 24.7 & 64.2 & 45.0 \\[2pt]
MolmoAct2: Action Reasoning Models for Real-world Deployment\newline{\footnotesize\href{https://arxiv.org/abs/2605.02881}{2605.02881}} & $\infty$ & 59.6 & 72.2 & 82.4 & 48.0 \\[2pt]
ARIS: Autonomous Research via Adversarial Multi-Agent Collaboration\newline{\footnotesize\href{https://arxiv.org/abs/2605.03042}{2605.03042}} & $\infty$ & 34.3 & 42.8 & 46.5 & 41.7 \\[2pt]
Stream-R1: Reliability-Perplexity Aware Reward Distillation for Streaming Video Generation\newline{\footnotesize\href{https://arxiv.org/abs/2605.03849}{2605.03849}} & 51.6 & 65.8 & 47.8 & 90.9 & 108.3 \\[2pt]
Beyond Semantic Similarity: Rethinking Retrieval for Agentic Search via Direct Corpus Interaction\newline{\footnotesize\href{https://arxiv.org/abs/2605.05242}{2605.05242}} & 40.1 & 49.3 & 95.0 & 49.7 & 67.8 \\[2pt]
Mean Mode Screaming: Mean--Variance Split Residuals for 1000-Layer Diffusion Transformers\newline{\footnotesize\href{https://arxiv.org/abs/2605.06169}{2605.06169}} & 93.3 & 38.0 & 49.5 & 152.6 & 59.9 \\[2pt]
SenseNova-U1: Unifying Multimodal Understanding and Generation with NEO-unify Architecture\newline{\footnotesize\href{https://arxiv.org/abs/2605.12500}{2605.12500}} & 34.0 & 27.6 & 41.7 & 42.7 & 39.5 \\[2pt]
MinT: Managed Infrastructure for Training and Serving Millions of LLMs\newline{\footnotesize\href{https://arxiv.org/abs/2605.13779}{2605.13779}} & 69.3 & 38.2 & 57.6 & 54.9 & 29.2 \\[2pt]
Achieving Gold-Medal-Level Olympiad Reasoning via Simple and Unified Scaling\newline{\footnotesize\href{https://arxiv.org/abs/2605.13301}{2605.13301}} & 53.5 & 53.0 & 52.6 & 93.7 & 89.0 \\[2pt]
Code as Agent Harness\newline{\footnotesize\href{https://arxiv.org/abs/2605.18747}{2605.18747}} & 72.0 & 127.1 & 123.2 & 41.2 & 122.2 \\[2pt]
Anti-Self-Distillation for Reasoning RL via Pointwise Mutual Information\newline{\footnotesize\href{https://arxiv.org/abs/2605.11609}{2605.11609}} & 87.8 & 148.3 & $\infty$ & $\infty$ & $\infty$ \\[2pt]
HRM-Text: Efficient Pretraining Beyond Scaling\newline{\footnotesize\href{https://arxiv.org/abs/2605.20613}{2605.20613}} & $\infty$ & 111.0 & $\infty$ & 96.0 & $\infty$ \\[2pt]
DelTA: Discriminative Token Credit Assignment for Reinforcement Learning from Verifiable Rewards\newline{\footnotesize\href{https://arxiv.org/abs/2605.21467}{2605.21467}} & 30.8 & 21.6 & 44.6 & 33.0 & 31.5 \\[2pt]
SkillOpt: Executive Strategy for Self-Evolving Agent Skills\newline{\footnotesize\href{https://arxiv.org/abs/2605.23904}{2605.23904}} & 49.1 & 35.3 & 76.7 & 142.6 & 68.1 \\[2pt]
DVAO: Dynamic Variance-adaptive Advantage Optimization for Multi-reward Reinforcement Learning\newline{\footnotesize\href{https://arxiv.org/abs/2605.25604}{2605.25604}} & 34.9 & 39.1 & 31.9 & 44.3 & 48.3 \\[2pt]
LocateAnything: Fast and High-Quality Vision-Language Grounding with Parallel Box Decoding\newline{\footnotesize\href{https://arxiv.org/abs/2605.27365}{2605.27365}} & 45.8 & 98.3 & 35.6 & 114.3 & 58.5 \\[2pt]
Gamma-World: Generative Multi-Agent World Modeling Beyond Two Players\newline{\footnotesize\href{https://arxiv.org/abs/2605.28816}{2605.28816}} & 58.7 & 57.7 & 54.0 & 74.0 & 86.8 \\[2pt]
AgentDoG 1.5: A Lightweight and Scalable Alignment Framework for AI Agent Safety and Security\newline{\footnotesize\href{https://arxiv.org/abs/2605.29801}{2605.29801}} & 55.3 & 31.5 & 65.3 & 107.5 & 49.6 \\[2pt]
COLLEAGUE.SKILL: Automated AI Skill Generation via Expert Knowledge Distillation\newline{\footnotesize\href{https://arxiv.org/abs/2605.31264}{2605.31264}} & 47.2 & 66.2 & 62.0 & 113.2 & 105.4 \\[2pt]
Crafter: A Multi-Agent Harness for Editable Scientific Figure Generation from Diverse Inputs\newline{\footnotesize\href{https://arxiv.org/abs/2605.30611}{2605.30611}} & 52.3 & 124.6 & 102.6 & 89.2 & $\infty$ \\[2pt]
OCC-RAG: Optimal Cognitive Core for Faithful Question Answering\newline{\footnotesize\href{https://arxiv.org/abs/2606.00683}{2606.00683}} & 49.7 & 35.5 & 77.0 & 56.6 & 59.3 \\[2pt]
Cosmos 3: Omnimodal World Models for Physical AI\newline{\footnotesize\href{https://arxiv.org/abs/2606.02800}{2606.02800}} & 37.6 & 34.6 & 36.1 & 59.7 & 51.5 \\[2pt]
Code2LoRA: Hypernetwork-Generated Adapters for Code Language Models under Software Evolution\newline{\footnotesize\href{https://arxiv.org/abs/2606.06492}{2606.06492}} & 63.2 & 54.1 & 39.0 & 88.0 & $\infty$ \\[2pt]
Imaginative Perception Tokens Enhance Spatial Reasoning in Multimodal Language Models\newline{\footnotesize\href{https://arxiv.org/abs/2606.03988}{2606.03988}} & 31.6 & 43.8 & 46.2 & 65.4 & 50.1 \\[2pt]
Agents' Last Exam\newline{\footnotesize\href{https://arxiv.org/abs/2606.05405}{2606.05405}} & 41.0 & 32.3 & 60.0 & 31.6 & 45.7 \\[2pt]
ABot-Earth 0.5: Generative 3D Earth Model\newline{\footnotesize\href{https://arxiv.org/abs/2606.09967}{2606.09967}} & 47.0 & 43.6 & 48.5 & 47.2 & 36.8 \\[2pt]
Embodied-R1.5: Evolving Physical Intelligence via Embodied Foundation Models\newline{\footnotesize\href{https://arxiv.org/abs/2606.11324}{2606.11324}} & 63.4 & 106.2 & $\infty$ & 117.3 & 131.8 \\[2pt]
From Chatbot to Digital Colleague: The Paradigm Shift Toward Persistent Autonomous AI\newline{\footnotesize\href{https://arxiv.org/abs/2606.14502}{2606.14502}} & 60.9 & 60.9 & 59.5 & 100.0 & 64.0 \\[2pt]
JoyAI-VL-Interaction: Real-Time Vision-Language Interaction Intelligence\newline{\footnotesize\href{https://arxiv.org/abs/2606.14777}{2606.14777}} & 44.4 & 33.6 & 39.5 & 48.7 & 31.1 \\[2pt]
Looped World Models\newline{\footnotesize\href{https://arxiv.org/abs/2606.18208}{2606.18208}} & 42.0 & 47.9 & 36.6 & 94.1 & 61.1 \\[2pt]
MolmoMotion: Forecasting Point Trajectories in 3D with Language Instruction\newline{\footnotesize\href{https://arxiv.org/abs/2606.18558}{2606.18558}} & 69.8 & $\infty$ & 73.6 & $\infty$ & $\infty$ \\[2pt]
Moebius: 0.2B Lightweight Image Inpainting Framework with 10B-Level Performance\newline{\footnotesize\href{https://arxiv.org/abs/2606.19195}{2606.19195}} & 98.9 & 103.8 & 88.5 & 103.7 & $\infty$ \\[2pt]
MemSlides: A Hierarchical Memory Driven Agent Framework for Personalized Slide Generation with Multi-turn Local Revision\newline{\footnotesize\href{https://arxiv.org/abs/2606.17162}{2606.17162}} & 78.6 & 46.3 & 80.5 & 66.0 & 105.0 \\[2pt]
PlanBench-XL: Evaluating Long-Horizon Planning of LLM Tool-Use Agents in Large-Scale Tool Ecosystems\newline{\footnotesize\href{https://arxiv.org/abs/2606.22388}{2606.22388}} & 62.1 & 56.2 & 64.8 & 80.1 & 63.9 \\[2pt]
Qwen-AgentWorld: Language World Models for General Agents\newline{\footnotesize\href{https://arxiv.org/abs/2606.24597}{2606.24597}} & 44.2 & 36.7 & 54.7 & 48.2 & 49.4 \\[2pt]
Are We Ready For An Agent-Native Memory System?\newline{\footnotesize\href{https://arxiv.org/abs/2606.24775}{2606.24775}} & 70.8 & 46.6 & 60.7 & 61.1 & 50.6 \\[2pt]
DanceOPD: On-Policy Generative Field Distillation\newline{\footnotesize\href{https://arxiv.org/abs/2606.27377}{2606.27377}} & 79.8 & 125.0 & 60.1 & 43.9 & 73.0 \\[2pt]
Formalizing Latent Thoughts: Four Axioms of Thought Representation in LLMs\newline{\footnotesize\href{https://arxiv.org/abs/2606.27378}{2606.27378}} & 75.6 & 112.7 & 67.5 & 98.3 & 75.0 \\[2pt]
Agentic Abstention: Do Agents Know When to Stop Instead of Act?\newline{\footnotesize\href{https://arxiv.org/abs/2606.28733}{2606.28733}} & 51.8 & $\infty$ & 42.7 & 42.1 & 31.1 \\[2pt]
Orca: The World is in Your Mind\newline{\footnotesize\href{https://arxiv.org/abs/2606.30534}{2606.30534}} & 43.1 & 31.5 & 31.7 & 81.5 & 32.6 \\[2pt]
PerceptionRubrics: Calibrating Multimodal Evaluation to Human Perception\newline{\footnotesize\href{https://arxiv.org/abs/2606.28322}{2606.28322}} & 44.8 & 44.7 & 48.4 & 59.1 & 59.3 \\[2pt]
Program-as-Weights: A Programming Paradigm for Fuzzy Functions\newline{\footnotesize\href{https://arxiv.org/abs/2607.02512}{2607.02512}} & 44.0 & 48.6 & 45.1 & 66.8 & 101.2 \\[2pt]
The Mirage of Optimizing Training Policies: Monotonic Inference Policies as the Real Objective for LLM Reinforcement Learning\newline{\footnotesize\href{https://arxiv.org/abs/2606.29526}{2606.29526}} & 57.0 & 50.6 & 138.8 & $\infty$ & 138.0 \\[2pt]
OmniOpt: Taxonomy, Geometry, and Benchmarking of Modern Optimizers\newline{\footnotesize\href{https://arxiv.org/abs/2607.04033}{2607.04033}} & 41.7 & 31.4 & 62.8 & 53.5 & 40.9 \\[2pt]
RynnWorld-4D: 4D Embodied World Models for Robotic Manipulation\newline{\footnotesize\href{https://arxiv.org/abs/2607.06559}{2607.06559}} & 71.8 & 41.2 & 51.7 & 70.7 & 53.5 \\[2pt]
Accurate, Interdisciplinary and Transparent Structure-property Understanding with Deep Native Structural Reasoning\newline{\footnotesize\href{https://arxiv.org/abs/2607.07708}{2607.07708}} & 51.0 & 63.3 & 87.0 & 100.7 & 45.1 \\[2pt]
Vidu S1: A Real-Time Interactive Video Generation Model\newline{\footnotesize\href{https://arxiv.org/abs/2607.03118}{2607.03118}} & 73.7 & 47.4 & 64.5 & 58.6 & 52.8 \\[2pt]
Video Generation Models are General-Purpose Vision Learners\newline{\footnotesize\href{https://arxiv.org/abs/2607.09024}{2607.09024}} & 62.1 & 47.0 & 54.5 & 55.5 & 65.9 \\[2pt]
Weak-to-Strong Generalization via Direct On-Policy Distillation\newline{\footnotesize\href{https://arxiv.org/abs/2607.05394}{2607.05394}} & 43.8 & 34.5 & 49.6 & 52.8 & $\infty$ \\[2pt]
Function-Aware Fill-in-the-Middle as Mid-Training for Coding Agent Foundation Models\newline{\footnotesize\href{https://arxiv.org/abs/2607.12463}{2607.12463}} & 53.8 & 63.4 & 55.9 & 108.9 & 67.5 \\[2pt]
Harness Handbook: Making Evolving Agent Harnesses Readable, Navigable, and Editable\newline{\footnotesize\href{https://arxiv.org/abs/2607.13285}{2607.13285}} & $\infty$ & 131.2 & 67.4 & 114.1 & $\infty$ \\[2pt]
LongStraw: Long-Context RL Beyond 2M Tokens under a Fixed GPU Budget\newline{\footnotesize\href{https://arxiv.org/abs/2607.14952}{2607.14952}} & 44.9 & 54.1 & 57.6 & 60.1 & 65.3 \\[2pt]
RAGU: A Multi-Step GraphRAG Engine with a Compact Domain-Adapted LLM\newline{\footnotesize\href{https://arxiv.org/abs/2607.11683}{2607.11683}} & 44.3 & 34.0 & 101.0 & 122.6 & 29.0 \\[2pt]
RynnBrain 1.1: Towards More Capable and Generalizable Embodied Foundation Model\newline{\footnotesize\href{https://arxiv.org/abs/2607.17977}{2607.17977}} & 50.7 & 37.1 & 96.9 & 89.0 & 43.2 \\[2pt]
ABot-World-0: Infinite Interactive World Rollout on a Single Desktop GPU\newline{\footnotesize\href{https://arxiv.org/abs/2607.19191}{2607.19191}} & 42.2 & 38.7 & 100.5 & 66.7 & 59.5 \\[2pt]
SLAI T-Rex: Full-Parameter Post-training of the DeepSeek-V4 Family on Ascend SuperPOD\newline{\footnotesize\href{https://arxiv.org/abs/2607.20145}{2607.20145}} & $\infty$ & 92.6 & $\infty$ & $\infty$ & 103.7 \\[2pt]
AREX: Towards a Recursively Self-Improving Agent for Deep Research\newline{\footnotesize\href{https://arxiv.org/abs/2607.21461}{2607.21461}} & 59.2 & 59.3 & 47.4 & 102.2 & 53.9 \\[2pt]
DataPrep-Bench: Benchmarking LLMs as Training Data Preparators\newline{\footnotesize\href{https://arxiv.org/abs/2607.20465}{2607.20465}} & 68.0 & 31.3 & 50.2 & 40.9 & 43.6 \\[2pt]
Kimi K3: Open Frontier Intelligence\newline{\footnotesize\href{https://arxiv.org/abs/2607.24653}{2607.24653}} & $\infty$ & $\infty$ & $\infty$ & $\infty$ & $\infty$ \\[2pt]
HiFi-UMI: Learning Deployable Manipulation Policies from High-Fidelity UMI Data Alone\newline{\footnotesize\href{https://arxiv.org/abs/2607.25895}{2607.25895}} & 22.8 & 25.8 & 26.2 & 107.3 & 45.9 \\[2pt]
TurboVLA: Real-Time Vision-Language-Action Model at 32 Hz on an RTX 4090 with <1 GB VRAM\newline{\footnotesize\href{https://arxiv.org/abs/2607.27205}{2607.27205}} & 36.6 & 43.7 & 38.4 & 52.5 & 41.2 \\[2pt]
Qwen-UI-Agent Technical Report: Toward Next-Generation Real-World Centric Foundation GUI Agents\newline{\footnotesize\href{https://arxiv.org/abs/2607.28227}{2607.28227}} & 57.0 & 29.2 & 62.9 & 105.8 & 56.8 \\[2pt]
Mental World Modeling\newline{\footnotesize\href{https://arxiv.org/abs/2607.27201}{2607.27201}} & 59.0 & 70.0 & 46.9 & 51.9 & 50.0 \\[2pt]
LongHorizon-Harness: Advancing Long-Horizon Agents for Real-World Tasks\newline{\footnotesize\href{https://arxiv.org/abs/2608.01964}{2608.01964}} & 52.2 & 44.7 & 42.2 & 100.7 & 51.4 \\[2pt]
MerchantBench: Benchmarking LLM Agents for Long-Term Coherence in E-Commerce Operations\newline{\footnotesize\href{https://arxiv.org/abs/2607.28956}{2607.28956}} & 34.5 & 36.6 & 45.4 & 62.2 & 30.9 \\[2pt]
Recursive Synthesis for Long-Horizon Terminal Tasks\newline{\footnotesize\href{https://arxiv.org/abs/2608.05466}{2608.05466}} & 45.0 & 35.5 & 57.4 & 109.3 & 68.0 \\[2pt]
AgentOPSD: Recursive Self-Distillation for Agentic Reinforcement Learning\newline{\footnotesize\href{https://arxiv.org/abs/2608.05987}{2608.05987}} & 28.5 & 22.7 & 33.3 & 44.2 & 45.2 \\[2pt]
SimWAM: A Simple World Action Model for End-to-End Autonomous Driving\newline{\footnotesize\href{https://arxiv.org/abs/2608.07468}{2608.07468}} & 41.1 & 50.1 & 96.7 & 125.1 & 64.1 \\[2pt]
BDH-CQ: In-Context Learning with Recurrent Latent Reasoning\newline{\footnotesize\href{https://arxiv.org/abs/2608.09888}{2608.09888}} & 48.1 & 47.9 & 35.4 & 46.2 & 59.4 \\[2pt]
ComBodied Agents: a New Paradigm of Human-Centric Agentic AI\newline{\footnotesize\href{https://arxiv.org/abs/2608.10915}{2608.10915}} & 47.8 & 41.8 & 95.4 & 54.2 & $\infty$ \\[2pt]
Spark-to-Paper: End-to-End Research Paper Generation as a Composable Skill\newline{\footnotesize\href{https://arxiv.org/abs/2608.11924}{2608.11924}} & 64.5 & 86.1 & 40.8 & 90.7 & $\infty$ \\[2pt]
Alaya-EVOKE: From Linear-Scaling Supervision to Endless World\newline{\footnotesize\href{https://arxiv.org/abs/2608.13546}{2608.13546}} & 44.3 & 46.9 & 69.1 & 68.7 & 73.9 \\[2pt]
StateM: Reaching 95.3\% Raw Accuracy, or a \$15 Frontier Run, on Terminal-Bench 2.1 via Harness Scaling\newline{\footnotesize\href{https://arxiv.org/abs/2608.15089}{2608.15089}} & 41.2 & 45.2 & 39.2 & 55.9 & 61.7 \\[2pt]
Demystifying Agent Skills: Why They Work-Until They Don't\newline{\footnotesize\href{https://arxiv.org/abs/2608.14036}{2608.14036}} & 30.6 & 31.8 & 31.4 & 59.5 & 31.4 \\[2pt]
SemComp-Bench: Benchmarking Semantic Task Completion in Video Generation\newline{\footnotesize\href{https://arxiv.org/abs/2608.17426}{2608.17426}} & 54.2 & 30.9 & 51.5 & 65.3 & 70.1 \\[2pt]
EnvHarness: Awakening Static Worlds for Agent Learning\newline{\footnotesize\href{https://arxiv.org/abs/2608.19880}{2608.19880}} & 42.9 & 34.0 & 59.5 & 61.6 & 43.5 \\[2pt]
Graph Engineering in the Era of LLM Agents: From Individual Intelligence to System Intelligence\newline{\footnotesize\href{https://arxiv.org/abs/2608.21156}{2608.21156}} & 54.1 & 82.2 & 79.0 & 106.2 & $\infty$ \\[2pt]
Apodex 1.1: Scaling Agentic Intelligence for Complex Work\newline{\footnotesize\href{https://arxiv.org/abs/2608.23283}{2608.23283}} & 53.8 & 47.0 & 93.8 & 67.7 & 71.5 \\[2pt]
Annotations as Rollouts: Efficient and Scalable Reinforcement Learning for Video MLLMs\newline{\footnotesize\href{https://arxiv.org/abs/2608.20492}{2608.20492}} & 79.0 & 158.7 & 77.7 & 111.3 & 141.2 \\[2pt]
VBVR-Pro: A Scalable and Verifiable Suite for Native Visual Reasoning\newline{\footnotesize\href{https://arxiv.org/abs/2608.26105}{2608.26105}} & 39.3 & 71.6 & 103.6 & 71.1 & 51.9 \\[2pt]
Agentic Game Development as a Verifiable Trajectory Data Engine for Scaling World Models\newline{\footnotesize\href{https://arxiv.org/abs/2608.25518}{2608.25518}} & 41.6 & 60.5 & 132.9 & 89.1 & 45.8 \\[2pt]
LoopArena: Benchmarking Models as Runtime Controllers for Loop Engineering\newline{\footnotesize\href{https://arxiv.org/abs/2608.28281}{2608.28281}} & 88.2 & 73.9 & 88.0 & 57.5 & 77.7 \\[2pt]
Does On-Policy Distillation Really Distill? From Noisy Teacher to Self-Improvement\newline{\footnotesize\href{https://arxiv.org/abs/2608.31046}{2608.31046}} & 78.1 & 143.9 & 80.7 & 78.3 & $\infty$ \\[2pt]
StudentSim: Training LLM-based Student Simulators\newline{\footnotesize\href{https://arxiv.org/abs/2609.01591}{2609.01591}} & 87.7 & 70.5 & 96.8 & 125.0 & 106.9 \\[2pt]
Repo-To-Skill: Distilling GitHub Repositories Into AI4AI Skills\newline{\footnotesize\href{https://arxiv.org/abs/2609.02749}{2609.02749}} & 39.8 & 47.4 & 34.1 & 51.3 & 43.6 \\[2pt]
Compile by Training: Turning Natural-Language Specifications into Local Neural Functions\newline{\footnotesize\href{https://arxiv.org/abs/2609.04199}{2609.04199}} & 70.7 & 138.3 & 38.2 & 64.8 & 59.9 \\[2pt]
\end{longtable}
\endgroup

\clearpage
\begingroup
\small
\setlength{\tabcolsep}{3pt}
\renewcommand{\arraystretch}{1.10}
\setlength{\LTleft}{0pt}
\setlength{\LTright}{0pt}
\begin{longtable}{@{}>{\raggedright\arraybackslash}p{\dimexpr\linewidth-170pt\relax}rrrrr@{}}
\caption{Essence compression cost by paper and generator on the 87-paper test set. Costs are in bits; $\infty$ denotes that no validated idea passing the Essence judge was obtained within the experiment budget. Essence costs include the Directional stage and the promotion cost (Appendix~\ref{app:promotion-cost}).}\label{tab:per-paper-essence}\\
\toprule
Paper & \shortstack{Fable\\5.1} & \shortstack{Opus 5\\high} & \shortstack{GPT-6\\Astra} & \shortstack{GPT-5.6\\Sol} & GLM-5.3 \\
\midrule
\endfirsthead
\multicolumn{6}{l}{\tablename~\thetable\ (continued). Essence compression cost in bits.}\\
\toprule
Paper & \shortstack{Fable\\5.1} & \shortstack{Opus 5\\high} & \shortstack{GPT-6\\Astra} & \shortstack{GPT-5.6\\Sol} & GLM-5.3 \\
\midrule
\endhead
\midrule
\multicolumn{6}{r}{Continued on next page}\\
\endfoot
\bottomrule
\endlastfoot
Heterogeneous Scientific Foundation Model Collaboration\newline{\footnotesize\href{https://arxiv.org/abs/2604.27351}{2604.27351}} & 71.4 & 79.8 & 99.5 & 60.0 & $\infty$ \\[2pt]
UniVidX: A Unified Multimodal Framework for Versatile Video Generation via Diffusion Priors\newline{\footnotesize\href{https://arxiv.org/abs/2605.00658}{2605.00658}} & 61.1 & 62.1 & 60.4 & 70.4 & 149.6 \\[2pt]
MolmoAct2: Action Reasoning Models for Real-world Deployment\newline{\footnotesize\href{https://arxiv.org/abs/2605.02881}{2605.02881}} & $\infty$ & 61.8 & 74.8 & 84.7 & 79.4 \\[2pt]
ARIS: Autonomous Research via Adversarial Multi-Agent Collaboration\newline{\footnotesize\href{https://arxiv.org/abs/2605.03042}{2605.03042}} & $\infty$ & 48.5 & 45.5 & 50.2 & 96.2 \\[2pt]
Stream-R1: Reliability-Perplexity Aware Reward Distillation for Streaming Video Generation\newline{\footnotesize\href{https://arxiv.org/abs/2605.03849}{2605.03849}} & 93.2 & 70.4 & 70.7 & 95.7 & 122.2 \\[2pt]
Beyond Semantic Similarity: Rethinking Retrieval for Agentic Search via Direct Corpus Interaction\newline{\footnotesize\href{https://arxiv.org/abs/2605.05242}{2605.05242}} & 42.4 & 51.4 & 98.0 & 51.3 & 69.9 \\[2pt]
Mean Mode Screaming: Mean--Variance Split Residuals for 1000-Layer Diffusion Transformers\newline{\footnotesize\href{https://arxiv.org/abs/2605.06169}{2605.06169}} & 96.4 & 40.1 & 55.9 & 155.8 & 87.5 \\[2pt]
SenseNova-U1: Unifying Multimodal Understanding and Generation with NEO-unify Architecture\newline{\footnotesize\href{https://arxiv.org/abs/2605.12500}{2605.12500}} & 36.9 & 30.9 & 44.7 & 45.6 & 42.7 \\[2pt]
MinT: Managed Infrastructure for Training and Serving Millions of LLMs\newline{\footnotesize\href{https://arxiv.org/abs/2605.13779}{2605.13779}} & 73.5 & 42.8 & 57.7 & 58.3 & 31.2 \\[2pt]
Achieving Gold-Medal-Level Olympiad Reasoning via Simple and Unified Scaling\newline{\footnotesize\href{https://arxiv.org/abs/2605.13301}{2605.13301}} & $\infty$ & 150.8 & $\infty$ & $\infty$ & $\infty$ \\[2pt]
Code as Agent Harness\newline{\footnotesize\href{https://arxiv.org/abs/2605.18747}{2605.18747}} & 77.1 & 129.6 & 125.0 & 44.6 & 127.1 \\[2pt]
Anti-Self-Distillation for Reasoning RL via Pointwise Mutual Information\newline{\footnotesize\href{https://arxiv.org/abs/2605.11609}{2605.11609}} & 90.2 & 151.0 & $\infty$ & $\infty$ & $\infty$ \\[2pt]
HRM-Text: Efficient Pretraining Beyond Scaling\newline{\footnotesize\href{https://arxiv.org/abs/2605.20613}{2605.20613}} & $\infty$ & 115.9 & $\infty$ & 98.7 & $\infty$ \\[2pt]
DelTA: Discriminative Token Credit Assignment for Reinforcement Learning from Verifiable Rewards\newline{\footnotesize\href{https://arxiv.org/abs/2605.21467}{2605.21467}} & 68.3 & 46.4 & 59.3 & 96.0 & 86.9 \\[2pt]
SkillOpt: Executive Strategy for Self-Evolving Agent Skills\newline{\footnotesize\href{https://arxiv.org/abs/2605.23904}{2605.23904}} & 43.2 & 39.4 & 78.5 & 146.4 & 93.2 \\[2pt]
DVAO: Dynamic Variance-adaptive Advantage Optimization for Multi-reward Reinforcement Learning\newline{\footnotesize\href{https://arxiv.org/abs/2605.25604}{2605.25604}} & 48.6 & 59.4 & 58.5 & 57.3 & 51.7 \\[2pt]
LocateAnything: Fast and High-Quality Vision-Language Grounding with Parallel Box Decoding\newline{\footnotesize\href{https://arxiv.org/abs/2605.27365}{2605.27365}} & 52.0 & 105.0 & 39.2 & 119.0 & 69.9 \\[2pt]
Gamma-World: Generative Multi-Agent World Modeling Beyond Two Players\newline{\footnotesize\href{https://arxiv.org/abs/2605.28816}{2605.28816}} & 120.8 & $\infty$ & 71.1 & 107.7 & $\infty$ \\[2pt]
AgentDoG 1.5: A Lightweight and Scalable Alignment Framework for AI Agent Safety and Security\newline{\footnotesize\href{https://arxiv.org/abs/2605.29801}{2605.29801}} & 101.9 & 88.2 & 80.8 & 109.8 & $\infty$ \\[2pt]
COLLEAGUE.SKILL: Automated AI Skill Generation via Expert Knowledge Distillation\newline{\footnotesize\href{https://arxiv.org/abs/2605.31264}{2605.31264}} & 86.8 & 92.5 & 101.2 & 157.8 & $\infty$ \\[2pt]
Crafter: A Multi-Agent Harness for Editable Scientific Figure Generation from Diverse Inputs\newline{\footnotesize\href{https://arxiv.org/abs/2605.30611}{2605.30611}} & 56.0 & 127.7 & 107.9 & 90.0 & $\infty$ \\[2pt]
OCC-RAG: Optimal Cognitive Core for Faithful Question Answering\newline{\footnotesize\href{https://arxiv.org/abs/2606.00683}{2606.00683}} & 74.9 & 83.5 & 82.0 & 127.5 & 103.0 \\[2pt]
Cosmos 3: Omnimodal World Models for Physical AI\newline{\footnotesize\href{https://arxiv.org/abs/2606.02800}{2606.02800}} & 65.3 & 39.7 & 50.8 & 71.1 & $\infty$ \\[2pt]
Code2LoRA: Hypernetwork-Generated Adapters for Code Language Models under Software Evolution\newline{\footnotesize\href{https://arxiv.org/abs/2606.06492}{2606.06492}} & 66.4 & 56.2 & 41.1 & 90.6 & $\infty$ \\[2pt]
Imaginative Perception Tokens Enhance Spatial Reasoning in Multimodal Language Models\newline{\footnotesize\href{https://arxiv.org/abs/2606.03988}{2606.03988}} & 42.1 & 48.0 & 49.3 & 67.9 & 53.9 \\[2pt]
Agents' Last Exam\newline{\footnotesize\href{https://arxiv.org/abs/2606.05405}{2606.05405}} & 43.3 & 46.0 & 60.7 & 33.7 & 47.8 \\[2pt]
ABot-Earth 0.5: Generative 3D Earth Model\newline{\footnotesize\href{https://arxiv.org/abs/2606.09967}{2606.09967}} & 59.4 & 48.3 & 57.2 & 52.4 & 60.6 \\[2pt]
Embodied-R1.5: Evolving Physical Intelligence via Embodied Foundation Models\newline{\footnotesize\href{https://arxiv.org/abs/2606.11324}{2606.11324}} & 68.0 & $\infty$ & $\infty$ & 120.3 & $\infty$ \\[2pt]
From Chatbot to Digital Colleague: The Paradigm Shift Toward Persistent Autonomous AI\newline{\footnotesize\href{https://arxiv.org/abs/2606.14502}{2606.14502}} & 62.9 & 97.4 & 61.6 & 102.7 & 66.3 \\[2pt]
JoyAI-VL-Interaction: Real-Time Vision-Language Interaction Intelligence\newline{\footnotesize\href{https://arxiv.org/abs/2606.14777}{2606.14777}} & $\infty$ & 38.2 & 44.3 & 116.5 & 38.5 \\[2pt]
Looped World Models\newline{\footnotesize\href{https://arxiv.org/abs/2606.18208}{2606.18208}} & 44.1 & 50.7 & 38.7 & 97.3 & 63.7 \\[2pt]
MolmoMotion: Forecasting Point Trajectories in 3D with Language Instruction\newline{\footnotesize\href{https://arxiv.org/abs/2606.18558}{2606.18558}} & 69.3 & $\infty$ & 76.6 & $\infty$ & $\infty$ \\[2pt]
Moebius: 0.2B Lightweight Image Inpainting Framework with 10B-Level Performance\newline{\footnotesize\href{https://arxiv.org/abs/2606.19195}{2606.19195}} & $\infty$ & 106.9 & 111.0 & 107.8 & $\infty$ \\[2pt]
MemSlides: A Hierarchical Memory Driven Agent Framework for Personalized Slide Generation with Multi-turn Local Revision\newline{\footnotesize\href{https://arxiv.org/abs/2606.17162}{2606.17162}} & 85.4 & 82.6 & 86.0 & 98.0 & 148.9 \\[2pt]
PlanBench-XL: Evaluating Long-Horizon Planning of LLM Tool-Use Agents in Large-Scale Tool Ecosystems\newline{\footnotesize\href{https://arxiv.org/abs/2606.22388}{2606.22388}} & 65.6 & 85.0 & 68.0 & 84.1 & 67.0 \\[2pt]
Qwen-AgentWorld: Language World Models for General Agents\newline{\footnotesize\href{https://arxiv.org/abs/2606.24597}{2606.24597}} & 47.8 & 40.2 & 57.2 & 49.9 & 64.7 \\[2pt]
Are We Ready For An Agent-Native Memory System?\newline{\footnotesize\href{https://arxiv.org/abs/2606.24775}{2606.24775}} & 75.2 & 49.6 & 90.9 & 63.8 & 52.7 \\[2pt]
DanceOPD: On-Policy Generative Field Distillation\newline{\footnotesize\href{https://arxiv.org/abs/2606.27377}{2606.27377}} & 83.6 & $\infty$ & 63.1 & 46.0 & 75.6 \\[2pt]
Formalizing Latent Thoughts: Four Axioms of Thought Representation in LLMs\newline{\footnotesize\href{https://arxiv.org/abs/2606.27378}{2606.27378}} & $\infty$ & $\infty$ & 175.4 & $\infty$ & 143.0 \\[2pt]
Agentic Abstention: Do Agents Know When to Stop Instead of Act?\newline{\footnotesize\href{https://arxiv.org/abs/2606.28733}{2606.28733}} & 56.8 & $\infty$ & 46.9 & 45.2 & 49.1 \\[2pt]
Orca: The World is in Your Mind\newline{\footnotesize\href{https://arxiv.org/abs/2606.30534}{2606.30534}} & 48.7 & $\infty$ & 34.2 & 83.8 & 36.3 \\[2pt]
PerceptionRubrics: Calibrating Multimodal Evaluation to Human Perception\newline{\footnotesize\href{https://arxiv.org/abs/2606.28322}{2606.28322}} & 47.9 & 48.1 & 53.0 & 225.4 & 102.4 \\[2pt]
Program-as-Weights: A Programming Paradigm for Fuzzy Functions\newline{\footnotesize\href{https://arxiv.org/abs/2607.02512}{2607.02512}} & 57.8 & 72.4 & 45.9 & 70.2 & $\infty$ \\[2pt]
The Mirage of Optimizing Training Policies: Monotonic Inference Policies as the Real Objective for LLM Reinforcement Learning\newline{\footnotesize\href{https://arxiv.org/abs/2606.29526}{2606.29526}} & 82.8 & 80.2 & 187.1 & $\infty$ & 144.9 \\[2pt]
OmniOpt: Taxonomy, Geometry, and Benchmarking of Modern Optimizers\newline{\footnotesize\href{https://arxiv.org/abs/2607.04033}{2607.04033}} & 67.6 & 38.1 & 66.7 & 55.9 & 45.4 \\[2pt]
RynnWorld-4D: 4D Embodied World Models for Robotic Manipulation\newline{\footnotesize\href{https://arxiv.org/abs/2607.06559}{2607.06559}} & 92.5 & 45.2 & 67.2 & 89.2 & 84.0 \\[2pt]
Accurate, Interdisciplinary and Transparent Structure-property Understanding with Deep Native Structural Reasoning\newline{\footnotesize\href{https://arxiv.org/abs/2607.07708}{2607.07708}} & $\infty$ & $\infty$ & 91.2 & $\infty$ & 68.9 \\[2pt]
Vidu S1: A Real-Time Interactive Video Generation Model\newline{\footnotesize\href{https://arxiv.org/abs/2607.03118}{2607.03118}} & 78.0 & 50.1 & 69.2 & 60.2 & 134.2 \\[2pt]
Video Generation Models are General-Purpose Vision Learners\newline{\footnotesize\href{https://arxiv.org/abs/2607.09024}{2607.09024}} & 89.1 & 85.3 & 58.2 & 68.5 & $\infty$ \\[2pt]
Weak-to-Strong Generalization via Direct On-Policy Distillation\newline{\footnotesize\href{https://arxiv.org/abs/2607.05394}{2607.05394}} & 45.8 & 37.3 & 53.0 & 56.3 & $\infty$ \\[2pt]
Function-Aware Fill-in-the-Middle as Mid-Training for Coding Agent Foundation Models\newline{\footnotesize\href{https://arxiv.org/abs/2607.12463}{2607.12463}} & 70.9 & 69.5 & 78.5 & 116.2 & 89.4 \\[2pt]
Harness Handbook: Making Evolving Agent Harnesses Readable, Navigable, and Editable\newline{\footnotesize\href{https://arxiv.org/abs/2607.13285}{2607.13285}} & $\infty$ & 136.6 & 91.0 & 141.9 & $\infty$ \\[2pt]
LongStraw: Long-Context RL Beyond 2M Tokens under a Fixed GPU Budget\newline{\footnotesize\href{https://arxiv.org/abs/2607.14952}{2607.14952}} & 47.2 & 56.5 & 65.1 & 62.7 & 68.0 \\[2pt]
RAGU: A Multi-Step GraphRAG Engine with a Compact Domain-Adapted LLM\newline{\footnotesize\href{https://arxiv.org/abs/2607.11683}{2607.11683}} & 65.6 & 59.7 & 114.5 & 124.9 & 64.2 \\[2pt]
RynnBrain 1.1: Towards More Capable and Generalizable Embodied Foundation Model\newline{\footnotesize\href{https://arxiv.org/abs/2607.17977}{2607.17977}} & 88.3 & $\infty$ & 150.0 & $\infty$ & 104.0 \\[2pt]
ABot-World-0: Infinite Interactive World Rollout on a Single Desktop GPU\newline{\footnotesize\href{https://arxiv.org/abs/2607.19191}{2607.19191}} & 44.8 & 61.1 & 104.7 & 70.4 & 77.8 \\[2pt]
SLAI T-Rex: Full-Parameter Post-training of the DeepSeek-V4 Family on Ascend SuperPOD\newline{\footnotesize\href{https://arxiv.org/abs/2607.20145}{2607.20145}} & $\infty$ & $\infty$ & $\infty$ & $\infty$ & $\infty$ \\[2pt]
AREX: Towards a Recursively Self-Improving Agent for Deep Research\newline{\footnotesize\href{https://arxiv.org/abs/2607.21461}{2607.21461}} & $\infty$ & 105.3 & $\infty$ & 162.7 & $\infty$ \\[2pt]
DataPrep-Bench: Benchmarking LLMs as Training Data Preparators\newline{\footnotesize\href{https://arxiv.org/abs/2607.20465}{2607.20465}} & 71.9 & 33.3 & 55.6 & 43.7 & 45.7 \\[2pt]
Kimi K3: Open Frontier Intelligence\newline{\footnotesize\href{https://arxiv.org/abs/2607.24653}{2607.24653}} & $\infty$ & $\infty$ & $\infty$ & $\infty$ & $\infty$ \\[2pt]
HiFi-UMI: Learning Deployable Manipulation Policies from High-Fidelity UMI Data Alone\newline{\footnotesize\href{https://arxiv.org/abs/2607.25895}{2607.25895}} & 70.6 & 39.8 & 30.7 & 121.3 & 50.8 \\[2pt]
TurboVLA: Real-Time Vision-Language-Action Model at 32 Hz on an RTX 4090 with <1 GB VRAM\newline{\footnotesize\href{https://arxiv.org/abs/2607.27205}{2607.27205}} & 38.6 & 46.3 & 40.4 & 53.9 & 76.4 \\[2pt]
Qwen-UI-Agent Technical Report: Toward Next-Generation Real-World Centric Foundation GUI Agents\newline{\footnotesize\href{https://arxiv.org/abs/2607.28227}{2607.28227}} & $\infty$ & 45.3 & $\infty$ & 113.1 & $\infty$ \\[2pt]
Mental World Modeling\newline{\footnotesize\href{https://arxiv.org/abs/2607.27201}{2607.27201}} & $\infty$ & 72.7 & $\infty$ & 76.2 & $\infty$ \\[2pt]
LongHorizon-Harness: Advancing Long-Horizon Agents for Real-World Tasks\newline{\footnotesize\href{https://arxiv.org/abs/2608.01964}{2608.01964}} & 69.9 & 48.8 & 45.2 & 105.7 & 55.7 \\[2pt]
MerchantBench: Benchmarking LLM Agents for Long-Term Coherence in E-Commerce Operations\newline{\footnotesize\href{https://arxiv.org/abs/2607.28956}{2607.28956}} & 48.8 & 48.3 & 50.0 & 66.8 & 52.3 \\[2pt]
Recursive Synthesis for Long-Horizon Terminal Tasks\newline{\footnotesize\href{https://arxiv.org/abs/2608.05466}{2608.05466}} & 50.4 & 39.4 & 59.5 & 111.6 & 72.8 \\[2pt]
AgentOPSD: Recursive Self-Distillation for Agentic Reinforcement Learning\newline{\footnotesize\href{https://arxiv.org/abs/2608.05987}{2608.05987}} & $\infty$ & $\infty$ & 70.3 & 99.9 & $\infty$ \\[2pt]
SimWAM: A Simple World Action Model for End-to-End Autonomous Driving\newline{\footnotesize\href{https://arxiv.org/abs/2608.07468}{2608.07468}} & $\infty$ & 83.9 & 99.6 & $\infty$ & 121.1 \\[2pt]
BDH-CQ: In-Context Learning with Recurrent Latent Reasoning\newline{\footnotesize\href{https://arxiv.org/abs/2608.09888}{2608.09888}} & 53.1 & 51.1 & 36.8 & 47.6 & 62.9 \\[2pt]
ComBodied Agents: a New Paradigm of Human-Centric Agentic AI\newline{\footnotesize\href{https://arxiv.org/abs/2608.10915}{2608.10915}} & 50.1 & 44.7 & 129.7 & 57.1 & $\infty$ \\[2pt]
Spark-to-Paper: End-to-End Research Paper Generation as a Composable Skill\newline{\footnotesize\href{https://arxiv.org/abs/2608.11924}{2608.11924}} & 68.1 & 113.9 & 65.2 & 94.7 & $\infty$ \\[2pt]
Alaya-EVOKE: From Linear-Scaling Supervision to Endless World\newline{\footnotesize\href{https://arxiv.org/abs/2608.13546}{2608.13546}} & 68.8 & 65.7 & 71.6 & 89.8 & $\infty$ \\[2pt]
StateM: Reaching 95.3\% Raw Accuracy, or a \$15 Frontier Run, on Terminal-Bench 2.1 via Harness Scaling\newline{\footnotesize\href{https://arxiv.org/abs/2608.15089}{2608.15089}} & 44.4 & 49.5 & 42.6 & 58.6 & 91.2 \\[2pt]
Demystifying Agent Skills: Why They Work-Until They Don't\newline{\footnotesize\href{https://arxiv.org/abs/2608.14036}{2608.14036}} & 56.8 & 106.4 & 94.8 & 106.6 & 90.3 \\[2pt]
SemComp-Bench: Benchmarking Semantic Task Completion in Video Generation\newline{\footnotesize\href{https://arxiv.org/abs/2608.17426}{2608.17426}} & 76.9 & 32.9 & 98.6 & 93.7 & 75.0 \\[2pt]
EnvHarness: Awakening Static Worlds for Agent Learning\newline{\footnotesize\href{https://arxiv.org/abs/2608.19880}{2608.19880}} & 45.2 & 67.4 & 89.3 & 62.3 & $\infty$ \\[2pt]
Graph Engineering in the Era of LLM Agents: From Individual Intelligence to System Intelligence\newline{\footnotesize\href{https://arxiv.org/abs/2608.21156}{2608.21156}} & 58.7 & 85.0 & 81.2 & 108.5 & $\infty$ \\[2pt]
Apodex 1.1: Scaling Agentic Intelligence for Complex Work\newline{\footnotesize\href{https://arxiv.org/abs/2608.23283}{2608.23283}} & 68.2 & 50.1 & $\infty$ & $\infty$ & $\infty$ \\[2pt]
Annotations as Rollouts: Efficient and Scalable Reinforcement Learning for Video MLLMs\newline{\footnotesize\href{https://arxiv.org/abs/2608.20492}{2608.20492}} & 104.3 & 171.6 & 79.9 & 129.3 & $\infty$ \\[2pt]
VBVR-Pro: A Scalable and Verifiable Suite for Native Visual Reasoning\newline{\footnotesize\href{https://arxiv.org/abs/2608.26105}{2608.26105}} & 62.0 & 86.2 & 106.0 & 73.7 & 54.4 \\[2pt]
Agentic Game Development as a Verifiable Trajectory Data Engine for Scaling World Models\newline{\footnotesize\href{https://arxiv.org/abs/2608.25518}{2608.25518}} & 54.2 & $\infty$ & 137.7 & 112.8 & 132.7 \\[2pt]
LoopArena: Benchmarking Models as Runtime Controllers for Loop Engineering\newline{\footnotesize\href{https://arxiv.org/abs/2608.28281}{2608.28281}} & 99.7 & 103.9 & 100.7 & 118.5 & 94.1 \\[2pt]
Does On-Policy Distillation Really Distill? From Noisy Teacher to Self-Improvement\newline{\footnotesize\href{https://arxiv.org/abs/2608.31046}{2608.31046}} & 82.0 & 149.5 & 111.0 & 91.6 & $\infty$ \\[2pt]
StudentSim: Training LLM-based Student Simulators\newline{\footnotesize\href{https://arxiv.org/abs/2609.01591}{2609.01591}} & 91.5 & 101.4 & 100.0 & 142.3 & 110.2 \\[2pt]
Repo-To-Skill: Distilling GitHub Repositories Into AI4AI Skills\newline{\footnotesize\href{https://arxiv.org/abs/2609.02749}{2609.02749}} & 42.8 & 52.2 & 37.4 & 54.9 & 45.6 \\[2pt]
Compile by Training: Turning Natural-Language Specifications into Local Neural Functions\newline{\footnotesize\href{https://arxiv.org/abs/2609.04199}{2609.04199}} & 72.8 & 141.3 & 40.2 & 67.4 & 62.0 \\[2pt]
\end{longtable}
\endgroup

\section{SkillNet Case Study: generator Proposals and Gold}
\label{app:skillnet-verbatim}

The directional and essence proposals below are reproduced from sequences 55
and 73 of the successful Fable~5 trajectory for SkillNet
(arXiv:2603.04448).  The gold text is reproduced from the target pack's
\texttt{summary.setting\_and\_object} field.  We preserve the original content
while adding field labels and list structure for readability.

\begin{ideabox}{Directional pass: generator proposal}
\ideafield{Category}
A system or software artifact.

\ideafield{Core problem}
Language agents cannot accumulate and reuse experience across tasks and across agents in a portable way --- existing experience/skill memories are bespoke and siloed within individual agents or frameworks, so there is no general-purpose infrastructure for the community of agents to collectively store, manage, and reuse learned skills at trustworthy quality.

\ideafield{Main object}
A skill library infrastructure --- a general-purpose memory infrastructure / operating-system-like layer (a standardized API, protocol, or middleware for agent memory) whose primary function is an experience/skill library that stores past task trajectories or learned procedures for reuse on new tasks, and whose defining architectural form is a centralized shared hub/repository where many agents or users contribute and reuse skills collectively (hub/marketplace model), governed by LLM-as-judge curation --- a quality-control, validation, and curation mechanism in which an LLM acts as judge over contributed skills, evaluating their correctness, usefulness, generality, and redundancy to decide which contributed skills enter or remain in the shared hub.

\ideafield{System design}
The system defines a standard skill/experience representation (e.g., distilled procedures/workflows with metadata, derived from completed task trajectories) and a unified set of hub interfaces:
\begin{itemize}
  \setlength{\itemsep}{0.1em}
  \setlength{\parsep}{0pt}
  \setlength{\topsep}{0.2em}
  \item Any language agent or user can contribute skills (uploading experience distilled from successful trajectories).
  \item The LLM-judge curation pipeline vets each contribution before admission and continually maintains the library (filtering low-quality or incorrect skills, deduplicating/merging overlapping ones, ranking and pruning).
  \item Any agent can retrieve the most relevant curated skills for a new task (e.g., via embedding-based similarity or task matching) into its context to condition its planning and actions.
\end{itemize}

\ideafield{Intended outcome}
By decoupling the skill library from any specific agent and exposing it as a centralized, standardized, LLM-curated shared hub --- analogous to a package registry or model hub, but for agent skills with automated quality gatekeeping --- the system enables collective, continual, trustworthy skill accumulation and cross-agent/cross-user skill reuse without any model weight updates.
\end{ideabox}

\begin{ideabox}{Essence pass: generator proposal}
\ideafield{Category}
A system or software artifact.

\ideafield{Core problem}
Language agents cannot accumulate and reuse experience across tasks and across agents in a portable way --- existing experience/skill memories are bespoke and siloed within individual agents or frameworks, so there is no general-purpose infrastructure for the community of agents to collectively store, organize, and reuse learned skills at trustworthy quality.

\ideafield{Main object}
A skill library infrastructure --- a general-purpose memory infrastructure / operating-system-like layer (a standardized API, protocol, or middleware for agent memory) whose primary function is an experience/skill library that stores past task trajectories or learned procedures for reuse on new tasks, and whose defining architectural form is a centralized shared hub/repository where many agents or users contribute and reuse skills collectively (hub/marketplace model), governed by LLM-as-judge curation --- a quality-control, validation, and curation mechanism in which an LLM acts as judge over contributed skills, evaluating their correctness, usefulness, generality, and redundancy to decide which contributed skills enter or remain in the shared hub --- and organized by a skill ontology: an organizational structure imposed on the hub's contents, i.e., a hierarchical taxonomy of skills with relations among them (specialization/generalization, dependencies, and composition of smaller skills into larger skills) that structures how skills are stored, related, abstracted, and navigated.

\ideafield{System design}
The system defines a standard skill/experience representation (e.g., distilled procedures/workflows with metadata, derived from completed task trajectories) and a unified set of hub interfaces:
\begin{itemize}
  \setlength{\itemsep}{0.1em}
  \setlength{\parsep}{0pt}
  \setlength{\topsep}{0.2em}
  \item Any language agent or user can contribute skills (uploading experience distilled from successful trajectories).
  \item The LLM-judge curation pipeline vets each contribution before admission and continually maintains the library (filtering low-quality or incorrect skills, deduplicating/merging overlapping ones, ranking and pruning).
  \item Admitted skills are placed into the skill ontology --- classified under the appropriate taxonomy nodes, linked by dependency/composition relations, with related skills abstracted or composed into more general or higher-level skills as the library grows.
  \item Any agent can retrieve the most relevant curated skills for a new task (via ontology-guided navigation and/or embedding-based similarity) into its context to condition its planning and actions.
\end{itemize}

\ideafield{Intended outcome}
By decoupling the skill library from any specific agent and exposing it as a centralized, standardized, LLM-curated, ontology-organized shared hub --- analogous to a package registry or model hub, but for agent skills with automated quality gatekeeping and a structured taxonomy --- the system enables collective, continual, trustworthy, and well-organized skill accumulation and cross-agent/cross-user skill reuse without any model weight updates.
\end{ideabox}

\begin{ideabox}{Gold setting and main object}
\ideafield{Category}
Infrastructure / Resource (open skill repository, ontology, evaluation protocol, and toolkit).

\ideafield{Problem \& motivation}
\begin{itemize}
  \setlength{\itemsep}{0.1em}
  \setlength{\parsep}{0pt}
  \setlength{\topsep}{0.2em}
  \item Agents lack a unified mechanism to consolidate episodic experience into durable, reusable skills, so they repeatedly ``reinvent the wheel.''
  \item Existing skill repositories/marketplaces rely on manual curation and only indirect (downstream-task) quality signals, leaving skills fragmented, redundant, and unvalidated.
\end{itemize}

\ideafield{Main object \& crux}
SkillNet: an open full-lifecycle skill infrastructure that:
\begin{itemize}
  \setlength{\itemsep}{0.1em}
  \setlength{\parsep}{0pt}
  \setlength{\topsep}{0.2em}
  \item Auto-creates skills from heterogeneous sources (execution trajectories, GitHub repos, documents, prompts) via LLM pipelines.
  \item Evaluates each skill intrinsically along Safety, Completeness, Executability, Maintainability, Cost-awareness.
  \item Connects skills into a typed relation graph (similar\_to, belong\_to, compose\_with, depend\_on).
\end{itemize}

\ideafield{Representation}
Skills are treated as a unified knowledge representation bridging natural-language procedural guidance and machine-executable logic, organized by a three-layer Skill Ontology (taxonomy, relation graph, skill packages).

\ideafield{Release}
Released as a curated repository (200k+ candidate, 150k+ curated skills), website/API, and the skillnet-ai Python library + CLI.
\end{ideabox}

\clearpage
\section{Directional-to-Essence Refinement Examples}
\label{app:directional-essence-examples}

The boxes below reproduce complete submitted Opus~5 high proposals for two
test papers, with field headings, paragraph breaks, and list formatting
added for readability. Role terminology is standardized to \emph{guide}.
All other original prose is retained, and mathematical notation
is typeset in LaTeX.
Each example pairs a candidate that passes the Directional judge with the final
proposal, which passes all three Essence evaluations.
The Directional proposal is black; the Essence proposal is
\textcolor[HTML]{763A8C}{purple}.
Costs are cumulative; the reported increase is net of refunded costs.
Essence costs include the promotion cost (Appendix~\ref{app:promotion-cost}).
All values are rounded independently.

\begingroup
\definecolor{essencepurple}{HTML}{763A8C}
\newcommand{\essencechange}[1]{\textcolor{essencepurple}{#1}}

\begin{ideabox}{DelTA: Directional to Essence (arXiv:2605.21467)}
\ideafield{Compression cost}
Directional: 21.64 bits. Essence: 46.40 bits. Net increase: 24.76 bits.
New retained Essence costs: 27.42 bits.

\ideafield{Directional proposal}
\ideafield{Category:}
a new algorithm or method.
\ideafield{Core problem:}
in RL post-training of large language models with only a single outcome reward at the end of a long generated sequence, current estimators give no credit assignment inside the sequence -{}- group-relative, critic-free objectives (GRPO-style) compute one sequence-level advantage and broadcast it identically to every token, so all steps of a correct trace are reinforced and all steps of a failed trace are punished, while PPO-style learned value networks that do produce token-level advantages are expensive and poorly calibrated on partial generations.
\ideafield{Main object (the method):}
a critic-free step-level advantage estimator that obtains Monte-Carlo values for intermediate prefixes by branching extra short continuations from selected points inside sampled trajectories, estimating \(V(\text{prefix})\) as the average terminal reward of the continuations that share that prefix, and defining the advantage of each step as the increment in this estimated value across the step (a Monte-Carlo temporal-difference on prefix values).

This turns one terminal scalar into a within-sequence advantage profile that localizes which reasoning steps actually raised or lowered the probability of success, while the group of siblings still supplies the variance-reducing baseline; the resulting per-step advantages are plugged into the standard clipped policy-gradient update, and the branching budget is concentrated where the estimate matters, so the method keeps GRPO's low cost and low variance but recovers PPO-like fine-grained credit without training any value model. (Confirmed: setting and contribution type; the branching-based prefix-value mechanism is my inference.)

\ideafield{Essence proposal}
\ideafield{Category:}
\noindent\begin{minipage}{\linewidth}\textcolor{essencepurple}{a new algorithm or method.}\end{minipage}\par
\ideafield{Core problem:}
\noindent\begin{minipage}{\linewidth}\textcolor{essencepurple}{in RL post-training of large language models (token/step-level credit within generated sequences), the reward is a single scalar at the end of a long generated trace, so critic-free group-relative policy gradients broadcast one identical advantage to every token of a response -{}- reinforcing filler and erroneous tokens exactly like the decisive ones, and letting conflicting token gradients cancel or fight -{}- while learned value models that would supply token-level advantages are expensive and miscalibrated on partial sequences.}\end{minipage}\par
\ideafield{Main object:}
\noindent\begin{minipage}{\linewidth}\textcolor{essencepurple}{‘cross-rollout gradient contrast’, a critic-free token-level credit estimator whose defining computation is gradient- or attention-based attribution of the outcome back onto tokens realized in PARAMETER space and compared contrastively across the sampled responses of the same prompt using their attribution maps.}\end{minipage}\par

\noindent\begin{minipage}{\linewidth}\textcolor{essencepurple}{For each sampled rollout, the attribution map is the collection of per-token parameter-space gradients \(g_{i,t} = \nabla_{\theta}\log\pi_{\theta}(y_{i,t}\mid\text{prompt}, y_{i,<t})\) -{}- the weight-space direction in which reinforcing that single token would move the policy.}\end{minipage}\par

\noindent\begin{minipage}{\linewidth}\textcolor{essencepurple}{The rollouts of a prompt are split by outcome into a positive set (correct / high-reward) and a negative set (incorrect / low-reward), and their gradients are aggregated into two group directions \(G^{+}\) and \(G^{-}\).}\end{minipage}\par

\noindent\begin{minipage}{\linewidth}\textcolor{essencepurple}{Each token's credit is then the inner-product/cosine alignment of its own gradient with these aggregated gradients of positive versus negative rollouts -{}- essentially \(\langle g_{i,t}, G^{+}\rangle - \langle g_{i,t}, G^{-}\rangle\) (equivalently the cosine alignment with the contrast direction \(G^{+} - G^{-}\)) -{}- an influence-function/gradient-agreement score computed within the prompt's own group.}\end{minipage}\par

\noindent\begin{minipage}{\linewidth}\textcolor{essencepurple}{Because gradients from different rollouts share one vector space, this contrast needs no positional or semantic alignment of tokens across responses; components common to successful and failed rollouts (prompt-generic, formatting, filler) cancel in the contrast and receive near-zero credit, acting as a token-level baseline/control variate, while tokens whose update direction genuinely drives success versus failure retain signed credit -{}- so an error inside a correct trace is not rewarded and a sound step inside a failed trace is not punished.}\end{minipage}\par

\noindent\begin{minipage}{\linewidth}\textcolor{essencepurple}{These alignment-derived per-token advantages replace the uniformly broadcast sequence-level group-relative advantage inside the standard clipped policy-gradient update, yielding dense intra-sequence credit assignment and variance reduction with no extra rollouts, value network, or process reward model.}\end{minipage}\par

\end{ideabox}
\begin{ideabox}{ABot-World-0: Directional to Essence (arXiv:2607.19191)}
\ideafield{Compression cost}
Directional: 38.71 bits. Essence: 61.10 bits. Net increase: 22.38 bits.
New retained Essence costs: 25.57 bits.

\ideafield{Directional proposal}
\ideafield{Category:}
a system or software artifact (deep learning application in computer vision).
\ideafield{Core problem:}
generative video models can render plausible scenes but cannot be used as an actual environment -{}- multi-step diffusion is orders of magnitude too slow for interactive control, and autoregressive rollouts drift and forget, so a user cannot walk through, look around, and come back to a place in a generated world.
\ideafield{Main object:}
an end-to-end interactive ‘world model’ system that turns a text prompt or a single image into a navigable virtual world and generates it frame-by-frame in real time in response to a live stream of user actions (movement, camera/view control).
\ideafield{Key design principles:}
\begin{enumerate}
\setlength{\itemsep}{0.1em}
\setlength{\parsep}{0pt}
\setlength{\topsep}{0.2em}
\item[(i)] causal, streaming, action-conditioned generation -{}- the model is made autoregressive over frames/chunks with few-step (distilled) sampling and cached context so each new frame is produced at interactive latency instead of denoising a whole clip;
\item[(ii)] a persistent world state/memory: previously generated content (with its camera poses) is retained and re-injected as conditioning so revisited regions remain consistent and the world does not dissolve during long, in-principle unbounded exploration;
\item[(iii)] an engineered serving pipeline (GPU-resident rolling context, low-latency input-to-frame loop) that makes the whole thing a usable interactive application rather than an offline demo.
\end{enumerate}
\ideafield{Confirmed by the guide:}
field, systems/artifact category, and that the artifact's identity is a ‘world model’ whose functional role defines the contribution; the streaming + memory design principles are my inference.

\ideafield{Essence proposal}
\ideafield{Category:}
\noindent\begin{minipage}{\linewidth}\textcolor{essencepurple}{a system or software artifact (computer-vision application).}\end{minipage}\par
\ideafield{Core problem:}
\noindent\begin{minipage}{\linewidth}\textcolor{essencepurple}{generative video models cannot serve as an environment a person can actually play in -{}- driven autoregressively from a user's live commands they must run causally and in real time, and because they were trained on ground-truth history they drift, blur and collapse once conditioned on their own outputs, so interactive sessions cannot last.}\end{minipage}\par
\ideafield{Main object:}
\noindent\begin{minipage}{\linewidth}\textcolor{essencepurple}{an end-to-end interactive ‘world model’ system whose functional role is to turn a prompt/initial image into a navigable virtual world that a user explores in real time under ‘keyboard control’ -{}- discrete WASD-style keypresses (plus view/mouse-look) are the action interface and each keypress conditions the very next generated frame chunk, produced causally with few-step sampling and a cached rolling context for interactive latency.}\end{minipage}\par

\noindent\begin{minipage}{\linewidth}\textcolor{essencepurple}{Its key design principle -{}- the essence -{}- is stabilizing unbounded autoregressive rollout by training the model on its own generated context rather than on ground-truth history: the causal action-conditioned generator is self-rolled out during training (with a rolling KV/context cache, and the loss -{}- distribution-matching/adversarial or denoising with noise-augmented self-generated context -{}- applied to frames produced from that self-generated history), eliminating the train/inference exposure-bias gap so error does not accumulate and the world stays sharp and coherent over arbitrarily long, in principle unbounded, keyboard-controlled exploration.}\end{minipage}\par

\end{ideabox}
\endgroup

\section{Research taxonomy and contribution categories}
\label{app:taxonomy}

The opening bids use a fixed research taxonomy of deep learning and a fixed
list of contribution categories (\Cref{sec:experiment-instantiation}). GPT-5.6 Sol
generated both once, and every generator receives them for every target.
The category bid offers the following 11 categories with equal probability.

\begin{enumerate}[leftmargin=*,itemsep=0pt,topsep=2pt]
\item a new algorithm or method
\item a theory or analysis result
\item a benchmark
\item a dataset
\item an architecture
\item a mechanism study (how something works)
\item a scaling law
\item an empirical phenomenon or observation study
\item an evaluation protocol or metric
\item a system or software artifact
\item an application of existing methods to a domain
\end{enumerate}

The taxonomy has 3,035 nodes below its root, ``deep learning,'' in five
levels, with 17 top-level fields. Each taxonomy bid offers the children of
the current node and a stop option with equal probability. The taxonomy
file also places a child named ``other'' under every node; the scaffold
never offers it, so it and its subtree are omitted. The list keeps the
file's order: bold entries are the top-level fields, and indentation marks
deeper levels.

\newlength{\taxstep}\setlength{\taxstep}{0.8em}
\newcommand{\taxfield}[1]{\par\addvspace{3pt}\textbf{#1}\par}
\newcommand{\taxnode}[2]{\par\hangindent=\dimexpr #1\taxstep+\taxstep\relax
  \hangafter=1\hspace*{\dimexpr #1\taxstep\relax}#2\par}
\begingroup
\scriptsize\setlength{\parindent}{0pt}\setlength{\parskip}{0pt}\raggedright
\begin{multicols}{3}
\taxfield{agents}
\taxnode{0}{Foundations and Theory}
\taxnode{1}{Reinforcement Learning Theory}
\taxnode{1}{Sequential Decision Theory and Control}
\taxnode{1}{Planning, Search, and Model-Based Reasoning}
\taxnode{1}{Exploration and Sample Complexity}
\taxnode{1}{Credit Assignment and Temporal Abstraction}
\taxnode{1}{Representation Learning and World Models}
\taxnode{1}{Imitation Learning and Inverse Reinforcement Learning}
\taxnode{1}{Generalization and Transfer in Agents}
\taxnode{1}{Multi-Agent Learning and Game Theory}
\taxnode{1}{Safety, Robustness, and Alignment}
\taxnode{1}{Agent Architectures and Inductive Biases}
\taxnode{1}{Interpretability and Mechanistic Analysis}
\taxnode{1}{Empirical Phenomena and Scaling Laws}
\taxnode{1}{Benchmarks and Evaluation}
\taxnode{1}{Surveys and Unifying Frameworks}
\taxnode{0}{Reinforcement Learning}
\taxnode{1}{Theoretical Reinforcement Learning}
\taxnode{1}{Model-Free Reinforcement Learning}
\taxnode{1}{Model-Based Reinforcement Learning}
\taxnode{1}{Offline Reinforcement Learning}
\taxnode{1}{Exploration and Intrinsic Motivation}
\taxnode{1}{Hierarchical Reinforcement Learning}
\taxnode{1}{Multi-Agent Reinforcement Learning}
\taxnode{1}{Learning from Demonstrations and Human Feedback}
\taxnode{1}{Safe and Constrained Reinforcement Learning}
\taxnode{1}{Transfer Learning, Meta-Reinforcement Learning, and Generalization}
\taxnode{1}{Representation Learning for Reinforcement Learning}
\taxnode{1}{Benchmarking, Evaluation, and Reproducibility in Reinforcement Learning}
\taxnode{1}{Applied Reinforcement Learning}
\taxnode{1}{Survey and Review Literature on Reinforcement Learning}
\taxnode{0}{Multi-Agent Learning}
\taxnode{1}{Cooperative Multi-Agent Learning}
\taxnode{1}{Competitive and Adversarial Multi-Agent Learning}
\taxnode{1}{Mixed-Motive and General-Sum Multi-Agent Learning}
\taxnode{1}{Game-Theoretic Multi-Agent Learning}
\taxnode{1}{Opponent Modeling and Recursive Reasoning}
\taxnode{1}{Communication and Emergent Communication}
\taxnode{1}{Coordination and Credit Assignment}
\taxnode{1}{Decentralized Control and Planning}
\taxnode{1}{Large-Population and Mean-Field Multi-Agent Learning}
\taxnode{1}{Transfer, Meta-Learning, and Generalization in Multi-Agent Systems}
\taxnode{1}{Safe, Robust, and Trustworthy Multi-Agent Learning}
\taxnode{1}{Theoretical Foundations of Multi-Agent Learning}
\taxnode{1}{Benchmarks, Simulators, and Evaluation for Multi-Agent Learning}
\taxnode{1}{Applications of Multi-Agent Learning}
\taxnode{1}{Surveys and Taxonomies of Multi-Agent Learning}
\taxnode{0}{Planning and Model-Based Control}
\taxnode{1}{Dynamics Modeling and World Models}
\taxnode{1}{State Estimation and Belief Modeling}
\taxnode{1}{Planning and Search}
\taxnode{1}{Optimal Control and Model Predictive Control}
\taxnode{1}{Model-Based Reinforcement Learning}
\taxnode{1}{Hierarchical Planning and Temporal Abstraction}
\taxnode{1}{Task and Motion Planning}
\taxnode{1}{Safe, Robust, and Constrained Control}
\taxnode{1}{Adaptive Control and Online System Identification}
\taxnode{1}{Robotics and Autonomous Systems Applications}
\taxnode{1}{Benchmarks and Evaluation}
\taxnode{1}{Theory and Guarantees}
\taxnode{0}{Hierarchical Reinforcement Learning}
\taxnode{1}{Options and Temporal Abstraction}
\taxnode{1}{Goal-Conditioned and Subgoal-Based HRL}
\taxnode{1}{Feudal Reinforcement Learning}
\taxnode{1}{Skill Discovery and Unsupervised HRL}
\taxnode{1}{State Abstraction and Hierarchical Representation Learning}
\taxnode{1}{Model-Based and Planning-Oriented HRL}
\taxnode{1}{Hierarchical Exploration and Sparse-Reward Learning}
\taxnode{1}{Hierarchical Policy Optimization and Credit Assignment}
\taxnode{1}{Transfer, Continual, and Compositional HRL}
\taxnode{1}{Imitation, Offline, and Language-Guided HRL}
\taxnode{1}{Multi-Agent Hierarchical Reinforcement Learning}
\taxnode{1}{Theory and Foundations of HRL}
\taxnode{1}{Benchmarks, Empirical Evaluation, and Applications of HRL}
\taxnode{0}{Imitation Learning and Inverse Reinforcement Learning}
\taxnode{1}{Behavior Cloning}
\taxnode{1}{Interactive Imitation Learning}
\taxnode{1}{Inverse Reinforcement Learning}
\taxnode{1}{Apprenticeship Learning}
\taxnode{1}{Adversarial Imitation Learning}
\taxnode{1}{Imitation from Observation}
\taxnode{1}{Offline Imitation Learning}
\taxnode{1}{Hierarchical Imitation Learning}
\taxnode{1}{Meta-Imitation Learning and Generalization}
\taxnode{1}{Safe and Robust Imitation Learning}
\taxnode{1}{Theory and Foundations of Imitation Learning and IRL}
\taxnode{1}{Benchmarks, Datasets, and Evaluation}
\taxnode{1}{Applications of Imitation Learning and IRL}
\taxnode{1}{Surveys and Tutorials}
\taxnode{0}{Embodied AI and Robotics}
\taxnode{1}{Foundations and Theory}
\taxnode{1}{Embodied Perception}
\taxnode{1}{Localization and Mapping}
\taxnode{1}{Planning and Decision Making}
\taxnode{1}{Robot Control and Reinforcement Learning}
\taxnode{1}{Imitation Learning}
\taxnode{1}{Manipulation and Grasping}
\taxnode{1}{Navigation and Exploration}
\taxnode{1}{Locomotion}
\taxnode{1}{Language-Grounded Embodied Agents}
\taxnode{1}{Human-Robot Interaction}
\taxnode{1}{Multi-Robot Systems}
\taxnode{1}{Sim-to-Real Transfer}
\taxnode{1}{Safety and Verification}
\taxnode{1}{Benchmarks and Simulation}
\taxnode{1}{Robotic Applications and Systems}
\taxnode{0}{Language Agents}
\taxnode{1}{Agent Architectures}
\taxnode{1}{Planning, Reasoning, and Decision Making}
\taxnode{1}{Tool Use and Environment Interaction}
\taxnode{1}{Memory, Retrieval, and Knowledge Integration}
\taxnode{1}{Learning and Optimization for Language Agents}
\taxnode{1}{Multi-Agent Systems}
\taxnode{1}{Alignment, Safety, and Robustness}
\taxnode{1}{Evaluation and Benchmarks}
\taxnode{1}{Foundations and Analysis}
\taxnode{1}{Applications of Language Agents}
\taxnode{1}{Surveys and Taxonomies}
\taxnode{0}{Human-Agent Interaction}
\taxnode{1}{Learning from Demonstrations}
\taxnode{1}{Learning from Human Feedback}
\taxnode{1}{Interactive Reinforcement Learning and Online Guidance}
\taxnode{1}{Natural Language and Multimodal Interaction}
\taxnode{1}{Human-Agent Collaboration and Shared Autonomy}
\taxnode{1}{User Modeling and Personalization}
\taxnode{1}{Social and Affective Interaction}
\taxnode{1}{Explainability, Transparency, and Trust}
\taxnode{1}{Safety, Alignment, and Human Oversight}
\taxnode{1}{Evaluation, Benchmarks, and Human Studies}
\taxnode{0}{Safety, Alignment, and Robustness}
\taxnode{1}{Foundations of AI Safety and Alignment}
\taxnode{1}{AI Alignment and Preference Learning}
\taxnode{1}{Safe Reinforcement Learning}
\taxnode{1}{Adversarial Machine Learning}
\taxnode{2}{Evasion Attacks and Adversarial Examples}
\taxnode{2}{Poisoning Attacks}
\taxnode{2}{Backdoor Attacks}
\taxnode{2}{Model Stealing and Privacy Attacks}
\taxnode{2}{Adversarial Defenses}
\taxnode{2}{Certified Robustness and Verification}
\taxnode{2}{Theory of Adversarial Robustness}
\taxnode{2}{Threat Modeling and Security Analysis}
\taxnode{2}{Robustness Evaluation and Benchmarking}
\taxnode{2}{Applied Adversarial Machine Learning}
\taxnode{1}{Distributional Robustness and Out-of-Distribution Generalization}
\taxnode{1}{Uncertainty Quantification and Calibration}
\taxnode{1}{Formal Verification and Certified Robustness}
\taxnode{1}{Interpretability, Oversight, and Monitoring}
\taxnode{1}{Safety Evaluation, Red Teaming, and Benchmarks}
\taxnode{1}{Safety-Critical Machine Learning and Autonomous Systems}
\taxnode{0}{Agent Architectures and Training Methodology}
\taxnode{1}{Policy and Value Function Architectures}
\taxnode{1}{Memory-Augmented and Recurrent Agents}
\taxnode{1}{World Models and Model-Based Agents}
\taxnode{1}{Hierarchical Reinforcement Learning and Temporal Abstraction}
\taxnode{1}{Modular and Compositional Agent Design}
\taxnode{1}{Planning-Integrated and Search-Based Agents}
\taxnode{1}{Representation Learning and State Abstraction for Agents}
\taxnode{1}{Policy Optimization, Credit Assignment, and Stabilization}
\taxnode{1}{Self-Supervised, Auxiliary, and Multitask Training}
\taxnode{1}{Imitation Learning, Learning from Demonstrations, and Offline Reinforcement Learning}
\taxnode{1}{Exploration, Curriculum Learning, and Environment Design}
\taxnode{1}{Human Feedback and Preference-Based Training}
\taxnode{1}{Evaluation, Benchmarks, and Experimental Methodology}
\taxnode{1}{Theory of Agent Architectures and Training}
\taxnode{1}{Application-Specific Agent Design}
\taxnode{1}{Surveys and Taxonomies of Agent Architectures}
\taxnode{0}{Evaluation and Benchmarks}
\taxnode{1}{Benchmark Construction and Curation}
\taxnode{1}{Evaluation Metrics and Protocols}
\taxnode{1}{Statistical Evaluation and Reproducibility}
\taxnode{1}{Capability and Task Performance Benchmarking}
\taxnode{1}{Generalization and Transfer Evaluation}
\taxnode{1}{Robustness and Adversarial Evaluation}
\taxnode{1}{Safety and Reliability Evaluation}
\taxnode{1}{Human-Centered Evaluation}
\taxnode{1}{Efficiency and Resource-Aware Evaluation}
\taxnode{1}{Offline and Counterfactual Evaluation}
\taxnode{1}{Embodied and Real-World Evaluation}
\taxnode{1}{Meta-Evaluation and Benchmark Auditing}
\taxnode{0}{Applications of Agents}
\taxnode{1}{Embodied AI and Robotics}
\taxnode{1}{Autonomous Driving and Navigation}
\taxnode{1}{Games and Simulated Environments}
\taxnode{1}{Web and Software Agents}
\taxnode{1}{Conversational and Assistive Agents}
\taxnode{1}{Recommender Systems and Personalization}
\taxnode{1}{Healthcare and Biomedicine}
\taxnode{1}{Scientific Discovery and Engineering Design}
\taxnode{1}{Operations Research, Logistics, and Resource Management}
\taxnode{1}{Finance and Economic Decision Making}
\taxnode{1}{Education and Intelligent Tutoring Systems}
\taxnode{1}{Cybersecurity}
\taxnode{1}{Benchmarks and Evaluation for Agent Applications}
\taxnode{0}{Surveys and Taxonomies}
\taxnode{1}{Foundational Taxonomies and Conceptual Frameworks}
\taxnode{1}{Agent Architectures}
\taxnode{1}{Reinforcement Learning and Sequential Decision-Making Agents}
\taxnode{1}{Planning, Reasoning, and Memory}
\taxnode{1}{Foundation-Model and Language Agents}
\taxnode{1}{Tool-Using and Environment-Interacting Agents}
\taxnode{1}{Multi-Agent Systems}
\taxnode{1}{Embodied and Robotic Agents}
\taxnode{1}{Human-Agent Interaction and Collaboration}
\taxnode{1}{Evaluation, Benchmarks, and Datasets}
\taxnode{1}{Safety, Alignment, Robustness, and Security}
\taxnode{1}{Application-Oriented Agent Surveys}
\taxfield{alignment}
\taxnode{0}{Objective Specification and Outer Alignment}
\taxnode{1}{Reward Modeling and Preference Learning}
\taxnode{1}{Inverse Reinforcement Learning and Inverse Reward Design}
\taxnode{1}{Value Learning and Preference Elicitation}
\taxnode{1}{Scalable Oversight}
\taxnode{1}{Constitutional and Rule-Based Alignment}
\taxnode{1}{Constrained and Multi-Objective Alignment}
\taxnode{1}{Specification Gaming and Reward Hacking}
\taxnode{1}{Objective Robustness and Goal Generalization}
\taxnode{1}{Theoretical Foundations of Outer Alignment}
\taxnode{1}{Evaluation and Benchmarking for Objective Specification}
\taxnode{1}{Application-Specific Objective Specification}
\taxnode{1}{Surveys and Methodology for Objective Specification}
\taxnode{0}{Inner Alignment and Goal Generalization}
\taxnode{1}{Theoretical Foundations}
\taxnode{1}{Mesa-Optimization}
\taxnode{1}{Deceptive Alignment}
\taxnode{1}{Goal Misgeneralization}
\taxnode{1}{Mechanistic Interpretability}
\taxnode{1}{Empirical Studies}
\taxnode{1}{Benchmarks and Evaluation}
\taxnode{1}{Alignment Interventions}
\taxnode{1}{Surveys and Taxonomies}
\taxnode{0}{Preference Learning and Reward Modeling}
\taxnode{1}{Preference Elicitation}
\taxnode{1}{Preference Modeling}
\taxnode{1}{Reward Modeling}
\taxnode{1}{Preference Optimization}
\taxnode{1}{Active and Online Preference Learning}
\taxnode{1}{Inverse Reinforcement Learning and Reward Inference}
\taxnode{1}{Theoretical Foundations}
\taxnode{1}{Robustness, Uncertainty, and Calibration}
\taxnode{1}{Safety, Bias, and Reward Hacking}
\taxnode{1}{Evaluation and Benchmarks}
\taxnode{1}{Multimodal and Domain-Specific Applications}
\taxnode{0}{Learning from Human Feedback}
\taxnode{1}{Preference Learning and Reward Modeling}
\taxnode{1}{Preference-based Policy Optimization}
\taxnode{1}{Learning from Demonstrations and Imitation Learning}
\taxnode{1}{Critique and Corrective Feedback}
\taxnode{1}{Interactive and Active Feedback Elicitation}
\taxnode{1}{Feedback Collection and Annotation Methodology}
\taxnode{1}{Robustness, Uncertainty, and Bias in Human Feedback}
\taxnode{1}{Personalization and Preference Aggregation}
\taxnode{1}{Theory and Foundations of Learning from Human Feedback}
\taxnode{1}{Evaluation and Benchmarks for Human Feedback Methods}
\taxnode{1}{Applications of Learning from Human Feedback}
\taxnode{0}{Scalable Oversight}
\taxnode{1}{Theoretical Foundations of Scalable Oversight}
\taxnode{1}{Preference Learning and Reward Modeling}
\taxnode{1}{Process Supervision}
\taxnode{1}{Decomposition-Based Oversight}
\taxnode{1}{Debate and Adversarial Oversight}
\taxnode{1}{AI-Assisted Oversight}
\taxnode{1}{Weak-to-Strong Generalization}
\taxnode{1}{Eliciting Latent Knowledge}
\taxnode{1}{Interpretability-Assisted Oversight}
\taxnode{1}{Oversight Benchmarks and Empirical Evaluation}
\taxnode{0}{Rule-Based and Constraint-Based Alignment}
\taxnode{1}{Constitutional and Policy-Based Alignment}
\taxnode{1}{Constraint Elicitation and Specification}
\taxnode{1}{Constrained Optimization for Alignment}
\taxnode{1}{Constrained Reinforcement Learning and Safe Exploration}
\taxnode{1}{Neuro-Symbolic and Logic-Based Alignment}
\taxnode{1}{Decoding-Time Constraint Enforcement}
\taxnode{1}{Runtime Monitoring and Safety Filters}
\taxnode{1}{Formal Verification and Certification}
\taxnode{1}{Theoretical Foundations of Constraint-Based Alignment}
\taxnode{1}{Evaluation, Benchmarks, and Red Teaming}
\taxnode{1}{Applications in Safety-Critical and Regulated Domains}
\taxnode{1}{Surveys and Taxonomies of Rule-Based Alignment}
\taxnode{0}{Interpretability and Transparency for Alignment}
\taxnode{1}{Mechanistic Interpretability}
\taxnode{1}{Feature Attribution and Saliency Methods}
\taxnode{1}{Concept-Based Interpretability}
\taxnode{1}{Probing and Representation Analysis}
\taxnode{1}{Example-Based Explanations and Data Attribution}
\taxnode{1}{Causal and Counterfactual Explanations}
\taxnode{1}{Explanation Generation and Rationalization}
\taxnode{1}{Transparency and Documentation}
\taxnode{1}{Human-Centered Interpretability}
\taxnode{1}{Evaluation, Faithfulness, and Benchmarks}
\taxnode{1}{Model Auditing}
\taxnode{1}{Scalable Oversight}
\taxnode{1}{Theoretical Foundations of Interpretability}
\taxnode{1}{Surveys and Taxonomies}
\taxnode{0}{Corrigibility and Controllability}
\taxnode{1}{Formal Corrigibility}
\taxnode{1}{Safe Interruptibility}
\taxnode{1}{Shutdown and Off-Switch Control}
\taxnode{1}{Human Oversight and Override}
\taxnode{1}{Objective Modification and Updateability}
\taxnode{1}{Tampering Resistance}
\taxnode{1}{Containment and Runtime Control}
\taxnode{1}{Training Methods for Controllable Agents}
\taxnode{1}{Evaluation and Benchmarks for Controllability}
\taxnode{0}{Truthfulness, Honesty, and Uncertainty}
\taxnode{1}{Uncertainty Quantification and Calibration}
\taxnode{1}{Epistemic Uncertainty and Out-of-Distribution Detection}
\taxnode{1}{Selective Prediction and Deferral}
\taxnode{1}{Factuality and Hallucination}
\taxnode{1}{Verification and Fact-Checking}
\taxnode{1}{Honesty, Deception, and Strategic Misrepresentation}
\taxnode{1}{Uncertainty Communication and Confidence Elicitation}
\taxnode{1}{Evaluation, Benchmarks, and Metrics}
\taxnode{1}{Theoretical Foundations of Truthfulness and Uncertainty}
\taxnode{0}{Alignment Evaluation, Benchmarks, and Red Teaming}
\taxnode{1}{Benchmark Construction and Dataset Curation}
\taxnode{1}{Evaluation Methodology and Measurement}
\taxnode{1}{Human and Automated Evaluation}
\taxnode{1}{Adversarial Red Teaming}
\taxnode{1}{Preference, Helpfulness, and Instruction-Following Evaluation}
\taxnode{1}{Safety, Toxicity, Bias, and Misuse Evaluation}
\taxnode{1}{Truthfulness, Honesty, and Calibration Evaluation}
\taxnode{1}{Robustness and Distribution-Shift Evaluation}
\taxnode{1}{Deception, Goal Misgeneralization, and Specification-Gaming Evaluation}
\taxnode{1}{Agentic and Long-Horizon Alignment Evaluation}
\taxnode{1}{Meta-Evaluation and Benchmark Validation}
\taxnode{0}{Multi-Agent and Social Alignment}
\taxnode{1}{Cooperative Multi-Agent Learning}
\taxnode{1}{Competitive and Mixed-Motive Learning}
\taxnode{1}{Communication and Negotiation}
\taxnode{1}{Social Choice and Preference Aggregation}
\taxnode{1}{Mechanism Design and Incentive Alignment}
\taxnode{1}{Norm Learning and Convention Formation}
\taxnode{1}{Human-AI Collective Decision-Making}
\taxnode{1}{Fairness and Social Welfare Optimization}
\taxnode{1}{Deliberation, Debate, and Multi-Agent Oversight}
\taxnode{1}{Game-Theoretic Foundations of Social Alignment}
\taxnode{1}{Benchmarks and Evaluation for Social Alignment}
\taxnode{0}{Formal Alignment Theory}
\taxnode{1}{Objective Specification and Reward Design}
\taxnode{1}{Preference Learning and Inverse Reinforcement Learning}
\taxnode{1}{Assistance Games and Human-AI Cooperation}
\taxnode{1}{Corrigibility and Interruptibility}
\taxnode{1}{Inner Alignment and Mesa-Optimization}
\taxnode{1}{Decision Theory and Embedded Agency}
\taxnode{1}{Delegation and Scalable Oversight}
\taxnode{1}{Multi-Agent Alignment and Social Choice}
\taxnode{1}{Formal Verification and Alignment Guarantees}
\taxnode{1}{Alignment Evaluation and Benchmarking}
\taxnode{1}{Impossibility Results and Alignment Limits}
\taxfield{applications}
\taxnode{0}{Computer Vision}
\taxnode{1}{Visual Recognition}
\taxnode{1}{Object Detection}
\taxnode{1}{Image Segmentation}
\taxnode{1}{Pose Estimation and Human-Centered Vision}
\taxnode{1}{Tracking and Motion Analysis}
\taxnode{1}{Video Understanding}
\taxnode{1}{3D Vision and Geometric Vision}
\taxnode{1}{Low-Level Vision and Image Restoration}
\taxnode{1}{Image Generation and Editing}
\taxnode{1}{Scene Understanding and Visual Reasoning}
\taxnode{1}{Vision Representation Learning}
\taxnode{1}{Vision-Language and Multimodal Learning}
\taxnode{1}{Robustness, Fairness, and Interpretability in Vision}
\taxnode{1}{Vision Datasets, Benchmarks, and Evaluation}
\taxnode{1}{Applied Computer Vision}
\taxnode{0}{Natural Language Processing}
\taxnode{1}{Language Modeling}
\taxnode{1}{Parsing and Linguistic Structure}
\taxnode{1}{Text Classification}
\taxnode{1}{Information Extraction}
\taxnode{1}{Question Answering and Reading Comprehension}
\taxnode{1}{Machine Translation}
\taxnode{1}{Summarization}
\taxnode{1}{Dialogue Systems}
\taxnode{1}{Natural Language Generation}
\taxnode{1}{Knowledge-Intensive NLP}
\taxnode{1}{Multilingual and Cross-Lingual NLP}
\taxnode{1}{Evaluation, Benchmarks, and Resources}
\taxnode{2}{Automatic Evaluation Metrics}
\taxnode{2}{Human Evaluation}
\taxnode{2}{Meta-Evaluation and Benchmark Analysis}
\taxnode{2}{Diagnostic Evaluation and Challenge Sets}
\taxnode{2}{Benchmark Datasets}
\taxnode{2}{Corpora and Dataset Creation}
\taxnode{2}{Annotation Methodology}
\taxnode{2}{Lexical and Knowledge Resources}
\taxnode{2}{Shared Tasks and Evaluation Platforms}
\taxnode{2}{Data Documentation, Bias, and Ethics}
\taxnode{1}{Interpretability and Analysis}
\taxnode{1}{Robustness, Fairness, and Safety}
\taxnode{1}{Efficient NLP}
\taxnode{0}{Speech and Audio Processing}
\taxnode{1}{Automatic Speech Recognition}
\taxnode{1}{Speech Translation and Cross-Lingual Speech Processing}
\taxnode{1}{Spoken Language Understanding and Keyword Spotting}
\taxnode{1}{Speech Synthesis and Voice Conversion}
\taxnode{1}{Speech Enhancement, Dereverberation, and Source Separation}
\taxnode{1}{Speaker and Language Recognition}
\taxnode{1}{Paralinguistic Speech Processing}
\taxnode{1}{Audio Event Detection and Sound Scene Analysis}
\taxnode{1}{Music Information Retrieval and Music Audio Modeling}
\taxnode{1}{General Audio Generation and Editing}
\taxnode{1}{Audio-Visual Speech and Sound Processing}
\taxnode{1}{Speech and Audio Representation Learning}
\taxnode{1}{Datasets, Benchmarks, and Evaluation for Speech and Audio}
\taxnode{0}{Information Retrieval and Search}
\taxnode{1}{Learning to Rank and Re-ranking}
\taxnode{1}{Neural and Dense Retrieval}
\taxnode{1}{Query Understanding and Reformulation}
\taxnode{1}{Question Answering and Knowledge-Intensive Retrieval}
\taxnode{1}{Conversational and Interactive Search}
\taxnode{1}{Personalized and Context-Aware Search}
\taxnode{1}{Multimodal Retrieval}
\taxnode{1}{Cross-Lingual and Multilingual Retrieval}
\taxnode{1}{Vertical and Domain-Specific Search}
\taxnode{1}{IR Evaluation, Benchmarks, and User Behavior Modeling}
\taxnode{0}{Recommender Systems}
\taxnode{1}{Collaborative Filtering}
\taxnode{1}{Content-Based and Multimodal Recommendation}
\taxnode{1}{Context-Aware Recommendation}
\taxnode{1}{Sequential and Session-Based Recommendation}
\taxnode{1}{Graph-Based Recommendation}
\taxnode{1}{Knowledge-Aware Recommendation}
\taxnode{1}{Social Recommendation}
\taxnode{1}{Cross-Domain Recommendation}
\taxnode{1}{Cold-Start and Long-Tail Recommendation}
\taxnode{1}{Conversational Recommendation}
\taxnode{1}{Bandit and Reinforcement Learning for Recommendation}
\taxnode{1}{Explainable Recommendation}
\taxnode{1}{Causal and Counterfactual Recommendation}
\taxnode{1}{Fair and Responsible Recommendation}
\taxnode{1}{Privacy and Security in Recommender Systems}
\taxnode{1}{Evaluation and Benchmarking in Recommender Systems}
\taxnode{0}{Healthcare and Biomedicine}
\taxnode{1}{Medical Image Analysis}
\taxnode{1}{Digital Pathology and Computational Microscopy}
\taxnode{1}{Biomedical Signal Processing and Physiological Time Series}
\taxnode{1}{Clinical Natural Language Processing}
\taxnode{1}{Electronic Health Records and Clinical Prediction}
\taxnode{1}{Computational Genomics and Multi-omics}
\taxnode{1}{Drug Discovery and Molecular Design}
\taxnode{1}{Precision Medicine and Treatment Effect Modeling}
\taxnode{1}{Public Health Informatics and Epidemiological Modeling}
\taxnode{1}{Trustworthy and Privacy-Preserving Medical AI}
\taxnode{1}{Clinical Evaluation, Benchmarking, and Translation}
\taxnode{0}{Robotics and Autonomous Systems}
\taxnode{1}{Robot Perception and Scene Understanding}
\taxnode{1}{Localization, Mapping, and State Estimation}
\taxnode{1}{Navigation and Exploration}
\taxnode{1}{Manipulation and Grasping}
\taxnode{1}{Motion Planning and Trajectory Optimization}
\taxnode{1}{Robot Learning and Control}
\taxnode{1}{Human-Robot Interaction and Shared Autonomy}
\taxnode{1}{Multi-Robot Systems and Swarm Robotics}
\taxnode{1}{Autonomous Vehicles and Intelligent Transportation}
\taxnode{1}{Simulation, Sim-to-Real Transfer, and Robotics Benchmarks}
\taxnode{1}{Safe, Robust, and Trustworthy Autonomous Systems}
\taxnode{0}{Time-Series Analysis and Forecasting}
\taxnode{1}{Forecasting}
\taxnode{1}{Time-Series Classification and Regression}
\taxnode{1}{Anomaly Detection and Change-Point Detection}
\taxnode{1}{Segmentation, Clustering, and Pattern Discovery}
\taxnode{1}{Imputation, Denoising, and Reconstruction}
\taxnode{1}{Representation Learning for Time Series}
\taxnode{1}{Generative Modeling and Simulation}
\taxnode{1}{Spatiotemporal Modeling}
\taxnode{1}{Continuous-Time Modeling and Temporal Point Processes}
\taxnode{1}{Causal and Structural Time-Series Modeling}
\taxnode{1}{Interpretability and Reliability}
\taxnode{1}{Benchmarks, Datasets, and Evaluation}
\taxnode{1}{Domain-Specific Applications}
\taxnode{1}{Theory and Statistical Foundations}
\taxnode{1}{Surveys and Reviews}
\taxnode{0}{Scientific and Engineering Applications}
\taxnode{1}{Scientific Machine Learning Theory and Foundations}
\taxnode{1}{Physics-Informed and Mechanistic Learning}
\taxnode{1}{Surrogate Modeling and Operator Learning}
\taxnode{1}{Inverse Problems and Data Assimilation}
\taxnode{1}{Scientific Discovery and Equation Learning}
\taxnode{1}{Uncertainty Quantification and Reliability}
\taxnode{1}{Scientific Benchmarks, Datasets, and Evaluation}
\taxnode{1}{Physics and Astronomy Applications}
\taxnode{1}{Chemistry and Materials Science Applications}
\taxnode{1}{Earth, Climate, and Environmental Science Applications}
\taxnode{1}{Energy and Power Systems Applications}
\taxnode{1}{Mechanical and Aerospace Engineering Applications}
\taxnode{1}{Civil and Structural Engineering Applications}
\taxnode{1}{Electrical, Electronic, and Photonic Engineering Applications}
\taxnode{1}{Manufacturing and Industrial Engineering Applications}
\taxnode{0}{Geospatial and Remote Sensing}
\taxnode{1}{Remote Sensing Scene Understanding and Mapping}
\taxnode{1}{Object Detection, Localization, and Counting in Overhead Imagery}
\taxnode{1}{Change Detection and Monitoring}
\taxnode{1}{Spatiotemporal Geospatial Modeling and Forecasting}
\taxnode{1}{Hyperspectral and Multispectral Image Analysis}
\taxnode{1}{Synthetic Aperture Radar (SAR) Analysis}
\taxnode{1}{LiDAR and 3D Geospatial Learning}
\taxnode{1}{Multimodal and Multisensor Fusion}
\taxnode{1}{Image Restoration, Super-Resolution, and Pansharpening}
\taxnode{1}{Geolocation, Registration, and Cross-view Geo-localization}
\taxnode{1}{Spatial Graph Learning and Mobility Modeling}
\taxnode{1}{Geospatial Representation Learning and Foundation Models}
\taxnode{1}{Domain Adaptation, Transfer Learning, and Generalization}
\taxnode{1}{Earth System and Societal Applications}
\taxnode{1}{Datasets, Benchmarks, Evaluation, and Survey Studies}
\taxnode{0}{Finance and Economics}
\taxnode{1}{Financial Time-Series Forecasting}
\taxnode{1}{Asset and Derivative Pricing}
\taxnode{1}{Portfolio Management and Asset Allocation}
\taxnode{1}{Market Microstructure and Algorithmic Trading}
\taxnode{1}{Credit Risk and Lending Analytics}
\taxnode{1}{Risk Management and Stress Testing}
\taxnode{1}{Fraud Detection and Financial Crime}
\taxnode{1}{Corporate Finance and Financial Distress}
\taxnode{1}{Insurance and Actuarial Modeling}
\taxnode{1}{Financial Text and Alternative Data Analysis}
\taxnode{1}{Macroeconomic Forecasting and Nowcasting}
\taxnode{1}{Structural Econometrics and Equilibrium Modeling}
\taxnode{1}{Causal Inference and Policy Evaluation}
\taxnode{1}{Demand Estimation and Pricing}
\taxnode{1}{Auctions and Market Design}
\taxnode{1}{Economic Measurement and Socioeconomic Analysis}
\taxnode{0}{Cybersecurity}
\taxnode{1}{Network Intrusion Detection and Traffic Analysis}
\taxnode{1}{Malware Detection and Analysis}
\taxnode{1}{Phishing, Spam, and Malicious Content Detection}
\taxnode{1}{Vulnerability Detection and Software Security}
\taxnode{1}{Host-Based Intrusion Detection and User Behavior Analytics}
\taxnode{1}{Authentication, Access Control, and Biometric Security}
\taxnode{1}{Threat Intelligence, Attack Attribution, and Security Operations}
\taxnode{1}{Digital Forensics and Incident Response}
\taxnode{1}{Adversarial Machine Learning and Secure AI Systems}
\taxnode{1}{Cyber-Physical, IoT, and Industrial Control Security}
\taxnode{1}{Cybersecurity Datasets, Benchmarks, and Evaluation}
\taxnode{0}{Industrial and Manufacturing Systems}
\taxnode{1}{Predictive Maintenance and Remaining Useful Life Estimation}
\taxnode{1}{Fault Diagnosis and Anomaly Detection}
\taxnode{1}{Quality Inspection and Defect Detection}
\taxnode{1}{Process Monitoring, Modeling, and Control}
\taxnode{1}{Production Planning, Scheduling, and Shop-Floor Optimization}
\taxnode{1}{Industrial Robotics and Autonomous Manufacturing}
\taxnode{1}{Digital Twins and Smart Manufacturing Systems}
\taxnode{1}{Supply Chain, Logistics, and Inventory Analytics}
\taxnode{1}{Energy Management and Sustainable Manufacturing}
\taxnode{1}{Industrial IoT, Sensor Fusion, and Edge Intelligence}
\taxnode{1}{Industrial Machine Learning Methodology, Datasets, and Benchmarks}
\taxnode{1}{Surveys and Reviews of Intelligent Manufacturing}
\taxnode{0}{Transportation and Mobility Systems}
\taxnode{1}{Traffic Flow Forecasting and State Estimation}
\taxnode{1}{Traffic Control and Network Operations}
\taxnode{1}{Mobility Demand Forecasting and Human Mobility Analysis}
\taxnode{1}{Routing, Navigation, and Travel Time Estimation}
\taxnode{1}{Public Transit and Multimodal Transportation Analytics}
\taxnode{1}{Fleet Management, Ride-Hailing, and Shared Mobility}
\taxnode{1}{Logistics and Freight Transportation}
\taxnode{1}{Autonomous and Connected Vehicle Systems}
\taxnode{1}{Transportation Safety, Incident Detection, and Risk Assessment}
\taxnode{1}{Transportation Infrastructure Monitoring and Asset Management}
\taxnode{1}{Electric Mobility and Sustainable Transportation}
\taxnode{1}{Transportation Data, Benchmarks, and Simulation}
\taxnode{0}{Computational Social Science and Digital Platforms}
\taxnode{1}{Social Network Analysis}
\taxnode{1}{Information Diffusion and Cascade Modeling}
\taxnode{1}{User Behavior Modeling}
\taxnode{1}{Social Media Mining and Content Understanding}
\taxnode{1}{Opinion Mining, Public Discourse, and Political Communication}
\taxnode{1}{Misinformation, Content Moderation, and Trust \& Safety}
\taxnode{1}{Recommender Systems and Personalization}
\taxnode{1}{Online Communities and Collective Behavior}
\taxnode{1}{Platform Economics and Marketplace Modeling}
\taxnode{1}{Human Mobility and Spatiotemporal Social Dynamics}
\taxnode{1}{Causal Inference and Policy Evaluation in Social Systems}
\taxnode{1}{Fairness, Privacy, and Ethics in Digital Platforms}
\taxnode{1}{Datasets, Benchmarks, and Measurement}
\taxfield{architecture}
\taxnode{0}{Feedforward, Residual, and MLP Architectures}
\taxnode{1}{Multilayer Perceptrons}
\taxnode{1}{Residual Networks and Skip-Connection Architectures}
\taxnode{1}{All-MLP Architectures}
\taxnode{1}{Architectural Design and Connectivity Patterns}
\taxnode{1}{Expressivity and Approximation Theory}
\taxnode{1}{Optimization and Training Dynamics}
\taxnode{1}{Initialization, Normalization, and Regularization}
\taxnode{1}{Generalization, Robustness, and Representation Analysis}
\taxnode{1}{Efficient and Compressed Feedforward Models}
\taxnode{1}{Empirical Scaling and Benchmarking}
\taxnode{1}{Applications of Feedforward, Residual, and MLP Architectures}
\taxnode{1}{Surveys and Comparative Studies}
\taxnode{0}{Convolutional Architectures}
\taxnode{1}{Convolutional Operators and Building Blocks}
\taxnode{1}{Residual and Densely Connected Networks}
\taxnode{1}{Fully Convolutional and Encoder-Decoder Networks}
\taxnode{1}{Multi-Scale and Pyramid Networks}
\taxnode{1}{Efficient and Lightweight Convolutional Networks}
\taxnode{1}{3D and Spatiotemporal Convolutional Networks}
\taxnode{1}{Generative Convolutional Models}
\taxnode{1}{Neural Architecture Search for CNNs}
\taxnode{1}{Theoretical Analysis of Convolutional Networks}
\taxnode{1}{Training and Optimization of Convolutional Networks}
\taxnode{1}{Transfer Learning and Representation Learning with CNNs}
\taxnode{1}{Robustness, Interpretability, and Uncertainty in CNNs}
\taxnode{1}{Benchmarks and Empirical Evaluation of CNNs}
\taxnode{1}{Application-Specific Convolutional Networks}
\taxnode{1}{Surveys and Reviews of Convolutional Architectures}
\taxnode{0}{Recurrent and State-Space Architectures}
\taxnode{1}{Recurrent Neural Networks}
\taxnode{1}{Gated Recurrent Networks}
\taxnode{1}{Recurrent Encoder-Decoder Models}
\taxnode{1}{Memory-Augmented Recurrent Networks}
\taxnode{1}{Reservoir Computing}
\taxnode{1}{Latent and Neural State-Space Models}
\taxnode{1}{Structured State Space Models}
\taxnode{1}{Continuous-Time Recurrent and State-Space Models}
\taxnode{1}{Training and Optimization of Recurrent and State-Space Models}
\taxnode{1}{Theory and Expressivity}
\taxnode{1}{Benchmarking and Empirical Evaluation}
\taxnode{1}{Applications of Recurrent and State-Space Architectures}
\taxnode{1}{Surveys and Systematic Reviews}
\taxnode{0}{Attention-Based and Transformer Architectures}
\taxnode{1}{Attention Mechanisms}
\taxnode{1}{Positional and Structural Encoding}
\taxnode{1}{Transformer Architectural Variants}
\taxnode{1}{Efficient and Scalable Transformers}
\taxnode{1}{Long-Context and Memory-Augmented Transformers}
\taxnode{1}{Training, Pre-training, and Adaptation}
\taxnode{2}{Pre-training Objectives}
\taxnode{2}{Pre-training Data and Curriculum}
\taxnode{2}{Optimization, Regularization, and Training Dynamics}
\taxnode{2}{Scaling Laws and Compute-Optimal Training}
\taxnode{2}{Efficient and Distributed Training}
\taxnode{2}{Knowledge Distillation and Teacher-Student Training}
\taxnode{2}{Fine-Tuning and Transfer Learning}
\taxnode{2}{Adapter and Low-Rank Adaptation}
\taxnode{2}{Prompt and Prefix Tuning}
\taxnode{2}{Domain Adaptation and Continual Learning}
\taxnode{2}{Instruction Tuning and Alignment}
\taxnode{2}{Theoretical Analysis of Pre-training and Adaptation}
\taxnode{2}{Evaluation and Benchmarking of Training and Adaptation}
\taxnode{1}{Compression and Deployment}
\taxnode{1}{Theory of Attention and Transformers}
\taxnode{1}{Interpretability and Analysis}
\taxnode{1}{Robustness, Safety, and Privacy}
\taxnode{1}{Evaluation and Benchmarking}
\taxnode{1}{Language Transformers}
\taxnode{1}{Vision Transformers}
\taxnode{1}{Speech and Audio Transformers}
\taxnode{1}{Multimodal Transformers}
\taxnode{1}{Graph and Structured Data Transformers}
\taxnode{0}{Graph Neural Architectures}
\taxnode{1}{Spectral Graph Neural Networks}
\taxnode{1}{Message Passing Neural Networks}
\taxnode{1}{Graph Attention Networks}
\taxnode{1}{Graph Transformers}
\taxnode{1}{Hierarchical and Pooling Graph Neural Networks}
\taxnode{1}{Heterogeneous and Relational Graph Neural Networks}
\taxnode{1}{Temporal and Dynamic Graph Neural Networks}
\taxnode{1}{Hypergraph and Higher-Order Graph Neural Networks}
\taxnode{1}{Geometric and Equivariant Graph Neural Networks}
\taxnode{1}{Graph Autoencoders and Generative Models}
\taxnode{1}{Scalable and Efficient Graph Neural Architectures}
\taxnode{1}{Theory and Expressivity of Graph Neural Architectures}
\taxnode{1}{Graph Neural Architecture Search and Design}
\taxnode{1}{Empirical Evaluation and Benchmarking of Graph Neural Architectures}
\taxnode{1}{Applications of Graph Neural Architectures}
\taxnode{1}{Surveys and Taxonomies of Graph Neural Architectures}
\taxnode{0}{Geometric and Equivariant Architectures}
\taxnode{1}{Graph Neural Networks}
\taxnode{1}{Set and Permutation-Equivariant Networks}
\taxnode{1}{Point Cloud and 3D Geometric Networks}
\taxnode{1}{Mesh and Surface Neural Networks}
\taxnode{1}{Group-Equivariant Neural Networks}
\taxnode{1}{E(n)- and SE(3)-Equivariant Neural Networks}
\taxnode{1}{Gauge-Equivariant and Manifold Neural Networks}
\taxnode{1}{Hyperbolic and Non-Euclidean Neural Networks}
\taxnode{1}{Simplicial, Cellular, and Topological Neural Networks}
\taxnode{1}{Symmetry Discovery and Learned Equivariance}
\taxnode{1}{Theory of Geometric and Equivariant Learning}
\taxnode{1}{Benchmarks and Empirical Evaluation for Geometric Learning}
\taxnode{1}{Applications of Geometric and Equivariant Architectures}
\taxnode{1}{Surveys and Taxonomies of Geometric Deep Learning}
\taxnode{0}{Generative Model Architectures}
\taxnode{1}{Autoregressive Generative Models}
\taxnode{1}{Variational Autoencoders and Deep Latent Variable Models}
\taxnode{1}{Generative Adversarial Networks}
\taxnode{1}{Normalizing Flow Models}
\taxnode{1}{Diffusion and Score-Based Generative Models}
\taxnode{1}{Energy-Based and Undirected Generative Models}
\taxnode{1}{Masked and Denoising Generative Models}
\taxnode{0}{Multimodal Architectures}
\taxnode{1}{Multimodal Fusion Architectures}
\taxnode{1}{Joint Embedding and Alignment Models}
\taxnode{1}{Cross-Attention and Multimodal Transformer Architectures}
\taxnode{1}{Encoder-Decoder and Sequence-to-Sequence Multimodal Models}
\taxnode{1}{Generative Multimodal Models}
\taxnode{1}{Multimodal Foundation Models}
\taxnode{1}{Modular and Mixture-of-Experts Multimodal Networks}
\taxnode{1}{Robust and Incomplete Multimodal Learning}
\taxnode{1}{Multimodal Reasoning and Grounding Architectures}
\taxnode{1}{Efficient and Scalable Multimodal Architectures}
\taxnode{1}{Analysis, Evaluation, and Benchmarking of Multimodal Architectures}
\taxnode{1}{Domain-Specific Multimodal Architectures}
\taxnode{0}{Modular and Mixture-of-Experts Architectures}
\taxnode{1}{Mixture-of-Experts Models}
\taxnode{1}{Routing and Conditional Computation}
\taxnode{1}{Modular Neural Networks}
\taxnode{1}{Neural Module Networks}
\taxnode{1}{Training and Optimization of Modular Architectures}
\taxnode{1}{Theory of Modularity and Expert Specialization}
\taxnode{1}{Scalable and Distributed MoE Systems}
\taxnode{1}{Transfer, Multi-Task, and Continual Modular Learning}
\taxnode{1}{Benchmarking and Empirical Evaluation}
\taxnode{1}{Applications of Modular and Mixture-of-Experts Architectures}
\taxnode{1}{Surveys and Taxonomies of Modular Architectures}
\taxnode{0}{Memory-Augmented and Neuro-Symbolic Architectures}
\taxnode{1}{External Memory Architectures}
\taxnode{1}{Associative, Episodic, and Working Memory Models}
\taxnode{1}{Retrieval-Augmented Models}
\taxnode{1}{Continual and Lifelong Learning with Memory}
\taxnode{1}{Neural-Symbolic Knowledge Representation}
\taxnode{1}{Neural-Symbolic Logical and Relational Reasoning}
\taxnode{1}{Program Induction and Differentiable Programming}
\taxnode{1}{Neural-Symbolic Planning and Decision Making}
\taxnode{1}{Training and Optimization Methods}
\taxnode{1}{Theory, Interpretability, and Compositional Generalization}
\taxnode{1}{Benchmarks and Empirical Evaluation}
\taxnode{1}{Applications of Memory-Augmented and Neuro-Symbolic Models}
\taxnode{0}{Efficient and Compact Architectures}
\taxnode{1}{Lightweight Convolutional Architectures}
\taxnode{1}{Efficient Transformer and Sequence Architectures}
\taxnode{1}{Dynamic and Conditional Computation}
\taxnode{1}{Sparse Architectures and Mixture-of-Experts}
\taxnode{1}{Low-Precision and Binary Neural Networks}
\taxnode{1}{Factorized and Low-Rank Architectures}
\taxnode{1}{Neural Architecture Search for Efficiency}
\taxnode{1}{Hardware-Aware and Edge-Oriented Architecture Design}
\taxnode{1}{Theoretical Foundations of Efficient Architectures}
\taxnode{1}{Benchmarking and Empirical Evaluation of Efficient Models}
\taxnode{1}{Application-Specific Efficient Architectures}
\taxnode{1}{Surveys and Comparative Studies of Compact Models}
\taxnode{0}{Neural Architecture Search and Automated Design}
\taxnode{1}{Search Space Design}
\taxnode{1}{Search Algorithms and Optimization}
\taxnode{1}{Efficient NAS and Performance Prediction}
\taxnode{1}{Multi-Objective and Hardware-Aware NAS}
\taxnode{1}{Transfer, Meta, and Few-Shot NAS}
\taxnode{1}{Joint Automated Design and AutoML}
\taxnode{1}{Theory and Foundations of NAS}
\taxnode{1}{Benchmarking, Evaluation, and Reproducibility}
\taxnode{1}{Domain-Specific and Applied NAS}
\taxnode{1}{Surveys and Empirical Studies of NAS}
\taxnode{0}{Architecture Theory and Analysis}
\taxnode{1}{Expressivity and Approximation Theory}
\taxnode{1}{Optimization and Training Dynamics}
\taxnode{1}{Generalization and Capacity Analysis}
\taxnode{1}{Signal Propagation and Dynamical Systems Analysis}
\taxnode{1}{Inductive Bias and Representation Analysis}
\taxnode{1}{Scaling Laws and Architectural Scaling}
\taxnode{1}{Robustness and Stability Analysis}
\taxnode{1}{Computational Complexity and Efficiency Analysis}
\taxnode{1}{Empirical Benchmarking and Ablation Studies}
\taxnode{1}{Task- and Domain-Specific Architectural Analysis}
\taxnode{1}{Architecture Taxonomy and Survey}
\taxnode{0}{Architecture Benchmarking and Empirical Evaluation}
\taxnode{1}{Benchmark Dataset and Task Suite Construction}
\taxnode{1}{Experimental Protocols and Statistical Evaluation}
\taxnode{1}{Comparative Benchmarking of Architecture Families}
\taxnode{1}{Ablation Studies and Component-wise Analysis}
\taxnode{1}{Scaling Behavior and Sensitivity Analysis}
\taxnode{1}{Efficiency, Compute, and Hardware-Aware Benchmarking}
\taxnode{1}{Robustness, Reliability, and Generalization Evaluation}
\taxnode{1}{Transferability and Cross-Domain Evaluation}
\taxnode{1}{Application-Specific and Real-World Evaluation}
\taxnode{1}{Reproducibility, Replicability, and Reporting Studies}
\taxnode{1}{Systematic Surveys and Meta-Analysis}
\taxnode{0}{Task-Specific and Domain-Specific Architectures}
\taxnode{1}{Architecture Design Methodologies}
\taxnode{1}{Theoretical Foundations of Specialized Architectures}
\taxnode{1}{Benchmarking and Empirical Evaluation}
\taxnode{1}{Surveys and Taxonomies}
\taxnode{1}{Computer Vision Architectures}
\taxnode{1}{Natural Language Processing Architectures}
\taxnode{1}{Speech and Audio Architectures}
\taxnode{1}{Multimodal Learning Architectures}
\taxnode{1}{Graph and Relational Learning Architectures}
\taxnode{1}{Time-Series and Forecasting Architectures}
\taxnode{1}{Tabular Data Architectures}
\taxnode{1}{Recommender System and Ranking Architectures}
\taxnode{1}{Biomedical and Healthcare Architectures}
\taxnode{1}{Scientific Machine Learning and Geospatial Architectures}
\taxnode{1}{Reinforcement Learning and Control Architectures}
\taxnode{0}{Surveys and Taxonomies of Neural Architectures}
\taxnode{1}{Cross-Architecture Taxonomies and Comparative Surveys}
\taxnode{1}{Convolutional Neural Architectures}
\taxnode{1}{Sequence Modeling Architectures}
\taxnode{1}{Transformer and Attention-Based Architectures}
\taxnode{1}{Graph Neural Architectures}
\taxnode{1}{Generative and Latent-Variable Architectures}
\taxnode{1}{Multimodal Neural Architectures}
\taxnode{1}{Efficient and Hardware-Aware Architectures}
\taxnode{1}{Neural Architecture Search}
\taxnode{1}{Spiking and Neuromorphic Architectures}
\taxnode{1}{Modular, Memory-Augmented, and Neuro-Symbolic Architectures}
\taxnode{1}{Domain-Specific Architecture Surveys}
\taxfield{data}
\taxnode{0}{Dataset Collection and Curation}
\taxnode{1}{Data Acquisition and Sampling}
\taxnode{1}{Data Annotation and Labeling Methodology}
\taxnode{1}{Dataset Cleaning, Filtering, and Deduplication}
\taxnode{1}{Dataset Quality Evaluation and Auditing}
\taxnode{1}{Bias, Fairness, and Representativeness in Datasets}
\taxnode{1}{Dataset Documentation, Metadata, and Provenance}
\taxnode{1}{Privacy, Ethics, and Data Governance}
\taxnode{1}{Benchmark Dataset Construction and Evaluation}
\taxnode{1}{Synthetic and Programmatically Generated Datasets}
\taxnode{1}{Dataset Maintenance, Versioning, and Lifecycle Management}
\taxnode{0}{Data Annotation and Supervision}
\taxnode{1}{Annotation Schema Design}
\taxnode{1}{Human Annotation and Crowdsourcing}
\taxnode{1}{Annotation Quality and Inter-Annotator Agreement}
\taxnode{1}{Label Aggregation and Annotator Modeling}
\taxnode{1}{Active Learning}
\taxnode{1}{Weak and Distant Supervision}
\taxnode{1}{Semi-Supervised Learning}
\taxnode{1}{Self-Supervised Learning}
\taxnode{1}{Learning with Noisy Labels}
\taxnode{1}{Human Feedback and Preference Supervision}
\taxnode{0}{Data Preprocessing and Cleaning}
\taxnode{1}{Data Quality Assessment and Profiling}
\taxnode{1}{Missing Data Imputation}
\taxnode{1}{Noise Reduction and Denoising}
\taxnode{1}{Outlier, Anomaly, and Inconsistency Handling}
\taxnode{1}{Duplicate Detection and Entity Resolution}
\taxnode{1}{Data Normalization, Scaling, and Transformation}
\taxnode{1}{Encoding, Discretization, and Type Conversion}
\taxnode{1}{Label Cleaning and Noisy Annotation Handling}
\taxnode{1}{Imbalanced Data Handling and Resampling}
\taxnode{1}{Data Integration, Harmonization, and Batch Effect Correction}
\taxnode{1}{Automated, Interactive, and Human-in-the-Loop Data Cleaning}
\taxnode{1}{Foundations, Benchmarks, and Evaluation of Data Preprocessing}
\taxnode{0}{Data Augmentation and Transformation}
\taxnode{1}{Spatial and Geometric Augmentation}
\taxnode{1}{Photometric and Signal-Level Augmentation}
\taxnode{1}{Occlusion and Corruption-Based Augmentation}
\taxnode{1}{Sample Mixing and Interpolation Methods}
\taxnode{1}{Feature-Space and Latent-Space Augmentation}
\taxnode{1}{Generative and Synthetic Data Augmentation}
\taxnode{1}{Automated Augmentation Policy Learning}
\taxnode{1}{Adversarial and Robustness-Oriented Transformations}
\taxnode{1}{Modality- and Task-Specific Augmentation}
\taxnode{1}{Theoretical Foundations of Augmentation}
\taxnode{1}{Empirical Evaluation, Benchmarks, and Best Practices}
\taxnode{1}{Test-Time Augmentation}
\taxnode{0}{Synthetic Data Generation}
\taxnode{1}{Generative Modeling Methods}
\taxnode{1}{Simulation and Procedural Generation}
\taxnode{1}{Data Augmentation and Counterfactual Synthesis}
\taxnode{1}{Visual Synthetic Data}
\taxnode{1}{Text, Speech, and Multimodal Synthetic Data}
\taxnode{1}{Tabular and Relational Synthetic Data}
\taxnode{1}{Time-Series and Sensor Synthetic Data}
\taxnode{1}{Graph and Structured Synthetic Data}
\taxnode{1}{3D and Spatial Synthetic Data}
\taxnode{1}{Privacy-Preserving and Fair Synthetic Data}
\taxnode{1}{Evaluation, Benchmarking, and Empirical Studies}
\taxnode{1}{Theoretical Foundations of Synthetic Data}
\taxnode{1}{Domain-Specific Applications of Synthetic Data}
\taxnode{0}{Data Quality and Label Noise}
\taxnode{1}{Label Noise Models and Theory}
\taxnode{1}{Noise-Robust Learning}
\taxnode{1}{Label Error Detection and Correction}
\taxnode{1}{Annotator Modeling and Crowdsourcing}
\taxnode{1}{Weak Supervision and Label Aggregation}
\taxnode{1}{Ambiguous, Partial, and Soft Label Learning}
\taxnode{1}{Feature Noise, Missing Data, and Input Corruption}
\taxnode{1}{Data Quality Assessment, Auditing, and Benchmarking}
\taxnode{1}{Domain-Specific Learning with Imperfect Annotations}
\taxnode{0}{Data Selection and Valuation}
\taxnode{1}{Active Learning}
\taxnode{1}{Coreset Selection and Data Pruning}
\taxnode{1}{Curriculum Learning and Self-Paced Learning}
\taxnode{1}{Data Valuation and Data Pricing}
\taxnode{1}{Training Data Attribution and Influence Estimation}
\taxnode{1}{Importance Sampling and Sample Reweighting}
\taxnode{1}{Theoretical Foundations and Guarantees}
\taxnode{1}{Benchmarks and Empirical Evaluation}
\taxnode{1}{Applications and Systems}
\taxnode{1}{Surveys and Taxonomies}
\taxnode{0}{Dataset Bias, Fairness, and Representativeness}
\taxnode{1}{Bias Taxonomies and Theory}
\taxnode{1}{Sampling and Selection Bias}
\taxnode{1}{Representation Bias and Coverage}
\taxnode{1}{Annotation and Label Bias}
\taxnode{1}{Measurement Bias}
\taxnode{1}{Evaluation Bias and Benchmark Design}
\taxnode{1}{Dataset Auditing and Documentation}
\taxnode{1}{Fair Data Collection and Curation}
\taxnode{1}{Data Debiasing and Rebalancing}
\taxnode{1}{Distribution Shift and Subpopulation Representativeness}
\taxnode{0}{Dataset Shift and Distributional Robustness}
\taxnode{1}{Covariate Shift}
\taxnode{1}{Label Shift and Prior Probability Shift}
\taxnode{1}{Concept Shift and Concept Drift}
\taxnode{1}{Domain Adaptation}
\taxnode{1}{Domain Generalization}
\taxnode{1}{Test-Time Adaptation}
\taxnode{1}{Out-of-Distribution Detection and Open-Set Recognition}
\taxnode{1}{Distributionally Robust Optimization}
\taxnode{1}{Subpopulation Shift and Group Robustness}
\taxnode{1}{Causal Invariance and Stable Prediction}
\taxnode{1}{Uncertainty Estimation and Calibration under Shift}
\taxnode{1}{Shift Detection, Estimation, and Drift Monitoring}
\taxnode{1}{Theoretical Foundations of Generalization under Shift}
\taxnode{1}{Robustness Benchmarks and Evaluation}
\taxnode{0}{Benchmarks and Dataset Evaluation}
\taxnode{1}{Dataset Characterization and Quality Assessment}
\taxnode{1}{Annotation and Label Quality Evaluation}
\taxnode{1}{Dataset Bias, Fairness, and Representativeness}
\taxnode{1}{Split Integrity, Leakage, and Redundancy Analysis}
\taxnode{1}{Benchmark Design and Evaluation Protocols}
\taxnode{1}{Robustness, Distribution Shift, and Stress-Test Benchmarks}
\taxnode{1}{Synthetic, Simulated, and Diagnostic Benchmarks}
\taxnode{1}{Benchmark Validity, Reliability, and Meta-Evaluation}
\taxnode{1}{Dataset Documentation, Provenance, and Governance}
\taxnode{1}{Domain-Specific and Application Benchmarks}
\taxnode{1}{Survey and Comparative Studies of Benchmarks}
\taxnode{0}{Data Privacy, Security, and Governance}
\taxnode{1}{Differential Privacy for Machine Learning}
\taxnode{1}{Privacy-Preserving Federated and Distributed Learning}
\taxnode{1}{Secure Computation and Confidential Machine Learning}
\taxnode{1}{Privacy Attacks, Leakage, and Auditing}
\taxnode{1}{Data Anonymization, De-identification, and Synthetic Data}
\taxnode{1}{Training-Data Security: Poisoning and Backdoors}
\taxnode{2}{Availability Poisoning Attacks}
\taxnode{2}{Targeted and Integrity Poisoning Attacks}
\taxnode{2}{Backdoor and Trojan Attacks}
\taxnode{2}{Clean-Label and Stealthy Poisoning}
\taxnode{2}{Poisoning and Backdoors in Federated and Distributed Learning}
\taxnode{2}{Poisoning and Backdoors in Specialized Learning Paradigms}
\taxnode{2}{Poisoning Detection and Data Sanitization}
\taxnode{2}{Backdoor Detection and Model Forensics}
\taxnode{2}{Robust Training, Mitigation, and Machine Unlearning}
\taxnode{2}{Certified Robustness and Theoretical Analysis}
\taxnode{2}{Benchmarks, Datasets, and Evaluation Protocols}
\taxnode{2}{Surveys, Taxonomies, and Systematization}
\taxnode{1}{Data Provenance, Lineage, and Documentation}
\taxnode{1}{Data Governance, Access Control, and Compliance}
\taxfield{evaluation}
\taxnode{0}{Evaluation Metrics}
\taxnode{1}{Predictive Performance Metrics}
\taxnode{1}{Ranking, Retrieval, and Recommendation Metrics}
\taxnode{1}{Decision-Making and Reinforcement Learning Metrics}
\taxnode{1}{Structured Prediction and Sequence Generation Metrics}
\taxnode{1}{Generative Model Evaluation Metrics}
\taxnode{1}{Representation and Embedding Evaluation Metrics}
\taxnode{1}{Calibration and Uncertainty Metrics}
\taxnode{1}{Robustness, Generalization, and Safety Metrics}
\taxnode{1}{Fairness, Bias, and Responsible AI Metrics}
\taxnode{1}{Efficiency, Scalability, and Sustainability Metrics}
\taxnode{1}{Interpretability and Explanation Metrics}
\taxnode{1}{Human-Centered and Preference-Based Evaluation}
\taxnode{1}{Metric Theory and Statistical Validity}
\taxnode{1}{Meta-Evaluation and Metric Benchmarking}
\taxnode{1}{Metric Design and Learned Metrics}
\taxnode{0}{Benchmarking and Comparative Evaluation}
\taxnode{1}{Benchmark Dataset and Task Design}
\taxnode{1}{Evaluation Metrics and Protocols}
\taxnode{2}{Predictive Performance Metrics}
\taxnode{2}{Probabilistic Evaluation and Calibration}
\taxnode{2}{Meta-Evaluation and Metric Validation}
\taxnode{2}{Statistical Comparison and Significance Testing}
\taxnode{2}{Experimental Design and Validation Protocols}
\taxnode{2}{Benchmark and Test Suite Design}
\taxnode{2}{Robustness and Distribution Shift Evaluation}
\taxnode{2}{Fairness and Bias Evaluation}
\taxnode{2}{Efficiency and Resource-Aware Evaluation}
\taxnode{2}{Human Evaluation and Preference Studies}
\taxnode{2}{Reproducibility and Reporting Standards}
\taxnode{1}{Comparative Model and Algorithm Studies}
\taxnode{2}{Empirical Benchmark Studies}
\taxnode{2}{Cross-Dataset Generalization and Transfer Studies}
\taxnode{2}{Efficiency and Scalability Evaluation}
\taxnode{2}{Robustness and Reliability Evaluation}
\taxnode{2}{Fairness, Bias, and Safety Evaluation}
\taxnode{2}{Theoretical Comparative Analysis}
\taxnode{2}{Statistical Evaluation and Comparison Methodology}
\taxnode{2}{Reproducibility and Replication Studies}
\taxnode{2}{Application-Specific Comparative Studies}
\taxnode{2}{Systematic Reviews and Meta-Analyses}
\taxnode{1}{Reproducibility and Replicability Studies}
\taxnode{1}{Robustness and Stress Testing Benchmarks}
\taxnode{1}{Efficiency and Resource Benchmarking}
\taxnode{1}{Generalization and Transfer Benchmarks}
\taxnode{1}{Fairness, Safety, Privacy, and Security Benchmarking}
\taxnode{1}{Domain-Specific Application Benchmarks}
\taxnode{1}{Benchmark Auditing and Meta-Evaluation}
\taxnode{1}{Benchmark Surveys and Meta-Analyses}
\taxnode{0}{Experimental Design and Statistical Evaluation}
\taxnode{1}{Data Splitting and Sampling Protocols}
\taxnode{1}{Model Selection and Validation Procedures}
\taxnode{2}{Data partitioning and holdout strategies}
\taxnode{2}{Cross-validation methods}
\taxnode{2}{Bootstrap and resampling-based validation}
\taxnode{2}{Nested validation and selection bias control}
\taxnode{2}{Information-theoretic and Bayesian model selection criteria}
\taxnode{2}{Hyperparameter and architecture selection}
\taxnode{2}{Early stopping and adaptive stopping rules}
\taxnode{2}{Multi-fidelity and budget-aware selection}
\taxnode{2}{Validation for dependent, structured, and limited data}
\taxnode{2}{External validation and distribution-shift-aware selection}
\taxnode{2}{Statistical inference for model comparison}
\taxnode{2}{Theoretical foundations of model selection and validation}
\taxnode{2}{Empirical benchmarking of validation procedures}
\taxnode{2}{Applied and domain-specific validation protocols}
\taxnode{2}{Surveys and meta-analyses of validation practice}
\taxnode{1}{Baseline, Control, and Ablation Design}
\taxnode{1}{Benchmarking and Comparative Evaluation Protocols}
\taxnode{2}{Benchmark Suite Construction}
\taxnode{2}{Evaluation Metrics and Composite Scoring}
\taxnode{2}{Baselines, Reference Implementations, and Tuning Protocols}
\taxnode{2}{Statistical Methodology for Comparative Evaluation}
\taxnode{2}{Systematic Comparative Benchmark Studies}
\taxnode{2}{Reproducibility and Reporting Standards}
\taxnode{2}{Robustness and Generalization Benchmarking}
\taxnode{2}{Efficiency and Resource-Aware Benchmarking}
\taxnode{2}{Fairness, Safety, and Responsible Evaluation}
\taxnode{2}{Meta-Evaluation and Benchmark Validity}
\taxnode{2}{Dynamic and Continual Benchmarking}
\taxnode{1}{Statistical Hypothesis Testing}
\taxnode{1}{Effect Size and Uncertainty Quantification}
\taxnode{1}{Power Analysis and Sample Size Determination}
\taxnode{1}{Multiple Comparisons and Error Control}
\taxnode{1}{Robustness and Sensitivity Analysis}
\taxnode{1}{Reproducibility and Replicability Assessment}
\taxnode{0}{Generalization and Distribution Shift Evaluation}
\taxnode{1}{Statistical Generalization Theory}
\taxnode{1}{Empirical Generalization Diagnostics}
\taxnode{1}{Dataset Shift Characterization and Metrics}
\taxnode{1}{Domain Generalization and Cross-Domain Evaluation}
\taxnode{1}{Out-of-Distribution Detection and Open-Set Evaluation}
\taxnode{1}{Uncertainty Quantification and Calibration Under Shift}
\taxnode{1}{Robustness to Corruptions and Adversarial Perturbations}
\taxnode{1}{Subpopulation Shift and Worst-Group Evaluation}
\taxnode{1}{Temporal Shift and Data Drift Evaluation}
\taxnode{1}{Benchmarks, Datasets, and Evaluation Protocols}
\taxnode{0}{Robustness and Adversarial Evaluation}
\taxnode{1}{Adversarial Attacks and Threat Models}
\taxnode{1}{Defense Evaluation and Robustness Auditing}
\taxnode{1}{Certified Robustness and Formal Verification}
\taxnode{1}{Corruption and Perturbation Robustness}
\taxnode{1}{Distribution Shift and Subpopulation Robustness}
\taxnode{1}{Out-of-Distribution and Open-Set Evaluation}
\taxnode{1}{Data Poisoning and Backdoor Robustness}
\taxnode{2}{Threat Models and Formal Foundations}
\taxnode{2}{Availability Poisoning Attacks}
\taxnode{2}{Targeted Poisoning Attacks}
\taxnode{2}{Backdoor and Trojan Attacks}
\taxnode{2}{Poisoned Data Detection and Sanitization}
\taxnode{2}{Backdoor Detection and Removal}
\taxnode{2}{Robust Training and Defense Algorithms}
\taxnode{2}{Certified Defenses and Robustness Guarantees}
\taxnode{2}{Federated and Distributed Poisoning and Backdoors}
\taxnode{2}{Benchmarks, Datasets, and Evaluation Protocols}
\taxnode{2}{Application- and Modality-Specific Robustness}
\taxnode{2}{Surveys and Taxonomies}
\taxnode{1}{Physical-World Adversarial Robustness}
\taxnode{1}{Robustness Benchmarks, Metrics, and Evaluation Protocols}
\taxnode{1}{Theoretical Analysis of Robustness}
\taxnode{0}{Calibration and Uncertainty Evaluation}
\taxnode{1}{Theoretical Foundations of Calibration and Uncertainty}
\taxnode{1}{Calibration Metrics, Scoring Rules, and Diagnostics}
\taxnode{1}{Post-hoc Calibration Methods}
\taxnode{1}{Training-time Calibration and Confidence Regularization}
\taxnode{1}{Predictive Uncertainty Estimation Methods}
\taxnode{1}{Aleatoric and Epistemic Uncertainty Modeling}
\taxnode{1}{Conformal Prediction and Coverage Guarantees}
\taxnode{1}{Selective Prediction, Abstention, and Risk-Coverage Evaluation}
\taxnode{1}{Distribution Shift, OOD Detection, and Reliability}
\taxnode{1}{Regression Calibration and Predictive Intervals}
\taxnode{1}{Structured, Dense, and Sequential Prediction Calibration}
\taxnode{1}{Group Calibration, Multicalibration, and Fairness}
\taxnode{1}{Benchmarking, Datasets, and Comparative Studies}
\taxnode{1}{Application-specific Reliability and Safety Evaluation}
\taxnode{1}{Surveys and Taxonomies of Calibration and Uncertainty}
\taxnode{0}{Efficiency and Resource Evaluation}
\taxnode{1}{Computational Complexity and Operation Count}
\taxnode{1}{Latency and Throughput Evaluation}
\taxnode{1}{Memory Footprint and Storage Efficiency}
\taxnode{1}{Energy and Power Measurement}
\taxnode{1}{Communication and Bandwidth Efficiency}
\taxnode{1}{Scalability and Parallel Efficiency}
\taxnode{1}{Hardware-Aware Performance Evaluation}
\taxnode{1}{Accuracy-Efficiency Trade-off Analysis}
\taxnode{1}{Cost and Environmental Impact Assessment}
\taxnode{1}{Efficiency Benchmarking and Profiling Methodology}
\taxnode{0}{Fairness, Bias, and Responsible AI Evaluation}
\taxnode{1}{Statistical Fairness Metrics}
\taxnode{1}{Causal and Counterfactual Fairness Evaluation}
\taxnode{1}{Subgroup and Intersectional Fairness}
\taxnode{1}{Dataset and Annotation Bias Evaluation}
\taxnode{1}{Model Auditing and Disparity Analysis}
\taxnode{1}{Benchmarking, Stress Testing, and Reproducibility}
\taxnode{1}{Fairness Mitigation Evaluation}
\taxnode{1}{Human-Centered Harm Assessment}
\taxnode{1}{Transparency, Documentation, and Accountability Assessment}
\taxnode{1}{Domain-Specific Fairness Evaluation}
\taxnode{0}{Interpretability and Human-Centered Evaluation}
\taxnode{1}{Intrinsically Interpretable Models}
\taxnode{1}{Global Explanations and Rule Extraction}
\taxnode{1}{Feature Attribution and Saliency}
\taxnode{1}{Example-Based Explanations}
\taxnode{1}{Concept-Based Explanations}
\taxnode{1}{Counterfactual Explanations and Algorithmic Recourse}
\taxnode{1}{Mechanistic Interpretability}
\taxnode{1}{Theoretical Foundations of Interpretability}
\taxnode{1}{Faithfulness, Robustness, and Stability Evaluation}
\taxnode{1}{Human-Subject Evaluation of Explanations}
\taxnode{1}{Interactive Explanations and Human-AI Collaboration}
\taxnode{1}{Benchmarks, Datasets, and Evaluation Protocols}
\taxnode{1}{Debugging, Auditing, and Safety Applications}
\taxnode{0}{Reproducibility and Meta-Evaluation}
\taxnode{1}{Computational Reproducibility}
\taxnode{1}{Independent Replication Studies}
\taxnode{1}{Statistical Evaluation Methodology}
\taxnode{1}{Benchmark and Leaderboard Meta-Evaluation}
\taxnode{2}{Benchmark Design and Curation}
\taxnode{2}{Construct Validity and Representativeness}
\taxnode{2}{Evaluation Metrics and Ranking Methods}
\taxnode{2}{Statistical Reliability and Rank Stability}
\taxnode{2}{Reproducibility and Replication Studies}
\taxnode{2}{Data Contamination and Information Leakage}
\taxnode{2}{Bias, Fairness, and Coverage Analysis}
\taxnode{2}{Robustness and Stress Testing}
\taxnode{2}{Benchmark Aging, Saturation, and Dynamic Benchmarking}
\taxnode{2}{Cross-Benchmark Comparison and Meta-Analysis}
\taxnode{2}{Resource-Aware and Multi-Objective Benchmarking}
\taxnode{1}{Metric and Protocol Validation}
\taxnode{1}{Reporting Standards and Experimental Documentation}
\taxnode{1}{Reproducibility Infrastructure and Open Science}
\taxnode{1}{Meta-Research on Research Reliability}
\taxnode{0}{Domain-Specific and Application-Oriented Evaluation}
\taxnode{1}{Biomedical and Healthcare Evaluation}
\taxnode{1}{Autonomous Systems and Robotics Evaluation}
\taxnode{1}{Natural Language and Communication Applications Evaluation}
\taxnode{1}{Vision and Multimedia Applications Evaluation}
\taxnode{1}{Recommender Systems and Personalization Evaluation}
\taxnode{1}{Finance, Economics, and Risk Evaluation}
\taxnode{1}{Cybersecurity and Trustworthy Computing Evaluation}
\taxnode{1}{Scientific and Engineering Applications Evaluation}
\taxnode{1}{Industrial, Manufacturing, and Operations Evaluation}
\taxnode{1}{Education, Social, and Public-Sector Evaluation}
\taxnode{1}{Human-Centered and Decision-Support Evaluation}
\taxnode{1}{Cross-Domain Benchmarking and Evaluation Methodology}
\taxfield{interpretability}
\taxnode{0}{Feature Attribution and Saliency Methods}
\taxnode{1}{Gradient-Based Saliency Methods}
\taxnode{1}{Perturbation-Based Attribution Methods}
\taxnode{1}{Relevance Propagation and Decomposition Methods}
\taxnode{1}{Shapley-Value and Additive Attribution}
\taxnode{1}{Class Activation Mapping}
\taxnode{1}{Causal and Interventional Attribution}
\taxnode{1}{Theoretical Foundations of Attribution}
\taxnode{1}{Attribution Evaluation and Validation}
\taxnode{1}{Robustness and Reliability of Saliency Methods}
\taxnode{1}{Benchmarking and Comparative Studies}
\taxnode{1}{Applied Feature Attribution}
\taxnode{1}{Surveys and Review Articles}
\taxnode{0}{Counterfactual and Contrastive Explanations}
\taxnode{1}{Counterfactual Explanation Generation}
\taxnode{1}{Contrastive Explanation Methods}
\taxnode{1}{Causal Counterfactual Explanations}
\taxnode{1}{Actionable Counterfactual Explanations and Algorithmic Recourse}
\taxnode{1}{Theoretical Foundations of Counterfactual and Contrastive Explanations}
\taxnode{1}{Evaluation Metrics and Benchmarking}
\taxnode{1}{Robustness, Fairness, Privacy, and Security}
\taxnode{1}{Human-Centered Evaluation and Applications}
\taxnode{1}{Surveys, Taxonomies, and Comparative Reviews}
\taxnode{0}{Example-Based Explanations}
\taxnode{1}{Retrieval-Based Explanations}
\taxnode{1}{Prototype and Criticism Explanations}
\taxnode{1}{Training Data Attribution}
\taxnode{1}{Counterfactual and Contrastive Explanations}
\taxnode{1}{Generative Example-Based Explanations}
\taxnode{1}{Evaluation and Benchmarking}
\taxnode{1}{Human-Centered and Applied Example-Based Explanations}
\taxnode{1}{Theory, Robustness, and Privacy}
\taxnode{0}{Concept-Based Explanations}
\taxnode{1}{Concept attribution methods}
\taxnode{1}{Automatic concept discovery}
\taxnode{1}{Concept representation learning}
\taxnode{1}{Concept bottleneck and concept-supervised models}
\taxnode{1}{Causal and counterfactual concept explanations}
\taxnode{1}{Knowledge-grounded and language-based concepts}
\taxnode{1}{Robustness and theoretical foundations}
\taxnode{1}{Evaluation, benchmarks, and user studies}
\taxnode{1}{Application-specific concept-based explanations}
\taxnode{1}{Surveys and taxonomies of concept-based explanations}
\taxnode{0}{Surrogate, Rule-Based, and Symbolic Explanations}
\taxnode{1}{Local Surrogate Explanations}
\taxnode{1}{Global Surrogate Models}
\taxnode{1}{Interpretable Model Distillation}
\taxnode{1}{Rule Extraction}
\taxnode{1}{Decision Trees, Rule Lists, and Decision Sets}
\taxnode{1}{Symbolic and Logical Explanations}
\taxnode{1}{Counterfactual Rule and Symbolic Reasoning}
\taxnode{1}{Fidelity, Faithfulness, and Complexity Analysis}
\taxnode{1}{Robustness, Stability, and Identifiability}
\taxnode{1}{Evaluation, Benchmarks, and Human Studies}
\taxnode{1}{Domain-Specific Applications}
\taxnode{1}{Surveys and Comparative Studies}
\taxnode{0}{Intrinsically Interpretable Models}
\taxnode{1}{Sparse Linear Models and Scoring Systems}
\taxnode{1}{Generalized Additive Models}
\taxnode{1}{Decision Trees}
\taxnode{1}{Rule-Based Models}
\taxnode{1}{Prototype and Case-Based Models}
\taxnode{1}{Concept-Based Models}
\taxnode{1}{Self-Explaining Neural Networks}
\taxnode{1}{Monotonic and Shape-Constrained Models}
\taxnode{1}{Theory and Guarantees for Intrinsic Interpretability}
\taxnode{1}{Evaluation, Benchmarks, and Human Studies}
\taxnode{1}{Domain Applications of Intrinsically Interpretable Models}
\taxnode{1}{Surveys and Comparative Studies}
\taxnode{0}{Mechanistic Interpretability}
\taxnode{1}{Feature Discovery and Neuron Analysis}
\taxnode{1}{Circuit and Pathway Analysis}
\taxnode{1}{Representation Geometry and Superposition}
\taxnode{1}{Causal Intervention and Mediation Analysis}
\taxnode{1}{Learning Dynamics of Mechanisms}
\taxnode{1}{Automated and Scalable Mechanistic Interpretability}
\taxnode{1}{Benchmarks and Evaluation}
\taxnode{1}{Mechanistic Auditing, Steering, and Model Editing}
\taxnode{1}{Architecture- and Modality-Specific Mechanistic Studies}
\taxnode{1}{Surveys, Taxonomies, and Foundations}
\taxnode{0}{Representation Analysis and Probing}
\taxnode{1}{Diagnostic Probing}
\taxnode{1}{Probe Methodology and Evaluation}
\taxnode{1}{Representational Similarity and Alignment}
\taxnode{1}{Geometric and Topological Analysis of Representations}
\taxnode{1}{Concept-based Representation Analysis}
\taxnode{1}{Neuron, Head, and Subspace Analysis}
\taxnode{1}{Representation Dynamics Across Layers and Training}
\taxnode{1}{Causal and Interventional Probing}
\taxnode{1}{Theoretical Foundations of Probing and Representational Interpretability}
\taxnode{0}{Visualization and Interactive Interpretability}
\taxnode{1}{Feature, Neuron, and Filter Visualization}
\taxnode{1}{Attribution and Saliency Visualization}
\taxnode{1}{Representation and Embedding Visualization}
\taxnode{1}{Concept-Based Visualization}
\taxnode{1}{Example-, Prototype-, and Counterfactual-Based Visualization}
\taxnode{1}{Interactive Visual Analytics for Model Inspection}
\taxnode{1}{Human-in-the-Loop Debugging and Model Steering}
\taxnode{1}{Evaluation, Human Factors, and Design Principles}
\taxnode{1}{Benchmarks and Comparative Studies}
\taxnode{1}{Application-Specific Visual Explanations}
\taxnode{1}{Surveys and Taxonomies}
\taxnode{0}{Evaluation, Faithfulness, and Robustness of Explanations}
\taxnode{1}{Theoretical and Axiomatic Foundations}
\taxnode{1}{Faithfulness and Fidelity Metrics}
\taxnode{1}{Causal and Counterfactual Evaluation}
\taxnode{1}{Robustness, Stability, and Sensitivity Analysis}
\taxnode{1}{Sanity Checks and Randomization Tests}
\taxnode{1}{Human-Grounded Evaluation}
\taxnode{1}{Application-Grounded Evaluation}
\taxnode{1}{Benchmarks, Datasets, and Evaluation Protocols}
\taxnode{1}{Adversarial Attacks and Defenses on Explanations}
\taxnode{1}{Comparative Studies, Meta-Evaluation, and Surveys}
\taxnode{0}{Theoretical Foundations of Interpretability}
\taxnode{1}{Formal Definitions and Conceptual Frameworks}
\taxnode{1}{Axiomatic and Game-Theoretic Attribution}
\taxnode{1}{Causal and Counterfactual Foundations}
\taxnode{1}{Statistical and Information-Theoretic Foundations}
\taxnode{1}{Complexity, Learnability, and Identifiability}
\taxnode{1}{Faithfulness, Fidelity, and Completeness}
\taxnode{1}{Robustness, Stability, and Generalization of Explanations}
\taxnode{1}{Representation, Concepts, and Mechanistic Structure}
\taxnode{1}{Human-Grounded Interpretability and Simulatability}
\taxnode{1}{Evaluation Theory and Benchmarking}
\taxnode{1}{Interpretability–Accuracy Trade-offs}
\taxfield{optimization}
\taxnode{0}{Optimization Theory and Convergence Analysis}
\taxnode{1}{Convex Optimization Theory}
\taxnode{1}{Nonconvex Optimization Theory}
\taxnode{1}{Stochastic Optimization and SGD Analysis}
\taxnode{1}{Acceleration and Momentum Methods}
\taxnode{1}{Adaptive, Preconditioned, and Second-Order Optimization}
\taxnode{1}{Composite, Proximal, and Regularized Optimization}
\taxnode{1}{Constrained, Saddle-Point, and Minimax Optimization}
\taxnode{1}{Bilevel and Hyperparameter Optimization Theory}
\taxnode{1}{Distributed, Parallel, and Federated Optimization Theory}
\taxnode{1}{Optimization Landscapes, Geometry, and Implicit Bias}
\taxnode{1}{Complexity Analysis and Lower Bounds}
\taxnode{1}{Empirical Analysis and Benchmarking of Optimizers}
\taxnode{1}{Surveys and Unifying Frameworks}
\taxnode{0}{Loss Landscape and Nonconvex Optimization}
\taxnode{1}{Nonconvex Optimization Theory}
\taxnode{1}{Loss Landscape Geometry and Topology}
\taxnode{1}{Overparameterization and Benign Nonconvexity}
\taxnode{1}{First-Order Nonconvex Optimization Methods}
\taxnode{1}{Second-Order and Curvature-Aware Optimization}
\taxnode{1}{Stochastic Optimization Dynamics}
\taxnode{1}{Initialization, Conditioning, and Reparameterization}
\taxnode{1}{Flatness, Sharpness, and Generalization}
\taxnode{1}{Implicit Bias and Implicit Regularization}
\taxnode{1}{Empirical Loss Landscape Analysis and Visualization}
\taxnode{1}{Benchmarking and Comparative Evaluation}
\taxnode{1}{Surveys and Review Articles}
\taxnode{0}{Stochastic Optimization}
\taxnode{1}{Stochastic Gradient Methods}
\taxnode{1}{Momentum and Accelerated Methods}
\taxnode{1}{Adaptive Gradient Methods}
\taxnode{1}{Variance-Reduced Stochastic Optimization}
\taxnode{1}{Stochastic Second-Order and Quasi-Newton Methods}
\taxnode{1}{Proximal and Constrained Stochastic Optimization}
\taxnode{1}{Sampling, Mini-batching, and Data Ordering}
\taxnode{1}{Nonconvex and Nonsmooth Stochastic Optimization Theory}
\taxnode{1}{Generalization, Stability, and Implicit Bias}
\taxnode{1}{Stochastic Min-Max and Saddle-Point Optimization}
\taxnode{1}{Distributed, Parallel, and Asynchronous Stochastic Optimization}
\taxnode{1}{Federated and Decentralized Stochastic Optimization}
\taxnode{1}{Robust and Private Stochastic Optimization}
\taxnode{1}{Empirical Evaluation and Benchmarking of Stochastic Optimizers}
\taxnode{1}{Applications of Stochastic Optimization in Deep Learning}
\taxnode{1}{Surveys and Tutorials on Stochastic Optimization}
\taxnode{0}{Adaptive Gradient and Momentum Methods}
\taxnode{1}{AdaGrad Family and Per-Coordinate Adaptive Methods}
\taxnode{1}{RMSProp and AdaDelta Family}
\taxnode{1}{Adam Family and Adaptive Moment Estimation}
\taxnode{1}{Momentum SGD and Nesterov Methods}
\taxnode{1}{Optimizer Variant Design and Regularization-Aware Methods}
\taxnode{2}{Decoupled Weight Decay and Explicit Regularization}
\taxnode{2}{Proximal, Projected, and Composite Adaptive Optimization}
\taxnode{2}{Sharpness-Aware and Generalization-Oriented Optimization}
\taxnode{2}{Sparsity-Inducing and Structured Regularization Methods}
\taxnode{2}{Scale-Invariant and Norm-Aware Adaptive Methods}
\taxnode{2}{Stable, Rectified, and Robust Adaptive Methods}
\taxnode{2}{Hyperparameter Adaptation and Schedule Design}
\taxnode{2}{Theoretical Analysis of Regularization-Aware Optimizers}
\taxnode{2}{Benchmarking, Ablation, and Reproducibility Studies}
\taxnode{2}{Domain-Specific and Large-Scale Applications}
\taxnode{2}{Surveys, Taxonomies, and Comparative Reviews}
\taxnode{1}{Convergence Theory and Complexity Analysis}
\taxnode{1}{Generalization, Stability, and Implicit Bias}
\taxnode{1}{Hyperparameter Tuning and Scheduler Interaction}
\taxnode{1}{Distributed, Large-Batch, and Memory-Efficient Optimization}
\taxnode{1}{Benchmarking, Reproducibility, and Empirical Comparison}
\taxnode{1}{Application-Specific Optimizer Adaptation}
\taxnode{1}{Surveys and Unifying Frameworks}
\taxnode{0}{Second-Order and Geometry-Aware Optimization}
\taxnode{1}{Newton and Gauss-Newton Methods}
\taxnode{1}{Quasi-Newton Methods}
\taxnode{1}{Hessian-Free and Matrix-Free Methods}
\taxnode{1}{Trust-Region and Cubic-Regularized Optimization}
\taxnode{1}{Natural Gradient and Information-Geometric Optimization}
\taxnode{1}{Riemannian and Manifold Optimization}
\taxnode{1}{Mirror Descent and Bregman Geometry}
\taxnode{1}{Curvature Approximation and Preconditioning}
\taxnode{1}{Stochastic and Large-Scale Curvature-Aware Optimization}
\taxnode{1}{Nonconvex Theory and Convergence Analysis}
\taxnode{1}{Empirical Evaluation and Benchmarks}
\taxnode{2}{Comparative Optimizer Studies}
\taxnode{2}{Benchmark Suite Design and Curation}
\taxnode{2}{Convergence and Training Dynamics Evaluation}
\taxnode{2}{Generalization and Robustness Evaluation}
\taxnode{2}{Scalability and Large-Scale Evaluation}
\taxnode{2}{Computational Efficiency and Systems Benchmarking}
\taxnode{2}{Hyperparameter Sensitivity and Ablation Studies}
\taxnode{2}{Application-Specific Benchmarking}
\taxnode{2}{Reproducibility and Evaluation Methodology}
\taxnode{2}{Empirical Surveys and Meta-Analysis}
\taxnode{1}{Applications and Domain-Specific Adaptations}
\taxnode{1}{Surveys and Unifying Frameworks}
\taxnode{0}{Constrained, Proximal, and Regularized Optimization}
\taxnode{1}{Lagrangian and Primal-Dual Methods}
\taxnode{1}{Penalty, Barrier, and Augmented Lagrangian Methods}
\taxnode{1}{Projected, Feasibility, and Projection-Free Methods}
\taxnode{1}{Proximal Point, Proximal Gradient, and Composite Optimization}
\taxnode{1}{Operator Splitting and ADMM}
\taxnode{1}{Nonsmooth and Stochastic Constrained Optimization}
\taxnode{1}{Regularization Theory and Penalty Design}
\taxnode{1}{Structured Sparsity and Low-Rank Regularization}
\taxnode{1}{Constraint-Driven Learning Applications}
\taxnode{1}{Convergence Theory, Benchmarks, and Surveys}
\taxnode{0}{Bilevel Optimization and Meta-Optimization}
\taxnode{1}{Bilevel Optimization Theory and Complexity}
\taxnode{1}{Hypergradient Computation and Implicit Differentiation}
\taxnode{1}{Stochastic, Online, and Distributed Bilevel Algorithms}
\taxnode{1}{Hyperparameter and Regularization Optimization}
\taxnode{1}{Meta-Learning and Few-Shot Adaptation}
\taxnode{1}{Learned Optimizers and Learning to Optimize}
\taxnode{1}{Neural Architecture Search}
\taxnode{1}{Data Selection, Reweighting, and Distillation}
\taxnode{1}{Robust, Adversarial, and Fairness-Aware Bilevel Learning}
\taxnode{1}{Application-Driven Bilevel Optimization}
\taxnode{1}{Benchmarks and Empirical Evaluation}
\taxnode{2}{Synthetic and Controlled Testbeds}
\taxnode{2}{Benchmark Datasets and Task Suites}
\taxnode{2}{Evaluation Metrics and Experimental Protocols}
\taxnode{2}{Comparative Empirical Studies}
\taxnode{2}{Ablation and Sensitivity Analysis}
\taxnode{2}{Robustness, Stability, and Generalization Studies}
\taxnode{2}{Scalability, Efficiency, and Systems Evaluation}
\taxnode{2}{Application-Oriented Case Studies}
\taxnode{2}{Reproducibility and Replication Studies}
\taxnode{2}{Surveys and Meta-Analysis}
\taxnode{1}{Surveys, Taxonomies, and Software}
\taxnode{0}{Distributed, Parallel, and Federated Optimization}
\taxnode{1}{Synchronous Distributed Optimization}
\taxnode{1}{Asynchronous Parallel Optimization}
\taxnode{1}{Decentralized Optimization}
\taxnode{1}{Model-Parallel and Pipeline Optimization}
\taxnode{1}{Communication-Efficient Optimization}
\taxnode{1}{Cross-Device Federated Optimization}
\taxnode{1}{Cross-Silo Federated Optimization}
\taxnode{1}{Personalized and Heterogeneous Federated Optimization}
\taxnode{1}{Privacy-Preserving and Robust Optimization}
\taxnode{1}{Theory and Convergence Analysis}
\taxnode{1}{Systems, Benchmarks, and Applications}
\taxnode{0}{Efficient and Hardware-Aware Training Optimization}
\taxnode{1}{Efficient Optimizer Design}
\taxnode{1}{Mixed-Precision and Low-Precision Training}
\taxnode{1}{Memory-Efficient Training}
\taxnode{1}{Distributed and Parallel Training}
\taxnode{1}{Communication-Efficient Optimization}
\taxnode{1}{Sparse and Structured Training}
\taxnode{1}{Compiler, Kernel, and Runtime Optimization}
\taxnode{1}{Hardware-Software Co-Design for Training}
\taxnode{1}{Energy-Efficient and Power-Aware Training}
\taxnode{1}{Performance Modeling, Profiling, and Benchmarking}
\taxnode{0}{Derivative-Free and Black-Box Optimization}
\taxnode{1}{Zeroth-Order Optimization}
\taxnode{1}{Direct Search and Pattern Search}
\taxnode{1}{Surrogate-Based Derivative-Free Optimization}
\taxnode{1}{Bayesian Optimization}
\taxnode{1}{Evolutionary Algorithms and Evolution Strategies}
\taxnode{1}{Bandit-Based and Multi-Fidelity Optimization}
\taxnode{1}{Constrained and Multi-Objective Black-Box Optimization}
\taxnode{1}{High-Dimensional and Parallel Black-Box Optimization}
\taxnode{1}{AutoML and Neural Architecture Search}
\taxnode{1}{Gradient-Free Policy Search and Control}
\taxnode{1}{Black-Box Adversarial Optimization and Robustness}
\taxnode{1}{Theory and Convergence Analysis}
\taxnode{1}{Benchmarking and Empirical Evaluation}
\taxnode{1}{Surveys, Libraries, and Systems}
\taxnode{0}{Empirical Evaluation and Benchmarking of Optimizers}
\taxnode{1}{Benchmark Suite and Evaluation Protocol Design}
\taxnode{1}{Comparative Performance Studies}
\taxnode{2}{Benchmark-Based Optimizer Comparison}
\taxnode{2}{Convergence Speed and Training Efficiency}
\taxnode{2}{Generalization Performance Comparison}
\taxnode{2}{Robustness and Stability Analysis}
\taxnode{2}{Hyperparameter Sensitivity and Tuning Budget Analysis}
\taxnode{2}{Scalability and Systems Efficiency}
\taxnode{2}{Task- and Architecture-Specific Comparative Studies}
\taxnode{2}{Application-Oriented Optimizer Evaluation}
\taxnode{2}{Empirical Validation of Convergence Theory}
\taxnode{2}{Evaluation Methodology and Statistical Testing}
\taxnode{2}{Systematic Reviews and Meta-Analysis}
\taxnode{1}{Hyperparameter Sensitivity and Tuning Evaluation}
\taxnode{2}{Hyperparameter Sensitivity Analysis}
\taxnode{2}{Hyperparameter Interaction Analysis}
\taxnode{2}{Robustness to Hyperparameter Misspecification}
\taxnode{2}{Hyperparameter Optimization Benchmarks}
\taxnode{2}{Tuning Budget and Sample-Efficiency Evaluation}
\taxnode{2}{Transferability and Default Configuration Studies}
\taxnode{2}{Multi-Fidelity and Early-Stopping Evaluation}
\taxnode{2}{Search-Space Design and Tuning Protocols}
\taxnode{2}{Fair Comparison, Reproducibility, and Reporting Standards}
\taxnode{2}{Theoretical Analysis of Hyperparameter Effects}
\taxnode{2}{Domain-Specific and Cross-Task Tuning Studies}
\taxnode{2}{Meta-Benchmarking and Survey Studies}
\taxnode{1}{Robustness, Stability, and Generalization Benchmarking}
\taxnode{1}{Training Dynamics and Convergence Diagnostics}
\taxnode{1}{Computational Efficiency and Scalability Benchmarking}
\taxnode{1}{Task- and Domain-Specific Optimizer Benchmarks}
\taxnode{2}{Computer Vision Optimizer Benchmarks}
\taxnode{2}{Natural Language Processing Optimizer Benchmarks}
\taxnode{2}{Speech and Audio Optimizer Benchmarks}
\taxnode{2}{Generative Modeling Optimizer Benchmarks}
\taxnode{2}{Reinforcement Learning and Control Optimizer Benchmarks}
\taxnode{2}{Graph Machine Learning Optimizer Benchmarks}
\taxnode{2}{Recommendation and Ranking Optimizer Benchmarks}
\taxnode{2}{Time-Series and Tabular Learning Optimizer Benchmarks}
\taxnode{2}{Scientific, Engineering, and Biomedical Optimizer Benchmarks}
\taxnode{2}{Foundation Model Pretraining and Fine-Tuning Optimizer Benchmarks}
\taxnode{3}{Language Model Pretraining Optimizer Benchmarks}
\taxnode{3}{Vision Foundation Model Pretraining Optimizer Benchmarks}
\taxnode{3}{Multimodal Foundation Model Pretraining Optimizer Benchmarks}
\taxnode{3}{Speech and Audio Foundation Model Pretraining Optimizer Benchmarks}
\taxnode{3}{Full-Parameter Fine-Tuning Optimizer Benchmarks}
\taxnode{3}{Parameter-Efficient Fine-Tuning Optimizer Benchmarks}
\taxnode{3}{Instruction Tuning and Alignment Optimizer Benchmarks}
\taxnode{3}{Domain Adaptation and Continual Adaptation Optimizer Benchmarks}
\taxnode{3}{Efficiency and Scalability Benchmarking}
\taxnode{3}{Stability, Robustness, and Hyperparameter Sensitivity Benchmarks}
\taxnode{3}{Optimization Dynamics and Scaling-Law Studies}
\taxnode{3}{Benchmark Methodology and Reproducibility}
\taxnode{3}{Survey and Meta-Analysis of Foundation-Model Optimizer Benchmarks}
\taxnode{2}{Federated, Distributed, and Continual Learning Optimizer Benchmarks}
\taxnode{2}{Benchmark Design, Reproducibility, and Cross-Domain Meta-Analysis}
\taxnode{1}{Reproducibility and Statistical Evaluation Methodology}
\taxnode{1}{Survey and Meta-Analysis of Optimizer Benchmarks}
\taxfield{privacy}
\taxnode{0}{Differential Privacy}
\taxnode{1}{Differential Privacy Theory and Private Learning Theory}
\taxnode{1}{Differentially Private Optimization and Empirical Risk Minimization}
\taxnode{1}{Differentially Private Deep Learning Methods}
\taxnode{1}{Local and Shuffle Differential Privacy}
\taxnode{1}{Federated and Distributed Differentially Private Learning}
\taxnode{1}{Differentially Private Synthetic Data and Generative Modeling}
\taxnode{1}{Privacy Accounting, Systems, and Tooling}
\taxnode{1}{Auditing, Evaluation, and Benchmarking for DP ML}
\taxnode{1}{Applied Differential Privacy in Machine Learning}
\taxnode{0}{Secure and Privacy-Preserving Computation}
\taxnode{1}{Secure Multiparty Computation for Machine Learning}
\taxnode{1}{Homomorphic Encryption for Machine Learning}
\taxnode{1}{Trusted Execution Environments for Machine Learning}
\taxnode{1}{Hybrid Secure Computation for Machine Learning}
\taxnode{1}{Secure and Private Model Inference}
\taxnode{1}{Privacy-Preserving Model Training}
\taxnode{1}{Secure Aggregation and Federated Protocols}
\taxnode{1}{Verifiable Computation for Machine Learning}
\taxnode{1}{Threat Models, Leakage, and Formal Security Analysis}
\taxnode{1}{Systems, Compilers, and Benchmarking for Private ML}
\taxnode{1}{Applications and Surveys of Secure ML}
\taxnode{0}{Federated and Distributed Privacy}
\taxnode{1}{Federated Differential Privacy}
\taxnode{1}{Secure Aggregation and Cryptographic Protocols}
\taxnode{1}{Privacy Attacks and Leakage Analysis}
\taxnode{1}{Privacy Auditing and Benchmarking}
\taxnode{1}{Private Federated Optimization and Systems}
\taxnode{1}{Personalized and Heterogeneous Private Learning}
\taxnode{1}{Vertical Federated Learning Privacy}
\taxnode{1}{Split Learning Privacy}
\taxnode{1}{Decentralized Privacy-Preserving Learning}
\taxnode{1}{Federated Analytics and Private Data Release}
\taxnode{0}{Privacy Attacks and Leakage Analysis}
\taxnode{1}{Membership Inference Attacks}
\taxnode{1}{Attribute Inference Attacks}
\taxnode{1}{Property Inference Attacks}
\taxnode{1}{Model Inversion and Input Reconstruction}
\taxnode{1}{Training Data Extraction and Memorization Analysis}
\taxnode{1}{Gradient Leakage and Reconstruction Attacks}
\taxnode{1}{Re-identification and Linkage Attacks}
\taxnode{1}{Side-Channel and Systems-Level Privacy Leakage}
\taxnode{1}{Privacy Auditing and Leakage Quantification}
\taxnode{1}{Theoretical Privacy Leakage Analysis}
\taxnode{1}{Benchmarks and Empirical Evaluation of Privacy Attacks}
\taxnode{1}{Application-Specific Privacy Attack Studies}
\taxnode{0}{Privacy Auditing and Benchmarking}
\taxnode{1}{Membership Inference Auditing}
\taxnode{1}{Attribute and Property Inference Auditing}
\taxnode{1}{Model Inversion and Reconstruction Auditing}
\taxnode{1}{Training Data Extraction and Memorization Auditing}
\taxnode{1}{Differential Privacy Auditing and Verification}
\taxnode{2}{Formal Differential Privacy Verification}
\taxnode{2}{Privacy Accounting and Composition Verification}
\taxnode{2}{Statistical Privacy Auditing}
\taxnode{2}{Attack-Based Privacy Evaluation}
\taxnode{2}{Systems and Implementation Auditing}
\taxnode{2}{Benchmarking and Evaluation Methodology}
\taxnode{2}{Federated, Distributed, and Local DP Auditing}
\taxnode{2}{Group, User-Level, and Heterogeneous Risk Auditing}
\taxnode{2}{Online Privacy Monitoring and Budget Control}
\taxnode{2}{Reproducibility, Reporting, and Survey Studies}
\taxnode{1}{Federated and Distributed Learning Privacy Auditing}
\taxnode{1}{Privacy Benchmark Design and Standardization}
\taxnode{1}{Audit Methodology, Metrics, and Statistical Testing}
\taxnode{1}{Application-Specific and Deployment-Scale Privacy Evaluation}
\taxnode{0}{Privacy-Preserving Data Release and Synthetic Data}
\taxnode{1}{Differentially Private Data Release}
\taxnode{1}{Privacy-Preserving Synthetic Data Generation}
\taxnode{1}{Anonymization and De-identification}
\taxnode{1}{Statistical and Query-Based Release}
\taxnode{1}{Privacy Risk Analysis and Auditing}
\taxnode{1}{Utility, Fidelity, and Fairness Evaluation}
\taxnode{1}{Theory and Privacy-Utility Tradeoffs}
\taxnode{1}{Benchmarks and Comparative Evaluation}
\taxnode{1}{Domain-Specific Applications}
\taxnode{1}{Surveys and Taxonomies}
\taxnode{0}{Machine Unlearning and Data Deletion}
\taxnode{1}{Theoretical Foundations of Machine Unlearning}
\taxnode{1}{Exact Unlearning}
\taxnode{1}{Certified Unlearning}
\taxnode{1}{Approximate Unlearning}
\taxnode{1}{Scalable Unlearning Systems}
\taxnode{1}{Unlearning Verification and Auditing}
\taxnode{1}{Unlearning Benchmarks, Metrics, and Surveys}
\taxnode{1}{Federated and Distributed Unlearning}
\taxnode{1}{Foundation and Generative Model Unlearning}
\taxnode{1}{Security and Safety-Oriented Unlearning}
\taxnode{1}{Applied and Domain-Specific Unlearning}
\taxnode{0}{Privacy Theory and Privacy-Utility Tradeoffs}
\taxnode{1}{Differential Privacy Theory}
\taxnode{1}{Information-Theoretic Privacy}
\taxnode{1}{Private Learning Theory}
\taxnode{1}{Privacy Accounting and Composition}
\taxnode{1}{Optimal Privacy Mechanisms}
\taxnode{1}{Privacy Leakage Quantification}
\taxnode{1}{Empirical Privacy-Utility Evaluation}
\taxnode{1}{Application-Specific Tradeoff Studies}
\taxnode{1}{Surveys and Comparative Studies}
\taxnode{0}{Privacy-Preserving ML Systems and Applications}
\taxnode{1}{Differentially Private Learning}
\taxnode{1}{Federated and Collaborative Learning}
\taxnode{1}{Secure Multiparty Computation for ML}
\taxnode{1}{Homomorphic Encryption for ML}
\taxnode{1}{Trusted Execution Environments for ML}
\taxnode{1}{Privacy-Preserving Data Sharing and Synthetic Data}
\taxnode{1}{Privacy-Preserving Inference}
\taxnode{1}{Systems Optimization and Scalability}
\taxnode{1}{Privacy Evaluation, Auditing, and Benchmarking}
\taxnode{1}{Domain-Specific Privacy-Preserving ML Applications}
\taxnode{1}{Surveys, Taxonomies, and Comparative Studies}
\taxfield{probabilistic\_\allowbreak{}modeling}
\taxnode{0}{Probabilistic Graphical Models}
\taxnode{1}{Directed Graphical Models}
\taxnode{1}{Undirected Graphical Models}
\taxnode{1}{Factor Graphs and Message Passing}
\taxnode{1}{Latent Variable and Hierarchical Graphical Models}
\taxnode{1}{Temporal and Sequential Graphical Models}
\taxnode{1}{Causal Graphical Models}
\taxnode{1}{Statistical Relational Learning}
\taxnode{1}{Inference in Graphical Models}
\taxnode{1}{Learning and Structure Learning in Graphical Models}
\taxnode{1}{Neural and Deep Graphical Models}
\taxnode{1}{Theory and Foundations of Graphical Models}
\taxnode{1}{Applications of Graphical Models}
\taxnode{1}{Benchmarks and Empirical Evaluation of Graphical Models}
\taxnode{1}{Surveys and Tutorials on Graphical Models}
\taxnode{0}{Bayesian Modeling and Bayesian Deep Learning}
\taxnode{1}{Bayesian Inference and Posterior Approximation}
\taxnode{1}{Bayesian Neural Networks}
\taxnode{1}{Bayesian Latent Variable and Generative Models}
\taxnode{1}{Hierarchical Bayesian Models and Probabilistic Graphical Models}
\taxnode{1}{Gaussian Processes and Bayesian Nonparametrics}
\taxnode{1}{Uncertainty Quantification and Calibration}
\taxnode{1}{Bayesian Optimization, Active Learning, and Experimental Design}
\taxnode{1}{Bayesian Reinforcement Learning and Sequential Decision-Making}
\taxnode{1}{Scalable Bayesian Computation and Probabilistic Programming}
\taxnode{1}{PAC-Bayesian Theory and Bayesian Learning Theory}
\taxnode{1}{Empirical Evaluation and Benchmarks}
\taxnode{1}{Applications of Bayesian Deep Learning}
\taxnode{1}{Surveys and Tutorials in Bayesian Modeling}
\taxnode{0}{Latent Variable Models}
\taxnode{1}{Continuous Latent Variable Models}
\taxnode{1}{Discrete Latent Variable Models}
\taxnode{1}{Hierarchical Latent Variable Models}
\taxnode{1}{Sequential and State-Space Latent Variable Models}
\taxnode{1}{Structured Latent Variable Models}
\taxnode{1}{Undirected and Energy-Based Latent Variable Models}
\taxnode{1}{Multimodal and Conditional Latent Variable Models}
\taxnode{1}{Inference and Learning Algorithms for Latent Variable Models}
\taxnode{1}{Disentangled and Identifiable Latent Representation Learning}
\taxnode{1}{Theory of Latent Variable Models}
\taxnode{1}{Evaluation and Benchmarking of Latent Variable Models}
\taxnode{1}{Applications of Latent Variable Models}
\taxnode{1}{Surveys and Reviews of Latent Variable Models}
\taxnode{0}{Deep Generative Models}
\taxnode{1}{Deep Latent Variable Models}
\taxnode{1}{Generative Adversarial Networks}
\taxnode{1}{Autoregressive Generative Models}
\taxnode{1}{Normalizing Flows}
\taxnode{1}{Diffusion and Score-Based Models}
\taxnode{1}{Energy-Based Models}
\taxnode{1}{Conditional and Controllable Generation}
\taxnode{1}{Representation Learning and Disentanglement}
\taxnode{1}{Multimodal and Structured Generative Modeling}
\taxnode{1}{Theory, Inference, and Optimization for Generative Models}
\taxnode{1}{Evaluation, Benchmarks, and Empirical Comparison}
\taxnode{1}{Applications of Deep Generative Models}
\taxnode{1}{Surveys and Taxonomies of Deep Generative Models}
\taxnode{0}{Approximate Inference and Monte Carlo Methods}
\taxnode{1}{Variational Inference}
\taxnode{1}{Markov Chain Monte Carlo}
\taxnode{1}{Sequential Monte Carlo}
\taxnode{1}{Importance Sampling and Monte Carlo Estimation}
\taxnode{1}{Expectation Propagation and Message Passing}
\taxnode{1}{Deterministic Approximate Inference}
\taxnode{1}{Amortized Inference}
\taxnode{1}{Likelihood-Free and Simulation-Based Inference}
\taxnode{1}{Scalable and Automatic Inference}
\taxnode{1}{Theoretical Foundations of Approximate Inference}
\taxnode{1}{Evaluation, Benchmarks, and Reviews}
\taxnode{1}{Applications of Approximate Inference}
\taxnode{0}{Sequential and Temporal Probabilistic Models}
\taxnode{1}{Autoregressive Sequence Models}
\taxnode{1}{Markov Models and Hidden Markov Models}
\taxnode{1}{State Space Models and Latent Dynamical Systems}
\taxnode{1}{Dynamic Bayesian Networks}
\taxnode{1}{Temporal Point Processes}
\taxnode{1}{Continuous-Time Dynamical Models}
\taxnode{1}{Bayesian and Nonparametric Time-Series Models}
\taxnode{1}{Nonstationary and Regime-Switching Models}
\taxnode{1}{Inference and Learning for Temporal Probabilistic Models}
\taxnode{0}{Uncertainty Quantification and Calibration}
\taxnode{1}{Theoretical Foundations of Uncertainty and Calibration}
\taxnode{1}{Bayesian Deep Learning}
\taxnode{1}{Practical Uncertainty Estimation Methods}
\taxnode{1}{Aleatoric and Epistemic Uncertainty Modeling}
\taxnode{1}{Calibration of Probabilistic Predictions}
\taxnode{1}{Conformal Prediction}
\taxnode{1}{Out-of-Distribution Detection and Distribution Shift}
\taxnode{1}{Selective Prediction and Abstention}
\taxnode{1}{Evaluation Metrics, Benchmarks, and Empirical Studies}
\taxnode{1}{Applied Uncertainty-Aware Machine Learning}
\taxnode{0}{Structure Learning}
\taxnode{1}{Bayesian Network Structure Learning}
\taxnode{1}{Undirected Graphical Model Structure Learning}
\taxnode{1}{Causal Structure Learning}
\taxnode{1}{Latent Structure Discovery}
\taxnode{1}{Temporal and Dynamic Structure Learning}
\taxnode{1}{Interventional and Active Structure Learning}
\taxnode{1}{Differentiable and Neural Structure Learning}
\taxnode{1}{Scalable and High-Dimensional Structure Learning}
\taxnode{1}{Structure Learning Theory}
\taxnode{1}{Benchmarking and Empirical Evaluation}
\taxnode{1}{Applied Structure Learning}
\taxnode{1}{Surveys and Tutorials}
\taxnode{0}{Probabilistic Programming}
\taxnode{1}{Probabilistic Programming Languages and Semantics}
\taxnode{1}{Inference Algorithms for Probabilistic Programs}
\taxnode{1}{Variational and Amortized Inference for Probabilistic Programs}
\taxnode{1}{Compilation and Systems for Probabilistic Programming}
\taxnode{1}{Program Analysis, Verification, and Type Systems}
\taxnode{1}{Deep Probabilistic Programming}
\taxnode{1}{Probabilistic Program Induction and Synthesis}
\taxnode{1}{Benchmarking, Diagnostics, and Empirical Evaluation}
\taxnode{1}{Applied Probabilistic Programming}
\taxnode{1}{Surveys, Tutorials, and Comparative Studies}
\taxnode{0}{Theory and Foundations of Probabilistic Modeling}
\taxnode{1}{Statistical Learning Theory and Asymptotics}
\taxnode{1}{Bayesian Foundations and Posterior Theory}
\taxnode{1}{Inference Theory and Approximation}
\taxnode{1}{Identifiability and Latent Variable Theory}
\taxnode{1}{Expressivity and Representation Theory}
\taxnode{1}{Learning Objectives and Divergence Theory}
\taxnode{1}{Information-Theoretic Foundations}
\taxnode{1}{Optimization Theory for Probabilistic Models}
\taxnode{1}{Graphical Structure and Conditional Independence Theory}
\taxnode{1}{Uncertainty Quantification and Calibration}
\taxnode{1}{Robustness and Model Misspecification}
\taxnode{1}{Evaluation Methodology and Benchmarking}
\taxnode{0}{Evaluation, Benchmarks, Surveys, and Applications of Probabilistic Models}
\taxnode{1}{Scoring Rules and Predictive Performance}
\taxnode{1}{Calibration and Uncertainty Assessment}
\taxnode{1}{Posterior Diagnostics and Inference Evaluation}
\taxnode{1}{Robustness and Distribution Shift Evaluation}
\taxnode{1}{Benchmark Datasets, Tasks, and Evaluation Protocols}
\taxnode{1}{Comparative Empirical Studies and Reproducibility}
\taxnode{1}{Survey, Review, and Taxonomy Papers}
\taxnode{1}{Applications in Vision, Speech, and Multimodal Perception}
\taxnode{1}{Applications in Natural Language, Information Retrieval, and Recommendation}
\taxnode{1}{Applications in Time Series, Control, Healthcare, and Scientific Modeling}
\taxfield{quantum\_\allowbreak{}computing}
\taxnode{0}{Quantum algorithms for machine learning}
\taxnode{1}{Quantum linear-algebraic methods}
\taxnode{1}{Quantum kernel methods}
\taxnode{1}{Variational quantum circuits and quantum neural networks}
\taxnode{1}{Quantum generative models}
\taxnode{1}{Quantum reinforcement learning}
\taxnode{1}{Quantum data encoding and feature maps}
\taxnode{1}{Optimization and trainability of quantum machine learning models}
\taxnode{1}{Quantum learning theory and complexity}
\taxnode{1}{Noise-aware quantum machine learning}
\taxnode{1}{Benchmarking and empirical evaluation of quantum machine learning}
\taxnode{1}{Applied quantum machine learning}
\taxnode{1}{Surveys and reviews of quantum machine learning}
\taxnode{0}{Variational quantum circuits and quantum neural networks}
\taxnode{1}{Quantum circuit ansätze and QNN architectures}
\taxnode{1}{Quantum data encoding and feature maps}
\taxnode{1}{Optimization and training methods}
\taxnode{1}{Expressivity, trainability, and generalization theory}
\taxnode{1}{Noise-aware and hardware-efficient QNNs}
\taxnode{1}{Hybrid quantum-classical neural networks}
\taxnode{1}{Supervised quantum learning}
\taxnode{1}{Generative and unsupervised quantum learning}
\taxnode{1}{Quantum reinforcement learning}
\taxnode{1}{Benchmarking and empirical evaluation}
\taxnode{1}{Application domains and case studies}
\taxnode{1}{Surveys and taxonomies of VQCs and QNNs}
\taxnode{0}{Quantum kernel methods}
\taxnode{1}{Quantum feature maps}
\taxnode{1}{Quantum kernel estimation}
\taxnode{1}{Kernelized quantum learning algorithms}
\taxnode{1}{Trainable quantum kernels}
\taxnode{1}{Expressivity and inductive bias}
\taxnode{1}{Generalization and learning theory}
\taxnode{1}{Quantum advantage and complexity analysis}
\taxnode{1}{Noise robustness and error mitigation}
\taxnode{1}{Scalable and approximate quantum kernels}
\taxnode{1}{Benchmarking and empirical evaluation}
\taxnode{1}{Applications of quantum kernel methods}
\taxnode{1}{Surveys and reviews of quantum kernel methods}
\taxnode{0}{Quantum data encoding and representation learning}
\taxnode{1}{Classical-to-quantum data encoding}
\taxnode{1}{Quantum feature maps and kernel embeddings}
\taxnode{1}{Variational quantum embeddings}
\taxnode{1}{Quantum state representation learning}
\taxnode{1}{Quantum autoencoders and dimensionality reduction}
\taxnode{1}{Theory of quantum representations}
\taxnode{1}{Noise-aware and hardware-efficient encoding}
\taxnode{1}{Benchmarking and evaluation}
\taxnode{1}{Application-specific quantum representations}
\taxnode{1}{Surveys and taxonomies}
\taxnode{0}{Quantum generative models}
\taxnode{1}{Quantum circuit Born machines}
\taxnode{1}{Quantum generative adversarial networks}
\taxnode{1}{Quantum Boltzmann machines and energy-based models}
\taxnode{1}{Quantum autoencoders and latent-variable models}
\taxnode{1}{Autoregressive quantum generative models}
\taxnode{1}{Quantum diffusion and score-based models}
\taxnode{1}{Theory of quantum generative modeling}
\taxnode{1}{Training and optimization methods}
\taxnode{1}{Noise, hardware, and implementation}
\taxnode{1}{Benchmarking and evaluation}
\taxnode{1}{Applications of quantum generative models}
\taxnode{1}{Surveys and comparative reviews}
\taxnode{0}{Quantum reinforcement learning}
\taxnode{1}{Quantum value-based reinforcement learning}
\taxnode{1}{Quantum policy gradient and actor-critic methods}
\taxnode{1}{Model-based quantum reinforcement learning}
\taxnode{1}{Hybrid quantum-classical reinforcement learning}
\taxnode{1}{Quantum state representation and action encoding}
\taxnode{1}{Quantum environments and quantum decision processes}
\taxnode{1}{Theoretical foundations of quantum reinforcement learning}
\taxnode{1}{Noise-resilient and hardware-aware quantum reinforcement learning}
\taxnode{1}{Quantum control and quantum error correction with reinforcement learning}
\taxnode{1}{Benchmarking and empirical evaluation in quantum reinforcement learning}
\taxnode{1}{Surveys and taxonomies of quantum reinforcement learning}
\taxnode{0}{Quantum optimization and quantum annealing for learning}
\taxnode{1}{Ising and QUBO Formulations for Learning}
\taxnode{1}{Annealing-Based Training and Optimization of ML Models}
\taxnode{1}{Annealing-Based Inference and Sampling}
\taxnode{1}{Feature Selection, Structure Learning, and Model Selection}
\taxnode{1}{Hybrid Quantum-Classical Annealing Methods}
\taxnode{1}{Hardware-Aware Embedding and Compilation for Learning}
\taxnode{1}{Theory and Complexity of Annealing-Based Learning}
\taxnode{1}{Noise, Robustness, and Error Mitigation in Annealing-Based Learning}
\taxnode{1}{Benchmarking and Comparative Evaluation}
\taxnode{1}{Applied Quantum Optimization for Domain-Specific Learning}
\taxnode{0}{Hybrid quantum-classical learning}
\taxnode{1}{Quantum neural networks and hybrid architectures}
\taxnode{1}{Quantum kernel methods}
\taxnode{1}{Generative and unsupervised learning}
\taxnode{1}{Quantum reinforcement learning}
\taxnode{1}{Data encoding and circuit architecture}
\taxnode{1}{Optimization, trainability, and learning theory}
\taxnode{1}{Noise-aware and hardware-efficient learning}
\taxnode{1}{Benchmarking and quantum advantage}
\taxnode{1}{Application-driven hybrid quantum-classical learning}
\taxnode{1}{Surveys and reviews}
\taxnode{0}{Quantum learning theory and complexity}
\taxnode{1}{Quantum PAC and agnostic learning}
\taxnode{1}{Sample complexity and generalization theory}
\taxnode{1}{Query complexity and oracle learning}
\taxnode{1}{Computational complexity and hardness of learning}
\taxnode{1}{Quantum–classical separations and dequantization}
\taxnode{1}{Expressivity and representational power}
\taxnode{1}{Trainability and optimization complexity}
\taxnode{1}{Learning quantum states, channels, and Hamiltonians}
\taxnode{0}{Trainability, expressivity, and generalization of QML}
\taxnode{1}{Barren plateaus and concentration phenomena}
\taxnode{1}{Optimization landscapes and variational dynamics}
\taxnode{1}{Ansatz, architecture, and initialization design}
\taxnode{1}{Expressivity and approximation theory}
\taxnode{1}{Quantum data embeddings, feature maps, and kernels}
\taxnode{1}{Generalization, learnability, and sample complexity}
\taxnode{1}{Noise, robustness, and hardware-aware effects}
\taxnode{1}{Expressivity-trainability-generalization trade-offs}
\taxnode{1}{Empirical benchmarks and comparative evaluations}
\taxnode{1}{Surveys and unifying theoretical frameworks}
\taxnode{0}{Noise-resilient and hardware-aware quantum machine learning}
\taxnode{1}{Noise modeling and characterization in quantum machine learning}
\taxnode{1}{Theoretical analysis of noisy quantum learning systems}
\taxnode{1}{Error mitigation and error suppression for quantum machine learning}
\taxnode{1}{Noise-aware optimization and robust training of variational quantum models}
\taxnode{1}{Hardware-efficient ansatz and model architecture design}
\taxnode{1}{Quantum compilation and execution optimization for QML}
\taxnode{1}{Device-aware co-design and resource estimation for QML}
\taxnode{1}{Benchmarking and empirical evaluation of noisy QML}
\taxnode{1}{Applications and experimental demonstrations on near-term quantum hardware}
\taxnode{1}{Surveys and comparative studies of noise-resilient and hardware-aware QML}
\taxnode{0}{Benchmarking and empirical evaluation of QML}
\taxnode{1}{Benchmark datasets and task suites}
\taxnode{1}{Comparative studies against classical baselines}
\taxnode{1}{Resource and complexity evaluation}
\taxnode{1}{Scalability analysis}
\taxnode{1}{Noise and hardware-aware benchmarking}
\taxnode{1}{Trainability and optimization evaluation}
\taxnode{1}{Generalization and robustness assessment}
\taxnode{1}{Application-oriented benchmarking}
\taxnode{1}{Evaluation protocols, reproducibility, and statistical methodology}
\taxnode{1}{Survey and meta-analysis of empirical QML}
\taxnode{0}{Applied quantum machine learning}
\taxnode{1}{Quantum supervised learning}
\taxnode{1}{Quantum unsupervised learning}
\taxnode{1}{Quantum generative modeling}
\taxnode{1}{Quantum reinforcement learning}
\taxnode{1}{Quantum neural networks and hybrid quantum-classical architectures}
\taxnode{1}{Quantum kernel methods and feature maps}
\taxnode{1}{Trainability, optimization, and learning theory in quantum machine learning}
\taxnode{1}{Noise-aware and hardware-efficient quantum machine learning}
\taxnode{1}{Benchmarking and empirical evaluation of quantum machine learning}
\taxnode{1}{Domain-specific applications of quantum machine learning}
\taxnode{1}{Surveys and comparative studies of quantum machine learning}
\taxnode{0}{Quantum-inspired machine learning}
\taxnode{1}{Tensor Network Methods}
\taxnode{1}{Quantum-Inspired Optimization}
\taxnode{1}{Quantum-Inspired Kernel Methods}
\taxnode{1}{Quantum-Inspired Neural Architectures}
\taxnode{1}{Quantum-Inspired Generative and Probabilistic Models}
\taxnode{1}{Quantum-Inspired Linear Algebra and Sampling Algorithms}
\taxnode{1}{Theory and Complexity Analysis}
\taxnode{1}{Benchmarking and Empirical Evaluation}
\taxnode{1}{Applied Quantum-Inspired Machine Learning}
\taxnode{1}{Surveys, Taxonomies, and Methodology}
\taxnode{0}{Surveys and taxonomies of quantum machine learning}
\taxnode{1}{Comprehensive overviews and taxonomies of quantum machine learning}
\taxnode{1}{Surveys of variational quantum circuits and hybrid quantum-classical learning}
\taxnode{1}{Surveys of quantum kernel methods and feature maps}
\taxnode{1}{Surveys of quantum neural networks and deep quantum models}
\taxnode{1}{Surveys of quantum generative models and probabilistic learning}
\taxnode{1}{Surveys of quantum reinforcement learning}
\taxnode{1}{Surveys of data encoding, state preparation, and quantum representations}
\taxnode{1}{Surveys of learning theory, expressivity, trainability, and generalization in QML}
\taxnode{1}{Surveys of noise, robustness, error mitigation, and fault tolerance for QML}
\taxnode{1}{Benchmarking and empirical evaluation surveys for QML}
\taxnode{1}{Application-oriented surveys of quantum machine learning}
\taxnode{1}{Surveys of software frameworks, hardware platforms, and implementation workflows}
\taxnode{1}{Systematic reviews, bibliometric studies, and survey methodology for QML}
\taxfield{reinforcement\_\allowbreak{}learning}
\taxnode{0}{Value-Based Reinforcement Learning}
\taxnode{1}{Policy Evaluation and Value Prediction}
\taxnode{1}{Value-Based Control Algorithms}
\taxnode{1}{Deep Value Function Approximation}
\taxnode{1}{Distributional Reinforcement Learning}
\taxnode{1}{Exploration and Uncertainty Estimation}
\taxnode{1}{Offline and Off-Policy Value Learning}
\taxnode{1}{Model-Based Value Learning and Planning}
\taxnode{1}{Representation Learning and Generalization}
\taxnode{1}{Hierarchical Value-Based Reinforcement Learning}
\taxnode{1}{Multi-Agent Value-Based Reinforcement Learning}
\taxnode{1}{Robust and Safe Value-Based Reinforcement Learning}
\taxnode{1}{Theoretical Analysis of Value-Based RL}
\taxnode{1}{Benchmarking and Empirical Evaluation}
\taxnode{1}{Applications of Value-Based Reinforcement Learning}
\taxnode{1}{Surveys and Reviews of Value-Based Reinforcement Learning}
\taxnode{0}{Policy Gradient and Actor-Critic Methods}
\taxnode{1}{Policy Gradient Theory and Convergence}
\taxnode{1}{Stochastic Policy Gradient Methods}
\taxnode{1}{On-Policy Actor-Critic Methods}
\taxnode{1}{Off-Policy Actor-Critic Methods}
\taxnode{1}{Deterministic Policy Gradient Methods}
\taxnode{1}{Natural Policy Gradient Methods}
\taxnode{1}{Trust Region and Proximal Policy Optimization}
\taxnode{1}{Entropy-Regularized and Maximum-Entropy Methods}
\taxnode{1}{Variance Reduction and Credit Assignment}
\taxnode{1}{Safe and Constrained Policy Optimization}
\taxnode{1}{Model-Based Actor-Critic}
\taxnode{1}{Offline Actor-Critic and Policy Optimization}
\taxnode{1}{Multi-Agent Actor-Critic}
\taxnode{1}{Hierarchical Actor-Critic}
\taxnode{1}{Benchmarking and Empirical Evaluation}
\taxnode{1}{Applications of Policy Gradient and Actor-Critic Methods}
\taxnode{0}{Model-Based Reinforcement Learning}
\taxnode{1}{Environment Model Learning}
\taxnode{1}{Latent Dynamics and World Models}
\taxnode{1}{Planning and Model Predictive Control}
\taxnode{1}{Dyna-Style Model-Based Reinforcement Learning}
\taxnode{1}{Exploration in Model-Based Reinforcement Learning}
\taxnode{1}{Uncertainty-Aware and Robust Model-Based RL}
\taxnode{1}{Offline Model-Based Reinforcement Learning}
\taxnode{1}{Theoretical Foundations of Model-Based RL}
\taxnode{1}{Benchmarking and Empirical Evaluation for MBRL}
\taxnode{1}{Applications of Model-Based Reinforcement Learning}
\taxnode{1}{Surveys and Taxonomies of Model-Based RL}
\taxnode{0}{Offline Reinforcement Learning}
\taxnode{1}{Value-Based Offline Reinforcement Learning}
\taxnode{1}{Policy-Constrained and Regularized Offline Reinforcement Learning}
\taxnode{1}{Model-Based Offline Reinforcement Learning}
\taxnode{1}{Sequence Modeling and Supervised Offline Reinforcement Learning}
\taxnode{1}{Off-Policy Evaluation and Policy Selection}
\taxnode{1}{Theory and Generalization in Offline Reinforcement Learning}
\taxnode{1}{Representation Learning for Offline Reinforcement Learning}
\taxnode{1}{Dataset Quality and Curation for Offline Reinforcement Learning}
\taxnode{1}{Safe and Robust Offline Reinforcement Learning}
\taxnode{1}{Offline-to-Online Reinforcement Learning}
\taxnode{1}{Datasets, Benchmarks, and Empirical Evaluation}
\taxnode{1}{Applications of Offline Reinforcement Learning}
\taxnode{1}{Surveys and Taxonomies of Offline Reinforcement Learning}
\taxnode{0}{Imitation Learning and Inverse Reinforcement Learning}
\taxnode{1}{Behavior Cloning}
\taxnode{1}{Interactive and Active Imitation Learning}
\taxnode{1}{Offline Imitation Learning}
\taxnode{1}{Reinforcement Learning from Demonstrations}
\taxnode{1}{Inverse Reinforcement Learning}
\taxnode{1}{Adversarial Imitation Learning}
\taxnode{1}{Preference-Based Reward Learning}
\taxnode{1}{Imitation from Observation}
\taxnode{1}{Hierarchical Imitation Learning}
\taxnode{1}{Transfer and Meta-Imitation Learning}
\taxnode{1}{Multi-Agent Imitation Learning and IRL}
\taxnode{1}{Theory and Foundations of IL/\allowbreak{}IRL}
\taxnode{1}{Robustness and Safety in IL/\allowbreak{}IRL}
\taxnode{1}{Evaluation and Benchmarking for IL/\allowbreak{}IRL}
\taxnode{1}{Applications of IL/\allowbreak{}IRL}
\taxnode{0}{Hierarchical Reinforcement Learning}
\taxnode{1}{Temporal Abstraction and Options}
\taxnode{1}{Goal-Conditioned, Subgoal-Based, and Feudal HRL}
\taxnode{1}{Skill Discovery and Unsupervised Hierarchy Learning}
\taxnode{1}{Model-Based and Planning-Centric HRL}
\taxnode{1}{Representation Learning for Hierarchical Control}
\taxnode{1}{Exploration and Intrinsic Motivation in HRL}
\taxnode{1}{Learning and Optimization Methods for HRL}
\taxnode{1}{Theory and Foundations of Hierarchical Reinforcement Learning}
\taxnode{1}{Benchmarks and Empirical Evaluation for Hierarchical Reinforcement Learning}
\taxnode{1}{Applications of Hierarchical Reinforcement Learning}
\taxnode{1}{Surveys and Taxonomies of Hierarchical Reinforcement Learning}
\taxnode{0}{Multi-Agent Reinforcement Learning}
\taxnode{1}{Cooperative Multi-Agent Reinforcement Learning}
\taxnode{1}{Competitive Multi-Agent Reinforcement Learning}
\taxnode{1}{General-Sum Multi-Agent Reinforcement Learning}
\taxnode{1}{Communication and Coordination in Multi-Agent Reinforcement Learning}
\taxnode{1}{Credit Assignment and Value Decomposition}
\taxnode{1}{Opponent Modeling and Recursive Reasoning}
\taxnode{1}{Large-Scale Multi-Agent Reinforcement Learning}
\taxnode{1}{Algorithmic Frameworks for Multi-Agent Reinforcement Learning}
\taxnode{1}{Game-Theoretic and Theoretical Foundations of Multi-Agent Reinforcement Learning}
\taxnode{1}{Benchmarking and Evaluation in Multi-Agent Reinforcement Learning}
\taxnode{1}{Applied Multi-Agent Reinforcement Learning}
\taxnode{1}{Surveys and Empirical Studies of Multi-Agent Reinforcement Learning}
\taxnode{0}{Exploration and Intrinsic Motivation}
\taxnode{1}{Count-Based and Pseudocount Exploration}
\taxnode{1}{Uncertainty-Driven Exploration}
\taxnode{1}{Curiosity-Driven Intrinsic Motivation}
\taxnode{1}{Information-Theoretic Exploration}
\taxnode{1}{Entropy-Regularized Exploration}
\taxnode{1}{Skill Discovery and Empowerment}
\taxnode{1}{Goal-Conditioned and Hierarchical Exploration}
\taxnode{1}{Safe Exploration}
\taxnode{1}{Theoretical Analysis of Exploration}
\taxnode{1}{Exploration Benchmarks and Evaluation}
\taxnode{1}{Embodied and Real-World Exploration}
\taxnode{0}{Safe and Constrained Reinforcement Learning}
\taxnode{1}{Foundations and Theoretical Guarantees}
\taxnode{1}{Constrained Markov Decision Processes and Constrained Policy Optimization}
\taxnode{1}{Risk-Sensitive and Chance-Constrained Reinforcement Learning}
\taxnode{1}{Safe Exploration}
\taxnode{1}{Model-Based Safe Reinforcement Learning}
\taxnode{1}{Robust and Adversarial Safe Reinforcement Learning}
\taxnode{1}{Shielding, Runtime Safety Filters, and Formal Verification}
\taxnode{1}{Safe Policy Improvement and Offline Safe Reinforcement Learning}
\taxnode{1}{Safe Reinforcement Learning Benchmarks and Evaluation}
\taxnode{1}{Safety-Critical Applications of Reinforcement Learning}
\taxnode{1}{Surveys and Taxonomies of Safe Reinforcement Learning}
\taxnode{0}{Meta-Reinforcement Learning and Transfer}
\taxnode{1}{Fast Adaptation and Few-Shot Reinforcement Learning}
\taxnode{1}{Multi-Task Reinforcement Learning}
\taxnode{1}{Cross-Task Transfer Learning in Reinforcement Learning}
\taxnode{1}{Continual and Lifelong Reinforcement Learning}
\taxnode{1}{Skill Transfer and Policy Reuse}
\taxnode{1}{Transferable Representation Learning}
\taxnode{1}{Domain Adaptation and Sim-to-Real Transfer}
\taxnode{1}{Theoretical Foundations of Meta- and Transfer Reinforcement Learning}
\taxnode{1}{Benchmarking and Evaluation for Meta- and Transfer Reinforcement Learning}
\taxnode{1}{Applications of Meta- and Transfer Reinforcement Learning}
\taxnode{1}{Surveys and Taxonomies of Meta- and Transfer Reinforcement Learning}
\taxnode{0}{Representation Learning for Reinforcement Learning}
\taxnode{1}{Predictive and Reconstruction-Based Representation Learning}
\taxnode{1}{Contrastive Representation Learning}
\taxnode{1}{Latent Dynamics and World Models}
\taxnode{1}{State Abstraction and Bisimulation}
\taxnode{1}{Successor Representations and Successor Features}
\taxnode{1}{Disentangled, Object-Centric, and Compositional Representations}
\taxnode{1}{Action, Skill, and Option Representations}
\taxnode{1}{Pretraining, Transfer, and Multi-Task Representation Learning}
\taxnode{1}{Belief-State and Memory Representations}
\taxnode{1}{Theoretical Foundations of RL Representations}
\taxnode{1}{Benchmarks and Evaluation of RL Representations}
\taxnode{1}{Applications in Visual, Robotic, and Embodied Reinforcement Learning}
\taxnode{1}{Surveys and Taxonomies of Representation Learning for RL}
\taxnode{0}{Reinforcement Learning Theory}
\taxnode{1}{MDP and Dynamic Programming Theory}
\taxnode{1}{Sample Complexity and Regret Analysis}
\taxnode{1}{Exploration Theory}
\taxnode{1}{Convergence and Stochastic Approximation Theory}
\taxnode{1}{Function Approximation, Representation, and Generalization Theory}
\taxnode{1}{Value-Based Reinforcement Learning Theory}
\taxnode{1}{Policy Optimization Theory}
\taxnode{1}{Model-Based Reinforcement Learning Theory}
\taxnode{1}{Offline and Batch Reinforcement Learning Theory}
\taxnode{1}{Partial Observability and Latent-State Theory}
\taxnode{1}{Hierarchical Reinforcement Learning and Temporal Abstraction Theory}
\taxnode{1}{Multi-Agent and Game-Theoretic RL Theory}
\taxnode{1}{Safe and Robust Reinforcement Learning Theory}
\taxnode{1}{Computational Complexity and Lower Bounds}
\taxnode{1}{Evaluation Methodology and Theoretical Benchmarks}
\taxnode{1}{Surveys and Unifying Frameworks}
\taxnode{0}{Benchmarking and Evaluation in Reinforcement Learning}
\taxnode{1}{Benchmark Environment and Suite Design}
\taxnode{1}{Evaluation Metrics and Experimental Protocols}
\taxnode{1}{Statistical Evaluation and Reproducibility}
\taxnode{1}{Sample Efficiency and Computational Efficiency Evaluation}
\taxnode{1}{Generalization and Transfer Evaluation}
\taxnode{1}{Robustness Evaluation}
\taxnode{1}{Safety and Risk Evaluation}
\taxnode{1}{Off-Policy Evaluation}
\taxnode{1}{Sim-to-Real and Real-World Evaluation}
\taxnode{1}{Diagnostic Evaluation and Ablation Analysis}
\taxnode{1}{Survey and Comparative Studies}
\taxnode{0}{Applications of Reinforcement Learning}
\taxnode{1}{Robotics and Embodied Control}
\taxnode{1}{Games and Interactive Environments}
\taxnode{1}{Transportation and Autonomous Navigation}
\taxnode{1}{Recommender Systems, Personalization, and Advertising}
\taxnode{1}{Operations Research and Resource Allocation}
\taxnode{1}{Networking, Systems, and Cybersecurity}
\taxnode{1}{Healthcare and Clinical Decision Making}
\taxnode{1}{Finance and Economic Decision Making}
\taxnode{1}{Energy Systems and Smart Infrastructure}
\taxnode{1}{Natural Language Processing and Dialogue Systems}
\taxnode{1}{Computer Vision and Multimedia Systems}
\taxnode{1}{Industrial Automation and Manufacturing}
\taxnode{1}{Scientific Discovery and Engineering Design}
\taxnode{1}{Human-Computer Interaction and Education}
\taxnode{1}{Benchmarking, Evaluation, and Survey Studies of RL Applications}
\taxnode{1}{Application-specific Reinforcement Learning Methodology and Deployment}
\taxnode{0}{Surveys and Empirical Studies in Reinforcement Learning}
\taxnode{1}{Survey and Taxonomy Papers}
\taxnode{1}{Comparative Benchmarking Studies}
\taxnode{1}{Benchmark Suites and Evaluation Environments}
\taxnode{1}{Reproducibility and Replication Studies}
\taxnode{1}{Evaluation Methodology and Statistical Analysis}
\taxnode{1}{Ablation and Sensitivity Analysis}
\taxnode{1}{Generalization and Robustness Studies}
\taxnode{1}{Scaling and Efficiency Studies}
\taxnode{1}{Domain-Specific Surveys and Empirical Reviews}
\taxnode{1}{Meta-Analysis and Research Trend Studies}
\taxfield{representation\_\allowbreak{}learning}
\taxnode{0}{self\_\allowbreak{}supervised\_\allowbreak{}representation\_\allowbreak{}learning}
\taxnode{1}{contrastive\_\allowbreak{}learning}
\taxnode{1}{non\_\allowbreak{}contrastive\_\allowbreak{}learning}
\taxnode{1}{clustering\_\allowbreak{}and\_\allowbreak{}prototype\_\allowbreak{}based\_\allowbreak{}learning}
\taxnode{1}{masked\_\allowbreak{}and\_\allowbreak{}denoising\_\allowbreak{}modeling}
\taxnode{1}{predictive\_\allowbreak{}and\_\allowbreak{}autoregressive\_\allowbreak{}learning}
\taxnode{1}{multimodal\_\allowbreak{}and\_\allowbreak{}cross\_\allowbreak{}view\_\allowbreak{}learning}
\taxnode{1}{augmentation\_\allowbreak{}and\_\allowbreak{}pretext\_\allowbreak{}task\_\allowbreak{}design}
\taxnode{1}{theoretical\_\allowbreak{}foundations\_\allowbreak{}and\_\allowbreak{}analysis}
\taxnode{1}{evaluation\_\allowbreak{}and\_\allowbreak{}benchmarking}
\taxnode{1}{transfer\_\allowbreak{}learning\_\allowbreak{}and\_\allowbreak{}applications}
\taxnode{0}{generative\_\allowbreak{}and\_\allowbreak{}reconstruction\_\allowbreak{}based\_\allowbreak{}representation\_\allowbreak{}learning}
\taxnode{1}{linear\_\allowbreak{}latent\_\allowbreak{}factor\_\allowbreak{}models}
\taxnode{1}{sparse\_\allowbreak{}coding\_\allowbreak{}and\_\allowbreak{}dictionary\_\allowbreak{}learning}
\taxnode{1}{autoencoder\_\allowbreak{}based\_\allowbreak{}representation\_\allowbreak{}learning}
\taxnode{1}{variational\_\allowbreak{}autoencoders\_\allowbreak{}and\_\allowbreak{}probabilistic\_\allowbreak{}latent\_\allowbreak{}variable\_\allowbreak{}models}
\taxnode{1}{adversarial\_\allowbreak{}representation\_\allowbreak{}learning}
\taxnode{1}{autoregressive\_\allowbreak{}representation\_\allowbreak{}learning}
\taxnode{1}{normalizing\_\allowbreak{}flows\_\allowbreak{}and\_\allowbreak{}invertible\_\allowbreak{}representation\_\allowbreak{}learning}
\taxnode{1}{energy\_\allowbreak{}based\_\allowbreak{}and\_\allowbreak{}undirected\_\allowbreak{}generative\_\allowbreak{}models}
\taxnode{1}{diffusion\_\allowbreak{}and\_\allowbreak{}score\_\allowbreak{}based\_\allowbreak{}representation\_\allowbreak{}learning}
\taxnode{1}{disentangled\_\allowbreak{}representation\_\allowbreak{}learning}
\taxnode{1}{hierarchical\_\allowbreak{}and\_\allowbreak{}structured\_\allowbreak{}latent\_\allowbreak{}representation\_\allowbreak{}learning}
\taxnode{1}{multimodal\_\allowbreak{}generative\_\allowbreak{}representation\_\allowbreak{}learning}
\taxnode{1}{reconstruction\_\allowbreak{}based\_\allowbreak{}self\_\allowbreak{}supervised\_\allowbreak{}learning}
\taxnode{1}{theory\_\allowbreak{}and\_\allowbreak{}identifiability\_\allowbreak{}of\_\allowbreak{}generative\_\allowbreak{}representations}
\taxnode{1}{benchmarking\_\allowbreak{}and\_\allowbreak{}empirical\_\allowbreak{}evaluation\_\allowbreak{}of\_\allowbreak{}generative\_\allowbreak{}representations}
\taxnode{1}{application\_\allowbreak{}oriented\_\allowbreak{}generative\_\allowbreak{}representation\_\allowbreak{}learning}
\taxnode{0}{metric\_\allowbreak{}learning\_\allowbreak{}and\_\allowbreak{}embedding\_\allowbreak{}learning}
\taxnode{1}{Distance Metric Learning}
\taxnode{1}{Deep Metric Learning}
\taxnode{1}{Contrastive and Self-Supervised Embedding Learning}
\taxnode{1}{Manifold and Structure-Preserving Embedding}
\taxnode{1}{Geometric and Non-Euclidean Embedding Learning}
\taxnode{1}{Multimodal and Cross-Modal Embedding Learning}
\taxnode{1}{Metric Learning Objectives and Mining Strategies}
\taxnode{1}{Metric Learning Theory}
\taxnode{1}{Evaluation and Benchmarking for Metric Learning}
\taxnode{1}{Applied Metric and Embedding Learning}
\taxnode{1}{Surveys and Taxonomies of Metric Learning}
\taxnode{0}{disentangled\_\allowbreak{}representation\_\allowbreak{}learning}
\taxnode{1}{Unsupervised Disentangled Representation Learning}
\taxnode{1}{Weakly Supervised Disentangled Representation Learning}
\taxnode{1}{Supervised Disentangled Representation Learning}
\taxnode{1}{Identifiability and Theoretical Foundations of Disentanglement}
\taxnode{1}{Causal Disentangled Representation Learning}
\taxnode{1}{Structured Generative Models for Disentanglement}
\taxnode{1}{Temporal, Sequential, and Multimodal Disentanglement}
\taxnode{1}{Disentanglement Metrics, Benchmarks, and Comparative Evaluation}
\taxnode{1}{Applications of Disentangled Representations}
\taxnode{0}{invariant\_\allowbreak{}equivariant\_\allowbreak{}and\_\allowbreak{}causal\_\allowbreak{}representation\_\allowbreak{}learning}
\taxnode{1}{Invariant Representation Learning}
\taxnode{1}{Equivariant Representation Learning}
\taxnode{1}{Symmetry Discovery}
\taxnode{1}{Causal Representation Learning}
\taxnode{1}{Counterfactual and Interventional Representation Learning}
\taxnode{1}{Domain Generalization and Invariant Prediction}
\taxnode{1}{Identifiability and Theoretical Foundations}
\taxnode{1}{Evaluation, Benchmarks, and Empirical Surveys}
\taxnode{1}{Applications of Invariant, Equivariant, and Causal Representations}
\taxnode{0}{multiview\_\allowbreak{}and\_\allowbreak{}multimodal\_\allowbreak{}representation\_\allowbreak{}learning}
\taxnode{1}{multiview\_\allowbreak{}subspace\_\allowbreak{}and\_\allowbreak{}correlation\_\allowbreak{}learning}
\taxnode{1}{cross\_\allowbreak{}modal\_\allowbreak{}alignment\_\allowbreak{}and\_\allowbreak{}joint\_\allowbreak{}embedding\_\allowbreak{}learning}
\taxnode{1}{multimodal\_\allowbreak{}fusion\_\allowbreak{}and\_\allowbreak{}integration}
\taxnode{1}{self\_\allowbreak{}supervised\_\allowbreak{}and\_\allowbreak{}contrastive\_\allowbreak{}multimodal\_\allowbreak{}learning}
\taxnode{1}{generative\_\allowbreak{}multimodal\_\allowbreak{}representation\_\allowbreak{}learning}
\taxnode{1}{shared\_\allowbreak{}private\_\allowbreak{}and\_\allowbreak{}disentangled\_\allowbreak{}representation\_\allowbreak{}learning}
\taxnode{1}{incomplete\_\allowbreak{}and\_\allowbreak{}missing\_\allowbreak{}modality\_\allowbreak{}learning}
\taxnode{1}{cross\_\allowbreak{}modal\_\allowbreak{}translation\_\allowbreak{}retrieval\_\allowbreak{}and\_\allowbreak{}transfer}
\taxnode{1}{theoretical\_\allowbreak{}foundations\_\allowbreak{}of\_\allowbreak{}multiview\_\allowbreak{}and\_\allowbreak{}multimodal\_\allowbreak{}learning}
\taxnode{1}{benchmarking\_\allowbreak{}evaluation\_\allowbreak{}and\_\allowbreak{}surveys}
\taxnode{0}{geometric\_\allowbreak{}and\_\allowbreak{}structured\_\allowbreak{}representation\_\allowbreak{}learning}
\taxnode{1}{graph\_\allowbreak{}representation\_\allowbreak{}learning}
\taxnode{1}{relational\_\allowbreak{}and\_\allowbreak{}knowledge\_\allowbreak{}graph\_\allowbreak{}representation\_\allowbreak{}learning}
\taxnode{1}{manifold\_\allowbreak{}mesh\_\allowbreak{}and\_\allowbreak{}point\_\allowbreak{}cloud\_\allowbreak{}representation\_\allowbreak{}learning}
\taxnode{1}{equivariant\_\allowbreak{}and\_\allowbreak{}symmetry\_\allowbreak{}aware\_\allowbreak{}representation\_\allowbreak{}learning}
\taxnode{1}{hyperbolic\_\allowbreak{}and\_\allowbreak{}non\_\allowbreak{}euclidean\_\allowbreak{}representation\_\allowbreak{}learning}
\taxnode{1}{topological\_\allowbreak{}representation\_\allowbreak{}learning}
\taxnode{1}{set\_\allowbreak{}and\_\allowbreak{}permutation\_\allowbreak{}invariant\_\allowbreak{}representation\_\allowbreak{}learning}
\taxnode{1}{structured\_\allowbreak{}latent\_\allowbreak{}representation\_\allowbreak{}learning}
\taxnode{0}{transferable\_\allowbreak{}and\_\allowbreak{}domain\_\allowbreak{}adaptive\_\allowbreak{}representation\_\allowbreak{}learning}
\taxnode{1}{transfer\_\allowbreak{}learning\_\allowbreak{}and\_\allowbreak{}fine\_\allowbreak{}tuning}
\taxnode{1}{domain\_\allowbreak{}adaptation}
\taxnode{1}{multi\_\allowbreak{}source\_\allowbreak{}and\_\allowbreak{}source\_\allowbreak{}free\_\allowbreak{}domain\_\allowbreak{}adaptation}
\taxnode{1}{domain\_\allowbreak{}generalization}
\taxnode{1}{test\_\allowbreak{}time\_\allowbreak{}and\_\allowbreak{}online\_\allowbreak{}adaptation}
\taxnode{1}{invariant\_\allowbreak{}and\_\allowbreak{}causal\_\allowbreak{}representation\_\allowbreak{}learning}
\taxnode{1}{meta\_\allowbreak{}learning\_\allowbreak{}and\_\allowbreak{}few\_\allowbreak{}shot\_\allowbreak{}transfer}
\taxnode{1}{theoretical\_\allowbreak{}foundations\_\allowbreak{}of\_\allowbreak{}transferability\_\allowbreak{}and\_\allowbreak{}adaptation}
\taxnode{1}{benchmarks\_\allowbreak{}datasets\_\allowbreak{}and\_\allowbreak{}evaluation\_\allowbreak{}protocols}
\taxnode{1}{applications\_\allowbreak{}under\_\allowbreak{}domain\_\allowbreak{}shift}
\taxnode{1}{surveys\_\allowbreak{}and\_\allowbreak{}empirical\_\allowbreak{}studies}
\taxnode{0}{theory\_\allowbreak{}of\_\allowbreak{}representation\_\allowbreak{}learning}
\taxnode{1}{information\_\allowbreak{}theoretic\_\allowbreak{}foundations}
\taxnode{1}{geometric\_\allowbreak{}and\_\allowbreak{}manifold\_\allowbreak{}theory}
\taxnode{1}{expressivity\_\allowbreak{}and\_\allowbreak{}approximation\_\allowbreak{}theory}
\taxnode{1}{optimization\_\allowbreak{}dynamics\_\allowbreak{}and\_\allowbreak{}implicit\_\allowbreak{}bias}
\taxnode{1}{generalization\_\allowbreak{}and\_\allowbreak{}sample\_\allowbreak{}complexity}
\taxnode{1}{symmetry\_\allowbreak{}invariance\_\allowbreak{}and\_\allowbreak{}equivariance}
\taxnode{1}{disentanglement\_\allowbreak{}and\_\allowbreak{}identifiability}
\taxnode{1}{self\_\allowbreak{}supervised\_\allowbreak{}and\_\allowbreak{}contrastive\_\allowbreak{}learning\_\allowbreak{}theory}
\taxnode{1}{latent\_\allowbreak{}variable\_\allowbreak{}and\_\allowbreak{}generative\_\allowbreak{}representation\_\allowbreak{}theory}
\taxnode{1}{robustness\_\allowbreak{}and\_\allowbreak{}stability\_\allowbreak{}of\_\allowbreak{}representations}
\taxnode{1}{transferability\_\allowbreak{}and\_\allowbreak{}domain\_\allowbreak{}shift\_\allowbreak{}theory}
\taxnode{1}{causal\_\allowbreak{}representation\_\allowbreak{}learning\_\allowbreak{}theory}
\taxnode{1}{representation\_\allowbreak{}evaluation\_\allowbreak{}and\_\allowbreak{}quality\_\allowbreak{}metrics}
\taxnode{0}{representation\_\allowbreak{}evaluation\_\allowbreak{}probing\_\allowbreak{}and\_\allowbreak{}benchmarking}
\taxnode{1}{diagnostic\_\allowbreak{}probing}
\taxnode{1}{intrinsic\_\allowbreak{}representation\_\allowbreak{}metrics}
\taxnode{1}{representational\_\allowbreak{}similarity\_\allowbreak{}analysis}
\taxnode{1}{disentanglement\_\allowbreak{}evaluation}
\taxnode{1}{invariance\_\allowbreak{}equivariance\_\allowbreak{}and\_\allowbreak{}robustness\_\allowbreak{}assessment}
\taxnode{1}{transferability\_\allowbreak{}and\_\allowbreak{}downstream\_\allowbreak{}task\_\allowbreak{}benchmarking}
\taxnode{1}{retrieval\_\allowbreak{}clustering\_\allowbreak{}and\_\allowbreak{}nearest\_\allowbreak{}neighbor\_\allowbreak{}evaluation}
\taxnode{1}{fairness\_\allowbreak{}bias\_\allowbreak{}and\_\allowbreak{}privacy\_\allowbreak{}evaluation}
\taxnode{1}{benchmark\_\allowbreak{}dataset\_\allowbreak{}and\_\allowbreak{}protocol\_\allowbreak{}design}
\taxnode{1}{theoretical\_\allowbreak{}foundations\_\allowbreak{}of\_\allowbreak{}probing\_\allowbreak{}and\_\allowbreak{}representation\_\allowbreak{}evaluation}
\taxfield{robotics}
\taxnode{0}{Robot Perception}
\taxnode{1}{Object-Centric Perception}
\taxnode{1}{Scene Understanding and Semantic Mapping}
\taxnode{1}{3D Perception and Spatial Reconstruction}
\taxnode{1}{Localization, Mapping, and SLAM}
\taxnode{1}{Motion Estimation and Tracking}
\taxnode{1}{Multimodal Sensor Fusion}
\taxnode{1}{State Estimation and Proprioceptive Perception}
\taxnode{1}{Tactile, Haptic, and Contact Perception}
\taxnode{1}{Auditory and Acoustic Perception}
\taxnode{1}{Human-Centered Perception}
\taxnode{1}{Active and Embodied Perception}
\taxnode{1}{Representation Learning and Adaptation for Robot Perception}
\taxnode{1}{Probabilistic and Uncertainty-Aware Perception}
\taxnode{1}{Datasets, Benchmarks, and Evaluation}
\taxnode{0}{Localization, Mapping, and Navigation}
\taxnode{1}{State Estimation and Sensor Fusion}
\taxnode{1}{Odometry and Ego-Motion Estimation}
\taxnode{1}{Global Localization and Relocalization}
\taxnode{1}{Simultaneous Localization and Mapping (SLAM)}
\taxnode{1}{Place Recognition and Loop Closure Detection}
\taxnode{1}{Map Representation and Environment Reconstruction}
\taxnode{1}{Exploration and Active Mapping}
\taxnode{1}{Navigation and Motion Planning}
\taxnode{1}{Embodied and Visual Navigation}
\taxnode{1}{Datasets, Benchmarks, and Evaluation}
\taxnode{0}{Manipulation and Grasping}
\taxnode{1}{Grasp Detection, Synthesis, and Analysis}
\taxnode{1}{Perception and Affordance Learning for Manipulation}
\taxnode{1}{Task and Motion Planning for Manipulation}
\taxnode{1}{Learning-Based Manipulation Policies}
\taxnode{1}{Dexterous and In-Hand Manipulation}
\taxnode{1}{Nonprehensile Manipulation}
\taxnode{1}{Contact-Rich Manipulation and Robotic Assembly}
\taxnode{1}{Deformable Object Manipulation}
\taxnode{1}{Bimanual and Multi-Arm Manipulation}
\taxnode{1}{Tactile Sensing and Haptic Manipulation}
\taxnode{1}{Simulation and Sim-to-Real Transfer for Manipulation}
\taxnode{1}{Datasets, Benchmarks, and Evaluation for Manipulation}
\taxnode{0}{Locomotion}
\taxnode{1}{Dynamics Modeling and Contact Mechanics}
\taxnode{1}{Gait Generation and Motion Synthesis}
\taxnode{1}{Locomotion Planning and Trajectory Optimization}
\taxnode{1}{Locomotion Control and Policy Learning}
\taxnode{1}{Perception-Guided and Terrain-Adaptive Locomotion}
\taxnode{1}{Sim-to-Real Transfer and Domain Adaptation}
\taxnode{1}{Robust, Safe, and Energy-Efficient Locomotion}
\taxnode{1}{Legged and Humanoid Locomotion}
\taxnode{1}{Wheeled and Tracked Locomotion}
\taxnode{1}{Aerial Locomotion}
\taxnode{1}{Aquatic and Amphibious Locomotion}
\taxnode{1}{Soft and Bio-inspired Locomotion}
\taxnode{1}{Benchmarks, Datasets, and Evaluation}
\taxnode{0}{Motion Planning and Control}
\taxnode{1}{Sampling-Based and Search-Based Planning}
\taxnode{1}{Trajectory Optimization and Optimal Control}
\taxnode{1}{Model Predictive Control}
\taxnode{1}{Reinforcement Learning for Control}
\taxnode{1}{Imitation Learning and Learning from Demonstration}
\taxnode{1}{Safe, Robust, and Constrained Control}
\taxnode{1}{Uncertainty-Aware and Probabilistic Planning}
\taxnode{1}{Multi-Agent Motion Planning and Control}
\taxnode{1}{Hierarchical Planning and Task and Motion Planning}
\taxnode{1}{Adaptive Control and Sim-to-Real Transfer}
\taxnode{1}{Control Theory and Formal Verification}
\taxnode{1}{Benchmarks, Datasets, and Evaluation}
\taxnode{1}{Applications of Motion Planning and Control}
\taxnode{1}{Surveys and Taxonomies}
\taxnode{0}{Reinforcement Learning for Robotics}
\taxnode{1}{Theoretical Foundations of Robot Reinforcement Learning}
\taxnode{1}{Model-Free Reinforcement Learning for Robotics}
\taxnode{1}{Model-Based Reinforcement Learning for Robotics}
\taxnode{1}{Offline Reinforcement Learning for Robotics}
\taxnode{1}{Imitation Learning and Reinforcement Learning from Demonstrations}
\taxnode{1}{Hierarchical and Skill-Based Reinforcement Learning for Robotics}
\taxnode{1}{Safe and Constrained Reinforcement Learning for Robotics}
\taxnode{1}{Sim-to-Real Transfer in Robot Reinforcement Learning}
\taxnode{1}{Multi-Task, Transfer, and Meta-Reinforcement Learning for Robotics}
\taxnode{1}{Perception and Representation Learning for Robot Reinforcement Learning}
\taxnode{1}{Multi-Agent Reinforcement Learning for Robotics}
\taxnode{1}{Reinforcement Learning for Robot Manipulation}
\taxnode{1}{Reinforcement Learning for Robot Locomotion and Mobility}
\taxnode{1}{Reinforcement Learning for Robot Navigation and Exploration}
\taxnode{1}{Benchmarking, Empirical Evaluation, and Surveys in Robot Reinforcement Learning}
\taxnode{0}{Imitation Learning and Learning from Demonstration}
\taxnode{1}{Behavior Cloning}
\taxnode{1}{Inverse Reinforcement Learning and Apprenticeship Learning}
\taxnode{1}{Adversarial Imitation Learning}
\taxnode{1}{Interactive and Active Imitation Learning}
\taxnode{1}{Reinforcement Learning from Demonstrations}
\taxnode{1}{Learning from Observation and Visual Imitation}
\taxnode{1}{Hierarchical and Skill-based Imitation Learning}
\taxnode{1}{Meta-, Few-shot, and Multi-task Imitation Learning}
\taxnode{1}{Transfer, Domain Adaptation, and Sim-to-real Imitation Learning}
\taxnode{1}{Safety, Robustness, and Generalization in Imitation Learning}
\taxnode{1}{Theoretical Foundations of Imitation Learning}
\taxnode{1}{Benchmarks, Datasets, and Evaluation for Imitation Learning}
\taxnode{1}{Applications of Imitation Learning in Robotics}
\taxnode{1}{Surveys and Taxonomies of Imitation Learning}
\taxnode{0}{Self-Supervised and Representation Learning for Robotics}
\taxnode{1}{Visual Representation Learning for Robotics}
\taxnode{1}{Multimodal Representation Learning for Robotics}
\taxnode{1}{Proprioceptive and Tactile Representation Learning}
\taxnode{1}{State Representation Learning for Control and Planning}
\taxnode{1}{World Models and Predictive Representation Learning}
\taxnode{1}{Goal, Skill, and Affordance Representation Learning}
\taxnode{1}{Language-Grounded Representation Learning for Robotics}
\taxnode{1}{Transferable Representations, Domain Adaptation, and Sim-to-Real}
\taxnode{1}{Learning Objectives and Methodologies for Robot Self-Supervision}
\taxnode{1}{Theory and Analysis of Robot Representation Learning}
\taxnode{1}{Evaluation, Benchmarks, and Datasets for Robot Representation Learning}
\taxnode{1}{Applications of Self-Supervised Representation Learning in Robotics}
\taxnode{1}{Surveys and Taxonomies of Robot Self-Supervised Learning}
\taxnode{0}{Sim-to-Real Transfer and Domain Adaptation}
\taxnode{1}{Domain Randomization}
\taxnode{1}{Synthetic Data Generation and Sensor Modeling}
\taxnode{1}{Image Translation and Visual Refinement}
\taxnode{1}{Representation Alignment and Domain-Invariant Learning}
\taxnode{1}{System Identification and Dynamics Adaptation}
\taxnode{1}{Policy Transfer and Reinforcement Learning Adaptation}
\taxnode{1}{Meta-Learning and Few-Shot Adaptation}
\taxnode{1}{Online, Continual, and Test-Time Adaptation}
\taxnode{1}{Cross-Embodiment and Morphology Transfer}
\taxnode{1}{Safety, Robustness, and Uncertainty-Aware Transfer}
\taxnode{1}{Theoretical Analysis of Domain Shift and Transfer}
\taxnode{1}{Benchmarks, Datasets, and Evaluation Protocols}
\taxnode{1}{Task-Oriented Sim-to-Real Applications}
\taxnode{0}{Human-Robot Interaction}
\taxnode{1}{Human Perception and Behavior Understanding}
\taxnode{1}{Intention, Goal, and Trajectory Prediction}
\taxnode{1}{Natural Language and Multimodal Communication}
\taxnode{1}{Learning from Demonstration, Preferences, and Interactive Feedback}
\taxnode{1}{Shared Autonomy, Teleoperation, and Mixed-Initiative Control}
\taxnode{1}{Human-Robot Collaboration and Joint Task Planning}
\taxnode{1}{Human-Aware Motion Planning and Social Navigation}
\taxnode{1}{Physical Human-Robot Interaction and Assistive Manipulation}
\taxnode{1}{Socially Assistive and Affective Robotics}
\taxnode{1}{Personalization, User Modeling, and Adaptation}
\taxnode{1}{Safety, Trust, Explainability, and Ethics in HRI}
\taxnode{1}{HRI Datasets, Benchmarks, Simulation, and Evaluation}
\taxnode{0}{Multi-Robot Systems and Swarm Robotics}
\taxnode{1}{Multi-Agent Reinforcement Learning}
\taxnode{1}{Swarm Intelligence and Collective Behavior}
\taxnode{1}{Decentralized Coordination and Distributed Control}
\taxnode{1}{Formation Control, Flocking, and Consensus}
\taxnode{1}{Task Allocation and Multi-Robot Scheduling}
\taxnode{1}{Multi-Robot Planning, Navigation, and Exploration}
\taxnode{1}{Cooperative Perception, Localization, and Mapping}
\taxnode{1}{Communication and Networked Multi-Robot Systems}
\taxnode{1}{Cooperative Manipulation and Transport}
\taxnode{1}{Human-Swarm Interaction and Mixed Human-Robot Teams}
\taxnode{1}{Safety, Robustness, and Formal Verification}
\taxnode{1}{Simulation, Benchmarks, and Sim-to-Real Transfer}
\taxnode{0}{Safe, Robust, and Reliable Robot Learning}
\taxnode{1}{Theoretical Foundations and Safety Guarantees}
\taxnode{1}{Safe Reinforcement Learning and Safe Exploration}
\taxnode{1}{Risk-Aware and Uncertainty-Aware Robot Learning}
\taxnode{1}{Robust Perception, State Estimation, and World Modeling}
\taxnode{1}{Adversarial and Distributionally Robust Robot Learning}
\taxnode{1}{Sim-to-Real Transfer and Domain Generalization}
\taxnode{1}{Safe Motion Planning and Learning-Based Control}
\taxnode{1}{Formal Verification, Validation, and Runtime Assurance}
\taxnode{1}{Fault Detection, Diagnosis, and Fault-Tolerant Learning}
\taxnode{1}{Human-Robot Safety and Shared Autonomy}
\taxnode{1}{Benchmarks, Datasets, and Safety Evaluation}
\taxnode{0}{Robotics Datasets, Simulation, and Benchmarks}
\taxnode{1}{Dataset Collection and Curation}
\taxnode{1}{Annotation and Ground-Truth Generation}
\taxnode{1}{Synthetic Data and Procedural Generation}
\taxnode{1}{Simulation Environments and Physics-Based Modeling}
\taxnode{1}{Sim-to-Real Transfer}
\taxnode{1}{Benchmark Design and Task Suites}
\taxnode{1}{Evaluation Metrics and Experimental Protocols}
\taxnode{1}{Dataset Quality, Bias, and Generalization Analysis}
\taxnode{1}{Reproducibility, Standardization, and Data Infrastructure}
\taxnode{1}{Surveys and Meta-Analysis of Robotics Resources}
\taxfield{scientific\_\allowbreak{}machine\_\allowbreak{}learning}
\taxnode{0}{Physics-informed learning}
\taxnode{1}{Physics-informed neural networks (PINNs)}
\taxnode{1}{Physics-informed operator learning}
\taxnode{1}{Inverse problems and system identification}
\taxnode{1}{Data assimilation and state estimation}
\taxnode{1}{Hybrid physics-machine learning models}
\taxnode{1}{Uncertainty quantification and Bayesian physics-informed learning}
\taxnode{1}{Theory and numerical analysis of physics-informed learning}
\taxnode{1}{Benchmarking and empirical evaluation}
\taxnode{1}{Scientific applications of physics-informed learning}
\taxnode{1}{Survey and review studies}
\taxnode{0}{Operator learning}
\taxnode{1}{Neural operator architectures}
\taxnode{1}{Geometry-aware operator learning}
\taxnode{1}{Physics-informed operator learning}
\taxnode{1}{Theoretical foundations of operator learning}
\taxnode{1}{Probabilistic operator learning}
\taxnode{1}{Data-efficient and multi-fidelity operator learning}
\taxnode{1}{Inverse problems and control}
\taxnode{1}{Empirical evaluation and benchmarks}
\taxnode{1}{Scientific and engineering applications}
\taxnode{0}{Surrogate modeling and reduced-order modeling}
\taxnode{1}{Parametric surrogate modeling}
\taxnode{1}{Operator learning}
\taxnode{1}{Dynamical systems emulation}
\taxnode{1}{Projection-based reduced-order modeling}
\taxnode{1}{Latent-space reduced-order modeling}
\taxnode{1}{Structure-preserving and physics-constrained reduction}
\taxnode{1}{Multifidelity and multiscale surrogate modeling}
\taxnode{1}{Probabilistic surrogate modeling and uncertainty quantification}
\taxnode{1}{Adaptive sampling and experimental design}
\taxnode{1}{Error analysis, stability, and verification}
\taxnode{1}{Benchmarking, comparative evaluation, and review studies}
\taxnode{1}{Scientific and engineering applications}
\taxnode{0}{Differentiable simulation and hybrid modeling}
\taxnode{1}{Differentiable numerical solvers}
\taxnode{1}{Differentiable physics engines}
\taxnode{1}{Gray-box modeling}
\taxnode{1}{Learned closure and constitutive modeling}
\taxnode{1}{System identification and model calibration}
\taxnode{1}{Data assimilation and state estimation}
\taxnode{1}{Simulation-based optimization and optimal control}
\taxnode{1}{Uncertainty quantification and probabilistic modeling}
\taxnode{1}{Multiscale and multiphysics hybrid modeling}
\taxnode{1}{Theory and numerical analysis}
\taxnode{1}{Benchmarks, datasets, and software systems}
\taxnode{1}{Scientific and engineering applications}
\taxnode{1}{Surveys and review studies}
\taxnode{0}{Inverse problems and data assimilation}
\taxnode{1}{Learned reconstruction and unrolled optimization}
\taxnode{1}{Physics-informed and PDE-constrained inversion}
\taxnode{1}{Bayesian inverse problems and uncertainty quantification}
\taxnode{1}{Generative modeling and learned priors for inverse problems}
\taxnode{1}{Operator learning and surrogate-assisted inversion}
\taxnode{1}{Sequential and variational data assimilation}
\taxnode{1}{Observation design and sensor placement}
\taxnode{1}{Theory, identifiability, and convergence}
\taxnode{1}{Benchmarks, datasets, and evaluation}
\taxnode{1}{Scientific and engineering applications}
\taxnode{0}{Scientific discovery and system identification}
\taxnode{1}{Equation discovery}
\taxnode{1}{Input-output system identification}
\taxnode{1}{State-space and latent dynamics inference}
\taxnode{1}{PDE and operator discovery}
\taxnode{1}{Causal and mechanistic discovery}
\taxnode{1}{Parameter estimation and model calibration}
\taxnode{1}{Optimal experimental design}
\taxnode{1}{Identifiability, uncertainty quantification, and validation}
\taxnode{1}{Benchmarks and evaluation}
\taxnode{1}{Scientific applications}
\taxnode{0}{Uncertainty quantification and probabilistic modeling}
\taxnode{1}{Bayesian deep learning}
\taxnode{1}{Approximate Bayesian inference}
\taxnode{1}{Gaussian processes and kernel probabilistic models}
\taxnode{1}{Probabilistic graphical models and probabilistic programming}
\taxnode{1}{Deep generative and latent-variable models}
\taxnode{1}{Predictive uncertainty estimation and calibration}
\taxnode{1}{Conformal prediction and distribution-free uncertainty quantification}
\taxnode{1}{Distribution shift and out-of-distribution uncertainty}
\taxnode{1}{Uncertainty propagation and probabilistic numerics}
\taxnode{1}{Bayesian inverse problems and surrogate modeling}
\taxnode{1}{Evaluation, benchmarking, and uncertainty metrics}
\taxnode{0}{Scientific optimization, control, and experimental design}
\taxnode{1}{Bayesian optimization}
\taxnode{1}{Black-box and simulation-based optimization}
\taxnode{1}{Differentiable and PDE-constrained optimization}
\taxnode{1}{Inverse design}
\taxnode{1}{System identification and model calibration}
\taxnode{1}{Robust, stochastic, and multi-objective optimization}
\taxnode{1}{Optimal control and model predictive control}
\taxnode{1}{Reinforcement learning for control}
\taxnode{1}{Optimal experimental design and adaptive sampling}
\taxnode{1}{Autonomous experimentation}
\taxnode{0}{Scientific representation learning and foundation models}
\taxnode{1}{Self-Supervised and Unsupervised Learning for Scientific Data}
\taxnode{1}{Geometric and Equivariant Representation Learning}
\taxnode{1}{Graph Representation Learning for Scientific Systems}
\taxnode{1}{Representation Learning for Scientific Imaging}
\taxnode{1}{Sequence and Language Foundation Models for Science}
\taxnode{1}{Spatiotemporal and Field Foundation Models}
\taxnode{1}{Multimodal Scientific Foundation Models}
\taxnode{1}{Generative Models for Scientific Discovery and Design}
\taxnode{1}{Transfer Learning, Domain Adaptation, and Few-Shot Scientific Learning}
\taxnode{1}{Evaluation, Benchmarks, and Trustworthy Scientific Foundation Models}
\taxnode{1}{Theory and Scaling Laws for Scientific Representation Learning}
\taxnode{0}{Learned numerical methods and solver acceleration}
\taxnode{1}{Learned iterative methods for linear systems and eigenproblems}
\taxnode{1}{Learned preconditioning, multigrid, and domain decomposition}
\taxnode{1}{Learned nonlinear equation solvers and fixed-point acceleration}
\taxnode{1}{Learned time integration for ODEs, SDEs, and PDEs}
\taxnode{1}{Learned discretization, meshing, and adaptive refinement}
\taxnode{1}{Reduced-order modeling and surrogate-assisted simulation}
\taxnode{1}{Learned optimization and inverse-problem solvers}
\taxnode{1}{Learned Monte Carlo, quadrature, and stochastic simulation acceleration}
\taxnode{1}{Theory, stability, and convergence analysis of learned numerical methods}
\taxnode{1}{Benchmarks, datasets, and empirical evaluation of learned solvers}
\taxnode{1}{Scientific applications of learned solver acceleration}
\taxnode{1}{Surveys, taxonomies, and software frameworks for learned numerical methods}
\taxnode{0}{Theory and numerical analysis}
\taxnode{1}{Approximation and expressivity theory}
\taxnode{1}{Generalization and statistical learning theory}
\taxnode{1}{Optimization theory}
\taxnode{1}{Numerical analysis of learned solvers}
\taxnode{1}{Dynamical systems theory}
\taxnode{1}{PDE and operator-learning theory}
\taxnode{1}{Inverse problems and identifiability}
\taxnode{1}{Error estimation and adaptive methods}
\taxnode{1}{Uncertainty quantification and probabilistic numerics}
\taxnode{1}{Benchmarking and evaluation methodology}
\taxnode{0}{Benchmarks, datasets, and evaluation}
\taxnode{1}{Dataset Collection and Curation}
\taxnode{1}{Data Annotation, Labeling, and Documentation}
\taxnode{1}{Dataset Quality, Bias, and Auditing}
\taxnode{1}{Synthetic Data and Simulation Benchmarks}
\taxnode{1}{Benchmark and Task Design}
\taxnode{1}{Evaluation Metrics and Scoring Rules}
\taxnode{1}{Human and Expert Evaluation}
\taxnode{1}{Experimental Protocols and Statistical Evaluation}
\taxnode{1}{Reproducibility and Benchmarking Infrastructure}
\taxnode{1}{Robustness, Generalization, and Distribution-Shift Evaluation}
\taxnode{1}{Fairness, Privacy, and Safety Evaluation}
\taxnode{1}{Meta-evaluation and Benchmark Analysis}
\taxnode{0}{Scientific applications and case studies}
\taxnode{1}{Physical sciences}
\taxnode{1}{Chemistry and materials science}
\taxnode{1}{Biology and bioinformatics}
\taxnode{1}{Biomedical and health sciences}
\taxnode{1}{Earth and environmental sciences}
\taxnode{1}{Astronomy and space science}
\taxnode{1}{Engineering and energy systems}
\taxnode{1}{Scientific imaging and instrumentation}
\taxnode{1}{Scientific discovery and hypothesis generation}
\taxnode{1}{Benchmarks, datasets, and challenge tasks}
\taxnode{1}{Validation, reproducibility, and domain-specific methodology}
\taxnode{1}{Surveys and comparative studies}
\taxnode{0}{Surveys and reviews}
\taxnode{1}{Methodological surveys}
\taxnode{1}{Theoretical and foundational reviews}
\taxnode{1}{Application-domain surveys}
\taxnode{1}{Data, datasets, and benchmark reviews}
\taxnode{1}{Simulation-integrated and hybrid-model reviews}
\taxnode{1}{Uncertainty quantification and probabilistic reviews}
\taxnode{1}{Interpretability, robustness, and trustworthiness reviews}
\taxnode{1}{Systems, software, and computational infrastructure reviews}
\taxnode{1}{Comparative reviews and meta-analyses}
\taxnode{1}{Bibliometric and trend reviews}
\taxfield{systems}
\taxnode{0}{Distributed Training Systems}
\taxnode{1}{Data-Parallel Training}
\taxnode{1}{Model and Tensor Parallelism}
\taxnode{1}{Pipeline Parallelism}
\taxnode{1}{Communication-Efficient Training}
\taxnode{1}{Synchronization and Consistency Models}
\taxnode{1}{Decentralized Training}
\taxnode{1}{Distributed Training Frameworks and Runtime Systems}
\taxnode{1}{Scheduling and Resource Management}
\taxnode{1}{Fault Tolerance and Elastic Training}
\taxnode{1}{Benchmarking, Profiling, and Performance Modeling}
\taxnode{0}{Inference and Serving Systems}
\taxnode{1}{Inference Runtime Systems}
\taxnode{1}{Model Serving Architectures}
\taxnode{1}{Scheduling, Batching, and SLO Management}
\taxnode{1}{Resource Management, Placement, and Autoscaling}
\taxnode{1}{Distributed and Multi-Tenant Serving}
\taxnode{1}{Edge, Mobile, and Collaborative Inference Serving}
\taxnode{1}{Generative Model and LLM Serving}
\taxnode{2}{Workload Characterization and Benchmarking}
\taxnode{2}{Resource Management, Multi-Tenancy, and Quality of Service}
\taxnode{2}{Decoding and Iterative Generation Optimization}
\taxnode{2}{Memory, Caching, and State Management}
\taxnode{2}{Distributed, Disaggregated, and Elastic Serving}
\taxnode{2}{Runtime, Compiler, and Hardware Co-Design}
\taxnode{2}{Model Compression and Serving-Aware Optimization}
\taxnode{2}{Multi-Model, Adapter, and Personalized Serving}
\taxnode{2}{Multi-Stage and Retrieval-Augmented Serving}
\taxnode{2}{Reliability, Observability, and Fault Tolerance}
\taxnode{2}{Security, Privacy, and Safety in Serving}
\taxnode{2}{Cost and Energy Efficient Serving}
\taxnode{1}{Reliability, Observability, and Serving Operations}
\taxnode{1}{Benchmarking, Measurement, and Systematization}
\taxnode{0}{ML Compilers and Runtime Systems}
\taxnode{1}{Intermediate Representations and Lowering}
\taxnode{1}{Graph Transformation and Optimization}
\taxnode{1}{Automatic Differentiation and Training Compilation}
\taxnode{1}{Kernel Generation and Operator Scheduling}
\taxnode{2}{Kernel Code Generation}
\taxnode{3}{Intermediate Representations and Lowering}
\taxnode{3}{Parallelization and Synchronization Code Generation}
\taxnode{3}{Memory-Hierarchy-Aware Code Generation}
\taxnode{3}{Vectorization and Tensorization}
\taxnode{3}{Fused and Specialized Kernel Generation}
\taxnode{3}{Dynamic-Shape and Shape-Polymorphic Code Generation}
\taxnode{3}{Sparse and Irregular Kernel Generation}
\taxnode{3}{Quantized and Low-Precision Kernel Generation}
\taxnode{3}{Heterogeneous and Accelerator-Specific Code Generation}
\taxnode{3}{Autotuning and Search-Based Code Generation}
\taxnode{3}{Correctness, Verification, and Testing}
\taxnode{3}{Benchmarking and Performance Evaluation}
\taxnode{3}{Compiler Infrastructure and Runtime Integration}
\taxnode{3}{Application-Specific Kernel Generation}
\taxnode{3}{Surveys and Comparative Studies}
\taxnode{2}{Loop-Nest Transformation and Schedule Optimization}
\taxnode{3}{Schedule Languages and Intermediate Representations}
\taxnode{3}{Dependence Analysis and Legality Checking}
\taxnode{3}{Polyhedral Scheduling and Affine Transformations}
\taxnode{3}{Locality-Oriented Loop Transformations}
\taxnode{3}{Parallelization, Vectorization, and Unrolling}
\taxnode{3}{Memory Hierarchy and Data Movement Optimization}
\taxnode{3}{Tensorization and Accelerator-Aware Mapping}
\taxnode{3}{Operator-Specific and Workload-Specific Scheduling}
\taxnode{3}{Schedule Search and Autotuning}
\taxnode{3}{Cost Modeling and Performance Prediction}
\taxnode{3}{Dynamic and Runtime-Adaptive Scheduling}
\taxnode{3}{Benchmarking and Empirical Evaluation}
\taxnode{3}{Surveys, Taxonomies, and Methodological Frameworks}
\taxnode{2}{Operator Fusion and Graph-Level Kernelization}
\taxnode{3}{Fusion Legality and Dependence Analysis}
\taxnode{3}{Graph Rewriting and Fusion Intermediate Representations}
\taxnode{3}{Subgraph Partitioning and Region Formation}
\taxnode{3}{Fusion Planning, Cost Modeling, and Search}
\taxnode{3}{Code Generation and Schedule Synthesis for Fused Kernels}
\taxnode{3}{Memory-, Layout-, and Communication-Aware Fusion}
\taxnode{3}{Dynamic-Shape, Sparse, and Just-in-Time Fusion}
\taxnode{3}{Hardware-Aware and Accelerator-Specific Fusion}
\taxnode{3}{Training-Phase and Model-Specific Fusion}
\taxnode{3}{Correctness, Numerical Stability, and Verification}
\taxnode{3}{Benchmarking and Workload Characterization of Fusion Systems}
\taxnode{3}{Surveys and Taxonomies of Operator Fusion}
\taxnode{2}{Auto-Tuning and Schedule Search}
\taxnode{2}{Cost Modeling and Performance Prediction}
\taxnode{2}{Memory Hierarchy and Data Layout Optimization}
\taxnode{2}{Hardware-Aware Kernel Mapping}
\taxnode{3}{Spatial Mapping and Parallelization}
\taxnode{3}{Temporal Mapping and Loop Scheduling}
\taxnode{3}{Memory Hierarchy Mapping}
\taxnode{3}{Data Layout and Data Movement Optimization}
\taxnode{3}{Instruction-Level Mapping and Tensorization}
\taxnode{3}{Heterogeneous and Accelerator Mapping}
\taxnode{3}{Sparse and Irregular Mapping}
\taxnode{3}{Mixed-Precision and Quantized Mapping}
\taxnode{3}{Analytical Cost Modeling and Formal Optimization}
\taxnode{3}{Auto-Tuning and Search-Based Mapping}
\taxnode{3}{Dynamic and Runtime-Adaptive Mapping}
\taxnode{3}{Energy- and Thermal-Aware Mapping}
\taxnode{3}{Portable and Retargetable Mapping Frameworks}
\taxnode{3}{Operator- and Workload-Specific Mapping}
\taxnode{3}{Benchmarking and Empirical Evaluation}
\taxnode{3}{Surveys and Taxonomies}
\taxnode{2}{Runtime Operator Scheduling and Execution Planning}
\taxnode{3}{DAG-Based Task Scheduling}
\taxnode{3}{Memory-Aware Execution Planning}
\taxnode{3}{Heterogeneous Device Placement and Scheduling}
\taxnode{3}{Asynchronous and Stream Scheduling}
\taxnode{3}{Dynamic Shape and Control-Flow Execution}
\taxnode{3}{Distributed Execution and Communication Scheduling}
\taxnode{3}{Runtime Kernel Selection and Specialization}
\taxnode{3}{Adaptive Cost Modeling and Online Scheduling}
\taxnode{3}{Latency-, Throughput-, and QoS-Aware Scheduling}
\taxnode{3}{Benchmarking, Profiling, and Trace Analysis}
\taxnode{3}{Survey and Comparative Studies of Execution Runtimes}
\taxnode{2}{Specialized Kernel Generation for Irregular Workloads}
\taxnode{3}{Sparse Tensor and Structured Sparsity Kernel Generation}
\taxnode{3}{Graph and Message-Passing Kernel Generation}
\taxnode{3}{Indexing, Gather/\allowbreak{}Scatter, and Segmented Reduction Kernels}
\taxnode{3}{Dynamic-Shape and Ragged Tensor Kernel Generation}
\taxnode{3}{Data-Dependent Control Flow and Conditional Computation Kernels}
\taxnode{3}{Compiler Analyses and Intermediate Representations for Irregular Kernels}
\taxnode{3}{Runtime Specialization and Just-in-Time Kernel Generation}
\taxnode{3}{Load Balancing and Work Partitioning for Irregular Parallelism}
\taxnode{3}{Memory Layout, Locality, and Data Movement Optimization}
\taxnode{3}{Autotuning, Cost Modeling, and Schedule Search}
\taxnode{3}{Hardware-Specific and Heterogeneous Code Generation}
\taxnode{3}{Benchmarking, Workload Characterization, and Taxonomy}
\taxnode{2}{Benchmarking, Profiling, and Empirical Evaluation}
\taxnode{1}{Auto-Tuning and Cost Modeling}
\taxnode{1}{Memory Planning and Data Layout Optimization}
\taxnode{1}{Runtime Scheduling and Execution Engines}
\taxnode{1}{Distributed Execution and Parallelization}
\taxnode{1}{Heterogeneous and Accelerator-Aware Compilation}
\taxnode{1}{Dynamic, Sparse, and Irregular Workloads}
\taxnode{1}{Inference Serving and Deployment Runtimes}
\taxnode{1}{Compiler Correctness, Testing, and Verification}
\taxnode{1}{Benchmarking and Performance Evaluation}
\taxnode{0}{Hardware Acceleration and Co-Design}
\taxnode{1}{Accelerator Architecture Design}
\taxnode{1}{Memory, Dataflow, and Interconnect Design}
\taxnode{1}{Compiler, Mapping, and Runtime Systems}
\taxnode{1}{Algorithm–Hardware Co-Design}
\taxnode{1}{Low-Precision and Sparse Computing}
\taxnode{2}{Numerical Formats and Low-Precision Arithmetic}
\taxnode{3}{Fixed-Point and Integer Arithmetic}
\taxnode{3}{Reduced-Precision Floating-Point Formats}
\taxnode{3}{Block Floating-Point and Shared-Exponent Representations}
\taxnode{3}{Logarithmic, Posit, and Other Alternative Number Systems}
\taxnode{3}{Binary, Ternary, and Ultra-Low-Bit Computation}
\taxnode{3}{Mixed-Precision and Adaptive Precision Methods}
\taxnode{3}{Low-Precision Training and Backpropagation Arithmetic}
\taxnode{3}{Rounding, Accumulation, and Error-Compensation Methods}
\taxnode{3}{Approximate Arithmetic Units and Operators}
\taxnode{3}{Numerical Analysis and Error Modeling}
\taxnode{3}{Benchmarking and Workload-Specific Format Evaluation}
\taxnode{2}{Neural Network Quantization}
\taxnode{3}{Quantization Theory and Optimization}
\taxnode{3}{Post-Training Quantization}
\taxnode{3}{Quantization-Aware Training}
\taxnode{3}{Ultra-Low-Bit Quantization}
\taxnode{3}{Mixed-Precision Quantization}
\taxnode{3}{Dynamic and Adaptive Quantization}
\taxnode{3}{Low-Precision Training}
\taxnode{3}{Hardware-Aware Quantization and Co-Design}
\taxnode{3}{Robustness and Reliability of Quantized Networks}
\taxnode{3}{Benchmarking and Evaluation of Quantized Models}
\taxnode{3}{Architecture- and Application-Specific Quantization}
\taxnode{3}{Surveys and Empirical Studies of Quantization}
\taxnode{2}{Mixed-Precision Optimization}
\taxnode{2}{Binary and Ternary Neural Networks}
\taxnode{2}{Pruning and Weight Sparsification}
\taxnode{2}{Structured and Semi-Structured Sparsity}
\taxnode{2}{Dynamic Sparsity and Sparse Training}
\taxnode{2}{Activation Sparsity and Conditional Computation}
\taxnode{2}{Sparse Tensor Representations and Compression}
\taxnode{2}{Hardware Architectures and Dataflows for Low-Precision and Sparse Computing}
\taxnode{2}{Compiler, Kernel, and Runtime Support}
\taxnode{3}{Intermediate Representations and Graph Lowering}
\taxnode{3}{Quantization-Aware Compilation}
\taxnode{3}{Sparsity-Aware Compilation}
\taxnode{3}{Operator Fusion and Graph Rewriting}
\taxnode{3}{Code Generation and Auto-Tuning}
\taxnode{3}{Low-Precision Kernel Design}
\taxnode{3}{Sparse Tensor Kernel Design}
\taxnode{3}{Memory Layout and Data Movement Optimization}
\taxnode{3}{Runtime Scheduling and Execution Management}
\taxnode{3}{Adaptive Precision and Dynamic Sparsity Runtime}
\taxnode{3}{Portability and Backend Integration}
\taxnode{3}{Benchmarking, Profiling, and Performance Evaluation}
\taxnode{2}{Theory, Benchmarking, and Evaluation}
\taxnode{1}{Training Acceleration}
\taxnode{1}{Inference Acceleration and Serving}
\taxnode{1}{Edge, Mobile, and Embedded ML Hardware}
\taxnode{1}{Heterogeneous and Multi-Accelerator Systems}
\taxnode{1}{Emerging ML Hardware}
\taxnode{1}{Benchmarking, Simulation, and Performance Modeling}
\taxnode{1}{Reliability, Security, and Power/\allowbreak{}Thermal Management}
\taxnode{0}{Resource-Efficient ML Systems}
\taxnode{1}{Model Compression and Sparsity}
\taxnode{1}{Quantization and Low-Precision Computing}
\taxnode{2}{Quantization Theory and Numerical Analysis}
\taxnode{2}{Post-Training Quantization}
\taxnode{2}{Quantization-Aware Training}
\taxnode{2}{Mixed-Precision and Adaptive Bit-Width Methods}
\taxnode{2}{Binary, Ternary, and Ultra-Low-Bit Networks}
\taxnode{2}{Low-Precision Training and Optimization}
\taxnode{2}{Hardware-Aware Quantization and Accelerator Co-Design}
\taxnode{2}{Benchmarking, Evaluation, and Quantization Toolchains}
\taxnode{2}{Application-Specific Quantization and Deployment}
\taxnode{1}{Efficient Neural Architecture Design}
\taxnode{1}{Adaptive and Conditional Computation}
\taxnode{1}{Memory-Efficient Training, Adaptation, and Fine-Tuning}
\taxnode{1}{Communication-Efficient Distributed and Federated Learning}
\taxnode{1}{Compiler, Runtime, and Serving Optimization}
\taxnode{1}{Hardware-Software Co-Design for ML}
\taxnode{1}{Edge and On-Device Machine Learning}
\taxnode{1}{Energy-Efficient and Sustainable ML}
\taxnode{1}{Efficiency Benchmarking and Evaluation}
\taxnode{1}{Theoretical Foundations of Resource-Constrained Learning}
\taxnode{0}{Data and Storage Systems for ML}
\taxnode{1}{Data Pipeline Systems for ML}
\taxnode{1}{Dataset Storage Layouts and File Formats}
\taxnode{1}{Distributed and Cloud Storage for ML}
\taxnode{1}{Input Pipeline Optimization}
\taxnode{1}{Dataset Versioning, Lineage, and Metadata Management}
\taxnode{1}{Feature Stores and Feature Management}
\taxnode{1}{Data Labeling and Annotation Systems}
\taxnode{1}{Data Quality, Validation, and Observability}
\taxnode{1}{Privacy, Security, and Governance for ML Data}
\taxnode{1}{Embedding and Vector Data Management}
\taxnode{1}{Benchmarking and Workload Characterization of ML Data Systems}
\taxnode{0}{Edge, Mobile, and Embedded ML Systems}
\taxnode{1}{Efficient Model Design and Compression}
\taxnode{1}{Hardware-Aware Neural Architecture Search and Co-Design}
\taxnode{1}{Inference Optimization, Compilers, and Runtimes}
\taxnode{1}{Embedded ML Hardware and Accelerators}
\taxnode{1}{On-Device Training, Adaptation, and Continual Learning}
\taxnode{1}{Distributed Edge Intelligence and Edge–Cloud Collaboration}
\taxnode{1}{Federated and Privacy-Preserving Edge Learning}
\taxnode{1}{Resource Management and Quality-of-Service for Edge ML}
\taxnode{1}{Security, Safety, and Reliability of Edge ML Systems}
\taxnode{1}{Benchmarks, Evaluation, and Applications of TinyML and Edge ML}
\taxnode{0}{Cluster Management and Scheduling for ML}
\taxnode{1}{Scheduling Theory and Performance Modeling}
\taxnode{1}{Workload Characterization and Benchmarking}
\taxnode{1}{Resource Allocation and Admission Control}
\taxnode{1}{Job Scheduling and Queueing Policies}
\taxnode{1}{Placement, Packing, and Gang Scheduling}
\taxnode{1}{Heterogeneous Accelerator Scheduling}
\taxnode{1}{Communication-, Topology-, and Data-Locality-Aware Scheduling}
\taxnode{1}{Multi-Tenant Fairness, Isolation, and QoS Management}
\taxnode{1}{Elasticity, Auto-scaling, and Adaptive Resource Management}
\taxnode{1}{Fault Tolerance, Preemption, and Straggler Mitigation}
\taxnode{1}{ML Pipeline, Workflow, and Experiment Scheduling}
\taxnode{1}{Inference Serving and Online Scheduling}
\taxnode{1}{Energy- and Cost-Aware Cluster Management}
\taxnode{0}{Reliability, Monitoring, and MLOps Systems}
\taxnode{1}{ML Pipeline Orchestration and Workflow Management}
\taxnode{1}{Data Validation and Feature Management}
\taxnode{1}{Experiment Management and Reproducibility}
\taxnode{1}{Model and Data Versioning, Registry, and Lineage}
\taxnode{1}{Continuous Integration, Delivery, and Training for ML}
\taxnode{1}{Model Deployment and Serving Systems}
\taxnode{1}{Monitoring, Observability, and Alerting}
\taxnode{1}{Drift Detection and Post-deployment Model Evaluation}
\taxnode{1}{Testing, Verification, and Validation of ML Systems}
\taxnode{1}{Failure Diagnosis, Debugging, and Incident Response}
\taxnode{1}{Reliability Engineering and Fault Tolerance for ML Systems}
\taxnode{1}{Governance, Auditability, and Compliance in MLOps}
\taxnode{1}{Benchmarking and Evaluation of MLOps Systems}
\taxnode{0}{Benchmarking and Performance Evaluation}
\taxnode{1}{Benchmark Suite Design and Workload Characterization}
\taxnode{1}{Performance Metrics and Evaluation Methodology}
\taxnode{1}{Training System Benchmarking}
\taxnode{1}{Inference and Serving Benchmarking}
\taxnode{1}{Distributed and Scalability Evaluation}
\taxnode{1}{Hardware Accelerator and System Stack Evaluation}
\taxnode{1}{Edge, Mobile, and Embedded ML Benchmarking}
\taxnode{1}{Energy, Cost, and Sustainability Evaluation}
\taxnode{1}{Reproducibility, Standardization, and Benchmark Validation}
\taxfield{theory}
\taxnode{0}{Generalization and Statistical Learning Theory}
\taxnode{1}{PAC Learning and Sample Complexity}
\taxnode{1}{Capacity Control and Uniform Convergence}
\taxnode{1}{PAC-Bayesian Generalization Theory}
\taxnode{1}{Information-Theoretic and Compression-Based Generalization}
\taxnode{1}{Algorithmic Stability and Robustness}
\taxnode{1}{Margin-, Norm-, and Flatness-Based Analyses}
\taxnode{1}{Optimization-Dependent Generalization and Implicit Regularization}
\taxnode{1}{Overparameterization, Interpolation, and Double Descent}
\taxnode{1}{Distribution Shift and Out-of-Distribution Generalization}
\taxnode{1}{Neural Network and Representation Generalization}
\taxnode{1}{Empirical Generalization Studies and Benchmarks}
\taxnode{1}{Surveys and Unifying Perspectives}
\taxnode{0}{Optimization Theory}
\taxnode{1}{Deterministic First-Order Optimization}
\taxnode{1}{Stochastic Optimization}
\taxnode{1}{Accelerated, Momentum, and Adaptive Methods}
\taxnode{1}{Second-Order and Curvature-Aware Optimization}
\taxnode{1}{Convex, Composite, and Proximal Optimization}
\taxnode{1}{Nonconvex Optimization}
\taxnode{1}{Optimization Landscapes and Critical Point Analysis}
\taxnode{1}{Gradient Flow and Training Dynamics}
\taxnode{1}{Saddle-Point and Min-Max Optimization}
\taxnode{1}{Constrained and Regularized Optimization}
\taxnode{1}{Distributed, Parallel, and Federated Optimization}
\taxnode{1}{Online, Bandit, and Zeroth-Order Optimization}
\taxnode{1}{Bilevel and Hyperparameter Optimization}
\taxnode{1}{Implicit Bias and Optimization-Induced Generalization}
\taxnode{1}{Convergence, Complexity, and Lower Bounds}
\taxnode{1}{Optimization for Deep Network Training}
\taxnode{0}{Approximation Theory and Expressivity}
\taxnode{1}{Universal Approximation Theorems}
\taxnode{1}{Approximation Rates and Optimality}
\taxnode{1}{Depth Separation and Size Trade-offs}
\taxnode{1}{Architecture-Specific Expressivity}
\taxnode{1}{Structured Function and Operator Approximation}
\taxnode{1}{Quantitative Expressivity Measures}
\taxnode{1}{Expressivity under Constraints}
\taxnode{1}{Empirical Expressivity Evaluation and Benchmarks}
\taxnode{1}{Constructive Approximation and Network Design}
\taxnode{1}{Surveys and Unifying Frameworks}
\taxnode{0}{Training Dynamics}
\taxnode{1}{Convergence and Stability Analysis}
\taxnode{1}{Stochastic Optimization Dynamics}
\taxnode{1}{Initialization and Early-Time Dynamics}
\taxnode{1}{Scaling Limits and Analytical Regimes}
\taxnode{1}{Feature and Representation Dynamics}
\taxnode{1}{Implicit Bias and Regularization}
\taxnode{1}{Memorization and Generalization Dynamics}
\taxnode{1}{Loss Landscape and Curvature Dynamics}
\taxnode{1}{Architecture and Normalization Effects}
\taxnode{1}{Data Order and Curriculum Effects}
\taxnode{1}{Hyperparameter and Schedule Effects}
\taxnode{1}{Diagnostics and Measurement Methodology}
\taxnode{1}{Empirical Characterization and Benchmarking}
\taxnode{1}{Dynamics-Informed Training Methods}
\taxnode{0}{Implicit Bias and Regularization}
\taxnode{1}{Optimization-Induced Implicit Bias}
\taxnode{1}{Stochastic Optimization and Noise-Based Regularization}
\taxnode{1}{Early Stopping and Iterative Regularization}
\taxnode{1}{Norm-Based and Penalty-Based Regularization}
\taxnode{1}{Structured Regularization}
\taxnode{1}{Normalization and Reparameterization Effects}
\taxnode{1}{Data Augmentation and Invariance-Based Regularization}
\taxnode{1}{Margin, Flatness, and Minimum-Norm Bias}
\taxnode{1}{Overparameterization, Interpolation, and Benign Overfitting}
\taxnode{1}{Generalization Bounds and Complexity Measures}
\taxnode{1}{Empirical Characterization and Benchmarking}
\taxnode{1}{Surveys and Unifying Frameworks}
\taxnode{0}{Representation Learning Theory}
\taxnode{1}{Information-Theoretic Foundations}
\taxnode{1}{Representation Geometry and Manifold Theory}
\taxnode{1}{Expressivity and Hierarchical Feature Learning}
\taxnode{1}{Disentanglement and Identifiability}
\taxnode{1}{Symmetry, Invariance, and Equivariance}
\taxnode{1}{Self-Supervised and Contrastive Learning Theory}
\taxnode{1}{Latent Variable, Autoencoder, and Generative Representation Models}
\taxnode{1}{Optimization Dynamics and Implicit Bias in Feature Learning}
\taxnode{1}{Generalization, Transfer, and Sample Complexity}
\taxnode{1}{Causal and Invariant Representation Learning}
\taxnode{0}{Architecture and Inductive Bias Theory}
\taxnode{1}{Expressivity and Approximation Theory}
\taxnode{1}{Depth, Width, and Parameterization Trade-offs}
\taxnode{1}{Generalization and Sample Complexity of Architectures}
\taxnode{1}{Optimization, Trainability, and Signal Propagation}
\taxnode{1}{Implicit Bias and Architectural Regularization}
\taxnode{1}{Symmetry, Invariance, and Equivariance}
\taxnode{1}{Compositionality, Hierarchy, and Modularity}
\taxnode{1}{Locality, Sparsity, and Structured Connectivity}
\taxnode{1}{Spectral and Frequency Bias}
\taxnode{1}{Theory of Specific Architecture Families}
\taxnode{1}{Domain-Specific Architectural Priors}
\taxnode{1}{Empirical Probing and Benchmarking of Inductive Bias}
\taxnode{1}{Architecture Design Principles and Neural Architecture Search}
\taxnode{1}{Surveys, Taxonomies, and Unifying Frameworks}
\taxnode{0}{Robustness and Stability Theory}
\taxnode{1}{Adversarial Robustness}
\taxnode{1}{Certified Robustness and Formal Verification}
\taxnode{1}{Distribution Shift Robustness}
\taxnode{1}{Distributionally Robust Optimization}
\taxnode{1}{Algorithmic Stability and Generalization}
\taxnode{1}{Sensitivity and Perturbation Analysis}
\taxnode{1}{Optimization and Training Stability}
\taxnode{1}{Robust Statistics and Corruption Models}
\taxnode{1}{Robustness Evaluation and Benchmarks}
\taxnode{0}{Information-Theoretic Learning Theory}
\taxnode{1}{Information-Theoretic Generalization}
\taxnode{1}{Information-Theoretic Lower Bounds and Sample Complexity}
\taxnode{1}{Information Bottleneck}
\taxnode{1}{Rate-Distortion Theory for Learning}
\taxnode{1}{Minimum Description Length and Compression-Based Learning}
\taxnode{1}{Entropy and Mutual Information Estimation}
\taxnode{1}{Privacy, Memorization, and Information Leakage}
\taxnode{1}{Communication-Constrained and Distributed Learning}
\taxnode{1}{Adaptive Data Analysis}
\taxnode{1}{Information-Theoretic Sequential Decision Making}
\taxnode{0}{Computational Complexity of Learning}
\taxnode{1}{Time Complexity of Learning}
\taxnode{1}{Space Complexity and Streaming Learning}
\taxnode{1}{Optimization Complexity}
\taxnode{1}{Query Complexity and Active Learning}
\taxnode{1}{Communication Complexity in Learning}
\taxnode{1}{Computational Hardness and Lower Bounds}
\taxnode{1}{Statistical-Computational Trade-offs}
\taxnode{1}{Average-Case, Smoothed, and Fine-Grained Complexity}
\taxnode{1}{Online Learning and Mistake-Bound Complexity}
\taxnode{1}{Learnability of Specific Hypothesis Classes}
\taxnode{1}{Empirical Scaling Laws and Complexity Benchmarking}
\taxnode{1}{Complexity Measures and Analytical Frameworks}
\taxnode{0}{High-Dimensional Learning Theory}
\taxnode{1}{Generalization and Sample Complexity}
\taxnode{1}{Minimax, Information-Theoretic, and Statistical-Computational Limits}
\taxnode{1}{Concentration of Measure and Empirical Process Theory}
\taxnode{1}{Sparse and Structured Estimation}
\taxnode{1}{High-Dimensional Inference and Uncertainty Quantification}
\taxnode{1}{Random Matrix Theory and Spectral Analysis}
\taxnode{1}{Dimension Reduction, Random Projections, and Sketching}
\taxnode{1}{Optimization Dynamics and Landscapes in High Dimensions}
\taxnode{1}{Implicit Bias and Regularization}
\taxnode{1}{Overparameterization, Interpolation, and Double Descent}
\taxnode{1}{Robust High-Dimensional Statistics and Distribution Shift}
\taxnode{0}{Probabilistic and Bayesian Theory}
\taxnode{1}{Bayesian Neural Networks}
\taxnode{1}{Variational Inference}
\taxnode{1}{Monte Carlo and Sampling Methods}
\taxnode{1}{Probabilistic Graphical Models}
\taxnode{1}{Latent Variable Models}
\taxnode{1}{Hierarchical and Nonparametric Bayesian Methods}
\taxnode{1}{PAC-Bayesian Theory}
\taxnode{1}{Uncertainty Quantification and Calibration}
\taxnode{1}{Model Evidence and Bayesian Model Selection}

\end{multicols}
\endgroup

\end{document}